\documentclass[twoside,11pt]{article}

\usepackage[utf8]{inputenc}
\usepackage{graphicx}
\usepackage{amsfonts} 
\usepackage{amsmath}
\usepackage{amsthm}
\usepackage{mathrsfs}
\usepackage{mathabx}
\usepackage{microtype}
\usepackage{mathtools}
\usepackage{xcolor}
\usepackage{caption}
\usepackage{subcaption}
\usepackage{stmaryrd} 
\usepackage{amssymb} 
\usepackage{mathtools}
\mathtoolsset{showonlyrefs}

\usepackage{tikz-cd}
\usepackage{tikz}
\tikzset{node distance=2cm, auto}
\usepackage[percent]{overpic}

\usepackage[preprint]{jmlr2e}

\usepackage{xspace}
\newcommand*{\eg}{e.g.\@\xspace}
\newcommand*{\ie}{i.e.\@\xspace}

\newcommand*{\cf}{cf.\@\xspace}

\newcommand{\RomanNumeralCaps}[1]
    {\MakeUppercase{\romannumeral #1}}
    
\newcommand{\mR}{\mathbb{R}}
\newcommand{\Pup}{\overline{P}}
\newcommand{\Plow}{\underline{P}}
\newcommand{\cvar}{\operatorname{CVar}_\alpha}
\newcommand{\cvarnoa}{\operatorname{CVar}}

\newcommand{\cR}{\mathcal{R}}
\newcommand{\cL}{\mathcal{L}}
\newcommand{\ronenorm}{\|\cdot\|_{\cR_1}}
\newcommand{\rtwonorm}{\|\cdot\|_{\cR_2}}
\newcommand{\ronenormy}{\|X\|_{\cR_1}}
\newcommand{\rtwonormy}{\|X\|_{\cR_2}}
\renewcommand{\d}{\,\operatorname{d}\!}

\DeclareMathOperator*{\argmin}{argmin}
\newcommand{\maxv}{\operatorname{MaxV}}
\newcommand{\du}{\operatorname{Du}}
\newcommand{\rim}{\operatorname{RIM}_{\alpha,\beta}}
\newcommand{\rimnoa}{\operatorname{RIM}}
\newcommand{\PTE}{\operatorname{PTE}}
\newcommand{\TErep}{\operatorname{TErep}}

\renewenvironment{proof}{\par\noindent{\bf Proof\ }}{\hfill\BlackBox\\[2mm]}

\usepackage{enumitem}

\providecommand{\varitem}{} 
\makeatletter
\newenvironment{axioms}[1]
 {\renewcommand\varitem[1]{\item[\textbf{#1\arabic{enumi}\rlap{$##1$}.}]%
    \edef\@currentlabel{#1\arabic{enumi}{$##1$}}}%
  \enumerate[label=\textbf{#1\arabic*.}, ref=#1\arabic*]}
 {\endenumerate}
\makeatother

\newcommand{\reals}{\mathbb{R}}
\newcommand{\argmax}{\operatornamewithlimits{arg\,{}max}}
\newcommand{\emb}{\hookrightarrow}
\newcommand{\embone}{\stackrel{1}{\emb}}
\newcommand{\Bfrak}{\mathfrak{B}}
\newcommand{\Fscr}{\mathscr{F}}
\newcommand{\Lscr}{\mathscr{L}}
\newcommand{\Pscr}{\mathscr{P}}
\newcommand{\Qscr}{\mathscr{Q}}
\newcommand{\Lcal}{\mathcal{L}}

\newcommand{\Xcal}{\mathcal{X}}
\newcommand{\Ycal}{\mathcal{Y}}

\usepackage{lastpage}
\jmlrheading{25}{2024}{1-\pageref{LastPage}}{6/22; Revised 1/24}{1/24}{22-0641}{Fröhlich and Williamson}
\ShortHeadings{Risk Measures and Upper Probabilities: Coherence and Stratification}{Fröhlich and Williamson}
\firstpageno{1}

\begin{document}
\title{Risk Measures and Upper Probabilities:\\ Coherence and Stratification}

\author{\name Christian Fröhlich \email christian.froehlich@uni-tuebingen.de \\
      \name Robert C. Williamson \email bob.williamson@uni-tuebingen.de \\
      \addr University of Tübingen\\
      and Tübingen AI Center\\
      Tübingen, Germany}
      
\editor{ }

\maketitle

\begin{abstract}
Machine learning typically presupposes classical probability theory which 
implies that aggregation is built upon expectation.  
There are now multiple reasons to motivate looking at richer 
alternatives to classical probability theory as a mathematical foundation for machine learning.
We systematically examine a powerful and rich class of alternative aggregation functionals, known 
variously as spectral risk measures, Choquet integrals or Lorentz norms. We present a range 
of characterization results, and demonstrate what makes this spectral family so special.
In doing so we 
arrive at a natural stratification of all coherent risk measures in terms of 
the upper probabilities that they induce 
by exploiting results from the theory of rearrangement invariant Banach spaces. 
We empirically demonstrate how this new approach to uncertainty 
helps tackling practical machine learning problems. 
\end{abstract}

\section{Introduction}
Machine learning (ML) typically presupposes classical probability theory. Recently, the assumption of a single stable probability distribution has been problematized, however.
Our motivation stems from the following ML problems:
in many cases, the empirical distribution of the data is not the ‘true’ one, so that some degree of distrust is
warranted. Under \textit{data set shift}, for instance, the learning method fails to generalize due to different distribution 
of the test data in the wild as compared to the well-controlled training environment. 
Furthermore, as machine learning is being increasingly deployed in sensitive domains (\eg medical problems, 
robot control), where failure can be catastrophic, demand for \textit{risk-averse} learning 
methods has arisen. 
This problem is often framed as aiming for \textit{distributional robustness} 
\citep{rahimian2019distributionally}, where the goal is to perform well with respect to perturbations of 
the reference distribution (the empirical distribution). A prima facie different problem is 
that of fair machine learning. In this setting, a machine learning system has direct bearing on 
ethically relevant individuals and we may hence ask for a system that does not discriminate 
between specified subgroups, consisting of ethically fungible individuals (\eg based on race, 
gender). This ethical ML problem of fairly distributing loss values is analogous to the 
‘technical’ ones discussed before and can be described in the same mathematical formalism.\footnote{This is one of multiple ways of mathematizing fairness, and requires formulating the loss function in a way that expresses the fairness-relevant aspects. See \citep{williamson2019fairness}.} 

The commonality that we identify among these problems is that they require rethinking the 
presumption that probability 
is merely about risk, but instead to realize a distinction between risk and (Knightian) uncertainty 
or ambiguity. Our method of inquiry is to take inspiration from other fields, 
where similar problems have received much more attention and treatment already. 
We find that there are numerous convergent strains of research scattered across the 
literature, with only a subset of their intricate interconnections laid out clearly 
so far. In particular, we consider ideas from imprecise probability, rational 
and social choice theory, finance, insurance, distributive justice and the theory of 
rearrangement invariant Banach spaces. Our main workhorse is the equivalence between 
coherent risk measures \citep{artzner1999coherent} from finance and coherent 
upper previsions \citep{walley1991statistical}, an influential approach to imprecise 
probability. These functionals can replace the expectation operator in 
expected risk minimization of a machine learning problem. 

The full generality of coherent risk measures is attractive, but we find that 
zooming in on a particular subclass, the \textit{spectral risk measures}, 
provides benefits such as clear interpretability. This subclass is particularly relevant, 
occupies a central place in the theory of coherent risk measures, and has been rediscovered 
numerous times by different authors with different motivations
\citep{yaari1987dual,schmeidler1989subjective,wang2000,acerbi2002spectral,quiggin2012generalized,buchak2013risk}. We explicate this relevance from various angles.
Essentially, spectral risk measures offer a systematic way to interpolate in the risk aversion spectrum.

We place the class of coherent risk measures in the broader framework of 
rearrangement invariant Banach function spaces \citep{Bennett:1988aa} 
and find that there, as well, the spectral risk measures occupy a prime position. 
This new connection also enables us to rederive the well-known Kusuoka representation theorem  
from a different angle, which states that any coherent risk measure has a 
representation in terms of spectral risk measures, thereby further underlining 
their centrality. Moreover, we leverage the theory to derive various 
characterization results of coherent risk measures in terms of their 
\textit{fundamental function}. Such a function specifies the underlying 
imprecise probability associated with a risk measure. To each such function,
we characterize the most optimistic and pessimistic extension from the imprecise 
probability to a risk measure. We explicate that the most significant distinctions 
of risk measures stem from their behaviour for tail events (an idea on which we have elaborated in a subsequent paper, see \citep{frohlich2023tailoring}).

Finally, we apply coherent and spectral risk measures to practical machine 
learning problems, and  find they lead to more robust and risk-averse 
solutions. We begin by outlining the risk and uncertainty distinction, 
which is the conceptual motivation for the following mathematical development.

\subsection{Risk and Uncertainty}
\label{sec:riskuncertainty}
\begin{quote}
    By “uncertain” knowledge, let me explain, I do not mean merely to distinguish
    what is known for certain from what is only probable. The game of roulette is
    not subject, in this sense, to uncertainty; nor is the prospect of a Victory bond
    being drawn. Or, again, the expectation of life is only slightly uncertain. Even
    the weather is only moderately uncertain. The sense in which I am using the
    term is that in which the prospect of a European war is uncertain, or the price
    of copper and the rate of interest twenty years hence, or the obsolescence of
    a new invention, or the position of private wealth-owners in the social system
    in 1970. About these matters there is no scientific basis on which to form any
    calculable probability whatever. We simply do not know.\\
    \hspace*{\fill} --- John Maynard \citet{keynes1937general}
\end{quote}
A distinction between risk and uncertainty has been around since Frank Knight's
 seminal 
work “Risk, Uncertainty and Profit” \citep{knight1921risk}, with precursors going back even to Adam Smith \citeyearpar{smith}. 
While it has received considerable attention in the economics literature, it has not yet been 
firmly established in the machine learning community. `Risk' refers to the benign situation, 
in which probabilities can be meaningfully associated to outcomes and full knowledge 
of the distribution is accessible; contrariwise, `uncertainty' refers to outcomes to 
which probabilities cannot be assigned. The meaning of ‘cannot’ here is subtle 
and warrants further discussion. Classical probability theory, based on 
Kolmogorov’s widely accepted axioms \cite{kolmogorov1950foundations}, is well-equipped 
to deal with the  former, but is arguably not an appropriate model for the latter. 
Along similar lines, Phil \citet{dawid2017individual} wrote 
\begin{quote}
    If you studied any Probability at school, it will have focused on the 
    behaviour of unbiased coins, well-shuffled packs of cards, perfectly 
    balanced roulette wheels, \textit{etc.}, \textit{etc.} --- in short, an excellent training for 
    a life misspent in the Casino. This is the ambit of \textit{Classical Probability}[.] [emphasis in original].
\end{quote} 
From a frequentist perspective, the crucial (problematic) assumption is that 
stochastic phenomena display stable relative frequencies in the limit. 
While often seemingly correct, this does not occur universally 
\citep{gorban2017statistical}. Frequentist probability is often given 
a metaphysical interpretation, by imagining an experiment which could be 
repeated infinitely many times to obtain independent outcomes 
\citep{dawid2017individual}. This is unlike the practical setting, 
where a ML system is deployed in a dynamically unfolding environment.
The failure to comply with a single stable probability distribution is then 
typically theorized using the notion of \textit{data set shift}
\citep{quinonero2008dataset}.

On the other hand, Bayesians assert that it is possible to supply a 
precise probability for any event or sequence of events. Such a precise 
credence (degree of belief) is then interpreted as your personal fair 
betting rate on an event \citep{de2017theory}. However, it is unclear 
whether you should have a precise betting rate on the event that right 
now \textit{24 men in Bulgaria are standing on their heads} 
\citep{schoenfield2012chilling}, as there is no evidence on which you 
could reasonably base your precise belief. Giving up on the insistence 
that you have a single betting rate, and instead positing that you 
have lower and upper betting rates, depending on whether you are 
required to bet for or against the event, yields imprecise probability 
\citep{walley1991statistical}, which we discuss in the next section.

A now classical challenge to probability theory is due to Daniel 
\citet{ellsberg1961risk}. Consider two urns, containing red and black balls.
In urn \RomanNumeralCaps{1}, there are 50 red and 50 black balls. 
Urn \RomanNumeralCaps{2} contains an unknown proportion of red and black balls,
adding up to 100 balls in total. On these four events (\RomanNumeralCaps{1}R, \RomanNumeralCaps{1}B, \RomanNumeralCaps{2}R, \RomanNumeralCaps{2}B), 
the subject may place a bet, which delivers \$100 if the ball is of the 
specified color and \$0 otherwise. Most subjects display the preference 
$\mbox{\RomanNumeralCaps{1}R}\sim \mbox{\RomanNumeralCaps{1}B} \succ \mbox{\RomanNumeralCaps{2}R}\sim \mbox{\RomanNumeralCaps{2}B}$, where $\sim$ denotes indifference and 
$\succ$ preference, so they prefer to bet on the first urn. 
We might call urn \RomanNumeralCaps{1} the risk urn, as probabilities 
can be precisely assigned as proportions of outcomes\footnote{%
    This holds when adopting the ``principal principle'' of 
    \cite{lewis1980subjectivist}, which asserts that knowledge of 
    chances requires that these be taken as subjective probabilities.}. 
Urn \RomanNumeralCaps{2} is an ambiguous urn, as the subject must 
entertain a whole set of possible urn compositions. The typical 
preference cannot be reconciled with probability theory and hence 
expected utility theory (in economic terms). If the subject is indifferent 
between a bet on red or on black for the ambiguous urn, it means in 
effect that she assigns the probability 0.5 to each color; but then she 
cannot strictly prefer betting on the first urn, where the probability is 
also 0.5. \citet{ellsberg1961risk} calls decision makers, 
which exhibit this paradoxical pattern, \textit{ambiguity-averse}. 
Here, ambiguity is to be understood as in-between risk and total 
Knightian uncertainty. After all, the subject supposes that the urn 
will exhibit stable relative frequencies, as opposed to an unstable real-world
process (\eg a machine learning system in a changing environment). 

Ellsberg’s urn paradoxes can be taken as purely descriptive, but are  
often interpreted and defended from a normative perspective: it is 
rationally permissible for subjects to be ambiguity-averse 
\citep{stefansson2019risk}. An education in classical probability may lead 
individuals to revise their initial preferences, after the inconsistency
has been pointed out, in order to conform with probabilistic reasoning. 
The challenge for such a response, however, is to give a non-circular 
justification for classical probability in the first place as the 
only permissible rational decision theory. An appeal to probability 
itself in such an argument is pointless. Contrariwise, we take Ellsberg’s
urns to be a serious challenge with normative appeal. Besides this 
thought experiment, a wealth of other challenges have been raised against 
classical probability theory \citep{allais1953comportement,walley1991statistical,
joyce2005probabilities,gilboa2009always,sep-imprecise-probabilities,isaacs2021non}. 
We do not attempt to summarize the vast literature on this topic.

While the above example may at first sight seem irrelevant to the 
practical concerns of a machine learning engineer, the challenge 
of ambiguity has in fact been recognized in the framework of 
distributionally robust ML \citep{rahimian2019distributionally}. 
The typical expected risk minimization problem
\begin{equation}
    \argmin_f \mathbb{E}_{P} \,\ell(f(X),Y)
\end{equation}
is there replaced by a worst-case attitude with respect to an 
ambiguity set of probability measures
\begin{equation}
    \argmin_f \sup_{Q \in \mathcal{Q}} \,\mathbb{E}_{Q}\,\ell(f(X),Y),
\end{equation}
where the ambiguity set $\{Q : d(Q,P) < \epsilon\}$ typically 
contains all probability measures in a specified 
$\epsilon$-neighbourhood of the base measure with respect to some divergence 
measure $d$ (\eg an $f$-divergence). One rationale for employing a 
distributionally robust (DR) approach is to account for the issue 
that the empirical distribution $\hat{P}_n$ is not the `true' one 
for finite sample size $n$. Instead, a whole set of probability 
distributions is considered and the most pessimistic, ambiguity-averse 
attitude is adopted by taking the supremum over the expected risks. 
We call the essence of this situation \textit{hallucinated ambiguity}: 
while the decision maker, \ie the machine learning engineer, 
is faced with a decision problem under risk, she also has good 
reason to believe that $\hat{P}_n$ does not coincide exactly with 
the `true' distribution. Therefore, she decides to introduce 
artificial ambiguity into the problem. In contrast, in an 
Ellsberg-like decision problem ambiguity arises naturally.

Distributionally robust optimization has proven to be useful for 
a range of machine learning problems. For instance, it has 
been used to counteract the possibility of adversarial 
attacks \citep{sinha2017certifying}. 
Due to its breadth, the DR framework can also tackle the problem of 
data set shift \citep{zhang2021coping}, where the training and test distributions
differ and which potentially yields diminished generalization performance. This is especially relevant since 
data set shift has been recognized as one of the most pressing 
problems in AI safety \citep{amodei2016concrete}. For example, 
\cite{kirschner2020distributionally} proposed a distributionally 
robust Bayesian optimization to deal with this phenomenon.

The line between ambiguity and risk can be blurry. If full access 
to the underlying distribution is available, the decision problem is 
one under risk. However, decision makers may have different rationally 
permissible attitudes towards such stochastic risk 
\citep{buchak2013risk}. When using expected risk minimization, 
the decision maker takes a neutral stance towards risk and 
cares merely about the average. Another decision maker might emphasize 
downside risk, in financial terms. These are losses which 
exceed the expected loss. This raises the question of how to 
systematically encode such an attitude. As machine learning is 
being increasingly deployed in sensitive domains, demand for \textit{risk-averse} 
learning methods has arisen. In such domains, tail risks, \ie 
unlikely events with highly negative impact, pose a threat to 
the system or even lead to human death. In reinforcement learning, 
risk-averse methods have been put forward \eg by
\cite{singh2020improving,urpi2021risk,dabney2018implicit,
tamar2015policy,vijayan2021}. 
To this end, these authors have employed coherent risk measures,
which we study in this paper. This effectively amounts to a
transformation of \textit{risk} to \textit{hallucinated ambiguity},
as we will show. To a first approximation, the mathematical approach
of coherent risk measures to handle risk in fact coincides with
the mathematics to handle ambiguity with imprecise probabilities.

In this paper, we consider distributional robustness in the general 
frameworks of risk measures and imprecise probability. 
This conceptual unification provides novel justification and 
interpretation and can guide further developments.

In summary, we take there to be a broad epistemic spectrum, 
where certainty, risk, (hallucinated) ambiguity and Knightian 
uncertainty lie. In this order, the adequateness of classical probability 
theory is increasingly challenged. We will argue that spectral risk 
measures are a distinguished subclass of coherent risk measures 
because they enable an interpolation between the two ends in a sensible manner.

\subsection{Contributions}
We elaborate the connection of coherent risk measures and 
imprecise probability, which has so far received little attention.
Thereby, we clarify the relation of risk (aversion) and ambiguity 
and what bearing this has on machine learning.  
In particular, we focus on the subclass of spectral risk measures.
Our goal is to bring as many different characterizations of them
as possible together in one place. On the way, we discover 
multiple new connections between theories.

We first summarize the existing theories of imprecise probability (Section~\ref{sec:walley}) and coherent risk measures (Section~\ref{sec:riskmeasures}). 
We embed the theory of coherent risk measures in the broader 
mathematical framework of rearrangement invariant Banach 
function spaces, thereby establishing an insightful connection (Section~\ref{sec:banach}). 
This enables the direct import of mathematical results, which are 
not yet known in the theory of coherent risk measures. 
Also, we can easily rederive the celebrated Kusuoka representation 
theorem \citep{kusuoka2001law} in this setting and provide an 
intuitive interpretation of it.
From this perspective, we derive various novel characterization 
results of coherent risk measures, typically in terms of 
their fundamental function, which corresponds to an imprecise probability.
We present some new results regarding the combination of two
risk measures, showing relationships to  the theory of interpolation 
of operators (Section~\ref{sec:combining}). In particular, we illustrate that one cannot avoid the 
element of choice in risk measure by appealing to an ``objective'' 
combination rule, because  the set of legitimate combination rules 
ends up being essentially as rich as the set of risk measures.
We conduct experiments, which demonstrate that spectral risk measures 
can encode risk aversion and robustness.  For our experiments (Section~\ref{sec:experiments}), we 
suggest two ways to evaluate the tail risk of a loss distribution. 
Specifically, we propose a graphical evaluation based on the 
\textit{conditional value at risk} and we employ Lorenz curves 
from the study of economic inequality. Throughout the paper, 
we translate results from different fields to a loss-based 
formulation, which aids the unification. As a consequence, 
when checking references, results might appear 
different from our statement of them.

\section{Coherent Lower and Upper Previsions}
\label{sec:walley}
The umbrella term \textit{imprecise probability} was popularized by \citet{walley1991statistical}, who offered a behavioural account of rational belief, which strictly generalizes probability theory. Walley takes inspiration from the work of the Bayesian  \citet{de2017theory}, who identified probability with personal fair betting rates. In contrast to de Finetti, however, Walley departs from the \textit{dogma of precision} and allows for a divergence of lower and upper betting rate. In this section we outline the basics of Walley's approach, with its main pillars of \textit{avoiding sure loss}, \textit{coherence} and \textit{natural extension}. While Walley's theory is formulated in terms of reward, we use a loss-based formulation throughout the paper, so that different theories can be directly related without tedious translations.

\subsection{Gambles and Previsions}
\label{sec:gamblesandprevisions}
Consider a possibility space $\Omega$, where $\omega \in \Omega$ represents a state of the world, including all information deemed relevant to the problem at hand. A \textit{gamble} is a bounded function $X: \Omega \rightarrow \mR$, yielding an uncertain loss $X(\omega)$ when the state $\omega$ is realized. In the ML context, such a gamble corresponds to a bounded loss function; here, we will allow negative loss values, too, which are then interpreted as reward. Gambles carry the obvious vector space structure with scalar multiplication $(\lambda X)(\omega) = \lambda X(\omega)$, $\lambda \in \mR$, and addition $(X+Y)(\omega) = X(\omega) + Y(\omega)$. Constant gambles $\alpha(\omega)=\alpha \text{ } \forall \omega$ are set in lowercase. We assume that a vector space $\mathcal{L}$ of gambles is given.

With simple axioms, we can characterize the set $\mathcal{D}$ of gambles which are desirable to the decision maker (i.e. the ML engineer).
A critical assumption is that loss lives on a bipolar linear measurement scale, where $0$ separates loss ($>0$, bad) from reward $(<0$, good). Then we can postulate the following structure:
\begin{axioms}{D}
  \item \label{item:D0} $\sup X < 0 \Rightarrow X \in \mathcal{D}$ 
  \item \label{item:D1} $\inf X > 0 \Rightarrow X \notin \mathcal{D}$ 
  \item \label{item:D2} $X \in \mathcal{D}, \lambda \in \mathbb{R}^+ \Rightarrow \lambda X \in \mathcal{D} $
  \item \label{item:D4} $X \in \mathcal{D}, Y \in \mathcal{D} \Rightarrow X+Y \in \mathcal{D}$
\end{axioms}
We may take these as axioms, but they are explained through the choice of a linear utility scale. As to \ref{item:D0}, certainly a gamble which yields only rewards is desirable. Conversely, a gamble which yields only loss is not desirable (\ref{item:D1}). Axioms \ref{item:D2} and \ref{item:D4} imply that the set $\mathcal{D}$ forms a convex cone, which due to \ref{item:D0} includes the interior of the negative orthant $\mathcal{L}^-$, and due to \ref{item:D1} excludes the interior of the positive orthant $\mathcal{L}^+$. We call a set $\mathcal{D}$ satisfying \ref{item:D0}--\ref{item:D4} a \textit{coherent set of desirable gambles}.

As such, this framework does not yet provide us with an evaluation of gambles which contain a mix of positive and negative outcomes. For this, we define a functional, called \textit{upper prevision} as follows:
\begin{equation}
\label{eq:upperprev}
    \overline{P}(X) \coloneqq \inf\{\alpha \in \mR: X - \alpha \in \mathcal{D} \}.
\end{equation}
We interpret $\Pup(X)$ as specifying the smallest amount of certain loss $\alpha$ that, when subtracted from the uncertain loss $X$, makes the resulting gamble desirable. In financial terms, this is the \textit{certainty equivalent} for $X$: the decision maker is willing to shoulder the risky position $X$ when offered the reward $-\alpha$ in exchange. Symetrically, we can define a \textit{lower prevision}:
\begin{align}
    \Plow(X) &\coloneqq -\Pup(-X)\\
    &= -\inf\{\alpha \in \mR: -X - \alpha \in \mathcal{D} \}\\
    &= \sup\{\alpha \in \mR: \alpha - X \in \mathcal{D} \},
\end{align}
which specifies the largest certain loss $\alpha$ we are willing to shoulder in exchange for giving away the uncertain $X$. In virtue of their conjugacy relation, we focus on the upper prevision in the following. When an upper prevision is defined from a coherent set of desirable gambles as in \eqref{eq:upperprev}, it can be shown to satisfy the properties \citep[p.\@\xspace 65]{walley1991statistical}:

\begin{enumerate}[label=\textbf{P\arabic*.}, ref=P\arabic*]
  \item \label{item:P1} $\Pup(X) \leq \sup(X)$ \quad (bounds)
  \item \label{item:P2} $\Pup(\lambda X) = \lambda \Pup(X) \text{, } \forall \lambda \in \mR^+$ \quad (positive homogeneity)
  \item \label{item:P3} $\Pup(X+Y) \leq \Pup(X) + \Pup(Y)$ \quad (subadditivity)
\end{enumerate}
We call a functional satisfying P1-P3 a \textit{coherent upper prevision}.
The corresponding coherent set of desirable gambles can be defined as $\mathcal{D} \coloneqq \{X : \Pup(X) \leq 0\}$, a definition which interacts well with \eqref{eq:upperprev}.
P2 and P3 together imply 
\begin{enumerate}[resume,label=\textbf{P\arabic*.}, ref=P\arabic*]
  \item \label{item:P4} $\Pup(\alpha X + (1-\alpha) Y) \leq \alpha \Pup(X) + (1-\alpha) \Pup(Y)\text{ } \forall \alpha \in [0,1]$ \quad (CX: convexity)
\end{enumerate}
but the converse is not true. In virtue of P2 and P3, $\Pup$ is a sublinear function and hence the support function of a closed convex set, a geometric fact which we will exploit later. Furthermore, P1-P3 imply \citep[p.\@\xspace 76]{walley1991statistical}
\begin{enumerate}[resume,label=\textbf{P\arabic*.}, ref=P\arabic*]
  \item \label{item:P5} $\Pup(c) = c \text{, } \forall c \in \mR$ \quad (agreement)
  \item \label{item:P6} $\Pup(X + c) = \Pup(X) + c \text{, } \forall c \in \mR$ \quad (translation equivariance)
  \item \label{item:P7} $X(\omega) \leq Y(\omega) \text{ } \forall \omega \in \Omega \Rightarrow \Pup(X) \leq \Pup(Y)$ \quad (monotonicity)
\end{enumerate}

An upper prevision generalizes the classical notion of the linear expectation $\mathbb{E}(X)$. In Walley's setting, a linear prevision is defined as a prevision which satisfies the self-conjugacy relation $\Pup(X) = -\Pup(-X)$. For a coherent prevision, it holds that $\Plow(X) \leq \Pup(X)$ and we may call the width of the interval $\left[\Plow(X),\Pup(X)\right]$ the \textit{degree of imprecision}. As a first simple example of a nonlinear upper prevision, consider the \textit{vacuous prevision} $\Pup(X) = \sup(X)$ and correspondingly $\Plow(X) = \inf(X)$. This prevision maximizes the degree of imprecision, while still being coherent. It is a model for complete ignorance, unlike a uniform distribution, which actually expresses precise beliefs \citep{konek2019epistemic}. On the other hand, the familiar expectation $\mathbb{E}$ is a precise, linear prevision. However, defining an expectation requires much more structure (a $\sigma$-algebra and a probability measure) than Walley imposes. 

To understand the structural implications of coherence, we first consider a strictly weaker rationality condition: \textit{avoiding sure loss}. In the Bayesian tradition, a typical justification for probability theory is based on \textit{Dutch book} arguments. A Dutch book is a collection of gambles, each of which is desirable to the decision maker, but the combination of which surely incurs a loss for the decision maker, no matter the outcome $\omega$. If it is not possible to find such a finite combination, we say that the prevision avoids sure loss. 
\begin{definition}
    A functional $\Pup$ defined on $\mathcal{L}$ avoids sure loss if
    \begin{equation}
\label{eq:asl}
    \forall n \in \mathbb{N} : \forall X_1,..,X_n \in \mathcal{L}: \sup_{\omega \in \Omega}\left[\sum_{j=1}^n \Pup(X_j) - X_j(\omega) \right] \geq 0.
\end{equation}
\end{definition}
Consider what happens if \eqref{eq:asl} fails. Then $\forall \omega \in \Omega:$ $\sum_{j=1}^n \Pup(X_j) < \sum_{j=1}^n X_j(\omega)$. This means that our risk assessments $\Pup(X_j)$ were too small, whatever the outcome. 
In the next section, we observe that the concept of avoiding sure loss has an approximate correspondence in the theory of risk measures as \textit{aversity}. A coherent upper prevision always avoids sure loss and is hence immune to Dutch books, but the converse is not generally true. In geometric terms, the above condition is equivalent to requiring that the set of desirable gambles excludes the interior of the positive orthant $\mathcal{L}^+$, where sure loss would occur. 

It can be shown \citep[p.\@\xspace 134] {walley1991statistical} that any upper prevision which avoids sure loss dominates at least one linear prevision pointwise, so the set $\mathcal{Q} = \{Q : Q(X) \leq \Pup(X) \text{ } \forall X \in \mathcal{L}, Q \text{ is linear prevision}\}$ is non-empty. We call such a set $\mathcal{Q}$ an \textit{envelope}.\footnote{In the literature on imprecise probabilities, the functional $\overline{E}$ in \eqref{eq:envelope} is called the envelope of $\mathcal{Q}$, whereas we use the term envelope for the set $\mathcal{Q}$ itself, in line with works such as \citep{rockafellar2015measures}.} Then we can construct a canonical coherent extension of $\Pup$ by forming the supremum over this set
\begin{equation}
\label{eq:envelope}
    \overline{E}(X) = \sup_{Q \in \mathcal{Q}} \, Q(X).
\end{equation}
This process is the \textit{natural extension} of $\Pup$ and yields a coherent upper prevision if and only if $\Pup$ avoids sure loss. If $\Pup$ was already coherent, then $\overline{E} = \Pup$. On the other hand, if $\Pup$ merely avoided sure loss, then the natural extension is the least committal extension from a behavioural perspective. This means that for any other coherent upper prevision $\overline{P'}$ which is dominated by $\Pup$, meaning $\overline{P'}(X) \leq \Pup(X) \text{ } \forall X \in \cL$, the natural extension lies in-between: $\overline{P'}(X) \leq \overline{E}(X) \leq \Pup(X) \text{ } \forall X \in \cL$. In this sense, the natural extension is the most pessimistic one, as it reduces the risk assessment by $\Pup$ just as little as necessary to achieve coherence.

Conversely, every coherent upper prevision can be written in the form of \eqref{eq:envelope} for some set $\mathcal{Q}$. Also, any representation of this form is automatically coherent. For the lower prevision, the infimum is taken over the same set. This provides a direct link to the ambiguity sets in DR optimization: any ambiguity set of linear previsions yields a coherent upper prevision. Note that until now, measure theory has not entered the picture, as Walley's theory is more general. Later we will identify linear previsions with $\mathbb{E}_{\mu_Q}[X]$ for some probability measure $\mu_Q$. 

In the imprecise probability literature, the envelope $\mathcal{Q}$ is called a \textit{credal set}. Figuratively, each linear prevision in the set corresponds to a member of a `credal committee' \citep{joyce2010defense}. Whereas each member holds a precise belief (credence) on the risk of $X$, their joint decision is based on a worst-case consideration and hence introduces imprecision. The question of desirability of a gamble consists in a unanimous vote of all credal members. By construction, Walley's theory thus encodes a maximally pessimistic attitude with respect to some envelope.

\subsection{Lower and Upper Probabilities}
So far we have focused on upper previsions, \ie nonlinear expectations, instead of probability. To obtain an imprecise probability on events, the prevision is applied on indicator gambles
\begin{equation}
    A \subseteq \Omega: \chi_A(\omega) \coloneqq 
    \begin{cases}
      1 &  \omega \in A\\
      0 &  \text{otherwise}
    \end{cases}    
\end{equation}
so that $\Pup(A)\coloneqq \Pup(\chi_A)$ is an \textit{upper probability} and $\Plow(A)\coloneqq 1-\Pup(\chi_{A^C})$ a \textit{lower probability}, where $A^C$ is the complement of $A$, i.e. $\Omega \setminus A$. These probabilities can be interpreted as a personal upper and lower betting rate, respectively, on the event that $A$ occurs. To verify coherence, the same criteria as for previsions may be used, but where the gambles are restricted to be indicator gambles. In the following, we assume $\Pup$ to be defined on a field\footnote{A field $(\Omega,\mathcal{F}$) consists of a set $\Omega$ and a family of subsets $\mathcal{F}$, which is closed under complements, finite unions and finite intersections. This is weaker than the definition of a $\sigma$-algebra, where closure under countable unions and intersections is assumed.} of events. Some consequences of coherence are then \citep[p.\@\xspace 84]{walley1991statistical}:
\begin{enumerate}[label=\textbf{P\alph*)}, ref=P\alph*]
  \item $0 \leq \Plow(A) \leq \Pup(A) \leq 1$
  \item $\Plow(\emptyset)=\Pup(\emptyset)=0;\quad \Plow(\Omega)=\Pup(\Omega)=1$
  \item \label{item:pmon}$A \subseteq B \Rightarrow \left(\Plow(A) \leq \Plow(B) \text{ and } \Pup(A) \leq \Pup(B)\right)$.

\end{enumerate}
Like for previsions, the width of the interval $[\Plow(A),\Pup(A)]$ is a natural measure for the degree of imprecision. An interesting interpretation for this comes from a comparison to modal logic \citep{augustin2014introduction}, where the possibility operator $\Diamond$ and the necessity operator $\Box$ stand in a similar conjugacy relation $\Diamond p = \neg \Box \neg p$ and likewise $\Box p = \neg \Diamond \neg p$. When the event $A$ is seen as a proposition, which represents incurring a unit loss, the lower probability quantifies the evidence that is certainly in favor of $A$ and likewise, the upper probability captures the evidence possibly in favor of $A$. Just as probability theory can be seen as an extension of propositional logic, imprecise probability theory extends modal logic. Similarly, a lower prevision gives the most optimistic (certain) assessment of the risk, whereas the upper prevision gives a more pessimistic (possible) assessment. 

In classical probability theory, there is a one-to-one correspondence between probability measures and the expectations they induce via Lebesgue integration. However, coherent upper probabilities in general do not uniquely determine a coherent upper prevision, which is why Walley focuses on previsions. The subclass of spectral risk measures we are particularly interested in, however, is based on upper probabilities in a one-to-one correspondence, which are then naturally extended to an upper prevision.

Similar to previsions, upper probabilities are characterized by the set of additive probabilities which they dominate. Additive probabilities are those which satisfy Kolmogorov's axioms, but with $\sigma$-additivity weakened to finite additivity:
\begin{enumerate}[label=\textbf{K\arabic*)}, ref=K\arabic*]
  \item $P(A) \geq 0$
  \item $P(\Omega)=1$
  \item $P(A\cup B) = P(A) + P(B)$, if $A\cap B = \emptyset$.
\end{enumerate}
An upper probability avoids sure loss if and only if it dominates an additive probability. It is furthermore coherent if and only if it is the envelope (cf.\@\xspace \eqref{eq:envelope}) of a set of additive probabilities. Hence, to extend an upper probability to an upper prevision, we may extend the additive probabilities in the envelope to linear previsions. This process, which may be complicated in general, is simplified for \textit{submodular} upper probabilities, which induce the class of spectral risk measures (Section~\ref{sec:distortion})\footnote{Technically, this is true if and only if the submodular probabilities are given as the composition of a concave function and a $\sigma$-additive probability.}. Due to their computationally convenient properties, submodular upper probabilities have received much attention in the imprecise probability literature \cite[see \eg][]{montes20182}.
They are also called \textit{2-alternating} \citep{miranda2003extreme} and the corresponding lower probabilities are \textit{2-monotone}. 
In the next section, we discuss coherent risk measures and relate the subclass of spectral risk measures to their corresponding submodular upper probabilities.

\section{Coherent Risk Measures}
\label{sec:riskmeasures}
The study of risk measures in financial mathematics aims to establish a systematic approach to the quantification of risk inherent in a portfolio. Such a portfolio, a collection of assets, will yield an uncertain future monetary loss or gain $X(\omega)$ when the state $\omega \in \Omega$ is realized. In this setting, risk is inherently asymmetrical: financial institutions are much more concerned with their downside risk, that is, returns below the expected value. Unexpectedly high gain is not a similar matter of concern. It is customary to view $X$ as a real-valued random variable, that is, a measurable function, on some underlying probability space $(\Omega,\mathcal{F},P)$. From the viewpoint of a regulating agency, the risk of this uncertain return must be quantified in order to arrive at a sensible capital requirement to prevent insolvency. Shouldering excessive risk without an appropriate capital requirement puts customers and the economy at risk. The failure to quantify risk properly has indeed been linked to the financial crisis, as discussed in 
\textit{The Turner Review}
\citep{turner}. Hence there has been increasing interest in risk measures which satisfy certain desiderata. 

\cite{artzner1999coherent} initiated the study of \textit{coherent risk measures}. They imposed axioms on \textit{acceptance sets}, which contain acceptable positions --- in Walley's terms, desirable gambles. The structure they imposed led to the corresponding risk functional having the following properties\footnote{For consistency, we work again with losses corresponding to positive real values, whereas it is common to work with monetary gains in the literature. In insurance, however, working with losses is common as well.}:
\begin{enumerate}[label=\textbf{C\arabic*.}, ref=C\arabic*]
  \item  $R(\lambda X) = \lambda R(X) \text{, } \forall \lambda \in \mR^+$ \quad (positive homogeneity)
  \item  $R(X+Y) \leq R(X) + R(Y)$ \quad (subadditivity)
  \item $R(X + c) = R(X) + c \text{, } \forall c \in \mR$ \quad (translation equivariance)
  \item $X(\omega) \leq Y(\omega) \text{ } \forall \omega \Rightarrow R(X) \leq R(Y)$ \quad (monotonicity)
\end{enumerate}
\cite{artzner1999coherent} justified these axioms from a financial perspective. Subadditivity is particularly interesting and much hinges on it. The rationale is that diversification should not be penalized. Intuitively, $X$ and $Y$ could act as a hedge against each other, thereby decreasing total risk. We will later discuss how subadditivity is related to ambiguity aversion. Translation equivariance is motivated as \textit{cash invariance}: adding a certain loss to a financial position $X$ should increase risk by exactly the same amount.

It has been observed \citep{pelessoni2003imprecise} that a risk measure corresponds to an upper prevision when random variables are bounded. This can be directly seen from the axioms of acceptance sets, which are equivalent to those for coherent sets of desirable gambles, but it is also instructive to relate the functional properties.

\begin{theorem} \label{theorem:coherencecoincidence}\citep{pelessoni2003imprecise}. Let $\mathcal{L}$ be a linear space of bounded real-valued random variables, containing all constants $c \in \mR$.
A functional $R$ is a coherent risk measure on $\mathcal{L}$ if and only if it is a coherent upper prevision on $\mathcal{L}$.
\end{theorem}
\begin{proof}
Let $R$ a coherent risk measure. We need to show only $R(X) \leq \sup(X)$. Since $X \leq \sup(X)$, we have by monotonicity and translation equivariance that $R(X) \leq R(\sup(X)) = R(0+\sup(X)) = R(0) + \sup(X) = \sup(X)$. Hence $R$ is a coherent upper prevision. For the converse direction, we refer to \citep[p.\@\xspace 76]{walley1991statistical} and \citep{pelessoni2003imprecise}. 
\end{proof}
As of now, the equivalence (barring the technicality of boundedness) of coherent risk measures and coherent upper previsions is a formal, mathematical observation. We assert, however, that it has profound philosophical consequences, which can be understood with regard to the risk and uncertainty spectrum, discussed in Section~\ref{sec:equivalence_consequences}.

\subsection{Boundedness and Law Invariance}
Recall that Walley's approach to imprecise probability does not require an underlying probability space, \ie a measure space, but instead presupposes boundedness of the gambles. From this, he derives, using the Hahn-Banach theorem, that any coherent upper prevision admits a representation of the form
\begin{equation}
    \Pup(X) = \sup_{Q \in \mathcal{Q}}\, Q(X),
\end{equation}
where $\mathcal{Q}$ is a set of linear previsions. Yet these linear previsions are merely finitely additive, instead of countably additive. Moreover, boundedness is inconvenient for theory.

On the other hand, coherent risk measures are typically introduced on an underlying probability space. A common choice for the space of random variables is then $\mathcal{L}^2(\Omega,\mathcal{F},P)$, which are those random variables with finite second moment. However, we will see in Section~\ref{sec:banach} that there is in fact a more natural space to work with. Then any coherent risk measure admits a representation of the form (Section~\ref{sec:envelopes})
\begin{equation}
    R(X) = \sup_{\mu_Q : Q \in \mathcal{Q}} \mathbb{E}_{\mu_Q}[X],
\end{equation}
where the $\mu_Q$ are \textit{countably additive} probability measures. To ensure that these are indeed valid probability measures, translation equivariance and monotonicity are key (Section~\ref{sec:envelopes}). Note that now the definition of monotonicity is adapted to the measure\footnote{\cite{artzner1999coherent} considered only finite $\Omega$, so they could define monotonicity as holding for all $\omega \in \Omega$.}:
\begin{equation}
    X \leq Y \thinspace\thinspace P\text{-a.s.} \Rightarrow R(X) \leq R(Y),
\end{equation}
where $P\text{-a.s}$ means almost surely ($P$-almost everywhere).
Mathematically, it is more convenient to work with countably additive probability measures and unbounded random variables. On the other hand, with the additional assumption of a single distinguished base measure this setup is less parsimonious than Walley's framework. We believe, however, that little generality is lost when doing so. Henceforth we will work with an underlying probability space in line with the risk measurement and machine learning community. The strength and usefulness of Walley's theory lies in the additional conceptual interpretations that it provides. For instance, that a coherent risk measure essentially relies on an underlying imprecise probability has not been appreciated widely. 

A much more restrictive, yet useful assumption is \textit{law invariance}. An upper prevision (coherent risk measure), defined on a probability space $(\Omega,\mathcal{F},P)$ is called law invariant if $\Pup(X)=\Pup(X')$ whenever $X$ and $X'$ share the same distribution with respect to $P$. Conceptually, this introduces reliance on a distinguished precise probability. For example, the expectation $\mathbb{E}$ is a law invariant coherent upper prevision. Law invariance encodes the idea that the fine structure of $\Omega$ does not actually matter: a decision maker cares only about the distribution of risk, not in which specific states $\omega$ it occurs. The property of law invariance, which is not even expressible in Walley's general framework, will be especially useful to us to characterize classes of coherent risk measures in Section~\ref{sec:banach}.

\subsection{Envelope Representations}
\label{sec:envelopes}
In virtue of positive homogeneity and subadditivity, a risk measure is a \textit{sublinear} functional and hence, assuming closedness\footnote{A function $R$ is closed if all sublevel sets $\{x \in \text{dom}(R): R(x) \leq c\}$, $c \in \mR$, are closed sets, \ie contain all limit points.} for technical reasons, the support function of a closed convex set. This geometric viewpoint provides direct insights into the structure of risk measures. We here work with the space $\mathcal{L}^p \coloneqq \mathcal{L}^p(\Omega,\mathcal{F},P)$ of random variables with finite $p$-th moment, $p \in [1,\infty]$. It is paired with the space $\cL^q$, $\frac{1}{p} + \frac{1}{q} = 1$, and the pairing is
\begin{equation}
    \langle X,Y \rangle = \int_{\Omega} X(\omega) Y(\omega) \d P(\omega), \quad \forall X \in \cL^p, Y \in \cL^q.
\end{equation}
In the case of $1 \leq p < \infty$, $\cL^q$ coincides with the dual space of $\cL^p$. The case of $p=\infty$ is complicated  \cite[see][]{schonherr2017pure}, but common practice is to pair it with $\cL^1$.

A standard result in convex analysis is that a canonical bijection between support functions $R$ and their supported sets $\mathcal{Q}$ is then given by
\begin{equation}
    R(X) = \sup_{Q \in \mathcal{Q}}\, \langle X, Q \rangle , \quad \mathcal{Q} = \left\{Q \in \cL^q: \langle X, Q \rangle \leq R(X) \text{ } \forall X \in \mathcal{L}^p\right\}.
\end{equation}
The following correspondences are known in the literature \citep{rockafellar2013fundamental,shapiro2013kusuoka,follmer2016stochastic,yuxi}:
\begin{enumerate}[label=\textbf{E\arabic*.}, ref=E\arabic*]
\item $R$ is monotone iff $\mathcal{Q} \subseteq \mathcal{L}_+^q$, where $\mathcal{L}_+^q=\{Q \in \mathcal{L}^q : Q \geq 0 \thinspace \thinspace P\text{-a.s.}\}$
\item $R$ is translation equivariant iff $\mathcal{Q} \subseteq \mathcal{E}_1$, where $\mathcal{E}_1=\{Q \in \mathcal{L}^q: \mathbb{E}(Q)=1\}$
\item $R(X) \geq \mathbb{E}(X)$ $\forall X \in \cL^p$ iff $1 \in \mathcal{Q}$, where $1(\omega)=1$ $\forall \omega \in \Omega$
\item \label{E4} If $R$ is a law invariant coherent risk measure, its envelope $\mathcal{Q}$ is invariant under measure-preserving transformations.
\end{enumerate}
For the technicalities regarding~\ref{E4} see \cite{shapiro2013kusuoka}. Intuitively, \ref{E4} means that under law invariance, if $Q$ and $Q'$ have the same distribution under the measure $P$, then either both are in the envelope or none.

Thus the envelope of a coherent risk measure satisfies $\mathcal{Q} \subseteq \mathcal{L}_+^q \cap \mathcal{E}_{1}$. Each $Q \in \mathcal{Q}$ defines a measure as
\begin{equation}
    \mu_Q(A) \coloneqq \int_{A} Q(\omega) \d P(\omega) = \mathbb{E}_P[\chi_A Q] \quad \forall A \in \mathcal{F}.
\end{equation}
Due to $Q \in \mathcal{L}_+^q$, $\mu_Q(A) \geq 0$ and due to $Q \in \mathcal{E}_1$, we have $\mu_Q(\Omega)=1$. Hence $\mu_Q$ is a probability measure and we can equivalently write the risk measure as
\begin{equation}
    R(X) = \sup_{\mu_Q : Q \in \mathcal{Q}} \mathbb{E}_{\mu_Q} \left[X \right].
\end{equation}
If $X$ is bounded, then from this representation it is clear that $R(X) \leq \sup(X)$ and therefore $R$ is a coherent upper prevision. Also, this representation provides the rationale for viewing a risk measure as a worst-case ``vote'' with respect to a set of probabilities. This is equivalent to Walley's formulation, where the set consists of linear previsions, which we can identify here as $\mathbb{E}_{\mu_Q}[\cdot]$. Whereas the expectation with respect to the base measure is $\mathbb{E}_P[\cdot]$, represented by the singleton envelope $\{1\}$, each probability measure $\mu_Q$ defines a legitimate linear prevision. The envelope, in Walley's terms, consists of the linear previsions which are dominated by $R$. Natural extension entails finding those linear previsions and taking the supremum over them, hence automatically enforcing both monotonicity and translation equivariance of the functional. Essentially, this relies on the fact that a closed sublinear function equals the supremum of the linear functions minorizing it \citep{hiriart2004fundamentals}.

\subsection{The Fundamental Coherent Risk Quadrangle}
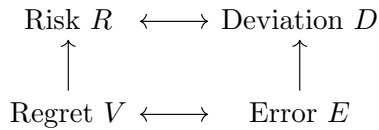
\begin{figure}
    \centering
    \setbox0\hbox{Regret $V$}
    \setbox2\hbox{Deviation $D$}
    \begin{tikzcd}
    \hbox to\wd0{\hfil Risk $R$\hfil} \arrow[r,leftrightarrow] & \copy2 \\
    \box0 \arrow[r,leftrightarrow] \arrow[u] & \hbox to\wd2{\hfil Error $E$\hfil} \arrow[u]
    \end{tikzcd}

    \caption{The fundamental risk quadrangle \citep{rockafellar2013fundamental}.}
    \label{fig:quadrangle}
\end{figure}

\cite{rockafellar2013fundamental} put the developments in the theory of risk measures in an even broader perspective by introducing the \textit{fundamental risk quadrangle}, depicted in Fig.~\ref{fig:quadrangle}. The authors made technical assumptions about certain limits, which were shown to be superfluous by \citet{rockafellar2015measures}, and which we therefore drop.

\citet{rockafellar2013fundamental} generally consider \textit{convex risk measures} \citep{follmer2016stochastic} on $\mathcal{L}^2(\Omega,\mathcal{F},P)$, where coherence is weakened by dropping subadditivity and positive homogeneity and only assuming convexity in its place. As a consequence, the acceptance set is then a convex set, but not in general a cone anymore. Since we are interested in coherence, we simplify their definitions and theorems to the coherent case. They further demand \textit{aversity}
\begin{equation}
     \forall c \in \mR: R(c)=c \text{ } \text{, but } R(X) > \mathbb{E}[X] \text{ for nonconstant } X, \thinspace \text{\ie} \thinspace\thinspace \thinspace P(\{X = c\})<1 \forall c \in \mR.
\end{equation}
For reference, we collect properties of averse coherent risk measures in the quadrangle:
\begin{enumerate}[label=\textbf{A\arabic*.}, ref=A\arabic*]
  \item  $R(\lambda X) = \lambda R(X) \text{, } \forall \lambda \in \mR^+$ \quad (positive homogeneity)
  \item  $R(X+Y) \leq R(X) + R(Y)$ \quad (subadditivity)
  \item $R(X + c) = R(X) + c \quad \forall c \in \mR$ \quad (translation equivariance)
  \item $X \leq Y \thinspace \thinspace P\text{-a.s. } \Rightarrow R(X) \leq R(Y)$ \quad (monotonicity)
  \item $R(c)=c \text{ } \forall c \in \mR \text{, but } R(X) > \mathbb{E}[X] \text{ for nonconstant } X$
  \item $R$ is closed, \ie it has closed sublevel sets $\{X \in \text{dom}(R) : R(X) \leq c\}$, $c \in \mR$. Here, $\text{dom}(R) \coloneqq \{X : R(X) < \infty\}$.
\end{enumerate}

In the top part of the quadrangle, there is a one-to-one correspondence between coherent risk measures and coherent deviation measures, given by the relation $R(X) = \mathbb{E}(X) + D(X)$. Such deviation measures are positively homogeneous, subadditive and closed and satisfy
\begin{equation}
    D(c)=0 \text{ } \forall c \in \mR \text{, but } D(X) > 0 \text{ for nonconstant } X
\end{equation}
\begin{equation}
    D(X) \leq \text{ess sup}(X) - \mathbb{E}[X], \quad \text{ess sup}(X) \coloneqq \inf\left\{\lambda \in \mR: P(X > \lambda) = 0\right\}
\end{equation}
\begin{equation}
    D(X+c)=D(X) \text { } \forall c \in \mR \quad \text{(translation invariance)}
\end{equation}
In practice, the variance is often employed to measure the deviation from the mean in a distribution. In the context of finance, this is the classical \textit{mean-variance analysis} \citep{markowitz}. However, the variance is not a coherent deviation measure, as it fails to be subadditive. Conceptually, the shortcoming is that the variance penalizes variability in both directions, but due to the loss/gain asymmetry, we like to emphasize the importance of losses exceeding the expectation.

In the bottom part of the quadrangle, there is a one-to-one correspondence between coherent regret measures $V$ and coherent error measures $E$, given by the relationship $V(X) = \mathbb{E}(X) + E(X)$.
By coherent regret measure, we mean a functional which is positively homogeneous, subadditive, monotonic, closed and averse in the sense that
\begin{equation}
\label{eq:regretaverse}
    V(0)=0 \text{, but } V(X) > \mathbb{E}[X] \text{ for nonzero }X\text{, i.e. } P(X = 0) < 1.
\end{equation}
According to \cite{rockafellar2013fundamental}, ``the role of a measure of regret, $V$, is to quantify the displeasure associated with the mixture of potential positive, zero and negative outcomes of a random variable $X$ that stands for an uncertain cost or loss.''. A coherent regret measure lacks only translation equivariance as compared to a coherent risk measure.

A coherent error measure quantifies the nonzeroness of $X$, is positively homogeneous, subadditive, closed and averse in the sense that
\begin{equation}
    E(0)=0 \text{, but } E(X)>0 \text{ for nonzero }X.
\end{equation}
Furthermore, we require
\begin{equation}
    E(X) \leq \mathbb{E}[-X] \text{ for } X \leq 0,
\end{equation}
which is equivalent to the monotonicity of the corresponding regret measure. A coherent error measure is hence fundamentally asymmetrical.
The bottom part of the quadrangle projects to the top part via the operations
\begin{equation}
    R(X) = \inf_{c \in \mR}\left\{V(X-c) + c\right\}, \quad D(X) = \inf_{c \in \mR}\left\{E(X-c)\right\}.
\end{equation}
For a coherent regret and error measure, respectively, the result will be a coherent risk and deviation measure. Note, however, that the backwards direction is not unique: one can find an infinity of regret/error measures which project to the same risk/deviation measure.

The projections can be understood as infimal convolution. 
Let $\sigma_\mathcal{Q}(X) \coloneqq \sup_{Q \in \mathcal{Q}}\,\langle X,Q \rangle$ be the support function of the set $\mathcal{Q} = \left\{Q \in \cL^q: \langle X, Q \rangle \leq \sigma_\mathcal{Q}(X) \text{ } \forall X \in \mathcal{L}^p\right\}$.
\begin{theorem}
\label{prop:infcon} \citep{sun2020risk}. 
Let $V=\sigma_\mathcal{Q}$ be a positively homogeneous, subadditive, monotonic and closed functional, \ie a coherent regret measure, and $\mathcal{Q}$ be its supported set. Suppose $\mathcal{Q}' \coloneqq \mathcal{Q} \cap \mathcal{E}_1 \neq \emptyset$. Then $R \coloneqq \sigma_{\mathcal{Q}'}$ is a coherent risk measure and $R(X) = \inf_{c \in \mR} V(X-c) + c$. 
\end{theorem}
This process can also be understood from Walley's perspective, where the projection from $V$ to $R$ is the natural extension (recall Section~\ref{sec:gamblesandprevisions}):

\begin{theorem}
A coherent regret measure $V$ avoids sure loss. Its natural extension coincides with  $R(X) = \inf_{c \in \mR} V(X-c) + c$.
\end{theorem}
\begin{proof}
Due to aversity of $V$
\begin{equation}
    \forall X_1,..,X_n \in \mathcal{L}^p: \sup_{\omega \in \Omega}\left[\sum_{j=1}^n V(X_j) - X_j(\omega) \right] \geq \sup_{\omega \in \Omega}\left[\sum_{j=1}^n \mathbb{E}[X_j] - X_j(\omega) \right] \geq  0,
\end{equation}
and therefore $V$ avoids sure loss. The converse implication (avoiding sure loss $\Rightarrow$ aversity) does not in general hold. Due to monotonicity and subadditivity and positive homogeneity we know that $V$ is the support function of some set $\mathcal{Q}$ and each $Q \in \mathcal{Q}$ is nonnegative almost everywhere. Computing the natural extension entails finding those linear previsions which are dominated by $V$ and forming the envelope of them, but these correspond to expectations $\mathbb{E}_{\mu_Q}[\cdot]$ induced by the set
\begin{equation}
    \{Q \in \mathcal{Q}: \mathbb{E}[Q]=1\} = \mathcal{Q} \cap \mathcal{E}_1,
\end{equation}
which is the supported set of $R(X) = \inf_{c \in \mR} V(X-c) + c$.
\end{proof}
The achievement of \cite{rockafellar2013fundamental} is to put risk in a broad conceptual framework and to establish a link between optimization ($R,V$) and estimation ($D,E$). Consider the archetypical regression problem, framed in terms of a coherent error measure:
\begin{equation}
    \text{minimize } E(Y - f(X_1,..,X_n)) \text{ over } f \in \mathcal{H}
\end{equation}
for random variables $X_1,..,X_n$, outcomes $Y$ and some hypothesis class $\mathcal{H}$. \cite{rockafellar2008risk} proved that under a mild technical assumption this problem can be equivalently phrased as
\begin{equation}
    \text{minimize } D(Y - f(X_1,..,X_n)) \text{ over } f \in \mathcal{H} \text{ s.t. } 0 \in \text{argmin}_{c \in \mR}\left\{E(Y - f(X_1,..,X_n) - c)\right\}.
\end{equation}
See also \citep{rockafellar2015measures}. This provides a new perspective on regression, where customized risk aversion is directly built in. 
In this paper, we mainly focus on coherent risk measures rather than  coherent error measures, since risk measures applied to a loss random variable are not constrained by a dependence on the $Y-f$ difference. However, our results in Section~\ref{sec:banach} are also linked to coherent regret (and thus error) measures.

\subsection{The Conditional Value at Risk}
We now examine a coherent quadrangle of particular interest, that of the \textit{conditional value at risk} $\cvar$, with parameter $\alpha \in [0,1)$. $\cvar$ is a special case of the larger class of spectral risk measures. In fact, we will see in Section~\ref{sec:rikusuoka} that the $\cvar$ are the basic building blocks not only of the spectral risk measures, but of \textit{all} law invariant coherent risk measures. Define the positive part of a random variable as $X^+ \coloneqq \max(X,0)$ and the negative part as $X^- \coloneqq \max(0,-X)$. For each $\alpha \in (0,1)$, a coherent quadrangle is given by:
\begin{align}
    R(X) &= \cvar(X), \quad\quad D(X)=\cvar(X-\mathbb{E}(X))\\
    V(X) &\coloneqq \frac{1}{1-\alpha}\mathbb{E}[X^+],\quad E(X)=\mathbb{E}\left[\frac{1}{1-\alpha}X^+ + X^-\right].
\end{align}
According to the projection from regret, $\cvar(X) = \min_c\{\frac{1}{1-\alpha}\mathbb{E}((X-c)^+) + c\}$.
We also define $\cvarnoa_{\alpha=0} \coloneqq \mathbb{E}$ in the same way, but this is only ``weakly'' averse in the degenerate sense that $\mathbb{E}\geq \mathbb{E}$.
The random variable $X$ has a right-continuous distribution function $F_X$ with generalized inverse\footnote{For consistency with Section~\ref{sec:banach}, we choose to work with the lower instead of the upper quantile.} 
$F_X^{-1}(q) = \sup\{\lambda \geq 0: F_X(\lambda) < q \}$. Then $\cvar$ can be equivalently expressed as an integral over quantiles
\begin{equation}
   \cvar(X) =  \frac{1}{1-\alpha} \int_\alpha^1 F_X^{-1}(q) \d q.
\end{equation}
If $F_X$ is continuous, this can be further written as
\begin{equation}
    \cvar(X) = \mathbb{E}\left[X | X \geq F_X^{-1}(\alpha)\right],
\end{equation}
and is also called \textit{expected shortfall}, \textit{tail conditional expectation} or \textit{superquantile} \citep{laguel2021superquantiles}. Intuitively, $\cvar$ takes the average of the $(1-\alpha)$-fraction of the worst outcomes and neglects the more fortunate outcomes completely. In one extreme, $\cvarnoa_{\alpha=0}$ corresponds to the expectation; in the other, $\cvarnoa_{\alpha\rightarrow 1} \coloneqq \lim_{\alpha \rightarrow 1} \cvar$ gives the essential supremum (worst-case) of $X$.

The envelope of $\cvar$ is known to be $\mathcal{Q} = \{Q : 0 \leq Q \leq \frac{1}{1-\alpha}, \mathbb{E}[Q]=1\}$, so coherence can be directly verified from the envelope. We can interpret the elements of the envelope as reweightings of the original distribution, where a reweighting of up to $1/(1-\alpha)$ is allowed. As a consequence, the supremum is achieved when that reweighting is fully concentrated on the $(1-\alpha)$-fraction of the largest losses. For $\alpha=0$, the supremum is clearly attained at $Q=1$, which corresponds to the expectation. On the other hand, for $\alpha \rightarrow 1$, the reweighting may be arbitrarily large and hence the worst-case will receive all of the weight (but the supremum will not be attained in general). 
To see that the above set is indeed the envelope of $\cvar$, consider the regret $V=\frac{1}{1-\alpha}\mathbb{E}[X^+]$. It is not hard to see that its envelope is the set $\{Q : 0 \leq Q \leq 1/(1-\alpha)\}$. The projection of $V$ to $\cvar$, \ie the natural extension, entails intersecting this set with the constraint $\mathbb{E}[Q]=1$.

\subsection{Spectral Risk Measures}
\label{sec:spectral}
$\cvar$ belongs to the family of spectral risk measures \citep{acerbi2002spectral}. We here work on $\cL^2(\Omega,\mathcal{F},P)$, but in Section~\ref{sec:banach} we show that the natural space to work on is in fact more subtle. Observe that a convex combination of coherent risk measures again yields a coherent risk measure. Given a probability measure $\lambda$ on $[0,1]$, this can be generalized to the form
\begin{equation}
\label{eq:spectralcvar}
    R^\lambda(X) \coloneqq \int_0^1 \cvar (X) \d \lambda(\alpha),
\end{equation}
which yields a coherent risk measure. We assume that the measure $\lambda$ does not have an atom at $1$, \ie $\lambda(\{1\})=0$. By expanding $\cvar$ as its integral representation and using Fubini-Tonelli, this can be rewritten as
\begin{equation}
\label{eq:spectralweight}
    R^\lambda(X) = R_{(w)}(X) \coloneqq \int_0^1 F_X^{-1}(q) w(q) \d q,
\end{equation}
with a spectral weighting function $w : [0,1] \rightarrow \mR^+$. This generates a coherent risk measure if and only if $w$ is nonnegative, monotonically increasing and $\int_0^1 w(q) \d q = 1$ \citep{acerbi2002spectral}. These properties are automatically satisfied when $w$ is induced by a probability measure $\lambda$ from \eqref{eq:spectralcvar}. The spectrum $w$ has a clear interpretation as a risk aversion profile. A monotonically increasing $w$ puts more weight on worse (highly positive) outcomes as a penalty. In the special case of $\cvar$, $\alpha \in [0,1)$, we have
\begin{equation}
    w(q) = 
    \begin{cases}
    0 & 0 < q < \alpha\\
    1/(1-\alpha) & q \geq \alpha
    \end{cases}
\end{equation}
hence all values below the $\alpha$-th quantile are ignored, and values above it receive the constant weight $1/(1-\alpha)\geq1$. Note that by demanding that $\lambda$ does not have an atom at $1$, we have excluded the supremum risk measure $\cvarnoa_{\alpha \rightarrow 1}$, which is represented by the Dirac measure at $1$. The corresponding weight function would be $0$ everywhere, rendering a representation of the form \eqref{eq:spectralweight} impossible. The supremum risk measure, while being in the ``closure'' of the family of spectral risk measures, cannot be considered a proper member due to its pathological properties. It will lead to additional technical complications in Section~\ref{sec:banach}.

For an arbitrary spectral risk measure $R_{(w)}$ with spectrum $w$, the envelope representation is \citep{pflug2006subdifferential}:
\begin{equation}
    R_{(w)}(X) = \sup_Q\left\{\langle X, Q \rangle : Q = w(U) \right\} \text{, where } U \text{ is uniformly distributed on } [0,1] \text{ wrt. }P,
\end{equation}
which requires that $\Omega$ is rich enough to support a uniform distribution. While this result may seem somewhat mysterious, we will obtain a different perspective on it in Section~\ref{sec:banach}, which also supplies intuition.

\subsection{Distortion Risk Measures}
\label{sec:distortion}
Assume again the space $\cL^2(\Omega,\mathcal{F},P)$. Equivalent to spectral risk measures are \textit{distortion risk measures} with concave distortions, which originate from distortion premium principles in actuarial science \citep{wang1997axiomatic,wang2000}. In insurance, the key challenge is to price a contingent claim. That is, from the viewpoint of the insurer, a random variable $X$ represents an uncertain loss that corresponds to a claim made by a policyholder. Given a probability model under which $X$ has distribution $F_X$, the question is how much should the insurer charge in exchange for shouldering the risk? This is called the insurance premium.
Consider an equivalent definition of the usual expectation:
\begin{align}
    \mathbb{E}[X] &= -\int_{-\infty}^{0} F_X(x) \d x + \int_0^\infty \left(1-F_X(x)\right) \d x\\
    &= \int_{-\infty}^{0} \left[S_X(x) - 1\right] \d x + \int_0^\infty S_X(x) \d x,\\
\end{align}
where we used the \textit{survival function} $S_X(x) \coloneqq 1-F_X(x) = P(X > x)$.
If the insurer simply charged the expectation as the premium, they could not make any profit and could face bankruptcy due to model misspecification. What if the specified probability does not accurately reflect the real risk? The idea of a distortion premium is to model risk aversion by instead calculating the expectation with respect to another distribution, given by the \textit{Choquet integral}:
\begin{align}
\label{eq:distortion}
    R_\phi(X) &\coloneqq \int_{-\infty}^{0} \left[\phi(S_X(x)) - 1\right]\d x + \int_0^\infty \phi\left(S_X(x)\right) \d x,
\end{align}
where $\phi: [0,1] \rightarrow [0,1]$ is a monotonically increasing concave function satisfying $\phi(0)=0$ and $\phi(1)=1$. In addition, we assume additionally that $\phi$ is continuous at $0$, which excludes the supremum risk measure but avoids technical issues. These boundary conditions ensure that $R(X)$ can be viewed as an expectation with respect to a valid (distorted) probability distribution. Concavity of $\phi$ models risk aversion in the sense that the higher the loss level, the higher the increase in the premium. Furthermore, the resulting functional $R$ is a coherent risk measure if and only if $\phi$ is concave \citep{gzyl2008relationship}. In the special case of $\phi(t)=t$, we obtain the expectation and for all other distortion risk measures we have $R_\phi(X) \geq \mathbb{E}(X)$. The difference $D(X)=R_\phi(X)-\mathbb{E}[X]$, Rockafellar and Uryasev's \citeyearpar{rockafellar2013fundamental} deviation measure, is also known as the \textit{risk premium}. For coherence the critical property is subadditivity of $R$, which corresponds to the concavity of the distortion. If instead $\phi$ is convex, the functional is superadditive\footnote{A functional $R$ is superadditive if $R(X+Y)\geq R(X) + R(Y)$.}. In Appendix~\ref{sec:choquet}, we consider functionals of the form \eqref{eq:distortion}, Choquet integrals, in more depth.
Here we observe that distortion risk measures are equivalent to spectral risk measures.
\begin{theorem}
\label{theorem:spectraldistortion} \citep{gzyl2008relationship,ridaoui2016choquet}. 
For any distortion risk measure $R_\phi$ with concave distortion $\phi$, with $\phi(0)=0$ and $\phi(1)=1$, there is an identical spectral risk measure $R_{(w)}=R_\phi$, with $\phi'(t)=w(1-t)$.
\end{theorem}
The proof is in Appendix ~\ref{app:spectraldistortion}. For example, in the case of $\cvar$ we have $\phi(t)=\int_0^t w(1-u) \d u = \min\{t/(1-\alpha),1\}$.

From a given base probability measure $P$ we obtain a distorted probability $\phi(P)$. We can interpret this as an upper probability in Walley's framework. Setting $\overline{\mu}(A) \coloneqq \phi(P(A)) \text{ } \forall A \in \mathcal{F}$ defines a \textit{capacity} on events. A capacity on $(\Omega,\mathcal{F})$ is a set function $\overline{\mu}: \mathcal{F} \rightarrow \mR$ with the normalization $\overline{\mu}(\emptyset)=0$, $\overline{\mu}(\Omega)=1$ and the monotonicity property $A \subseteq B \Rightarrow \overline{\mu}(A) \leq \overline{\mu}(B)$. In our case this is satisfied because $\phi(0)=0$ and $\phi(1)=1$ and $\phi$ is monotonically increasing.
A \textit{submodular capacity}, sometimes called concave capacity, is a capacity which satisfies the inequality
\begin{equation}
    \overline{\mu}(A\cup B) +\overline{\mu}(A\cap B) \leq \overline{\mu}(A)+\overline{\mu}(B) \quad \forall A,B \in \mathcal{F}.
\end{equation}

If and only if the function $\phi$ is concave and $\phi(0)=0$, $\overline{\mu}$ is a submodular capacity \citep[Prop.\@\xspace 4.7.5]{bednarski1981solutions,follmer2016stochastic}.
Submodularity is not only convenient from a mathematical point of view, but as we will show in Section~\ref{sec:choquet} it has a rich interpretation in terms of systematic risk aversion. It is known that a monotone submodular capacity is always coherent\footnote{The Choquet integral is convex if and only if the capacity is submodular \citep{alfonsi2015simple}. Furthermore, it is monotone and translation equivariant. Restricting the Choquet integral to events hence yields a coherent upper probability, which coincides with the submodular capacity on events.}, due to its envelope being non-empty:
\begin{align}
    \operatorname{core}(\overline{\mu}) &= \left\{P : P(A) \leq \overline{\mu}(A) \text{ } \forall A \in \mathcal{F} \text{, } P \text{ probability measure}\right\}\\
    \overline{\mu}(A) &= \sup_{P \in \operatorname{core}(\overline{\mu})}{P(A)}
\end{align}
In the general case of a capacity, the envelope is also called the `core' and consists of finitely additive probability measures. However, we have defined our capacity on a $\sigma$-algebra and therefore the core consists of countably additive measures. Employing the core, the Choquet integral \eqref{eq:distortion} for a submodular capacity is equivalently given by
\begin{equation}
\label{eq:coreint}
    R_\phi(X) = \sup_{P \in \operatorname{core}(\overline{\mu})}\left\{\int_{-\infty}^{\infty} X \d P\right\}.
\end{equation}
Thus, for a submodular capacity, Walley's natural extension coincides with the Choquet integral. This can be understood by viewing \eqref{eq:coreint} as forming the linear extensions via integration of the additive probabilities which are dominated by the capacity. This is yet another argument that demonstrates the specialness of distortion/spectral risk measures: they are natural extensions of coherent upper probabilities.

\begin{figure}
    \centering
    \begin{subfigure}[b]{0.35\textwidth}
    \includegraphics[width=\textwidth]{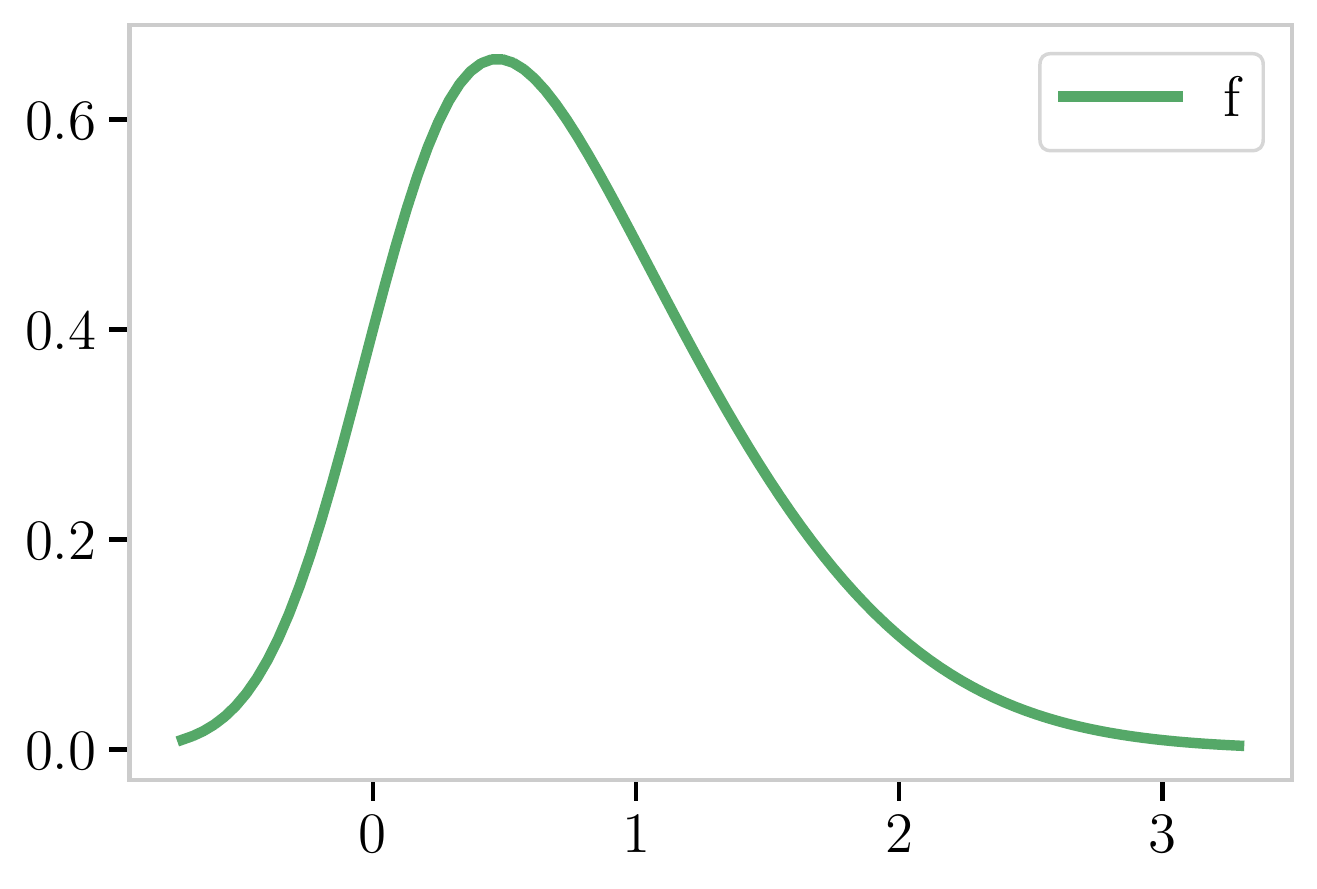}
    \end{subfigure}
    \begin{subfigure}[b]{0.35\textwidth}
    \includegraphics[width=\textwidth]{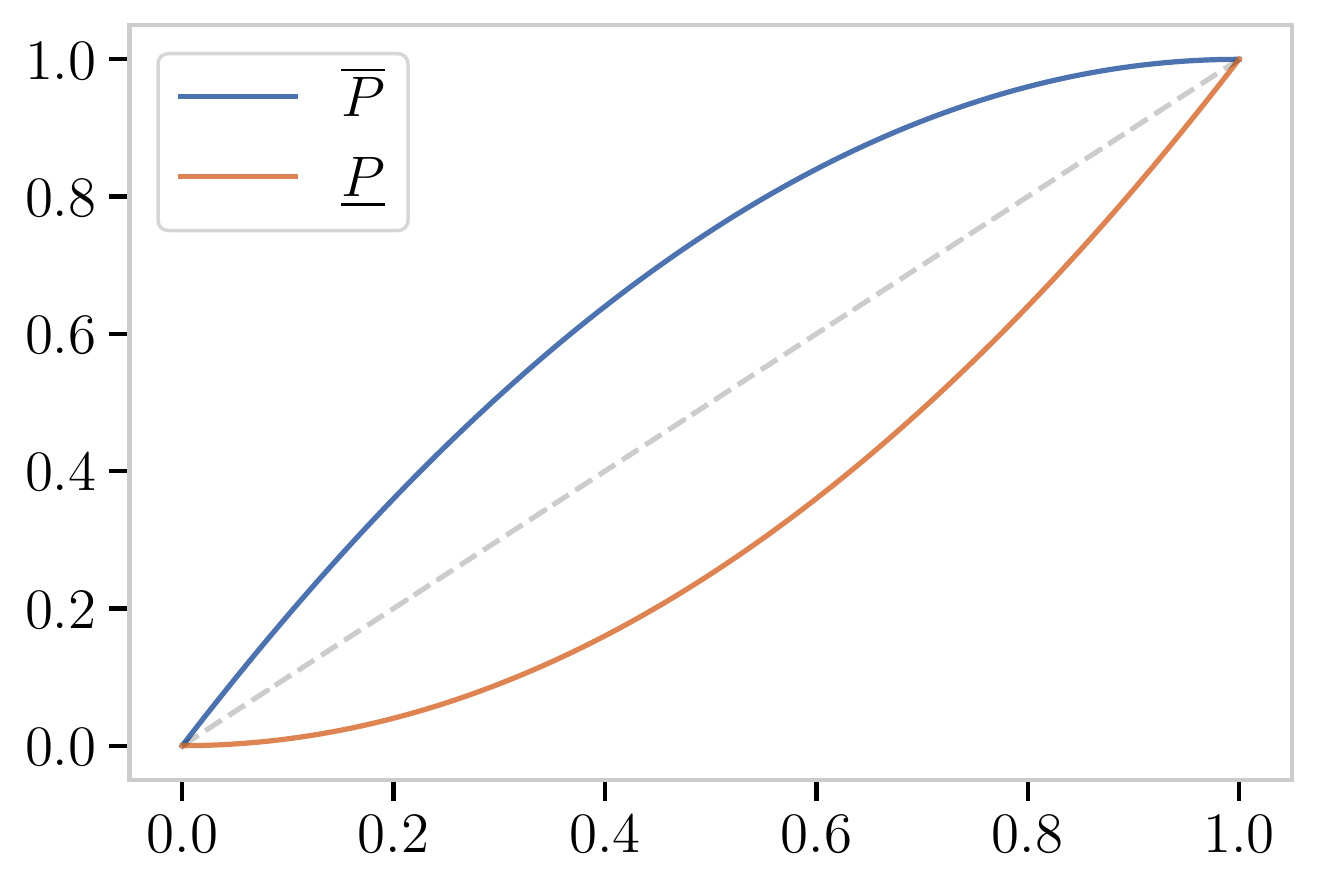}
    \end{subfigure}
    \begin{subfigure}[b]{0.35\textwidth}
    \includegraphics[width=\textwidth]{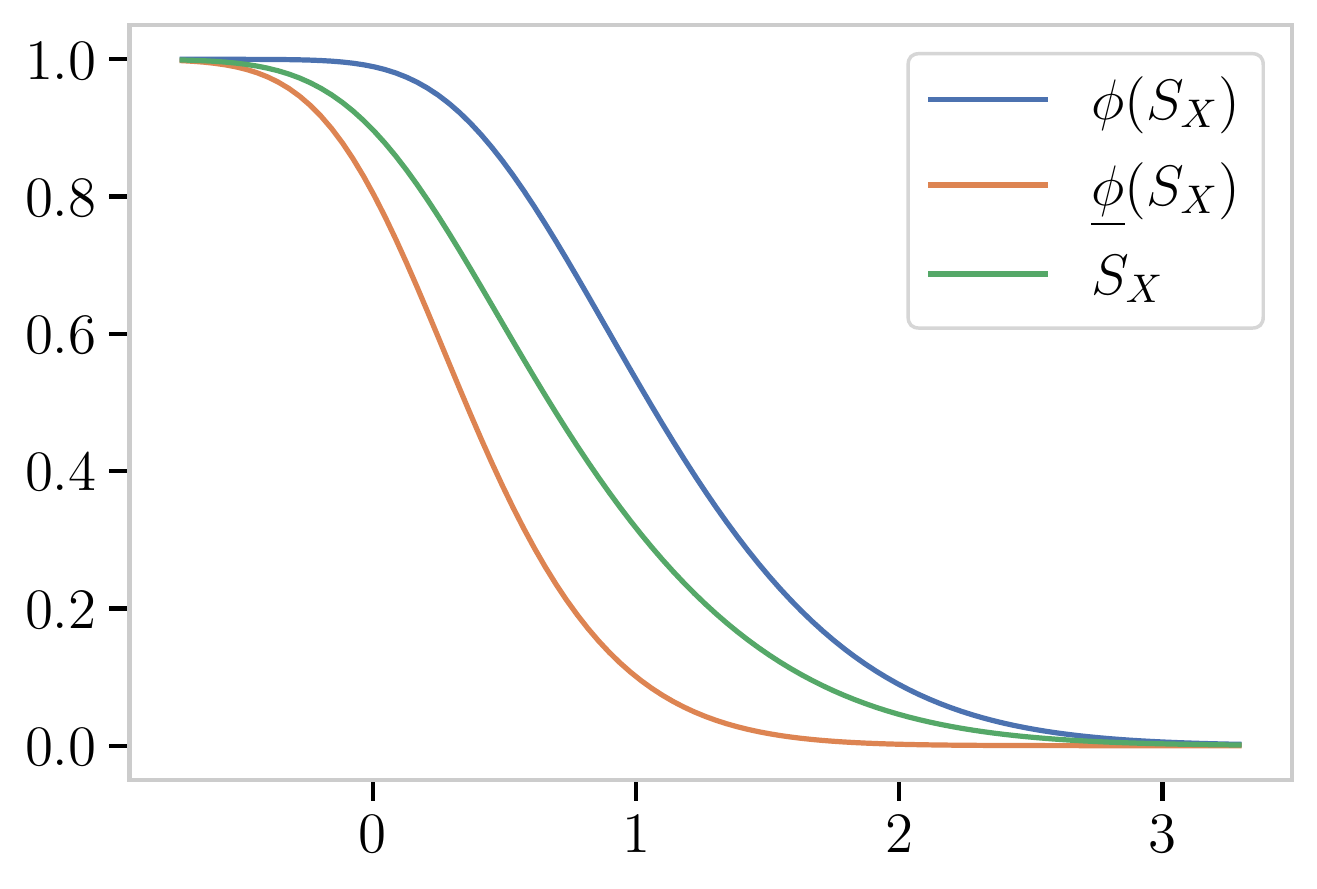}
    \end{subfigure}
    \begin{subfigure}[b]{0.35\textwidth}
    \includegraphics[width=\textwidth]{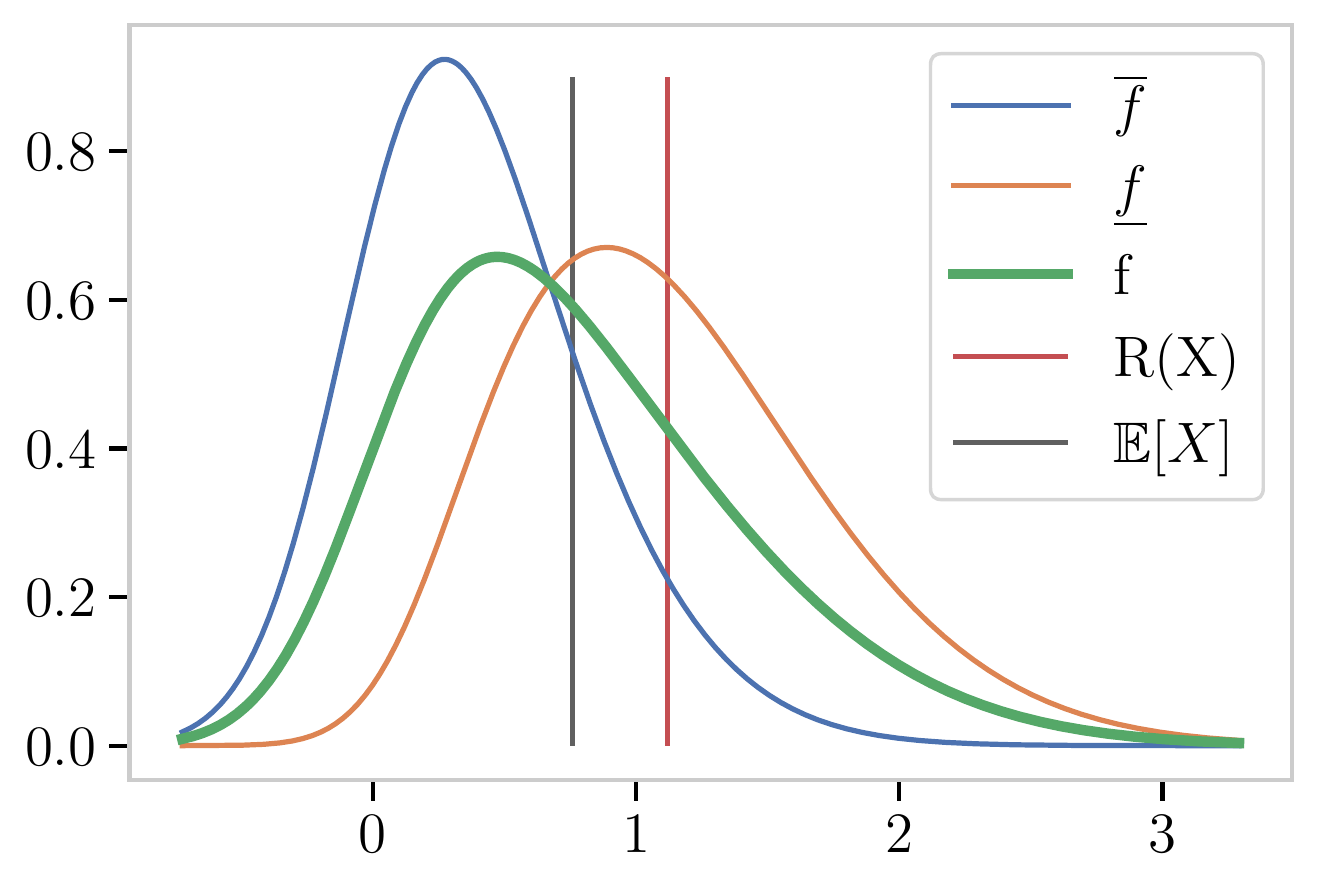}
    \end{subfigure}
    \caption{Top left: the density of an exemplary skew-normal distribution, belonging to some random variable $X$. Top right: lower and upper probabilities with distortion function $\phi(t)=1-(1-t)^2$. Bottom left: lower and upper distortion of the survival function, corresponding to the exemplary distribution. Bottom right: lower and upper densities, resulting from the distortion. The vertical lines indicate the expectation and the distortion risk. Note that $R_\phi(X)$ is substantially greater than $\mathbb{E}[X]$.}
    \label{fig:spectral_probs}
\end{figure}

To a given submodular capacity $\overline{\mu}$, we can define a corresponding dual capacity $\underline{\mu}(A) \coloneqq 1-\overline{\mu}(A^C)$ $\forall A \in \mathcal{F}$. This capacity is supermodular:
\begin{equation}
    \underline{\mu}(A\cup B) +\underline{\mu}(A\cap B) \geq \underline{\mu}(A)+\underline{\mu}(B) \quad \forall A,B \in \mathcal{F}
\end{equation}
and is a coherent lower probability. In our case, we can identify it as $\underline{\mu}=\underline{\phi} \circ P$, with the convex function $\underline{\phi}(t) = 1 - \phi(1-t)$. The upper and lower distribution functions are then defined as follows \citep[p. 130]{walley1991statistical}:
\begin{align}
    \overline{F}_X(x) &\coloneqq \Pup(X \leq x) = 1 - \Plow(X > x)\\
    \underline{F}_X(x) &\coloneqq \Plow(X \leq x) = 1 - \Pup(X > x).
\end{align}
Hence we can compute upper and lower densities as $\overline{f}_X = \overline{F}_X'$ and $\underline{f}_X = \underline{F}_X'$. This terminology is, however, somewhat unfortunate: the upper distribution function owes its name to the fact that it lies above the lower distribution function, but the upper distribution function is obtained from the lower survival function.
Figure~\ref{fig:spectral_probs} gives an intuition about the lower and upper probabilities, survival functions and densities for an exemplary distortion. To compute the distortion risk of $X$, \ie the upper prevision, one computes the expectation with respect to the upper survival function $\phi(S_X)$ or the lower density $\underline{f}$.

We subsequently use the term \textit{distortion risk measure} to refer to a distortion risk measure with a concave distortion $\phi$. Hence we may use the term interchangeably with \textit{spectral risk measure}.

\subsection{Coherent Measures of Risk or Uncertainty?}
\label{sec:equivalence_consequences}
In the finance context, the theory of coherent risk measures has been advanced as a theory about risk, as the name suggests. There is still a conceptual reliance on a single ``true'' probability measure and the goal is to embody a risk-averse attitude by specifying a more conservative (pessimistic) summary of a distribution. Due to the envelope representation, we can however interpret a risk measure as taking the worst-case decision with respect to a set of probability measures. This amounts to introducing artificial ``hallucinated'' ambiguity into a  decision under risk. Hence a connection to Walley's theory of imprecise probability is established and the mathematical equivalence is given an interpretation. A key conceptual difference is whether a distinguished base measure can still be identified, as in the case of risk measures, or whether one deals with a credal set consisting of various linear previsions, as in Walley's case.

A decision maker who uses a law invariant coherent risk measure has an underlying probability measure, but discounts her own belief in it. As an important example of this line of thinking, following the financial crisis, ``the Turner Review points to an excessive reliance on a single probabilistic model $P$ derived from past observations'' \citep{follmer2015axiomatic}. As a response, coherent risk measures have received increasing attention.
Using such a risk measure, a decision maker transforms the risky situation into an ambiguous situation by considering other similar probability measures, as well. If she further employs a spectral risk measure which is based on a distortion of the original distribution, we can conclude that she is coherent if and only if she assigns bigger weights to worse cases \citep{acerbi2002spectral}. Thus she exhibits a systematic risk aversion attitude, which is encoded in the spectrum (or equivalently, the distortion function).

The coincidence of the coherence concept in imprecise probability and the finance literature on risk measures (Theorem~\ref{theorem:coherencecoincidence}) is particularly interesting because the axioms are motivated in different fashion. \cite{walley1991statistical} provides a behavioral justification for coherence, tailored to the situation in which a decision maker finds herself when facing uncertainty. Walley's \citeyearpar{walley1991statistical} goal is to provide a guide to rational decision making. In finance, the agent is an institution or a regulator. For instance, subadditivity is then motivated as encouraging diversification; translation equivariance, in this context called ``cash-invariance'', is motivated by requiring that adding a certain amount of cash (negative loss) should decrease risk by exactly that amount. That the extensions of these two coherence concepts coincides is remarkable and serves as a corroboration of their groundedness.

Coherent risk measures are also intimately connected with generalized utility theories in rational choice theory, situated in the context of economics. These theories offer formal axiomatic bases for rational decision making under uncertainty. In Appendix~\ref{app:utility}, we explore the connections between risk measures and (non)-expected utility theories. In particular, the class of spectral risk measures has been reinvented in this setting as \textit{Choquet expected utility} or, more precisely, as \textit{rank dependent expected utility}.

In machine learning, the ``excessive reliance on a single probabilistic model $P$ derived from past observations'', in the words of \cite{follmer2015axiomatic}, is problematized in the context of data set shift and more generally, it is problematic due to having only a finite amount of training data, from which the ``true'' distribution can only be approximated. By putting \textit{true} in quotes, we wish to emphasize that the assumption of a single underlying probability measure is itself a questionable one, although it has received little attention yet. Data from the real world may exhibit unstable relative frequencies over time \citep{gorban2017statistical} and hence, at least from a frequentist perspective, cannot be based on a single probability distribution (see \citet{frohlich2023towards}). Furthermore, predictions can even influence the outcomes they aim to predict -- a phenomenon known as \textit{performative prediction} \citep{perdomo2020performative}.
Our goal is to contribute to tackling such problems by demonstrating how coherent risk measures, in particular spectral risk measure, can be helpful as a generalized theory of uncertainty.

\section{Rearrangement Invariant Banach Function Spaces}
\label{sec:banach}
In this section, we show that coherent risk measures are an incarnation of \textit{rearrangement invariant Banach function norms} and are hence embedded in a rich mathematical literature. This connection is, to the best of our knowledge, previously unknown and enables us to obtain novel characterization results. While some authors have studied norms related to risk measures (\eg \citet{pichler2013natural}, \citet{mafusalov2016cvar} and \citet{gotoh2016two}), we here present a broader picture.
We follow mainly the technical setup of \cite{Bennett:1988aa}. For a more accessible introduction we refer to \citep{rubshtein2016foundations}, who use the term ``symmetric spaces'' instead. Throughout, we work with the probability space $\Omega=[0,1]$ with the Lebesgue measure $\mu$, so $\mu(\Omega)=1$. This space is a standard probability space and all such non-atomic standard spaces are Borel isomorphic \citep[see \eg][]{bauerle2006stochastic}. Hence this introduces no loss of generality for our setting but allows for a cleaner exposition. Let $\mathcal{M}$ denote the class of Lebesgue measurable functions from $\Omega$ to $\mR$ and $\mathcal{M}^+$ the subset of Lebesgue measurable functions with values in $[0,\infty]$, \ie nonnegative random variables. (In)equalities between elements of $\mathcal{M}$ are to be understood as holding $\mu$-almost everywhere. Often, we will drop writing $X \in \mathcal{M}$ for brevity, since we have no concern for non-measurable functions throughout.

\begin{definition}\label{def:rifunctionnorm}
A functional $R : \mathcal{M}^+ \rightarrow [0,\infty]$ is called \textit{Banach function norm} if the following conditions hold for all $X_n,X \in \mathcal{M}^+$ and measurable $E \subseteq \Omega$:
\begin{enumerate}[label=\textbf{R\arabic*.}, ref=R\arabic*]
  \item \label{item:norm} $R(X)=0 \Leftrightarrow X = 0$; \quad $R(\lambda X) = \lambda R(X)$ $\forall \lambda \geq 0$; \quad $R(X+Y) \leq R(X) + R(Y)$
  \item \label{item:mon} $0 \leq X \leq Y \Rightarrow R(X) \leq R(Y)$
  \item \label{item:fatou} $0 \leq X_n \uparrow X \thinspace \thinspace \mu\text{-a.e.} \Rightarrow R(X_n) \uparrow R(X)$
  \item \label{item:technical} $R(\chi_E) < \infty$; \quad $\int_E X \d \mu < c_E R(X)$ for some $0 < c_E < \infty$ depending only on $E$ and $R$.
\end{enumerate}
\end{definition}

Since we work with a finite measure space, we also impose $R(\chi_\Omega)=1$ without loss of generality throughout the paper. Due to positive homogeneity, $R(\chi_\Omega)=c$ would simply correspond to a scaling of the function norm. Observe that a function norm $R$ is defined only on the positive cone of measurable functions. However, it induces a norm on the space $\mathcal{R} = \left\{X : R(|X|) < \infty\right\}$ by setting
\begin{equation}
\label{eq:absval}
    \|X\|_\mathcal{R} \coloneqq R(|X|).
\end{equation}
Then it can be shown that the pair $(\cR, \|\cdot\|_\cR)$ forms a \textit{Banach space}, \ie a complete normed vector space. For completeness of the space, the key axiom is the Fatou property \ref{item:fatou}. In the context of risk measures, however, it is undesirable to extend a function norm from the positive cone to the whole space by stipulating \eqref{eq:absval}. The reason is that we want to treat negative values (gain) as different from positive values (loss). Hence we restrict ourselves to nonnegative random variables $\mathcal{M}^+$ in the following discussion. Then, in virtue of \ref{item:norm} and \ref{item:mon}, a coherent risk measure can be viewed as a valid Banach function norm, if it also satisfies the mild technical axioms \ref{item:fatou} and \ref{item:technical}. A coherent risk measure further satisfies translation equivariance, however. We discuss the subtle role of (non)negativity and translation equivariance in Section~\ref{sec:te} below.

We are specifically interested in \textit{rearrangement invariance} of norms, which corresponds to the law-invariance property of risk measures. The idea is that such a norm only attends to the distribution of a function and hence respects the base measure $\mu$ in a suitable way, thereby disregarding the order in which the values are arranged. To this end, one defines the \textit{distribution function} $\mu_X : \mR^+ \rightarrow [0,1]$ of $X \in \mathcal{M}$ as
\begin{equation}
\label{eq:dist}
    \mu_X(\lambda) \coloneqq \mu\left\{\omega \in \Omega: |X(\omega)| > \lambda\right\}.
\end{equation}
For nonnegative random variables, this decreasing (non-increasing) and right-continuous function is just the survival function $S_X = 1 - F_X$. Two functions $X$ and $Y$ are called \textit{equimeasurable} if their distribution functions coincide, \ie $\mu_X(\lambda)=\mu_Y(\lambda)$ $\forall \lambda \geq 0$.
\begin{definition}
\label{def:rifunctionnormdef}
A Banach function norm $R$ is called rearrangement invariant if $R(X)=R(Y)$ for every equimeasurable $X,Y$. The space $\cR$ is then called a rearrangement invariant Banach space.
\end{definition}
From now on we abbreviate rearrangement invariant as \textit{ri} and call $\cR$ an \textit{ri space}. For each $X \in \mathcal{M}$, we obtain a canonical equimeasurable function $X^* : [0,1] \rightarrow \mR^+$ as the generalized inverse of its distribution function:
\begin{align}
\label{eq:decr}
    X^*(\omega) &\coloneqq \inf\{\lambda \geq 0: \mu_X(\lambda) \leq \omega \}, \quad \omega \in [0,1)\\
    X^*(1) &\coloneqq \lim_{\omega \uparrow 1} X^*(\omega).
\end{align}
$X^*$ is called the \textit{decreasing rearrangement} of $X$, as it arranges the (absolute) values of $X$ in decreasing order. It is therefore the continuous analog of sorting a list in descending order. $X^*$ is clearly decreasing and right-continuous. In the context of standard probability theory, this corresponds to the lower ``backwards'' quantile of $|X|$:
\begin{align}
    X^*(\omega) &= \inf\{\lambda \geq 0: \mu_X(\lambda) \leq \omega \} \\
    &= \sup\{\lambda \geq 0: \mu_X(\lambda) > \omega \} \\
    &= \sup\{\lambda \geq 0: 1- \mu_X(\lambda) < 1-\omega \}\\
    &= \sup\{\lambda \geq 0: F_{|X|}(\lambda) < 1-\omega \}\\
    &= F_{|X|}^{-1}(1-\omega).
\end{align}
The rationale for working with a decreasing, instead of an increasing rearrangement, is that the ri Banach space theory generally considers spaces of potentially infinite measure; hence a plot of an increasing rearrangement might not show anything interesting until $+\infty$.
For an ri function norm $R$, we have in particular $R(X)=R(X^*)$. A law invariant coherent risk measure induces an ri function norm, which is furthermore translation equivariant. 

\subsection{Duality and the Associate Space}
Banach spaces have an interesting duality aspect, which we will connect to the envelope representation. The \textit{dual space} $\cR^*$ of a Banach space $\cR$ consists of all linear, continuous and bounded functionals $u: \cR \rightarrow \mR$, and is equipped with the norm \citep[p.\@\xspace 83]{rubshtein2016foundations}
\begin{equation}
    \|u\|_{\cR^*} = \sup\left\{|u(X)| : \|X\|_\cR \leq 1\right\} < \infty.
\end{equation}
There exists a close relationship between the dual space and the \textit{associate space} to a function norm $R$, which is of more practical interest. For an ri norm $R$, the associate (function) norm is defined by
\begin{align}
    R'(X) &\coloneqq \sup\left\{\int_0^1 X^*(\omega) Y^*(\omega) \d \omega : R(Y) \leq 1, Y \in \mathcal{M}^+\right\}\\
    \|X\|_{\cR'} &\coloneqq \sup\left\{\int_0^1 X^*(\omega) Y^*(\omega) \d \omega : \|Y\|_\cR \leq 1, Y \in \cR \right\}.
\end{align}
With this pairing, the associate space $\cR'$ is canonically isometrically isomorphic to a closed ``norm-fundamental'' subspace of $\cR^*$ \citep[p.\@\xspace 13]{Bennett:1988aa}. For our purposes, we may ignore the subtle distinction between $\cR'$ and $\cR^*$.
An important aspect of the associate pairing is that Hölder's inequality holds. If $X \in \cR$ and $Y \in \cR'$ then
\begin{equation}
    \int_\Omega |XY| \d \mu \leq \|X\|_\cR \|Y\|_{\cR'}.
\end{equation}
Also, we have that $\cR=(\cR')'$ under the assumption of the Fatou property \ref{item:fatou}.

The prime example of ri spaces are the Lebesgue spaces $\cL^p$. A family of function norms is defined as
\begin{equation}
    R^p(X) \coloneqq \begin{cases}
    \left(\int_0^1 X^p \d \mu \right)^\frac{1}{p} &  1 \leq p < \infty\\
    \text{ess sup}(X) & p=\infty,
    \end{cases}
\end{equation}
where $\text{ess sup}(X) \coloneqq \inf\left\{\lambda \geq 0: \mu_X(\lambda) = 0\right\}$. We label the space induced by $R^p$ as $\cL^p$. The associate space of $\cL^p$ is $\cL^q$, where $\frac{1}{p} + \frac{1}{q} = 1$. For example, $R^1=\mathbb{E}[\cdot]$ is paired with its associate $R_\infty=\text{ess sup}$. On the other hand, the associate of $R^\infty$ is $R^1$, but the dual space is more subtle and in this case, the canonical embedding of $\cL^1$ into $(\cL^\infty)^*$ is strict. For a systematic treatment of this dual space, see \citep{schonherr2017pure}.

\subsection{The Embedding Theorem}
Given ri spaces $\cR$ and $\mathcal{S}$, where $\mathcal{S} \subseteq \cR$, there exists a constant $c$ such that \cite[p.\@\xspace 7]{Bennett:1988aa}
\begin{equation}
    \|X\|_\cR \leq c \|X\|_{\mathcal{S}} \quad \forall X \in \mathcal{S}.
\end{equation}
In this case, $\mathcal{S}$ continuously embeds into $\cR$, which we denote as $\mathcal{S} \hookrightarrow \cR$ and refer to a feasible $c$ as embedding constant (not unique). Let $\cR$ be any ri space. The following is known \citep[][p.\@\xspace 77, specialized to $\mu(\Omega)=1$]{Bennett:1988aa}:
\begin{equation}
\cL^\infty \hookrightarrow \cR \hookrightarrow \cL^1,
\end{equation}
and $1$ is a feasible embedding constant:
\begin{equation}
    \|X\|_{\cL^1} \leq \|X\|_{\cR} \quad \forall X \in \cR, \quad \|X\|_{\cR} \leq \|X\|_{\cL^\infty} \quad \forall X \in \cL^\infty.
\end{equation}
Therefore, $\cL^1$ and $\cL^\infty$ are special as they are the extremes of all ri spaces. This implies in particular that any law invariant coherent risk measure ``lives between'' the expectation and the essential supremum, which stand in an associate relationship. 
This distinguished status is also visible from their envelope representations (Section~\ref{sec:envelopes}): the envelope of the expectation is the singleton $\{1\}$ (a singleton envelope ${c}$ yields a constant multiple of the expectation\footnote{However, requiring that $R(1_\Omega)=1$ precludes such ri norms for constants $c \neq 1$.}), whereas the envelope of the worst-case risk measure consists of all probability measures. 

Note also that $\mathbb{E}[\cdot] = \operatornamewithlimits{CVar}_{\alpha=0}$ and $\text{ess sup} = \operatornamewithlimits{CVar}_{\alpha \shortrightarrow 1}$, hence $\cvar$ in a sense interpolates between the smallest and largest ri function norms. We will later state a more refined embedding theorem, which situates any law invariant coherent risk measure between the spectral risk measure corresponding to its upper probability and the \textit{Marcinkiewicz norm}.

\subsection{Hardy-Littlewood's Inequality}
For nonnegative real sequences $(x_i)$ and $(y_i)$, Hardy-Littlewood's inequality asserts that
\begin{equation}
    \sum_{i=1}^n x_i y_i \leq \sum_{i=1}^n x_i^* y_i^*,
\end{equation}
where $(x_i^*)$ and $(y_i^*)$ are the sequences where the elements of $(x_i)$, respectively $(y_i)$, are arranged in decreasing order. This inequality carries over to the continuous case.
If $X$ and $Y$ are finite $\mu$-almost everywhere, then \citep[p.\@\xspace 44]{Bennett:1988aa}:
\begin{equation}
\label{eq:hardy}
    \int_\Omega |XY| \d \mu \leq \int_0^1 X^*(\omega) Y^*(\omega) \d \omega.
\end{equation}
While this inequality has been employed in the study of Kusuoka representations and envelopes \citep[see \eg][]{pichler2015premiums}, the connection to the general theory of ri spaces has not yet been made.

When $Y$ is the indicator of a measurable set $E$ with $\mu(E)=t>0$, this specializes to
\begin{equation}
    \frac{1}{t} \int_E |X| \d \mu \leq \frac{1}{t} \int_0^t X^*(\omega) \d \omega.
\end{equation}

This suggests the definition of the \textit{maximal function} \citep[pp.\@\xspace 52-53]{Bennett:1988aa}:
\begin{equation}
    X^{**}(t) \coloneqq \frac{1}{t} \int_0^t X^*(\omega) \d \omega
    = \frac{1}{t}\sup\left\{\int_E |X| \d \mu : \mu(E)=t \right\}, \quad t>0,
\end{equation}
where the latter equality is here stated without proof. The maximal function achieves the highest average of the function $X$ over sets of measure $t$. This is done by integrating quantiles backwards:
\begin{align}
    \forall t \in (0,1]: X^{**}(t) &= \frac{1}{t} \int_0^t X^*(\omega) \d \omega\\
    &= \frac{1}{t} \int_0^t F_{|X|}^{-1}(1-\omega) \d \omega \\
    &= \frac{1}{1-t} \int_{1-t}^{1} F_{|X|}^{-1}(\omega) \d \omega \\
    &= \operatorname{CVar}_{1-t}(|X|).
\end{align}
The special behaviour of $\cvar$ is due to the fact that it integrates the function only in its $1-\alpha$ tail, where the function values are highest (recall that $X^*$ is decreasing). We observe the following remarkable fact \citep[p.\@\xspace 61]{Bennett:1988aa}
\begin{equation}
    \left(\forall \alpha \in [0,1): \cvar(|X|) \leq \cvar(|Y|) \right)\implies \|X\|_{\cR} \leq \|Y\|_{\cR}
\end{equation}
for any ri norm $\|\cdot\|_{\cR}$.
We will later see that the special behaviour of $\cvar$ is in some sense shared by the wider class of spectral risk measures (Theorem~\ref{theorem:lorentzlargest}). To this end, we need to introduce the \textit{fundamental function} of an ri space.

\subsection{The Fundamental Function}
\begin{definition}
Let $\cR$ be an ri space with function norm $R$. For each measurable subset $E \subseteq \Omega$ with measure $\mu(E)=t$, we define the \textit{fundamental function} $\phi_\cR: [0,1] \rightarrow \mR^+$ as
\begin{equation}
    \phi_\cR(t) \coloneqq \|\chi_E\|_\cR = R(\chi_E),
\end{equation}
where the latter equality comes from the nonnegativity of indicator functions.
\end{definition}
When the space is clear from the context, we drop the subscript. Due to the ri property, the choice of the set $E$ does not matter. If $R$ is a law invariant coherent risk measure, \ie an upper prevision, $\phi(t)$ specifies a coherent upper probability and then\footnote{A coherent risk measure satisfies translation equivariance, which also implies $R(c)=c$, hence $R(\chi_\Omega)=\phi(1)=1$.} $\phi(1)=1$. Since we stipulated $R(\chi_\Omega)=1$ for any ri function norm, it always holds that $\phi(1)=1$. The value of $t$ corresponds to the underlying probability with respect to the base measure, $\mu(E)$, which is then distorted through $R$. For example, the expectation has the fundamental function $\phi_{\cL^1}(t)=t$, whereas the $\text{ess sup}$ has fundamental function $\phi_{\cL^\infty}(t)=\chi_{(0,1]}$, so that $\phi_{\cL^\infty}(0)=0$ and $\phi_{\cL^\infty}(t)=1$ otherwise. For any ri space, the fundamental function is \textit{quasiconcave}, that is, it satisfies \citep[p.\@\xspace 67]{Bennett:1988aa}:
\begin{align}
    &\phi \text{ is non-decreasing and }\phi(0)=0\\
    &t \mapsto \phi(t)/t \text{ is decreasing}\\
    & \phi \text{ is continuous except perhaps at the origin.}
\end{align}
However, we focus on concave fundamental functions. Every concave function is also quasiconcave, but the converse is not necessarily true. The conceptual reason for our restriction is that the fundamental function models risk aversion on events: the indicator function $\chi_E$ represents the uncertain unit loss with probability $\mu(E)$ and $0$ loss with probability $1-\mu(E)$. Then $\phi(t)=\phi(\mu(E))$ is our risk assessment for this simple random variable. We argued before that a reasonable risk aversion profile is always concave, as it then puts more weight on worse outcomes in a systematic way. Hence $\phi \circ \mu$ defines a submodular capacity on events.

Mathematically, the restriction to concave fundamental functions is also not significant since it can be shown that an ri space with quasiconcave fundamental function $\phi$ can always be equivalently renormed to have a concave fundamental function, the least concave majorant of $\phi$ \citep[p.\@\xspace 71]{Bennett:1988aa}. Henceforth we always assume $\phi$ to be concave. We denote the class of concave functions $\phi: [0,1] \rightarrow [0,1]$ with $\phi(0)=0$ and $\phi(1)=1$ as $\Phi$. The right limit at $0$ is $\phi(0+)$. If the additional condition of continuity at $0$ is also satisfied, \ie $\phi(0+)=0$ and $\phi \in \Phi$, we write $\phi \in \Phi_{0+}$. 

Since $\phi$ only encodes the behavior of $R$ on events, \ie an upper probability, there is some freedom left in specifying a corresponding risk measure. However, we will show that $\phi$ still imposes significant structure (Section~\ref{sec:families},~\ref{sec:equivalences}). As an example, we consider $\phi(t)=1-(1-t)^2=2t-t^2$. Two different risk measures, which share this fundamental function, are \textit{MaxVar} \citep{cherny2009new} ($\maxv$) and the \textit{Dutch risk measure} \citep{van1992dutch} ($\du$):
\begin{align}
\label{eq:dutchminvar}
    \maxv(X) &\coloneqq \mathbb{E}\left[\max(X_1,X_2)\right], \thinspace \thinspace X_1,X_2 \stackrel{\text{ind}}{\sim} X \quad \forall X \in \mathcal{M}^+\\
    \du(X) &\coloneqq \mathbb{E}\left[\max(X,\mathbb{E}[X])\right] \quad \forall X \in \mathcal{M}^+,
\end{align}
where $X_1,X_2 \stackrel{\text{ind}}{\sim} X$ means that the random variables are independent and share the same distribution. $\maxv$ is indeed the spectral risk measure corresponding to the distortion $\phi$. Let $X \in \mathcal{M}^+$:
\begin{align}
    \maxv(X) &= \int_0^\infty 1-(1-S_X(\omega))^2 \d \omega
    = \int_0^\infty 1-(1-(1-F_X(\omega)))^2 \d \omega\\
    &= \int_0^\infty 1-F_X^2(\omega) \d \omega.
\end{align}
This is just the expectation of a random variable with distribution function $F_Y = F_X^2$, \ie $Y = \max(X_1,X_2)$, $X_1,X_2 \stackrel{\text{ind}}{\sim} X$.
The Dutch risk measure, on the other hand, is also law invariant but not spectral. It is also easy to see that $\mathbb{E}[X] \leq \du(X) \leq \maxv(X) $ $\forall X \in \mathcal{M}^+$ by applying Jensen's inequality in \eqref{eq:dutchminvar}. In this way, the Dutch risk measure is more optimistic than MaxVar on general random variables, even if they share the same risk aversion profile on indicator functions.
This is no coincidence: we now show that, given an arbitrary concave fundamental function, spectral risk measures correspond to the most pessimistic extension of $\phi$ to all (nonnegative) random variables. In contrast, we observe in Theorem~\ref{theorem_tm} that the Dutch risk measure is the most optimistic extension for its specific $\phi$.

\subsection{The Lorentz and Marcinkiewicz Norms}
Given any concave fundamental function $\phi \in \Phi$, the \textit{Lorentz norm} of $X \in \mathcal{M}$ is defined as
\begin{align}
    \|X\|_{\Lambda_\phi} &\coloneqq \int_0^1 X^*(\omega) \d \phi(\omega)\\
    &= X^*(0) \phi(0+) + \int_0^1 X^*(\omega)  \phi'(\omega) \d \omega\\
    &= X^*(0) \phi(0+) + \int_0^1 F_{|X|}^{-1}(1-\omega)  \phi'(\omega) \d \omega,
\end{align}
where we immediately recognize the correspondence to the distortion (spectral) risk measure \eqref{eq:spectral} on the positive cone with distortion $\phi$, if $\phi \in \Phi_{0+}$, \ie if $\phi$ is continuous at $0$. To the best of our knowledge, this connection has not been reported yet. It is easy to check that the Lorentz norm indeed has fundamental function $\phi$. In particular, $\|\cdot\|_{\cL^1}$ and $\|\cdot\|_{\cL^\infty}$ are both Lorentz norms for their respective fundamental functions. While $\phi_{\cL^1} \in \Phi_{0+}$, we have $\phi_{\cL^\infty} \notin \Phi_{0+}$. The effect of $\phi(0+) > 0$ is to put a fixed weight on the supremum, meaning that further decreasing its probability would not further decrease its weight. In the extreme case of $\phi_{\cL^\infty}$, all the weight is put on $X^*(0)=\text{ess sup}(|X|)$. In practice, we see little motivation for choosing a $\phi \in \Phi\setminus \Phi_{0+}$.

Another important norm, the \textit{Marcinkiewicz norm} of $X \in \mathcal{M}$, is defined as
\begin{align}
    \|X\|_{M_\phi} &= \sup_{0 < t \leq 1}\left\{\phi(t) X^{**}(t)\right\}\\
    &= \sup_{0 < t \leq 1}\left\{\phi(t) \frac{1}{t} \int_0^t X^*(\omega) \d \omega \right\}\\
    &= \sup_{0 < t \leq 1}\left\{\phi(t) \cvarnoa_{1-t}(|X|) \right\}
\end{align}
and also has fundamental function $\phi$. It is clear that both the Lorentz and the Marcinkiewicz norms are rearrangement invariant, as they are defined in terms of $X^*$. For the proof that they are indeed valid ri norms, we refer to \cite[pp.\@\xspace 116, 143]{rubshtein2016foundations}.

\begin{theorem}\citep[p.\@\xspace 72]{Bennett:1988aa}.
\label{theorem:sandwich}
Let $\Lambda_\phi$ (resp.~$M_\phi$) be the  ri spaces of the functions for which the Lorentz (resp.~Marcinkiewicz) norm is finite. For any other ri space $\cR$ with fundamental function $\phi \in \Phi$ we have the embedding 
\begin{equation}
    \Lambda_\phi \hookrightarrow \cR \hookrightarrow M_\phi
\end{equation}
and $1$ is a feasible embedding constant:
\begin{equation}
    \|X\|_{M_\phi} \leq \|X\|_R \quad \forall X \in \cR, \quad \|X\|_R \leq \|X\|_{\Lambda_\phi} \quad \forall X \in \Lambda_\phi.
\end{equation}
\end{theorem}
Since an ri space consists of those functions for which the norm is finite, the largest norm yields the smallest space and vice versa. We here state the theorem without proof. In Section \ref{sec:families} we provide a novel proof, which also gives an intuition for the \textit{why} behind the result.
This ``sandwiching'' result justifies the name \textit{fundamental function}: it indeed captures a fundamental aspect of an ri norm and confines all coherent risk measures with a given fundamental function to live between the Marcinkiewicz and the Lorentz norm of that fundamental function. From this it follows, for example, that $\du(X) \leq \maxv(X)$ $\forall X \in \mathcal{M}^+$, as the $\maxv$ is the Lorentz norm and they have the same fundamental function.
This result has direct behavioural implications for a decision maker: given a law invariant coherent upper probability, the natural extension, which coincides with its spectral risk measure (Section~\ref{sec:distortion}), hence the Lorentz norm, is the most pessimistic in the sense that it assigns the highest risk to random variables, while being compatible with the specified upper probability. On the other hand, the Marcinkiewicz norm is its most optimistic extension. However, in contrast to the Lorentz norm, the Marcinkiewicz norm is not in general translation equivariant (see Section~\ref{sec:te},\ref{sec:families}) and thus not in general a coherent risk measure (on $\mathcal{M}^+$).

In fact, the Lorentz and the Marcinkiewicz norm stand in a dual relationship. The dual fundamental function to $\phi$ is $\phi^*(t) \coloneqq t/\phi(t)$ and can be shown to be the fundamental function of the associate space\footnote{This holds true generally \citep[p.\@\xspace 135]{rubshtein2016foundations}, not restricted to the Lorentz/Marcinkiewicz duality. Note that if $\phi$ is concave, the dual fundamental function might only be quasiconcave.}. Then we can write the Lorentz norm as
\begin{equation}
    \|X\|_{\Lambda_\phi} = \sup\left\{\int_0^1 X^*(\omega) Y^*(\omega) \d \omega : \|Y\|_{M_{\phi^*}} \leq 1, Y \in \mathcal{M}^+ \right\},
\end{equation}
using the Marcinkiewicz norm with the dual fundamental function as its associate norm.
The other direction is more complicated: if $\phi$ is concave, $\phi^*$ might in general only be quasiconcave. However, the Lorentz norm is only a norm for concave fundamental functions. It can be shown that the dual of the Marcinkiewicz norm then is the Lorentz norm with respect to the least concave majorant of $\phi^*$ \citep[p.\@\xspace 147]{rubshtein2016foundations}.
For example, the associate relationship of $\|\cdot\|_{\cL^\infty}$ and $\|\cdot\|_{\cL^1}$ is due to the Marcinkiewicz-Lorentz duality.

It has been observed \citep[][p.\@\xspace 157]{rubshtein2016foundations} that in some special cases $\|\cdot\|_{\Lambda_\phi}=\|\cdot\|_{M_\phi}$, that is, a coincidence of the Lorentz and the Marcinkiewicz norm for the same fundamental function. As a consequence, the space of all ri norms collapses to a point due to the embedding theorem: there is then only a single ri norm with the given fundamental function. For instance, this holds true for $\cL^1$:
Let $\phi(t)=t$. Then $\|X\|_{\Lambda_\phi}=\|X\|_{\cL^1}$. Also,
\begin{align}
    \|X\|_{M_\phi} &= \sup_{0 < t \leq 1}\left\{\phi(t) X^{**}(t)\right\} 
    = \sup_{0 < t \leq 1}\left\{t \cdot  X^{**}(t)\right\}\\
    &= \sup_{0 < t \leq 1}\left\{\int_0^t X^*(\omega) \d \omega \right\}
    = \int_0^1 X^*(\omega) \d \omega = \|X\|_{\cL^1}.
\end{align}
Similarly, one easily checks that the coincidence also holds for $\cL^\infty$. Hence $\cL^1$ and $\cL^\infty$ are distuingished spaces as they allow only a single ri norm. We prove a novel result in Section~\ref{sec:families} (Theorem~\ref{theorem_mlcvar}): the Lorentz and Marcinkiewicz norm coincide if and only if the fundamental function is of the form $\phi(t)=\min(t/(1-\alpha),1)$ for some $\alpha \in [0,1)$ or $\alpha \rightarrow 1$, \ie for CVar-type fundamental functions. This further underlines the particularity of CVar, as it is the single coherent risk measure with this fundamental function (upper probability).

\subsection{Nonnegativity and Translation Equivariance}
\label{sec:te}
In the literature on ri spaces, a function norm is only defined on functions taking values in $[0,\infty]$. Recall that to obtain a valid Banach space, this is then extended to a norm by $\|X\| = R(|X|)$ using the absolute value. In our context, this is undesirable, as we want to distinguish loss from gain. Furthermore, we are interested in translation equivariant functionals. One possibility to resolve this tension is to postulate that all random variables are bounded from below --- in the context of machine learning, losses are often bounded from below by $0$. If the lower bound is negative, we can compute in the presence of translation equivariance:
\begin{equation}
    R(X) = R(X+c)-c,
\end{equation}
for some constant $c$ so that $\text{ess inf}(X+c) \geq 0$. It is then sufficient to define the norm only for nonnegative random variables. However, the  definition of translation equivariance itself requires dealing with potentially negative random variables. We instead propose the following restricted definition of positive translation equivariance (PTE).
\begin{definition}
An ri function norm $R$ is called \emph{PTE} if
\begin{equation}
\label{eq:PTE-def}
    \forall X \in \mathcal{M}^+, c \in \mR \text{ s.t. } X+c \geq 0: R(X+c)=R(X)+c.
\end{equation}
We also call an ri norm \emph{PTE} if it is induced by an ri function norm which is \emph{PTE}, and similarly we call an ri space \emph{PTE} if it carries an ri norm which is \emph{PTE}.
\end{definition}
The constant $c$ can potentially be negative but we require that $X+c$ is nonnegative.
We now show that PTE is equivalent to the possibility of reducing the representation of a function norm via its associate to dual variables with $\mathbb{E}[Y] = R'(Y) = 1$. Recall that \textit{any} ri function norm admits a representation of the form:
\begin{equation}
R(X) = \sup\left\{\int_0^1 X^*(\omega) Y^*(\omega) \d \omega : R'(Y) \leq 1, Y \in \mathcal{M}^+\right\}.
\end{equation}

\begin{theorem}
\label{theorem:minrep}
An ri function norm $R$ can be represented in the following reduced form if and only if it is positive translation equivariant $(\PTE)$:

\begin{align}
\label{eq:minrep}
    R(X) = \sup\left\{\int_0^1 X^*(\omega) Y^*(\omega) \d \omega : \mathbb{E}[Y]=R'(Y)=1, Y \in \mathcal{M}^+\right\}.
\end{align}
\end{theorem}
We call this the positive translation equivariant representation of $R$. The proof is in Appendix~\ref{app:terep}.
\begin{remark}\normalfont
When risk measures $R$ are defined on the whole $\mathcal{M}$, translation equivariance \textit{requires} that for any $Y$ in any envelope representation $\mathcal{Y}$ of $R$, it holds $\mathbb{E}[Y]=1$. Standard proofs for this (see Section~\ref{sec:envelopes}) rely on negative values. With the restriction to the positive cone, we are only able to make the weaker statement that $R$ \textit{allows} such a representation, not that any representation needs to be of this form.
\end{remark}
\begin{example}\normalfont
Consider $R^1=\mathbb{E}$. When it is defined on the whole space, the only envelope representation of $R$ is the singleton $\{1\}$. When $R$ is restricted to the positive cone $\mathcal{M}^+$, the set $\{Y : R^\infty(Y) \leq 1\}$ is also a valid representation. To see that this set is not a valid envelope on the whole space, consider the case of negative $X$ and $Y=0$.
\end{example}
\begin{example}
\label{ex:PTE}\normalfont
It is easy to see that the Lorentz norm $\|\cdot\|_{\Lambda_\phi}$ is PTE for any $\phi \in \Phi$. In contrast, the Marcinkiewicz norm $\|\cdot\|_{M_\phi}$ is PTE if and only if $\phi(t)=\min\left\{t/(1-\alpha),1\right\}$ for some $\alpha \in [0,1)$ or for $\alpha \rightarrow 1$, $\phi(t)  = \chi_{(0,1]}$ (Theorem~\ref{theorem_mlcvar}).
\end{example}

As an example of the above, consider the function norm $R^1$, which can be written as
\begin{equation}
\label{eq:l1nonneg}
    R^1(X) = \sup\left\{\int_0^1 X^*(\omega) Y^*(\omega) \d \omega : R^\infty(Y) \leq 1, Y \in \mathcal{M}^+ \right \} \quad \forall X \in \mathcal{M}^+.
\end{equation}
For nonnegative functions $X \in \mathcal{M}^+$, this is the expectation $\mathbb{E}[X]$. However, if we we want to extend the above definition to work on general $X \in \mathcal{M}$, we must use its positive translation equivariant representation
\begin{equation}
    R^1(X) = \sup\left\{\int_0^1 X^*(\omega) Y^*(\omega) \d \omega : \mathbb{E}[Y] = R^\infty(Y) = 1, Y \in \mathcal{M}^+ \right\} \quad \forall X \in \mathcal{M}^+.
\end{equation}
In fact, the singleton $\{1\}$ is sufficient to represent this function norm.
The two representations are equivalent for nonnegative $X$, since in this case, the supremum will always be attained for the constant $Y=1$. The second representation, but not the first, can easily be extended to work on potentially negative $X \in \mathcal{M}$. We define the generalized distribution function $\mu_X^{-} : \mR \rightarrow [0,1]$ and the generalized decreasing rearrangement $X^{*-} : [0,1] \rightarrow \mR$ as (\cf \eqref{eq:dist}, \eqref{eq:decr})
\begin{align}
    \mu_X^-(\lambda) &\coloneqq \mu\left\{\omega \in \Omega: X(\omega) > \lambda\right\}\\
    X^{*-}(\omega) &\coloneqq \inf\{\lambda \in \mR : \mu_X^-(\lambda) \leq \omega \}.
\end{align}
Clearly, $X^{*-}(\omega)=F_X^{-1}(1-\omega)$.
\begin{theorem}
Let $R$ be an ri function norm which has a positive translation equivariant representation. Then the extended functional
\begin{equation}
\label{eq:extension}
    R^{-}(X) \coloneqq \sup\left\{\int_0^1 X^{*-}(\omega) Y^*(\omega) \d \omega : \mathbb{E}[Y] = R'(Y) = 1, Y \in \mathcal{M}^+ \right\} \quad \forall X \in \mathcal{M}
\end{equation}
is a coherent law invariant risk measure, which coincides with $R$ on nonnegative random variables and is translation equivariant (for all $c \in \mR$).
\end{theorem}
\begin{proof}
For $X \in \mathcal{M}^+$, the coincidence is obvious, since then $X^{*-}=X^*$. Law invariance is obvious since the definition is only in terms of the distribution of $X$. The other properties of a coherent risk measure are easily checked, but also follow from Kusuoka's theorem discussed in the next section.
\end{proof}
Therefore, there is no real loss in generality when restricting ourselves to nonnegative functions in the following discussion. All comparison results obtained for positive translation equivariant ri norms defined on nonnegative functions, which are essentially bounded from below, carry directly over to the extended functionals. Note, however, that for instance the Marcinkiewicz norm is not in general positive translation equivariant, hence cannot be extended to a translation equivariant functional. But see Theorem~\ref{theorem_tm} for the construction of a PTE norm related to the Marcinkiewicz norm.

\subsection{Kusuoka Representations}
\label{sec:rikusuoka}
The celebrated Kusuoka representation theorem \citep{kusuoka2001law} states that CVar's are the basic building blocks of any law invariant coherent risk measure. \citet{kusuoka2001law} proved the theorem on $\cL^\infty$ for law invariant coherent risk measures satisfying the Fatou property (akin to \ref{item:fatou}) and it has been subsequently extended to $\cL^p$ spaces \cite[see \eg][]{pflug2007modeling}. In general, Kusuoka representations require an \textit{atomless}\footnote{A set $B \subseteq \mathcal{F}$ on $(\Omega,\mathcal{F},P)$ is an \textit{atom} if $P(B)>0$ and $A \subsetneq B \Rightarrow P(A)=0$. A probability space is \textit{atomless} if it has no atoms.} probability space. We continue to work on the atomless standard probability space $[0,1]$ with the Lebesgue measure and restrict ourselves to the positive cone to draw the connection to ri function norms. 
\begin{theorem}
Every ri function norm which is $\PTE$ admits a representation on the form
\begin{equation}
\label{eq:kusuoka}
    R(X) = \sup_{\lambda \in \mathfrak{M}}\left\{\int_0^1 \cvar(X) \d \lambda(\alpha) \right\} \quad \forall X \in \mathcal{M}^+
\end{equation}
for some set $\mathfrak{M}$ of probability measures on $[0,1]$. 
\end{theorem}
\begin{proof}
We observe that in the framework of ri spaces, the Kusuoka representation is a direct corollary of the representation of a norm via its associate norm:
\begin{equation}
\label{eq:arbitrarykusuoka}
    R(X) = \sup\left\{\int_0^1 X^*(\omega) Y^*(\omega) \d \omega : R'(Y) \leq 1, Y \in \mathcal{M}^+ \right\} \quad \forall X \in \mathcal{M}^+.
\end{equation}
This is a particular instantiation of the result in convex analysis that a norm is the support function of the unit ball of its dual norm. 
Here, the $Y^*$ are nonnegative and decreasing. If $R$ is a coherent risk measure, it is positive translation equivariant. Therefore it suffices to restrict ourselves to the subset of dual variables $\mathcal{Y}_{1} \coloneqq \{Y^* : R'(Y)=\mathbb{E}[Y]=1\}$ (Theorem~\ref{theorem:minrep}), that is $\int_0^1 Y^*(\omega) \d \omega = 1$. Then we can write this as ($Y \in \mathcal{M}^+$)
\begin{equation}
\label{eq:rikusuoka}
    R(X) = R_{\mathcal{Y}_1}(X) = \sup\left\{\int_0^1 F_X^{-1}(1-\omega) Y^*(\omega) \d \omega  : R'(Y) = \mathbb{E}[Y] = 1\right\} \quad \forall X \in \mathcal{M}^+.
\end{equation}
We recognize a supremum over a set of spectral risk measures, as each $w(\omega) \coloneqq Y^*(1-\omega)$ is a legitimate spectral weighting function (Section~\ref{sec:spectral}). Thus we can also write this as a supremum over distortion risk measures with concave distortions $\mathcal{Z} \coloneqq \{t \mapsto \int_0^t Y^*(\omega) \d \omega : Y \in \mathcal{Y}_1\}$. For each $Z \in \mathcal{Z}$, $Z : [0,1] \rightarrow [0,1]$, we have $Z(0)=0$, $Z(1)=1$ (due to PTE), since $Z(1)=\int_0^1 Y^*(\omega) \d \omega$. Therefore we obtain a representation as the supremum over Choquet integrals:
\begin{equation}
    R_{\mathcal{Y}_1}(X) = R_{\mathcal{Z}}(X) = \sup\left\{\int_0^\infty Z(\mu_X(\omega)) \d \omega : Z \in \mathcal{Z} \right\} \quad \forall X \in \mathcal{M}^+.
\end{equation}

We call either of the sets $\mathcal{Z}$ or $\mathcal{Y}_1$ a \textit{Kusuoka set} of $R$, since either fully characterizes the risk measure; in general, we notate dual variables as $Y$ and the integrals of their decreasing rearrangements as $Z$. In subsequent discussions, we shall also use the term ``Kusuoka set'' when PTE is not satisfied and the representation therefore describes a general ri function norm, without the constraint that $Z(1)=1 \Leftrightarrow \int_0^1 Y^*(\omega) \d \omega = 1$.
Finally, each spectral weighting function $w(\omega)=Y^*(1-\omega)$ can be associated with a probability measure $\lambda_w$ on $[0,1]$, via the relationship \citep[see \eg][]{pichler2015premiums}
\begin{equation}
    \lambda_w(E) \coloneqq w(0) \delta_0(E) + \int_E 1-\alpha \d w(\alpha) \quad E \text{ measurable},
\end{equation}
where $\delta_0$ is the Dirac measure at $0$. With this family of measures $\mathfrak{M} \coloneqq \{\lambda_w : w(\omega)=Y^*(1-\omega), Y^* \in \mathcal{Y}_1\}$, the representation \eqref{eq:kusuoka} is recovered. We remark that Kusuoka representations need not be unique in general.
Conversely, we can specify a functional $R_\mathcal{Y}$ directly as
\begin{equation}
    R_\mathcal{Y}(X) = \sup\left\{\int_0^1 X^*(\omega) Y^*(\omega) \d \omega : Y \in \mathcal{Y} \right\} \quad \forall X \in \mathcal{M}^+.
\end{equation}
For any set of nonnegative decreasing functions $\mathcal{Y}$, we are guaranteed to obtain a valid ri norm\footnote{The normalization $R_\mathcal{Y}(1)=1$ may not hold in general when no constraints on $\mathcal{Y}$ are imposed.}, since the supremum over a set of Lorentz norms preserves the relevant properties (Lemma~\ref{lemma:supoverrinorms}). However, $\mathcal{Y}$ need not be the maximal envelope.

When PTE is satisfied, the domain of $R$ can be extended from $\mathcal{M}^+$ to $\mathcal{M}$ to yield a coherent risk measure via the extension in Section~\ref{sec:te}, so that the original Kusuoka representation on the whole space is recovered. Observe that if only a single $Y^*$ suffices to represent the risk measure (of course, $Y^*$ which never obtain the supremum may be added to an envelope representation) and the normalization $\int_0^1 Y^*(\omega) \d \omega = 1$ holds, then \eqref{eq:rikusuoka} reduces to the definition of a Lorentz norm with $\phi(0+)=0$, \ie a spectral risk measure. This relates to the observation made by \cite{pichler2012uniqueness} that if the Kusuoka set is generated by a single element (modulo equimeasurability), then the risk measure is spectral. On the other hand, when starting with a single measure $\lambda$ on $[0,1]$, the situation is technically more subtle. If $\lambda$ does not have an atom at $1$, \ie $\lambda(\{1\})=0$, then it is equivalent to a concave distortion, which is continuous at $0$, and hence a spectral risk measure. Consider for instance the supremum risk measure $R^\infty$, represented by the Dirac measure at $1$. This cannot be expressed as a single distortion function, when demanding continuity at $0$. However, it can be represented as a supremum over a family of such functions.
\end{proof}
This way of arriving at the Kusuoka representation provides new insights as compared to standard proofs. First, it reveals that the natural space to work on with coherent risk measures is not in general an $\cL^p$ space, but rather a specific ri space. In the case of a spectral risk measure, this is a Lorentz space. A similar observation has been made by \cite{pichler2013natural}, but the author did not establish the link to the general theory of ri spaces. Moreover, it reveals that the Kusuoka representation is nothing more than the representation of a norm via its dual norm in the presence of the ri property. In general, a Banach function norm which is not ri can be written as
\begin{equation}
    R(X) = \sup\left\{\int_0^1 X(\omega) Y(\omega) \d \omega : R(Y) \leq 1 \right\} \quad \forall X \in \mathcal{M}^+.
\end{equation}
Under the ri property, $X$ and $Y$ can be replaced by any distributionally equivalent choice. The Hardy-Littlewood inequality \eqref{eq:hardy} tells us that the supremum is achieved for $X^*$ and $Y^*$. Another statement is in terms of \textit{Fréchet bounds}. Let $H$ be a bivariate distribution function with marginals $F$ and $G$, where ``distribution function'' is in the classical probabilistic sense. Then it holds (see \eg \citet[Section 1.2.2]{pflug2005measuring}):
\begin{equation}
    \max\{F(x) + G(y) - 1, 0\} \leq H(x,y) \leq \min\{F(x),G(y)\}, \quad \forall x,y \in \mR.
\end{equation}
Let $X$ have distribution $F$ and $Y$ have distribution $G$.
The lower bound is achieved when $X$ and $Y$ are \textit{antimonotone}; the upper bound is achieved when they are \textit{comonotone}.
Comonotonicity means that any of the following equivalent conditions hold:
\begin{enumerate}[label=\textbf{C\arabic*)}, ref=C\arabic*]
  \item $H(x,y) = \mu(X \leq x, Y \leq y) = \min(F(x),G(y))$
  \item $(X(\omega)-X(\omega'))(Y(\omega)-Y(\omega')) \geq 0 \quad \forall \omega,\omega' \in \Omega$
  \item $\exists Z \in \mathcal{M}$, non-decreasing functions $f,g$ such that $X=f(Z)$, $Y=g(Z))$.
\end{enumerate}
Comonotone $X$ and $Y$ have perfect rank correlation. The definition of antimonotonicity is the opposite, perfect negative rank correlation\footnote{Antimonotonicity means that $(X(\omega)-X(\omega'))(Y(\omega)-Y(\omega')) \leq 0 \quad \forall \omega,\omega' \in \Omega$.}. It is known that
\begin{equation}
    \mathbb{E}[XY] \leq \mathbb{E}[\tilde{X}\tilde{Y}],
\end{equation}
where $\tilde{X}$ and $\tilde{Y}$ are coupled in a comonotone way, but with the same marginals as $X$ and $Y$, respectively. Note that for any pair of random variables, $X^*$ and $Y^*$ are comonotone (C2 obviously holds). Therefore we recover Hardy-Littlewoods inequality \eqref{eq:hardy}.

The concept of comonotonicity is relevant both from a financial as well as from a purely uncertainty-motivated perspective. In finance, a desirable property for a risk measure is additivity for comonotone risks. That is, if $X$ and $Y$ are comonotone:
\begin{equation}
    R(X+Y)=R(X)+R(Y).
\end{equation}
The rationale is that in the presence of perfect rank correlation, $X$ cannot work as a hedge against $Y$ and vice versa. Essentially, comonotone gambles are bets on the same event in the world (due to C3). A decision maker under uncertainty may reason in an analogous way that when adding two comonotone uncertain losses, no hedge against uncertainty is possible. 

\begin{theorem}
\label{theorem:comon}
\citep{kusuoka2001law} on $\mathcal{L}^\infty$. A law invariant coherent risk measure with the Fatou property (akin to \ref{item:fatou}) is comonotonically additive if and only if it has a Kusuoka representation by a single probability measure on $[0,1]$.
\end{theorem}
That is, only spectral risk measures and its pathological neighbours (\eg the supremum risk measure) are comonotonically additive.

\subsection{Families of Fundamental Functions}
\label{sec:families}
The Kusuoka representation suggests a new way to characterize an ri norm fully by a family of fundamental functions. Recall that any ri norm has the representation
\begin{equation}
    R(X) = \sup\left\{\int_0^1 X^*(\omega) Y(\omega) \d \omega : Y \in \mathcal{Y} \right\}, \quad \mathcal{Y} = \{Y^* : Y \in \mathcal{M}^+, R'(Y) \leq 1\}
\end{equation}
for a set $\mathcal{Y}$ of nonnegative decreasing functions $Y$. Let $E\subseteq \Omega$ be a measurable subset with $\mu(E)=t$. The fundamental function induced by $R$ is
\begin{equation}
    \phi(t) = R(\chi_E) = \sup\left\{\int_0^t Y(\omega) \d \omega : Y \in \mathcal{Y}\right\} = \sup_{Z \in \mathcal{Z}}{Z(t)},
\end{equation}
where $\mathcal{Z} = \{t \mapsto \int_0^t Y(\omega) \d \omega : Y \in \mathcal{Y}\}$. Depending on the context, we may call either $\mathcal{Y}$ or $\mathcal{Z}$ a Kusuoka set, as either fully describes the representation\footnote{We typically use $\mathcal{Y}$ as the primary objects, as they appear directly in the ri space version of the Kusuoka representation. When starting with $\mathcal{Z}$, we need that their derivatives be defined almost everywhere. If $R$ is PTE, then the set of probability measures $\mathfrak{M}$ offers yet another representation.}. The fundamental function $\phi \in \Phi$ can be expressed as a supremum of concave distortions $Z$, each of which can be seen as the fundamental function of a spectral risk measure. Conceptually, we may say that ambiguity about the risk aversion spectrum exhausts the whole space of coherent risk measures. From this angle, it is possible to derive intuitive and instructive proofs, for instance for the extremal status of the Marcinkiewicz and the Lorentz norm. Furthermore, we construct the smallest translation equivariant norm, given an arbitrary concave fundamental function. First, we need a technical lemma, which we specialize to the domain $(0,1]$.
\begin{lemma}
    Hardy's lemma \citep{Bennett:1988aa}. Let $Y_1$ and $Y_2$ be nonnegative measurable functions on $(0,1]$ and 
\begin{equation}
    \int_0^t Y_1(\omega) \d \omega \leq \int_0^t Y_2(\omega) \d \omega \quad \forall t\in(0,1] .
\end{equation}
If $\eta$ is any nonnegative decreasing function on $(0,1]$, then
\begin{equation}
    \int_0^1 \eta(\omega) Y_1(\omega) \d \omega \leq  \int_0^1 \eta(\omega) Y_2(\omega) \d \omega.
\end{equation}
\end{lemma}
As an example, in our context this implies that if a concave distortion $Z_1(t) = \int_0^t Y_1(\omega) \d \omega$ majorizes another $Z_2$ pointwise ($Z_1,Z_2 \in \Phi_{0+}$), then the Lorentz norm corresponding to $Z_1$ majorizes the one corresponding to $Z_2$ for all random variables (set $\eta = X^*$). This does not apply to Lorentz norms, where the distortion is not continuous at $0$, however, as such a distortion cannot be represented by an integral of the form $Z_1(t) = \int_0^t Y_1(\omega) \d \omega$.

\begin{theorem}
\label{theorem:lorentzlargest}\citep[p.\@\xspace 72]{Bennett:1988aa}. 
    Given any ri norm $R$ with fundamental function $\phi \in \Phi_{0+}$, we have $R(X) \leq \|X\|_{\Lambda_\phi}$ $\forall X \in \mathcal{M}^+$.
\end{theorem}
\begin{proof}
Let $ R(X) = \sup_{Y \in \mathcal{Y}}\{\int_0^1 X^*(\omega) Y(\omega) \d \omega \}$.
    We know that $\phi(t)=\sup_{Y \in \mathcal{Y}}\{\int_0^t Y(\omega) \d \omega \}$. Recall that $\|X\|_{\Lambda_\phi} =  \int_0^1 X^*(\omega) \phi'(\omega) \d \omega$ if $\phi \in \Phi_{0+}$. But since $\phi$ majorizes all elements of $\{t \mapsto \int_0^t Y(\omega) \d \omega : Y \in \mathcal{Y}\}$ pointwise and $\phi \in \Phi_{0+}$, we immediately obtain from the qualification in Hardy's lemma
    \begin{equation}
        \int_0^t \phi'(\omega) \d \omega \geq \int_0^t Y(\omega) \d \omega  \quad \forall 0<t\leq 1 \quad \forall Y \in \mathcal{Y}
    \end{equation}
    that it holds
    \begin{align}
        R(X) &= \sup\left\{\int_0^1 X^*(\omega) Y(\omega) \d \omega : Y \in \mathcal{Y}\right\}\\
        &\leq \sup\left\{\int_0^1 X^*(\omega) \phi'(\omega) \d \omega : Y \in \mathcal{Y}\right\}
        = \|X\|_{\Lambda_\phi}.
    \end{align}
The statement also holds true if $\phi \in \Phi \setminus \Phi_{0+}$ \citep{rubshtein2016foundations}.
\end{proof}
\begin{theorem}
\label{theorem_m}\citep[p.\@\xspace 70]{Bennett:1988aa}. 
Given any ri norm $R$ with fundamental function $\phi \in \Phi$, we have $\|X\|_{M_\phi} \leq R(X)$ $\forall X \in \cR$.
\end{theorem}
\begin{proof}
Write
\begin{equation}
     R(X) = \sup_{Z_\gamma \in \mathcal{Z}}\left\{\int_0^1 X^*(\omega) Z'_\gamma(\omega) \d \omega \right\}, \quad \phi(t) = \sup_{Z_\gamma \in \mathcal{Z}}{Z_\gamma(t)}
\end{equation}
for some Kusuoka set $\mathcal{Z}$ of $R$. Each $Z_\gamma$ is concave since it is the integral of a nonnegative decreasing function. Hence, $Z_\gamma$ pointwise majorizes all of the piecewise linear functions\footnote{Let $Z_\gamma: [0,1] \rightarrow \mR^+$ be concave, \ie $\forall \alpha \in [0,1]: Z_\gamma(\alpha t + (1-\alpha) x) \geq \alpha Z_\gamma(t) + (1-\alpha) Z_\gamma(x).$ Choosing $x=0$ yields $Z_\gamma(\alpha t) \geq \alpha Z_\gamma(t)$. For any $x \leq t$ hence $x=\alpha t$ for some $\alpha \in [0,1]$, we obtain $Z_\gamma(x) \geq \frac{x}{t} Z_\gamma{t}$. For $x > t$ the statement is obvious, as the concave $Z_i$ has nonnegative derivative, whereas $Z_{\gamma,t}$ has zero derivative.}
\begin{equation}
    \forall t \in (0,1]: Z_{\gamma,t}(x) \coloneqq 
    \begin{cases}
    Z_\gamma(t) \frac{x}{t} & ,\text{ } x \leq t\\
    Z_\gamma(t) & , \text{ }x > t.
    \end{cases}
\end{equation}
The $Z_{\gamma,t}$ are constructed as the integrals of the functions\footnote{The ``derivatives'' $Z_{\gamma,t}'$ are here the primary objects. The derivative of $Z_{\gamma,t}$ at a kink may not exist, but we have defined $Z_{\gamma,t}'$ as prior to $Z_{\gamma,t}$. This enables us to apply Hardy's lemma, which works with the integrals of $Z_{\gamma,t}'$ directly.}
\begin{equation}
    \forall t \in (0,1]:  Z'_{\gamma,t} \coloneqq
    \begin{cases}
    \frac{Z_\gamma(t)}{t} & , \text{ }x \leq t\\
    0 & , \text{ }x > t.
    \end{cases}
\end{equation}
Applying Hardy's lemma yields
\begin{align}
\label{eq:mnorm_smallest}
    R(X) &= \sup_{Z_\gamma \in \mathcal{Z}}\left\{\int_0^1 X^*(\omega) Z'_\gamma(\omega) \d \omega \right\}\\
    &\geq \sup_{Z_{\gamma,t}}\left\{\int_0^1 X^*(\omega) Z'_{\gamma,t}(\omega) \d \omega \right\}\\
    &= \sup_{Z_{\gamma,t}}\left\{\frac{Z_\gamma(t)}{t} \int_0^t X^*(\omega) \d \omega\right\}\\
    &\geq \sup_{0 < t \leq 1}\left\{
    \sup\left\{
    \frac{Z_\gamma(t)}{t} : Z_\gamma \in \mathcal{Z} \right\}
    \int_0^t X^*(\omega) \d \omega
    \right\}\\
    &=  \sup_{0 < t \leq 1}\left\{\frac{\phi(t)}{t} \int_0^t X^*(\omega) \d \omega\right\} = \|X\|_{M_\phi},\\
\end{align}
since we have $\phi(t') = \sup_{Z_\gamma}{Z_\gamma(t')} = \sup_{Z_{\gamma}}{Z_{\gamma}(t') \frac{t'}{t'}} = \sup_{Z_{\gamma}}{Z_{\gamma,t'}(t')}$.
\end{proof}
\begin{remark}\normalfont
Throughout the paper, we often use shorthand notation to avoid explicitly writing the set over which a supremum ranges, when it is clear from context. For instance, the notation $\sup_{Z_{\gamma,t}} Z_{\gamma,t}(\cdot)$ means $\sup\{Z_{\gamma,t}(\cdot) : Z_{\gamma} \in \mathcal{Z}, t \in (0,1]\}$. In general, if not stated otherwise explicitly, we always take the supremum over all the respective defined quantities.
\end{remark}
We constructed the functions $Z_{\gamma,t}$ so that they are linear up to $t$ and then constant. This yields the Marcinkiewicz norm. A slight extension, where the functions are piecewise linear and reach $Z_{\gamma,t}(1)=1$ yields the smallest positive translation equivariant ri norm.

\begin{theorem}
\label{theorem_tm}
Given any concave fundamental function $\phi \in \Phi$, we can construct the smallest positive translation equivariant ri norm as:
\begin{equation}
    \|X\|_{TM_\phi} = \sup_{0 < t < 1}\left\{\frac{\phi(t)}{t}\int_0^t X^*(\omega) \d \omega
    + \frac{\phi(t)-1}{t-1}\int_t^1 X^*(\omega) \d \omega\right\}.
\end{equation}
For any other $\PTE$ ri norm $R$ with fundamental function $\phi$ we have $\|X\|_{M_\phi} \leq \|X\|_{TM_\phi} \leq R(X)$ $\forall X \in \cR$. We call $TM_\phi$ the \textit{positive translation equivariant Marcinkiewicz norm}.
\begin{figure}
    \centering
    \begin{subfigure}[b]{0.32\textwidth}
    \includegraphics[width=\textwidth]{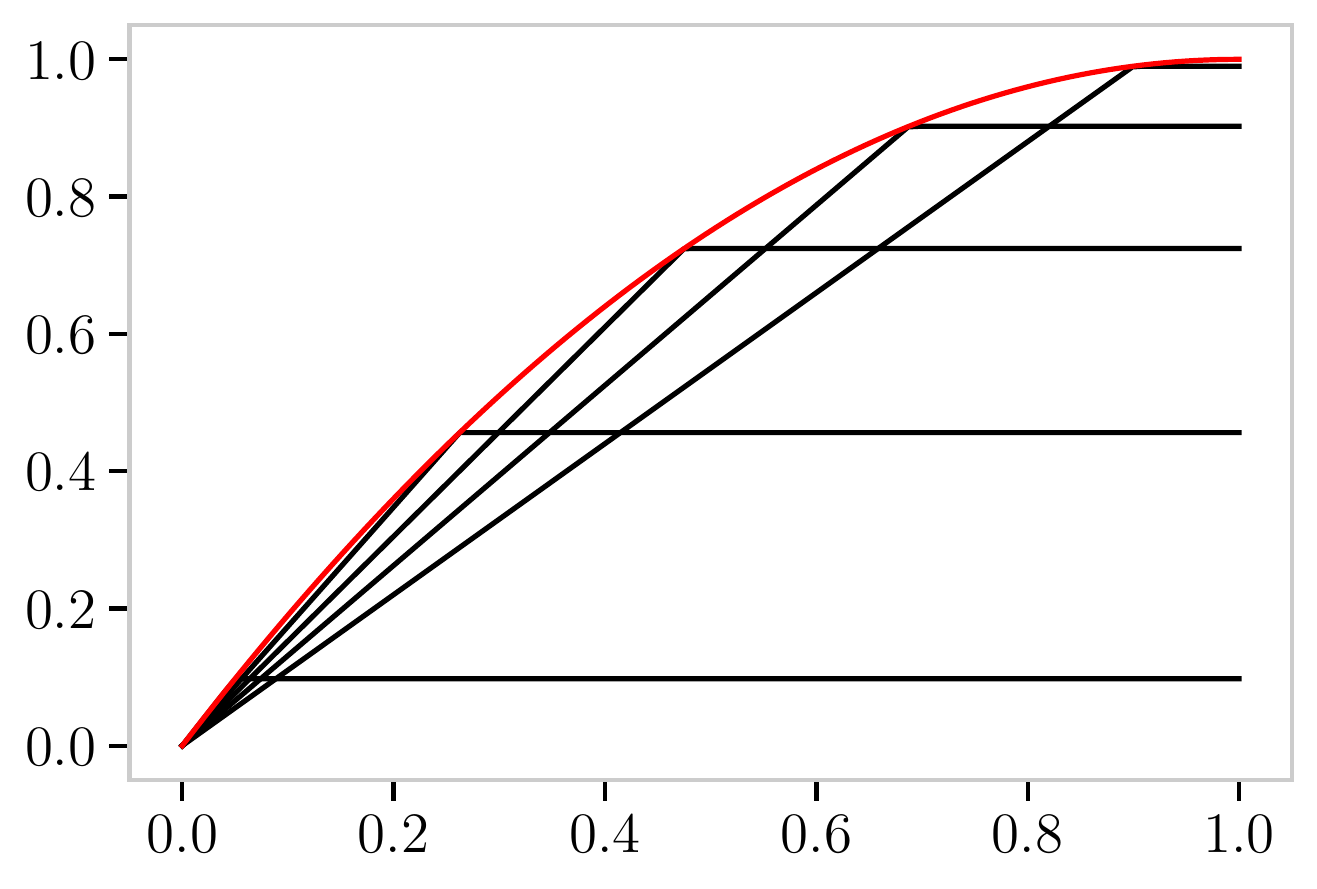}
    \end{subfigure}
    \begin{subfigure}[b]{0.32\textwidth}
    \includegraphics[width=\textwidth]{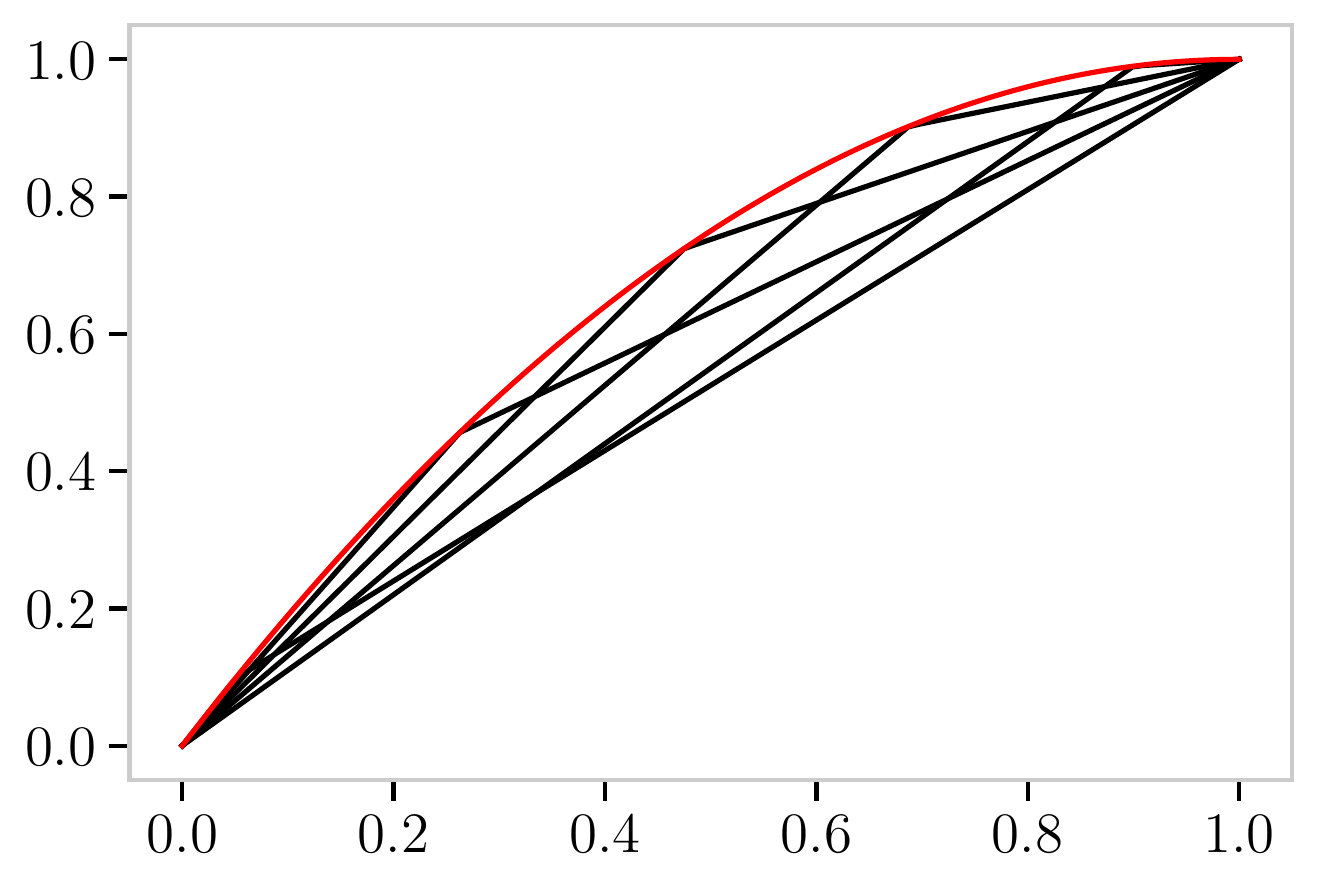}
    \end{subfigure}
    \caption{The red curve is the fundamental function $\phi(t)=1-(1-t)^2$. Left: the black lines correspond to five selected $\phi_t$ in the Marcinkiewicz norm construction. Right: the black lines correspond to five selected $\phi_t$ in the positive translation equivariant Marcinkiewicz norm construction. In this particular case, the latter yields the Dutch risk measure. Due to PTE, the $\phi_t$ need to reach $1$ at $t=1$. In both cases, the supremum over the (infinite) family of black lines recovers the red line, \ie the fundamental function $\phi$.}
    \label{fig:mtm_ffs}
\end{figure}
\end{theorem}
\begin{example}
\label{example:dutch}\normalfont
Recall that both the Dutch risk measure and the spectral MaxVar share the fundamental function $\phi(t)=2t-t^2$. Then: $\|X\|_{TM_\phi} = \du(|X|)$. This result implies that given this fundamental function, the Dutch risk measure is the most optimistic coherent risk measure, whereas MaxVar is the most pessimistic one.
\end{example}
The proof is in Appendix~\ref{app:tmnorm}.
Next, we show when equality of all ri norms for a given fundamental function holds.
\begin{theorem} 
\label{theorem_mlcvar}
Given any concave fundamental function $\phi \in \Phi$, it holds that 
\begin{equation}
\|X\|_{M_\phi} = R(X) = \cvar(X) = \|X\|_{\Lambda_\phi} \quad \forall X \in \mathcal{M}^+
\end{equation}for all ri function norms $R$ with fundamental function $\phi$ if and only if $\phi(t)=\min\left\{t/(1-\alpha),1\right\}$ for some $\alpha \in [0,1)$ or for $\alpha \rightarrow 1$, $\phi(t) = \phi_\infty(t) \coloneqq \chi_{(0,1]}(t)$. For $\alpha \rightarrow 1$, $R(X)=\|X\|_{\cL^\infty}$.
\end{theorem}

The proof is in Appendix~\ref{app:theorem_mlcvar}.
This result implies that for $\cvar$-type fundamental functions (including $\cL^1$ and $\cL^\infty$ as special cases\footnote{More precisely,  $\cL^1$ and $\cL^\infty$ are in fact the only two spaces for which the Marcinkiewicz and Lorentz norm coincide. This is due to the fact that for $\alpha \in [0,1)$, the space induced by $\cvar$ \textit{is} $\cL^1$, whereas $\alpha \rightarrow 1$ yields the $\cL^\infty$ space. See Section~\ref{sec:equivalences}.}), there is only a \textit{single} ri norm and hence a single law invariant coherent risk measure.
However, there is another interesting class of fundamental functions, for which all law invariant coherent risk measures coincide, but not all ri norms.
\begin{theorem}\label{theorem:rimwithdef}
Let $\phi(t) = \beta t + (1-\beta) \min(1,t/(1-\alpha))$ for any $\alpha,\beta \in [0,1)$. Then the Lorentz norm coincides with the positive translation equivariant Marcinkiewicz norm
\begin{equation}\|X\|_{\Lambda_\phi} = \|X\|_{TM_\phi} = \beta \mathbb{E}[X] + (1-\beta)\cvar(X) \eqqcolon \rim(X) \quad \forall X \in \mathcal{M}^+.
\end{equation}
\end{theorem}
\begin{proof}
The Lorentz norm is easily computed 
\begin{align}
    \|X\|_{\Lambda_\phi} &= \int_0^{1-\alpha} X^*(\omega) \left(\beta + (1-\beta) \frac{1}{1-\alpha}\right) \d \omega + \int_{1-\alpha}^1 X^*(\omega) \beta \d \omega \\
    &= \beta \mathbb{E}[|X|] + (1-\beta)\cvar(|X|).
\end{align}
A Kusuoka set of concave functions for the PTE Marcinkiewicz norm is
\begin{equation}
    \forall t \in (0,1): \phi_{TM,t}(x) \coloneqq 
    \begin{cases}
    \phi(t) \frac{x}{t} & , \text{ }x \leq t\\
    \frac{1-\phi(t)}{1-t} x + \frac{\phi(t)-t}{1-t} & ,\text{ } x > t
    \end{cases}
\end{equation}
Observe that $\phi$ is piecewise linear with a kink at $t=1-\alpha$, irrespective of the value of $\beta$, which adjusts the slope. Choose $t=1-\alpha$. Tedious calculation reveals what is obvious, that $\phi(x) = \phi_{TM,1-\alpha}(x)$. Therefore this $\phi_{TM,1-\alpha}$ dominates all other $\phi_{TM,t}$ and the supremum in the Kusuoka representation is in fact attained. But then $\|X\|_{TM_\phi} = \int_0^1 X^*(\omega) \phi_{TM,1-\alpha}'(\omega) \d \omega = \|X\|_{\Lambda_\phi}$.
\end{proof}
As a consequence, for this family of fundamental functions, the space of law invariant coherent risk measures collapses to a point. Moreover, the result is a spectral risk measure which is useful in practice as it can be easily computed. This function norm is called the \textit{risk measure for integrated risk measurement} \citep{pflug2005measuring}. The parameters $\alpha,\beta$ are intuitive knobs to adjust the tradeoff of tail-sensitivity (risk aversion) and globality (taking the full range of risk into account). Note also the close relation to the Dutch risk measure. From the representation
\begin{equation}
\du(X) = \sup_{0 < \beta < 1} \left\{\beta\mathbb{E}[X] + (1-\beta) \cdot \cvarnoa_{\beta}(X) \right\} \quad \forall X \in \mathcal{M}^+
\end{equation}
we observe that the Dutch risk measure can be seen as an ambiguity set over $\rim$s, where $\alpha=\beta$. This can be interpreted as a combination of risk and ambiguity aversion. To understand the relationship, consider the fundamental function $\phi(t)=2t-t^2$ of the Dutch risk measure and construct the corresponding $\phi_{TM,t}$. Each such $\phi_{TM,t}$ can be written as a $\phi_{TM,t}(x) = \beta x + (1-\beta) \min(1,x/(1-\alpha))$, where $t=1-\alpha=1-\beta$.
More generally, let $\phi(x)=1-(1-x)^n$ for some natural number $n \geq 2$. As $n$ increases, risk aversion increases. Then:
\begin{equation}
    \|X\|_{TM_\phi} = \sup_{0<\beta< 1} \beta^{n-1} \mathbb{E}[|X|] + (1-\beta^{n-1}) \cdot \cvarnoa_{\beta}(|X|) \quad \forall X \in \mathcal{M}.
\end{equation}
The corresponding Lorentz norm is \citep{cherny2009new}
\begin{equation}
    \|X\|_{\Lambda_\phi} = \mathbb{E}\left[\max(X_1,..,X_n)\right], \thinspace \thinspace X_1,..,X_n \stackrel{\text{ind}}{\sim} |X| \quad \forall X \in \mathcal{M}.
\end{equation} 

We can generalize this further to find that the PTE Marcinkiewicz norm has a family of $\rimnoa$s as its basic building blocks.

\begin{theorem}
Let $\phi \in \Phi_{0+}$ a fundamental function. Then
\begin{align}
    \|X\|_{TM_\phi} &= \sup_{0<t < 1} \frac{1-\phi(1-t)}{t} \mathbb{E}[|X|] +  \left(1-\frac{1-\phi(1-t)}{t}\right) \cdot \cvarnoa_{t}(|X|)\\
    &= \sup_{0<t < 1} \rimnoa_{\alpha(t),\beta(t)}(|X|), \quad \text{where } \alpha(t)=t, \beta(t) = \frac{1-\phi(1-t)}{t}.
\end{align}
\end{theorem}
\begin{proof}
Set $\phi_{TM,1-\alpha}(x) = \phi_{\rimnoa_{\alpha,\beta}}(x) = \beta x + (1-\beta) \min\{1, x/(1-\alpha)\}$. Both $\phi_{TM,1-\alpha}$ and  $\phi_{\rimnoa_{\alpha,\beta}}$ are piecewise linear with a kink at $1-\alpha$. A piecewise calculation shows that $\beta = \frac{1 - \phi(1-\alpha)}{\alpha}$ is a solution for both pieces. Then $\|X\|_{TM_\phi}$ has a Kusuoka representation in terms of spectral risk measures corresponding to the family of $\phi_{\rimnoa_{\alpha,\beta}}$, but these are just the $\rimnoa_{\alpha,\beta}$.
\end{proof}
We remark that $\rim$ admits the following variational representation:
\begin{equation}
    \rim(X) = \inf_{\mu \in \mR}{\mu + \mathbb{E} v(X-\mu)} \quad \forall X \in \mathcal{M},
\end{equation}
where the regret function $v$ is given by the piecewise linear function 
\begin{equation}
    v(t) = \begin{cases}
    \beta t & \text{ } t \leq 0\\
    \frac{\beta \alpha - 1}{\alpha-1} t &\text{ } t > 0.
    \end{cases}
\end{equation}
For the proof, see Appendix~\ref{app:rim}, which translates a result from \cite{pflug2005measuring}. Note that $V(X)=\mathbb{E}v(X)$ fulfills the requirements of a coherent regret measure in the quadrangle (Figure~\ref{fig:quadrangle}).

\subsection{Norm Equivalences and Tail Risk}
\label{sec:equivalences}
The fundamental function $\phi$, which corresponds to a coherent upper probability, imposes substantial structure on the compatible norms, which have this $\phi$ as their fundamental function. In this section, we expand on this claim by proving several (non)-equivalence results based on the derivative of $\phi$ at the origin. Recall that two norms $\ronenorm$ and $\rtwonorm$ are said to be equivalent if $\exists c_1, c_2 > 0: c_1 \ronenormy \leq \rtwonormy \leq c_2 \cdot \ronenormy$ $\forall X \in \cR_1 = \cR_2$. While a theoretical norm equivalence does not imply equivalence from a practical standpoint, it is nevertheless interesting how much of the norm behaviour is controlled by $\phi'(0)$ already. From our findings we conclude that the theoretically most essential differences between ri norms concern their behaviour with regard to tails of the random variables, an observation which we further develop in \citep{frohlich2023tailoring}.

We take inspiration from a result for coherent risk measures by \cite{pichler2013natural}.
Here we restate it in terms of ri norms.
\begin{theorem}
\label{theorem:equivalence}
Let $\ronenorm$ be an ri norm with Kusuoka set $\mathcal{Z}_1 = \{\phi_{1\gamma}\}$ and $\rtwonorm$ another ri norm with Kusuoka set $\mathcal{Z}_2 = \{\phi_{2\zeta}\}$, where $\gamma$ and $\zeta$ are from some arbitrary index sets. Denote the corresponding Banach spaces of functions, on which the norms are finite, as $\cR_1$ and $\cR_2$. Then if the constant
\begin{equation}
\label{eq:K}
    C \coloneqq \sup_{\phi_{2\zeta} \in \mathcal{Z}_2  } \inf_{\phi_{1\gamma} \in \mathcal{Z}_1} \sup_{0 <\alpha \leq 1} \frac{\phi_{2\zeta}(\alpha)}{\phi_{1\gamma}(\alpha)}
\end{equation}
is finite, we have the relationship
\begin{equation}
    \rtwonormy \leq C \cdot \ronenormy \quad \forall X \in \cR_1
\end{equation}
and $\cR_1 \subseteq \cR_2$, therefore $\cR_1 \hookrightarrow \cR_2$. If furthermore $\exists c>0: c \cdot \ronenormy \leq  \rtwonormy \forall Y \in \cR_2$, then $\cR_1 = \cR_2$ and we say that the norms $\ronenorm$ and $\rtwonorm$ are equivalent. 
\end{theorem}
The proof is in Appendix~\ref{app:theorem_equivalence}, where we also discuss a subtle issue with the original result.
In contrast, if $C=\infty$, we cannot make a statement for general ri norms (possibly, $\cR_1 \subseteq \cR_2$ or $\cR_1 \not\subseteq \cR_2$). At first sight one might conjecture that $C=\infty$ implies nonequivalence, but we provide a counterexample in Theorem~\ref{theorem:mtmareequiv}. However, we can state the following slightly refined result.

\begin{theorem}
Let the quantities $\ronenorm$, $\rtwonorm$, $\cR_1$, $\cR_2$, $\mathcal{Z}_1 = \{\phi_{1\gamma}\}$, $\mathcal{Z}_2 = \{\phi_{2\zeta}\}$,  be defined as in Theorem~\ref{theorem:equivalence}, so that $\phi_1(t) = \sup_{\phi_{1\gamma} \in \mathcal{Z}_1} \phi_{1\gamma}(t)$ and $\phi_2(t) = \sup_{\phi_{2\gamma} \in \mathcal{Z}_2} \phi_{2\gamma}(t)$ are the respective fundamental functions. If the constant
\begin{equation}
\label{eq:K-prime-prime}
    C' \coloneqq \sup_{\alpha \rightarrow 0} \sup_{\phi_{2\zeta} \in \mathcal{Z}_2  }  \inf_{\phi_{1\gamma} \in \mathcal{Z}_1}  \frac{\phi_{2\zeta}(\alpha)}{\phi_{1\gamma}(\alpha)}
\end{equation}
is infinite, then the norms are not equivalent; we have $\cR_1 \not\subseteq \cR_2$ and
\begin{equation}
\label{eq:eqfinitec}
    \nexists c: \rtwonormy \leq c \cdot \ronenormy \quad \forall X \in \cR_1 .
\end{equation}
\end{theorem}
\begin{proof}
Suppose $C'=\infty$. The norm of the identity embedding $\cR_1 \hookrightarrow \cR_2$ is
\begin{equation}
    \|\operatorname{id}\| = \sup\left\{\frac{\rtwonormy}{\ronenormy} : X \in \cR_1 \right\}.
\end{equation}
We restrict the supremum to measurable indicator functions and obtain:
\begin{equation}
    \|\operatorname{id}\| \geq \sup_{\chi_A }{\frac{\|\chi_A\|_{\cR_2}}{\|\chi_A\|_{\cR_1}}} = \sup_{\alpha \rightarrow 0} \sup_{\phi_{2\zeta} \in \mathcal{Z}_2  } \frac{\phi_{2\zeta}(\alpha)}{\sup_{\phi_{1\gamma} \in \mathcal{Z}_1} \phi_{1\gamma}(\alpha)} = \sup_{\alpha \rightarrow 0} \sup_{\phi_{2\zeta} \in \mathcal{Z}_2  }  \inf_{\phi_{1\gamma} \in \mathcal{Z}_1}  \frac{\phi_{2\zeta}(\alpha)}{\phi_{1\gamma}(\alpha)} = C' = \infty.
\end{equation}
Since $\|\operatorname{id}\|$ is unbounded, the norms are not equivalent.
\end{proof}
Note, however, that this criterion is not useful to test for non-equivalence of two norms with the same fundamental function, since in this case $C'=1$.

Throughout this section, we focus on those fundamental functions with $\phi(0+)=0$ since otherwise both the Marcinkiewicz $M_\phi$ and the Lorentz space $\Lambda_\phi$ are equal to $\cL^\infty$ \cite[p. 164]{rubshtein2016foundations}. 
We now give various characterization results in terms of $\phi'(0)$. To intuitively understand why this particular value is of interest, consider the Lorentz norm ($\phi \in \Phi_{0+})$:
\begin{equation}
    \|X\|_{\Lambda_\phi} =  \int_0^1 X^*(\omega) \phi'(\omega) \d \omega,
\end{equation}
and recall that these are the basic building blocks of any ri norm (Section~\ref{sec:rikusuoka}). Since $X^*$ are the backwards quantiles, $\phi'(0)$ is the highest weight which the most extreme loss receives. Due to concavity of $\phi$, its derivative $\phi'$ is nonnegative and decreasing. Risk measures fundamentally differ with respect to their tail behaviour: for instance, the expectation is maximally \textit{insensitive} to tails, as all quantiles receive constant weight $1$. On the other hand, for $\cvar$ we have $\phi'(0)=1/(1-\alpha)$, which for $\alpha \rightarrow 1$ may grow arbitrarily large. In general, the most benign situation occurs when $\phi'(0)$ is finite.

\begin{theorem}
\label{theorem:lmequiv}
Let $\phi \in \Phi_{0+}$. If the derivative of $\phi$ at $0$, \ie $\phi'(0)$, is finite, then there exists a constant $K$ such that
\begin{equation}
    \|X\|_{\Lambda_\phi} \leq K \cdot \|X\|_{M_\phi} \quad \forall X \in M_\phi.
\end{equation}
In view of the embedding theorem, we then have $\Lambda_\phi = M_\phi$, \ie equivalence of the Marcinkiewicz and the Lorentz norm. This implies in particular that given such a fundamental function, all law invariant coherent risk measures are equivalent. Moreover, a feasible constant is $K=1/(\phi(\frac{1}{\phi'(0)}))$.
\end{theorem}
The proof is in Appendix~\ref{app:normequivalences:lmequiv}.
Depending on the value of  $\phi'(0)$, the constant $K$ can be relatively small: as an example, for the fundamental function $\phi(t)=1-(1-t)^2$ of the Dutch risk measure and MaxVar, $\phi'(0)=2$, the constant is only $K=\frac{4}{3}$, implying that
\begin{equation}
    \|X\|_{M_\phi} \leq \du(|X|) \leq \maxv(|X|) \leq \frac{4}{3} \|X\|_{M_\phi} \quad \forall X \in M_\phi.
\end{equation}
The smallest $K$, however is achieved for $\cvar$: $K=1/\phi(\frac{1}{\phi'(0)})=1$ $\forall \alpha \in [0,1)$, a sanity check for Theorem~\ref{theorem_mlcvar}.

\begin{remark}\normalfont
Assume $R_1$ and $R_2$ are coherent risk measures. Then $R_1(X) \leq K \cdot R_2(X)$ $\forall X \in \mathcal{M}$ implies $K=1$ necessarily due to translation equivariance \citep{pichler2017quantitative}. Therefore, to obtain interesting and useful comparisons, we must restrict ourselves to the positive cone $\mathcal{M}^+$. Working with the norm $\|\cdot\|$ instead of the function norm has this effect.
\end{remark}

\begin{theorem}
\label{theorem:mlnotequiv}
Let $\phi \in \Phi_{0+}$ with $\phi'(0)=\infty$. Then the Marcinkiewicz and the Lorentz norm are not equivalent. We have
\begin{equation}
    \|X\|_{M_\phi} \leq \|X\|_{\Lambda_\phi} \forall X\in \Lambda_\phi \quad \text{ but } \nexists K: \|X\|_{\Lambda_\phi} \leq K \cdot \|X\|_{M_\phi} \forall X \in M_\phi.
\end{equation}
\end{theorem}
The proof is in Appendix~\ref{app:normequivalences:mlnotequiv}.

\begin{theorem}
\label{theorem:equivtol1}
Given any two ri norms $\ronenorm$,$\rtwonorm$ with possibly different fundamental functions $\phi_1,\phi_2 \in \Phi_{0+}$. If $\phi_1'(0)$ and $\phi_2'(0)$ are finite, then the norms are equivalent. In particular, all such norms are equivalent to the $\cL^1$ norm, \ie the expectation of a nonnegative random variable. 
\end{theorem}
The proof is in Appendix~\ref{app:normequivalences:equivtol1}.

\begin{corollary}
Given any two ri norms $\ronenorm$,$\rtwonorm$ with possibly different fundamental functions $\phi_1,\phi_2 \in \Phi_{0+}$. If $\phi_1'(0)=\infty$ but $\phi_2'(0)$ is finite, then they are not equivalent.
\end{corollary}
\begin{proof}
We use the following result from \cite[p. 164]{rubshtein2016foundations}: If $\phi_1'(0)=\infty$ then $M_{\phi_1} \subsetneq \cL^1$. On the other hand, we have shown before that all norms with $\phi_2'(0)$ finite are equivalent to $\cL^1$. Altogether, using the embedding theorem, we have
\begin{equation}
    \cR_1 \subseteq M_{\phi_1} \subsetneq \cL^1 = \cR_2.
\end{equation}
If the spaces do not coincide, the norms cannot be equivalent \cite[p.\@\xspace 7]{Bennett:1988aa}.

\end{proof}

\begin{theorem}
\label{theorem:mtmareequiv}
Given any $\phi \in \Phi_{0+}$ with $\phi'(0)=\infty$. Then the Marcinkiewicz norm $\|\cdot\|_{M_\phi}$ is equivalent to the positive translation equivariant Marcinkiewicz norm $\|\cdot\|_{TM_\phi}$, even though $C=\infty$.
\end{theorem}
The proof is in Appendix~\ref{app:normequivalences:mtmareequiv}. 

\begin{corollary}
Let $\phi \in \Phi_{0+}$ with $\phi'(0)=\infty$. In view of the embedding theorem, Theorem~\ref{theorem:mlnotequiv} and Theorem~\ref{theorem:mtmareequiv}, we have the embeddings:
\begin{equation}
     \Lambda_\phi \subsetneq TM_\phi = M_\phi 
\end{equation}
for the spaces induced by the Lorentz, $\PTE$ Marcinkiewicz and the Marcinkiewicz norm.
This means that if $\phi'(0)=\infty$ not all coherent risk measures are equivalent; however, the smallest ri norm is equivalent to the smallest coherent risk measure.
On the other hand, we have shown that if $\phi'(0)<\infty$, all of these are equivalent.
\end{corollary}

We have seen that $\phi'(0)$ plays an important role. If $\phi'(0)$ is bounded, such as for the fundamental function of the Dutch risk measure and MaxVar, the space of compatible law invariant coherent risk measures is very ``small'': from a theoretical perspective, they are all equivalent. We may summarize the role of $\phi'(0)$ by stating that \textit{it's all about the tails}. On a coarse level, $\phi(0+)$ controls how much weight is given to the most extreme event (the supremum), hence a risk measure with $\phi(0+)>0$ mimics the supremum (or if $\phi(0+)=1$, it \textit{is} the supremum). On a more fine grained level, $\phi'(0)$ is the weight that the extreme tails receive in the Lorentz norm. For an arbitrary ri norm, the interpretation of $\phi'(0)$ is more subtle due to the involved supremum in the Kusuoka representation.

\subsection{Rearrangement Invariant Norms and Risk}
We have seen that, on the positive cone, a law invariant coherent risk measure can be seen as a rearrangement invariant Banach function norm with the additional property of positive translation equivariance. Philosophically, this implies agreement about a base probability measure; however, decision makers may disagree about their risk aversion attitudes or they might want to introduce `hallucinated' ambiguity to account for a degree of distrust in the base measure. This is the specification of a fundamental function, a coherent upper probability. After a fundamental function is specified, there exist in general many different compatible norms. Among them, the Marcinkiewicz norm and the positive translation equivariant Marcinkiewicz norm are distinguished as the most optimistic extensions (subject to a constraint of requiring positive translation equivariance or not). Diametrically opposed, the Lorentz norms (spectral risk measures) are the most pessimistic extensions. In virtue of the Kusuoka representation, any other ri norm can be understood as being formed from an ambiguity set over spectral risk measures. Hence ambiguity about risk aversion exhausts the whole space of coherent risk measures. When there is no specific motivation for such a construction, the Lorentz norm is however the natural extension (indeed, also in Walley's terms) of the fundamental function: it is the only ri norm with a singleton Kusuoka set (if $\phi \in \Phi_{0+}$) and therefore fully described by its risk aversion profile.

We remark that ri function norms, which are not positive translation equivariant, are candidates for regret measures in the risk quadrangle \citep{rockafellar2013fundamental}. First, the ri function norm needs to be extended to the whole space, including potentially negative functions. Given an arbitrary ri norm $V$, a similar extension to \eqref{eq:extension} can be constructed:
\begin{equation}
    V^{-}(X) \coloneqq \sup\left\{\int_0^1 X^{*-}(\omega) Y^*(\omega) \d \omega : V'(Y) \leq 1, Y \in \mathcal{M}^+ \right\} \quad \forall X \in \mathcal{M}
\end{equation}
It is easy to check that this fulfills the desiderata of a coherent regret measure except perhaps aversity\footnote{However, a ``weak aversity'' condition $V^{-}(X) \geq \mathbb{E}[X]$ follows from the embedding theorem (essentially from law invariance). Note that we presupposed $R(1_\Omega)=1$ for any ri norm $R$. Many regret measures will not satisfy this. Monotonicity holds since $Y \in \mathcal{M}^+$, \cf Section~\ref{sec:envelopes}. Positive homogeneity and subadditivity are easily checked.}. The corresponding coherent risk measure can then be obtained by infimal convolution (Theorem~\ref{prop:infcon}). We believe that this opens up room for future research concerning risk-averse regression in the quadrangle, where the fundamental function offers fine control over the degree and shape of risk aversion.

\section{Creating New Risk Measures from Old}
\label{sec:combining}

In this section we investigate how one can combine several risk measures (ri function norms) to create new ones. 
Our motivation is two-fold. First, one can develop a better understanding 
of a ``thing'' by understanding the various  transformations of the thing --- 
a heuristic known as Grothendieck's relative method. 
We shall see, for example, that by 
considering the result of combining two risk measures reinforces the 
importance of the fundamental function. 
Second, risk measures can be not only used to encode risk aversion attitudes, 
but also fairness requirements 
\citep{williamson2019fairness}, and since  people will sometimes 
disagree on the right notion of fairness for a given situation 
(fairness being a prototypical example of an ``essentially contested concept'' 
\citep{gallie1955essentially}), a means is needed to reach a 
compromise between two distinct views on fairness, as codified by 
choices of risk measures.  The same argument applies sans fairness where two people
have different risk aversion attitudes.

\subsection{Properties of Quasiconcave Functions}

We will first present some elementary results concerning quasiconcave
functions and the effect of various combinations of ri Banach function
norms on the corresponding fundamental functions.
\begin{lemma}
\label{lemma:qcminmax}
	Suppose $\phi\colon\reals_{\ge 0}\rightarrow\reals_{\ge 0}$
	and $\phi(1)=1$.  If $\phi$ is quasiconcave then
	\begin{equation} \label{eq:bounds-on-quasiconcave-phi}
		\forall t\ge 0,\ \  \  \ 1\wedge t \le \phi(t) \le 1\vee t.
	\end{equation}
\end{lemma}
The proof is in Appendix~\ref{app:quasiconcave:qcminmax}.
\begin{lemma}\label{lemma:qc-dual}\citep[p.\@\xspace 127]{rubshtein2016foundations}.
The function $\phi\colon\reals_{\ge 0} \rightarrow\reals_{\ge0}$ is
quasiconcave if and only if $t\mapsto t/\phi(t)$ is quasiconcave.
\end{lemma}
Lemma \ref{lemma:qc-dual} has a natural
interpretation in terms of fundamental functions:
\begin{lemma}\label{lemma:ff-duals}\citep[p.\@\xspace 106]{Krein:1982aa}.
If $\phi$ is the fundamental function of an
ri space $\Xcal$, then $t\mapsto t/\phi(t)$ is the fundamental
function of the associate space $\Xcal'$.
\end{lemma}
Quasiconcavity is preserved under pointwise minima and maxima:
\begin{lemma}\label{lemma:qc-min-max}
	Suppose $\phi_i\colon\reals_{\ge 0}\rightarrow\reals_{\ge 0}$ are
	quasiconcave, $i\in[n]$.    Then $\bigwedge_{i\in[n]} \phi_i$ and
	$\bigvee_{i\in[n]}\phi_i$ are quasiconcave.
\end{lemma}
The proof is in Appendix~\ref{app:quasiconcave:qc-min-max}.
Lemma \ref{lemma:qc-min-max} suggests the question as to what other
combinations of quasiconcave functions are guaranteed to be quasiconcave.
We now show that quasiconcavity is preserved under a range of binary
operations induced by another quasiconcave function. 
\begin{definition}
	\label{def:perspective-Pcal}
	Suppose $\psi\colon\reals_{\ge 0}\rightarrow\reals_{\ge 0}$ is quasiconcave
	and $\psi(1)=1$. The \emph{perspective} of $\psi$ is the function
	\[
		\breve{\psi}\colon \reals_{\ge 0}\times\reals_{\ge 0}\ni
		(x,y)\mapsto y\psi(x/y).
	\]
	Let $\Pscr$ denote the set of functions  $\reals_{\ge 0}\times\reals_{\ge 0}
	\rightarrow\reals_{\ge 0}$ which are positively homogeneous,
	non-zero (except at $(0,0)$) and
	nondecreasing (in both arguments), and let $\Qscr$ denote the set
	of quasiconcave functions on $\reals_{\ge 0}$.
\end{definition}
\begin{lemma}
\label{lemma:perspective}
	Suppose $\psi\colon\reals_{\ge 0}\rightarrow\reals_{\ge 0}$.
	The perspective  $\breve{\psi}\in\Pscr$
if and only if $\psi\in\Qscr$.
\end{lemma}
The proof is in Appendix~\ref{app:quasiconcave:perspective}.
The following lemma shows that combining two quasiconcave functions using
$\breve{\psi}$ is guaranteed to result in a quasiconcave function, and that
this is the only way to ensure such a preservation of quasiconcavity.
\begin{lemma}
	\label{lemma:composition-of-quasiconcave}
	Suppose  $\phi_0,\phi_1,\psi\colon\reals_{\ge
	0}\rightarrow\reals_{\ge 0}$, 
	and let 
	$f_{\phi_0,\phi_1}(t)\coloneqq\breve{\psi}(\phi_0(t),\phi_1(t))$, $t\ge 0$. Then
	$[f_{\phi_0,\phi_1}\in\Qscr,\ \forall\phi_1,\phi_2\in\Qscr]$  
	if and only if $\psi\in\Qscr$.
\end{lemma}
The proof is in Appendix~\ref{app:quasiconcave:composition-of-quasiconcave}.
Observe that if $\psi(1)=1$, and $\phi_1=\phi_0$, then
\begin{equation}
	\label{eq:perpective-of-identical}
	\breve{\psi}(\phi_0(t),\phi_1(t))=\phi_0(t)\psi(\phi_0(t)/\phi_0(t))=\phi_0(t).
\end{equation}

\subsection{Interpolation Spaces}
The creation of new ri norms from given norms can be viewed as the
construction of an ``interpolation space'' \citep[pp.\@\xspace
99ff]{Bennett:1988aa}.   Given two ri spaces $\Xcal_0$ and
$\Xcal_1$, embedded in some separable linear topological space,
let $\Delta(\Xcal_0,\Xcal_1)\coloneqq \Xcal_0\cap \Xcal_1$
and $\Sigma(\Xcal_0,\Xcal_1)\coloneqq \Xcal_0+\Xcal_1$ with the corresponding norms
\begin{align}
	\|f\|_{\Delta(\Xcal_0, \Xcal_1)} & \coloneqq\|f\|_{\Xcal_0}\vee
	\|f\|_{\Xcal_1}\label{eq:Delta-def} \\
	\|f\|_{\Sigma(\Xcal_0,\Xcal_1)}& \coloneqq\inf\{\|f_0\|_{\Xcal_0}+\|f_1\|_{\Xcal_1}\colon f=f_0+f_1,\
	f_0\in \Xcal_0,\ f_1\in \Xcal_1\}.\label{eq:Sigma-def}
\end{align}

The spaces $\Delta(\Xcal_0,\Xcal_1)$ (resp.~$\Sigma(\Xcal_0,\Xcal_1)$) are the smallest
(resp.~largest) intermediate spaces between $\Xcal_0$ and $\Xcal_1$ in the sense
that any \emph{intermediate space} $\Xcal$ is continuously embedded  between them:
\[
	\Delta(\Xcal_0,\Xcal_1) \hookrightarrow \Xcal \hookrightarrow \Sigma(\Xcal_0,\Xcal_1).
\]
(This serves as a definition of intermediate space).
If $1$ is a feasible embedding constant, which we notate by $\embone$,  and 
which can always be ensured by simple
scaling, for any intermediate space $\Xcal$, for all $f\in \Xcal_0+\Xcal_1$,
\[
	\|f\|_{\Sigma(\Xcal_0,\Xcal_1)} \le \|f\|_\Xcal \le \|f\|_{\Delta(\Xcal_0, \Xcal_1)}.
\]

In order to appeal to results in the literature, we need to make some assumptions regarding 
the measure spaces $(\Omega,\mu)$ upon which our ri spaces are defined. We can restrict ourselves to finite measures spaces 
$\mu(\Omega)<\infty$, and in fact will 
assume $\mu(\Omega)=1$; for example, $\Omega=[0,1]$ with the Lebesgue measure as in Section~\ref{sec:banach}.  All of the results below then hold for any 
measure space that is purely non-atomic, 
or completely atomic which all atoms having equal measure.  
(This is a consequence of \cite[Theorem II.2.7]{Bennett:1988aa}.) Recall that we denote the associate space of $\mathcal{X}$ as $\mathcal{X}'$.
\begin{lemma}\label{lemma:dual-intersection}\citep[Theorem 2.7.1]{Bergh:1976aa}.
	Suppose $\Xcal_0$ and $\Xcal_1$ are ri spaces and
	$\Delta(\Xcal_0, \Xcal_1)$ is dense in both $\Xcal_0$ and $\Xcal_1$. Then
	$\Delta(\Xcal_0, \Xcal_1)'=\Sigma(\Xcal_0',\Xcal_1')$ and
	$\Sigma(\Xcal_0,\Xcal_1)'=\Delta(\Xcal_0', \Xcal_1')$.
\end{lemma}
We subsequently have the following analog of Lemma \ref{lemma:qc-min-max}
in terms of fundamental functions.
\begin{lemma}
	\label{lemma:ff-sum-intersect}
	Suppose $\Xcal_0$ and $\Xcal_1$ are ri spaces over a
	measure space $(\Omega,\mu)$ with
	corresponding fundamental functions $\phi_{\Xcal_0}$ and $\phi_{\Xcal_1}$.
	Then 
	$ \phi_{\Delta(\Xcal_0, \Xcal_1)}=\phi_{\Xcal_0}\vee \phi_{\Xcal_1}$
	and $\phi_{\Sigma(\Xcal_0,\Xcal_1)} =\phi_{\Xcal_0}\wedge\phi_{\Xcal_1}$.
\end{lemma}
\begin{proof}
	For $t\ge 0$, let $E_t\subset \Omega$ be such that $\mu(E_t)=t$.
	Equation \ref{eq:Delta-def} implies that for all $t\ge 0$,
	\[
		\phi_{\Delta(\Xcal_0, \Xcal_1)}(t)=\|\chi_{E_t}\|_{\Delta(\Xcal_0,
		\Xcal_1)} =
		\|\chi_{E_t}\|_{\Xcal_0}\vee\|\chi_{E_t}\|_{\Xcal_1} =
		\phi_{\Xcal_0}(t)\vee\phi_{\Xcal_1}(t).
	\]
	Lemmas \ref{lemma:ff-duals} and \ref{lemma:dual-intersection}
	together imply that for all $t\ge 0$,
	\begin{align}
		\phi_{\Sigma(\Xcal_0,\Xcal_1)}(t)&=\frac{t}{\phi_{\Sigma(\Xcal_0,\Xcal_1)'}(t)} =
		\frac{t}{\phi_{\Delta(\Xcal_0', \Xcal_1')}(t)} =
		\frac{t}{\phi_{\Xcal_0'}(t)\vee \phi_{\Xcal_1'}(t)}\\
		&=\frac{t}{\frac{t}{\phi_{\Xcal_0}(t)}\vee\frac{t}{\phi_{\Xcal_1}(t)}}
		= \phi_{\Xcal_0}(t)\wedge \phi_{\Xcal_1}(t).
	\end{align}
\end{proof}
Since $\breve{\psi}(x,1)=\psi(x)$, we see that the $\psi$ functions from
Lemma \ref{lemma:composition-of-quasiconcave} corresponding to max and min
are $\psi_{\mathrm{max}}(x)=\max(x,1)$ and
$\psi_{\mathrm{min}}(x)=\min(x,1)$, which are indeed both quasiconcave.

\subsection{Interpolation Functors and their Fundamental Functions}
We make use of a number of definitions and results of \cite{Brudnyi:1986aa}.  
Given the pair $(\Xcal_0,\Xcal_1)$ and an intermediate space $\Xcal$ for this pair, the
triple $((\Xcal_0,\Xcal_1);\Xcal)$ is called an \emph{interpolation triple}.
The triple $((\Xcal_0,\Xcal_1);\Xcal)$ is called an \emph{interpolation triple relative
to} the triple $((\Ycal_0,Y_1);\Ycal)$ if any bounded linear operator from the pair
$(\Xcal_0,\Xcal_1)$ to $(\Ycal_0,\Ycal_1)$ maps $\Xcal$ into $\Ycal$.  When that occurs, there
exists $c>0$ such that for any linear operator $T\in\Lscr(\Xcal,\Ycal)$,
$\|T\|_{\Xcal\rightarrow \Ycal} \le c\|T\|_{(\Xcal_0,\Xcal_1)\rightarrow (\Ycal_0,\Ycal_1)}$,
where the operator norm $\|T\|_{\Xcal\rightarrow \Ycal}= 
\sup\{\|T X\|_\Ycal\colon X\in\Xcal\mbox{\ and\ } \|X\|_\Xcal\le 1\}$.
If
$c\le 1$ then $((\Xcal_0,\Xcal_1);\Xcal)$ is called a \emph{normal interpolation triple
relative to the triple} $((\Ycal_0,\Ycal_1);\Ycal)$.  Let $\Bfrak$ denote the category
of Banach spaces and $\bar{\Bfrak}$ denote the category of Banach pairs (for what follows it suffices 
to just consider these as sets). 
\begin{definition}
	An \emph{interpolation functor} is a functor
	$\Fscr\colon\bar{\Bfrak}\rightarrow\Bfrak$  which assigns to each
	Banach pair $\bar{\Xcal} =(\Xcal_0,\Xcal_1)$ a Banach space $\Fscr(\bar{\Xcal})$
	intermediate between $\Xcal_0$ and $\Xcal_1$, and to each operator
	$T\in\Lcal(\bar{\Xcal},\bar{\Ycal})$ it assigns the restriction to the
	space $\Fscr(\bar{\Xcal})$.
\end{definition}

The triples $(\bar{\Xcal};\Fscr(\bar{\Xcal}))$ and $(\bar{\Ycal};\Fscr(\bar{\Ycal}))$ are
interpolation triples relative to each other.  If for any pairs $\bar{\Xcal}$
and $\bar{\Ycal}$ the resulting triples are normalised then $\Fscr$ is said to
be a \emph{normalised interpolation functor}.  The functors $\Delta$ and
$\Sigma$ introduced in (\ref{eq:Delta-def}) and (\ref{eq:Sigma-def}) are
both normalised interpolation functors.

For $\alpha>0$, the space $\alpha\reals$ is the set $\reals$ along with
norm given by $\|x\|_{\alpha \reals}=\alpha |x|$, for $x\in\reals$. Suppose
$\alpha,\beta>0$. Given an interpolation functor $\Fscr$, if we apply it to
the Banach pair $(\alpha\reals,\beta\reals)$ we obtain
$\Fscr(\alpha\reals,\beta\reals) =\phi_\Fscr(\alpha,\beta)\reals$, where
the constant $\phi_\Fscr(\alpha,\beta)$ is known as the \emph{fundamental
function of the functor $\Fscr$}~\citep{Brudnyi:1991aa}. (Sometimes
$\phi_\Fscr$ is called the \emph{characteristic  function of the functor}
$\Fscr$~\citep{Brudnyi:1986aa}, but such terminology conflicts with the
characteristic function $\chi_E$ of a set $E$ which we make considerable use of.) For
any functor $\Fscr$, $(\alpha,\beta)\mapsto\phi_\Fscr(\alpha,\beta)$ is
positive, positively homogeneous, and nondecreasing in each argument.  The
\emph{dual fundamental function of the functor $\Fscr$} is given by
$\phi_\Fscr^*(\alpha,\beta)=\frac{1}{\phi(1/\alpha,1/\beta)}$.
If $\Fscr$ is normalised, $\phi_\Fscr(1,1)=1$.
Normalised interpolation functors, when restricted to the Banach pair
$(\alpha\reals,\beta\reals)$ are characterised by their fundamental
function $\phi_\Fscr$.   

Given a Banach pair $\bar{\Xcal}=(\Xcal_0,\Xcal_1)$, the \emph{$K$-functional} is defined as
\[
	K(s_0,s_1,X,\bar{\Xcal}) \coloneqq \inf \{s_0\|X_0\|_{\Xcal_0}+s_1\|X_1\|_{\Xcal_1}\colon 
	     X_0\in \Xcal_0, X_1\in \Xcal_1 \mbox{\ s.t.\ } X=X_0+X_1 \}, \
	\ \ \ s_0,s_1\ge 0
\]
Pick an arbitrary function $\phi\in\Pscr$ (recall definition
\ref{def:perspective-Pcal}), and let $\Fscr$ denote the
interpolation functor on $(\alpha\reals,\beta\reals)$ with fundamental
function $\phi_\Fscr=\phi$. On one dimensional spaces, the interpolation functor 
is entirely determined by
its fundamental function; taking $\phi$ as given, then $\Fscr_\phi$ is given as
$\Fscr(\alpha\reals,\beta\reals)=\phi(\alpha,\beta)$.  If one defines an interpolation 
functor on one dimensional spaces, then it can be extended in many ways to 
arbitrary pairs of spaces. It turns out \citep[Section 1.16]{Brudnyi:1986aa} 
that there is a lower $\underline{\Fscr}$ 
and upper extension $\overline{\Fscr}$ such that for all pairs of Banach spaces 
$\overline{\Xcal}=(\Xcal_0,\Xcal_1)$ and all interpolation functors $\Fscr$,
\[
    \underline{\Fscr}(\overline{\Xcal}) \embone 
    \Fscr(\overline{\Xcal}) \embone \overline{\Fscr}(\overline{\Xcal}).
\]
The lower and upper extensions are characterised by the following
\citep[Section 1.17]{Brudnyi:1986aa}:
\begin{lemma}
	\label{lemma:characteristic-function-functor}
	Let $\bar{\Xcal}=(\Xcal_0,\Xcal_1)$ be an arbitrary Banach pair. The lower and upper extensions
	$\underline{\Fscr}(\bar{\Xcal})=\Lambda_\phi(\bar{\Xcal})$ 
	and $ \overline{\Fscr}(\bar{\Xcal})=M_\phi(\bar{\Xcal}))$ correspond to the
	space of all elements of $\Xcal_0+\Xcal_1$ with (respectively)  finite norms
	\begin{equation}
		\|X\|_{\Lambda_\phi(\bar{\Xcal})} \coloneqq \inf \sum_k
		\phi(\|X_k\|_{\Xcal_0}, \|X_k\|_{\Xcal_1}),
	\end{equation}
	where the infimum is taken over all representations of $X$ of the
	form $X=\sum_k X_k$, with $X_k\in \Xcal_0+\Xcal_1$ for all $k$; and
	\begin{equation}
		\|X\|_{M_\phi(\bar{\Xcal})} \coloneqq \sup_{s_0,s_1}
		\frac{K(s_0,s_1,X,\bar{\Xcal})}{\phi^*(s_0,s_1)}.
	\end{equation}
\end{lemma}
The spaces $\Lambda_\phi(\bar{\Xcal})$ and $M_\phi(\bar{\Xcal})$ are called the
\emph{abstract Lorentz space} and \emph{abstract Marcinkiewicz space}
respectively\footnote{
	That these define norms is obvious enough except perhaps for the
	convexity of $\|\cdot\|_{\Lambda_\phi(\bar{\Xcal})}$.   Since $\phi$ is
	 positively homogeneous and thus obviously 
	 $\|\cdot\|_{\Lambda_\phi(\bar{\Xcal})}$ is, it suffices to demonstrate
	subadditivity, namely that  (writing $Z=X+Y$),
	\begin{align*}
		& \|X\|_{\Lambda_{\phi}(\bar{\Xcal})}+
		\|Y\|_{\Lambda_{\phi}(\bar{\Xcal})} \ge 
		\|Z\|_{\Lambda_{\phi}(\bar{\Xcal})}\\
	\Leftrightarrow\ & \inf_{\sum_{k_1} X_{k_1}=X} \sum_{k_1}
		\phi(\|X_{k_1}\|_{\Xcal_0}, \|X_{k_1}\|_{\Xcal_1}) + 
		\inf_{\sum_{k_2} Y_{k_2}=Y} \sum_{k_2}
		\phi(\|Y_{k_2}\|_{\Xcal_0}, \|Y_{k_2}\|_{\Xcal_1})  
		\ge 
		\inf_{\sum_{k} Z_{k}=Z} \sum_{k}
		\phi(\|Z_{k}\|_{\Xcal_0}, \|Z_{k}\|_{\Xcal_1}) \\
	\Leftrightarrow\  &  
	\inf_{\substack{\sum_{k_1} X_{k_1}=X\\ \sum_{k_2} Y_{k_2}=Y}}
		 \left(\sum_{k_1} \phi(\|X_{k_1}\|_{\Xcal_0}, \|X_{k_1}\|_{\Xcal_1}) + 
			\sum_{k_2}
			\phi(\|Y_{k_2}\|_{\Xcal_0}, \|Y_{k_2}\|_{\Xcal_1})  \right)
		\ge \inf_{\sum_{k} Z_{k}=Z} \sum_{k}
		\phi(\|Z_{k}\|_{\Xcal_0}, \|Z_{k}\|_{\Xcal_1}) ,
	\end{align*}
	which holds since 
	 $\sum_{k_1} X_{k_1} +\sum_{k_2} Y_{k_2} = X+Y=Z$ and
	the infimum on the left is taken over a smaller set 
	since the $X_{k_1}$s have to sum to $X$ and separately the $Y_{k_2}$s
	have to sum to $Y$, but on the right this choice is also available plus additional 
	ones where no subset of the $Z_k$ are constrained to sum to $X$, and thus its infimum is less than 
	or equal to that on the left.
}, and the functors $\Lambda_\phi$ and $M_\phi$ are known as the
\emph{lower and upper extensions} of the functor $\Fscr_\phi$ defined on
$(\alpha\reals,\beta\reals)$ in terms of the fundamental function
$\phi\in\Pscr$
because they bound the behaviour of interpolation
functors with a given fundamental function:
\begin{lemma}
	\label{lemma:interp-functor-sandwich}\citep[Section 1.17]{Brudnyi:1986aa}.
	$\bar{\Xcal}=(\Xcal_0,\Xcal_1)$ be an arbitrary pair of Banach spaces.
	Let $\Xcal$ be a normal interpolation space between $\Xcal_0$ and $\Xcal_1$,
	and let $\Fscr$ be some normalised interpolation functor for which
	$\Fscr(\bar{\Xcal})=\Xcal$ with fundamental function $\phi$.  Then
	\begin{equation}
		\Lambda_\phi(\bar{\Xcal}) \embone \Xcal \embone M_\phi(\bar{\Xcal}),
	\end{equation}
	and thus
	\begin{equation}
		\label{eq:functor-interpolation-norms}
		\|X\|_{M_\phi(\bar{\Xcal})} \le \|X\|_\Xcal\le
		\|X\|_{\Lambda_\phi(\bar{\Xcal})}.
	\end{equation}
\end{lemma}

Recall that the Lorentz space $\Lambda_\phi$ is always positive translation equivariant (PTE) 
(Example \ref{ex:PTE}).
We might thus conjecture that the interpolation functor 
$\Lambda_\phi\colon \bar{\Bfrak}\rightarrow\Bfrak$ would preserve PTE; indeed that is the case as the lemma below shows.
\begin{lemma}
\label{lemma:ptefunctor}
    Suppose  $\Xcal_0,\Xcal_1$ are $\PTE$, and $\phi\in\Qscr$, then $\Lambda_\phi(\Xcal_0,\Xcal_1)$ is $\PTE$.
\end{lemma}

The proof is in Appendix~\ref{app:interpolation:lemma:ptefunctor}.
Lemma \ref{lemma:ff-sum-intersect} implies that
if $\phi$ is the fundamental
function of $\Xcal$, an intermediate space between $\Xcal_0$ and $\Xcal_1$, 
we have for all $t\ge 0$,
\begin{equation}
	\label{eq:phi-sandwiched-between-min-and-max}
	\phi_0(t)\wedge\phi_1(t) \le \phi(t)\le \phi_0(t)\vee\phi_1(t).
\end{equation}
Observe that if $\phi_0=\phi_1$, this means if $\Xcal$ is an intermediate space
between $\Xcal_0$ and $\Xcal_1$ with fundamental function $\phi$,  then
$\phi_0\le\phi\le\phi_0$ and hence
$\phi=\phi_0$.  
We formalise this observation as follows.
\begin{lemma}
	\label{lemma:ff-sandwich}
	Suppose $\bar{\Xcal}=(\Xcal_0,\Xcal_1)$ is an arbitrary pair of ri spaces, that $\Xcal_0$ and $\Xcal_1$ have the same 
	fundamental function $\phi$, and that 
	$\Xcal$ is an intermediate space of $\bar{\Xcal}$ with feasible embedding constant $1$: $\Xcal_0 \embone \Xcal \embone \Xcal_1$. Then $\phi_\Xcal=\phi$.
\end{lemma}
\begin{proof}
	The embedding ensures that for all $X\in \Xcal$, we have $\|X\|_{\Xcal_1}
	\le \|X\|_\Xcal\le \|X\|_{\Xcal_0}$ and thus choosing $X=\chi_{[0,t]}$ for
	some arbitrary $t>0$ we obtain
	\[
		\phi(t)=\|\chi_{[0,t]}\|_{\Xcal_1}\le
		\|\chi_{[0,t]}\|_\Xcal=\phi_\Xcal(t)\le \|\chi_{[0,t]}\|_{\Xcal_0}=\phi(t).
	\]
	Thus $\phi_\Xcal(t)=\phi(t)$ for all $t>0$.
\end{proof}
This illustrates the significance of the fundamental
function and justifies its name:  \textit{interpolation between two spaces with 
the same fundamental
function does not change the fundamental function}. Thus the fundamental
function provides a natural stratification of all possible ri spaces and their associated norms. 
When the fundamental functions of $\Xcal_0$ and $\Xcal_1$ differ,
Lemma \ref{lemma:ff-sum-intersect} shows a simple functional dependence of
the fundamental functions of $\Sigma(\Xcal_0,\Xcal_1)$  and $\Delta(\Xcal_0,\Xcal_1)$ on the
fundamental functions of $\Xcal_0$ and $\Xcal_1$.   We now develop a general
result along these lines that appears to be new.  We need some additional lemmas first.

\begin{lemma}
	\label{lemma:ff-abstract-marcinkiewicz}
	Suppose $\bar{\Xcal}=(\Xcal_0,\Xcal_1)$ is an arbitrary pair of ri spaces over a  measure space $(\Omega,\mu)$, 
	with corresponding fundamental functions $\phi_0$
	and $\phi_1$.   Let $\bar{\phi}\in\Pscr$. Then
	the fundamental function of $M_{\bar{\phi}}(\bar{\Xcal})$ satisfies 
	\[
		\phi_{M_{\bar{\phi}}(\bar{\Xcal})}(t)=\bar{\phi}(\phi_0(t),\phi_1(t)),\ \ \ \ t>0.
	\]
\end{lemma}
The proof is in Appendix~\ref{app:interpolation:lemma:ff-abstract-marcinkiewicz}.
We have an analogous (one-sided) result for $\Lambda_\phi(\bar{\Xcal})$:
\begin{lemma}
	\label{lemma:ff-abstract-lorentz}
	Suppose $\bar{\Xcal}=(\Xcal_0,\Xcal_1)$ is an arbitrary pair of ri spaces with corresponding fundamental functions $\phi_0$
	and $\phi_1$.   Let $\bar{\phi}\in\Pscr$. Then
	the fundamental function of $\Lambda_{\bar{\phi}}(\bar{\Xcal})$ satisfies 
	\[
		\phi_{\Lambda_{\bar{\phi}}(\bar{\Xcal})}(t)\le\bar{\phi}(\phi_0(t),\phi_1(t)),\ \ \ \ t>0.
	\]
\end{lemma}
The proof is in Appendix~\ref{app:interpolation:lemma:ff-abstract-lorentz}.
Lemmas \ref{lemma:characteristic-function-functor},
\ref{lemma:ff-abstract-marcinkiewicz}, and
\ref{lemma:ff-abstract-lorentz} 
combined with (\ref{eq:functor-interpolation-norms}) from
Lemma~\ref{lemma:interp-functor-sandwich} imply that for all $t>0$,
\begin{equation}
	\bar{\phi}(\phi_0(t),\phi_1(t))=\phi_{M_{\bar{\phi}}(\bar{\Xcal})}=
	\|\chi_{E_t}\|_{M_{\bar{\phi}}(\bar{\Xcal})} \le
	\|\chi_{E_t}\|_\Xcal=\phi_{\Lambda_{\bar{\phi}}(\bar{\Xcal})}(t)\le
	\|\chi_{E_t}\|_{\Lambda_{\bar{\phi}}(\bar{\Xcal})}\le \bar{\phi}(\phi_0(t),\phi_1(t)),
\end{equation}
and thus by Lemma~\ref{lemma:ff-sandwich}, we have for all $t>0$,
\begin{equation}
\label{eq:phi-Lambda-bar-X}
	\phi_{\Lambda_{\bar{\phi}}(\bar{\Xcal})}(t)=\bar{\phi}(\phi_0(t),\phi_1(t)).
\end{equation}
We have thus proved:
\begin{theorem}
	\label{theorem:interpolation-fundamental-functions}
	Suppose $(\Xcal_0,\Xcal_1)$ is an arbitrary pair of ri spaces and that the fundamental functions of $\Xcal_0$ and
    $\Xcal_1$ are $\phi_{0}$ and $\phi_{1}$ respectively. Suppose $\Fscr$ is a
	normalized interpolation functor  with fundamental function $\phi_\Fscr$. 
	Then $\Fscr(\Xcal_0,\Xcal_1)$ is an intermediate space for $(\Xcal_0,\Xcal_1)$ and its
	fundamental function  satisfies
	\begin{equation}
	\label{eq:fundamental-function-functors}
		\phi_{\Fscr(\Xcal_0,\Xcal_1)}
		=\phi_\Fscr(\phi_{0},\phi_{1}).
	\end{equation}
\end{theorem}
\subsection{Implications and Examples}
The ``fundamental function'' $\phi_\Xcal$ of an ri space
$\Xcal$  is justified in name from  a
mathematical viewpoint, as a risk aversion profile,  and from an ethical perspective. Mathematically, we have seen 
that combining two ri norms with the same
fundamental function will always result in another norm
with the same fundamental function. Thus the function $\phi_\Xcal$
really does pick out something
fundamental. From the risk aversion perspective, it captures 
the broad brush features of a decision maker's risk aversion.
From an ethical perspective, if we conceive of our function 
$X\colon\Omega\rightarrow\reals$ as representing some `bad' 
over a population of people $\Omega$, 
then the fundamental function $t\mapsto\phi(t)$ of a norm $\|\cdot\|$ 
captures a coarse aspect of the ethical implications of our choice: 
it tells us what value we ascribe to assigning  a bad of 1 to 
fraction $t$ of the population and a no bad to the rest. 
(Recall we are consistently adopting the loss perspective 
in this paper, where larger values are worse.)
The extreme cases of $\|\cdot\|_{\Lcal^1}$ and 
$\|\cdot\|_{\Lcal^\infty}$ then correspond to the ethical 
choices of John Harsanyi and John Rawls respectively; 
see \citep{williamson2019fairness} for an elaboration of this.

Since the choice of fundamental function is a personal 
choice (risk aversion, or ethical), different
designers will likely make different choices. This begs the
question of how a compromise between different choices can
be made. An obvious approach is to interpolate between the
two choices using an exact interpolation functor.  But
which functor?  The results and arguments above show that
the choice of functor has just as wide a scope as the
original choice of ri norm.  Essentially, the choice in
both cases is as large as the set of quasiconcave
functions. Thus there is no easy mechanical method to achieve 
a compromise between two distinct ethical positions (as 
encoded by two fundamental functions $\phi_1$ and $\phi_2$) 
because the result of interpolation between the two norms 
is dependent upon the choice of the interpolation functor, 
which is stratified by precisely the same class of functions 
as the original norms\footnote{
    The perspective developed here can be compared to that of Semmes, 
    who considered  geodesics in the Banach space of all Banach
    spaces \citep{Semmes:1988aa}  and
    argued that  we should not restrict our thinking about interpolation
    between normed spaces to the notion of interpolation of operators.
    In finite dimensional spaces, the question of interpolation can be
    posed in terms of constructing families of centre symmetric convex
    bodies (i.e. norm balls) ``in-between'' two given norm balls.  
    Similarly, the ``interpolation'' between two or more proper losses is essentially controlled by another proper loss \citep{williamson2023geometry}.
    }.

This perspective is further strengthened by noting the fact that 
\emph{every ri space is an exact interpolation space between 
$\Lcal^1$ and $\Lcal^\infty$} \cite[Theorem III.2.2]{Bennett:1988aa}.
Notwithstanding the previous somewhat negative conclusion,
the use of interpolation functors to create new ri norms
is valuable from a practical and computational perspective
in offering a wider choice of explicitly parametrised and
easily computable norms. Some examples  of special cases of
Theorem~\ref{theorem:interpolation-fundamental-functions} are given below.

\begin{example}\normalfont
	When $\phi_{\Xcal_0}=\phi_{\Xcal_1}$, Theorem 
		\ref{theorem:interpolation-fundamental-functions} implies 
		$\phi_\Xcal(t)=\phi(\phi_{\Xcal_1}(t),\phi_{\Xcal_1}(t))=
		\phi_{\Xcal_1}(t)$ using 
		(\ref{eq:perpective-of-identical})
		and the fact  that $\phi(1,1)=1$ since
		$\Fscr$ is a \emph{normalised}
		interpolation functor. This agrees with the
		observation made earlier following
		(\ref{eq:phi-sandwiched-between-min-and-max}).
\end{example}

\begin{example}\normalfont
	Consider two interpolation functors from Lemma 
		\ref{lemma:ff-sum-intersect}:
		$\Delta(\Xcal_0,\Xcal_1)=\Xcal_0\cap \Xcal_1$ and
		$\Sigma(\Xcal_0,\Xcal_1)=\Xcal_0+\Xcal_1$. For $\alpha,\beta>0$, we have 
		$ \|X\|_{\Delta(\alpha\reals,\beta\reals)}=
		\alpha|X|\vee\beta|X|=(\alpha\vee\beta)|X|$ and thus
		$\phi_{\Delta}(\alpha,\beta)=\alpha\vee\beta$.  
		Similarly, we have 
		$ \|X\|_{\Sigma(\alpha\reals,\beta\reals)}=
		\inf\{\alpha|X_0|+\beta|X_1|\colon X_0+X_1=X\}$.
		The infimum is achieved by choosing $X_0=X$ and $X_1=0$
		when $\alpha\le \beta$ and $X_0=0$ and $X_1=X$ when
		$\alpha\ge\beta$, in which case
		$\|X\|_{\Sigma(\alpha,\beta)}=(\alpha\wedge\beta)|X|$
		and thus $\phi_\Sigma(\alpha,\beta)=\alpha\wedge\beta$.
		Both cases correspond to the elementary results of 
		Lemma~\ref{lemma:ff-sum-intersect}.
\end{example}
    
\begin{example}\normalfont
    	Simple mean: $\|\cdot\|_{\Xcal_0\tilde{+} \Xcal_1}\coloneqq
		\frac{1}{2}(\|\cdot\|_{\Xcal_0}+\|\cdot\|_{\Xcal_1})$.   It is immediate
		that $\Xcal_0\tilde{+} \Xcal_1$ is rearrangement invariant if both $\Xcal_0$
		and $\Xcal_1$ are.
		We have 
		$ \|f\|_{\Xcal_0+\Xcal_1}\le \|f\|_{\Xcal_0\tilde{+}\Xcal_1} \le
			\|f\|_{\Xcal_0\cap \Xcal_1}, $
		where the first inequality follows from the fact that the formula for the
		norm $\|f\|_{\Xcal_0+\Xcal_1}$ takes the infimum over all additive
		decompositions $f=f_0+f_1$ but that for $\|f\|_{\Xcal_0\tilde{+} \Xcal_1}$
		chooses $f_1=f_2=f/2$.  The second inequality is a consequence of
		the mean of two numbers being no greater than their maximum.
		Thus $\Xcal_0\tilde{+} \Xcal_1$ is an intermediate space between $\Xcal_0$ and
		$\Xcal_1$.  It is immediate then that for all $t\ge 0$, we have
		\[
			\phi_{\Xcal_0\tilde{+} \Xcal_1}(t) =
			\frac{1}{2}\left(\phi_{\Xcal_0}(t)+\phi_{\Xcal_1}(t)\right).
		\]
\end{example}
\begin{example} \label{ex:simple-norm}\normalfont
	 More generally, let $\rho$ be a norm on $\reals^2$ normalised
		such that $\rho(1,1)=1$, and define
		$\|\cdot\|_{\rho(\Xcal_0,\Xcal_1)}\coloneqq
		\rho(\|\cdot\|_{\Xcal_0},\|\cdot\|_{\Xcal_1})$. Obviously
		$\rho(\Xcal_0,\Xcal_1)$ is rearrangement invariant if both $\Xcal_0$
		and $\Xcal_1$ are.    It is 
		standard that $\alpha+\beta\le\rho(\alpha,\beta)\le \alpha\vee \beta$ and it immediately
		follows from the definition that for all $t\ge 0$,
		\[
			\phi_{\rho(\Xcal_0,\Xcal_1)}(t)=\rho(\phi_0(t),\phi_1(t)).
		\]
\end{example}

\begin{example}\normalfont
    A special case of theorem \ref{theorem:interpolation-fundamental-functions}  
    (albeit stated for the interpolation of $N$ distinct ri spaces, 
    and not just 2)  is presented by \cite{Cobos:2017aa}, who considered a particular 
    family of interpolation functors $\Fscr^\theta$ ``of exponent $\theta$'', 
    $\theta\in(0,1)$.  They  showed that for all $t\ge 0$,
    \[
    	\phi_{\Fscr^{\theta}(\Xcal_0,\Xcal_1)}(t)
    	=\phi_{\Xcal_0}^{1-\theta}(t)\phi_{\Xcal_1}^\theta(t),
    \]
    which can be seen to be of the form of  (\ref{eq:fundamental-function-functors}).
    An analogous result is shown in
    \citep{Fernandez-Cabrera:2017aa} for a related but more
    complex interpolation method.  
    A related result (for ``envelopes,'' which are inversely related to 
    fundamental functions \cite[section 3.3]{haroske2006envelopes}) 
    was presented in  \cite{Haroske:2007aa}.
\end{example}

\begin{example} 
\label{ex:lorentz}\normalfont
    Consider two Lorentz norms $\Lambda_{\phi_0}$ and $\Lambda_{\phi_1}$
    and an arbitrary interpolation functor $\Fscr$. A natural question to ask is
    when (if ever) is $\Fscr(\Lambda_{\phi_0},\Lambda_{\phi_1})$ a Lorentz space $\Lambda_\phi$, 
    and when it is, is there some nice formula expressing $\phi=\psi(\phi_0,\phi_1)$?
    We conjecture that when $\Fscr$ corresponds to the abstract Lorentz norm 
    this is true.
    
    We do know that if $\Fscr=\Lambda_{\bar{\phi}}(\cdot,\cdot)$ then $\phi_\Fscr=\bar{\phi}$,
    since $\Lambda_{\bar{\phi}}$ is the extension of the functor defined on one 
    dimensional spaces with fundamental function $\bar{\phi}$. 
    We also know by (\ref{eq:phi-Lambda-bar-X}) that 
    $\phi_{\Lambda_{\bar{\phi}}(\Xcal_0,\Xcal_1)}=\bar{\phi}(\phi_0,\phi_1)$, 
    where $\phi_0=\phi_{\Xcal_0}$ and $\phi_1=\phi_{\Xcal_1}$. 
    By  \ref{theorem:sandwich}, we thus have for all $X$,
    \begin{equation}
    \|X\|_{\Lambda_{\bar{\phi}}(\Xcal_0,\Xcal_1)} \le \|X\|_{\Lambda_{\bar{\phi}(\phi_0,\phi_1)}}.
    \end{equation}
    We conjecture, but do not know, that the above inequality is in fact an equality.
    
    Regardless of whether our conjecture is true, it does suggest a 
    simple means of interpolating  between a pair of Lorentz spaces 
    (i.e.~spectral risk measures)  $\Lambda_{\phi_0}$ and 
    $\Lambda_{\phi_1}$ by choosing $\psi\in\Qscr$ 
    and letting $\bar{\phi}=\breve{\psi}$ and then constructing the space 
    $\Lambda_{\bar{\phi}(\phi_0,\phi_1)}$ --- all one needs to do 
    is to combine the fundamental functions $\phi_0$ and $\phi_1$ 
    via $\phi(t)=\bar{\phi}(\phi_0(t),\phi_1(t))$.
    In combining two spectral risk measures in this fashion, 
    one may wish to ensure that $\bar{\phi}$ is symmetric 
    (for equity reasons, so that the order in which the spaces are provided 
    will not affect the outcome). 
    Fortunately this has a simple characterisation. 
    In order that $\breve{\psi}(x,y)=\breve{\psi}(y,x)$ for all $x,y >0$, 
    it is necessary and sufficient that $\psi^\diamond(t)=\psi(t)$ for 
    all $t\in(0,1]$, where $\psi^\diamond(t)\coloneqq t\psi(1/t)$ is the 
    Csisz\'{a}r conjugate of $\psi$. 
    One can thus readily construct symmetric $\bar{\phi}=\breve{\psi}$ by 
    choosing an arbitrary quasiconcave function $\psi$ on $[0,1]$ and 
    extending it to $[1,\infty)$ via $\psi(t)=t\psi(1/t)$ for $t\ge 1$, 
    the resulting  ${\psi}$ is then guaranteed to satisfy the Csisz\'{a}r 
    conjugate condition and thus the induced perspective $\breve{\psi}$ is
    guaranteed symmetric.
    
    Observe that in contrast to the method in Example~\ref{ex:simple-norm},
    the present method enables the construction of an interpolated norm 
    $\|\cdot\|$ from $\|\cdot\|_{\Xcal_0}$ and $\|\cdot\|_{\Xcal_1}$  
    that can give finite values to $\|X\|$ even when one of the 
    values of $\|X\|_{\Xcal_0}$ or $\|X\|_{\Xcal_1}$ is infinite 
    (because of the tail behaviour of $X$).
    
    Interpolation of certain (classical) Lorentz spaces was considered 
    in \cite[Section\@\xspace 5]{cobos2005interpolation} and 
    an analogous question for Marcinkiewicz spaces   
    in \cite[Section\@\xspace 5]{Fernandez-Cabrera:2017aa}, however the 
    form of results is different to those which we sought here. 
\end{example}
    		
\begin{example}\normalfont
In order to provide some insight into (\ref{eq:fundamental-function-functors}),
especially for its use in Example \ref{ex:lorentz}, in Figure \ref{fig:qc-interp}
we illustrate the interpolation between two given fundamental functions, 
and show how the choice of the functor (in particular \emph{its}
fundamental function) affects the interpolation.
\begin{figure}[t]
\begin{center}
\includegraphics[width=10cm]{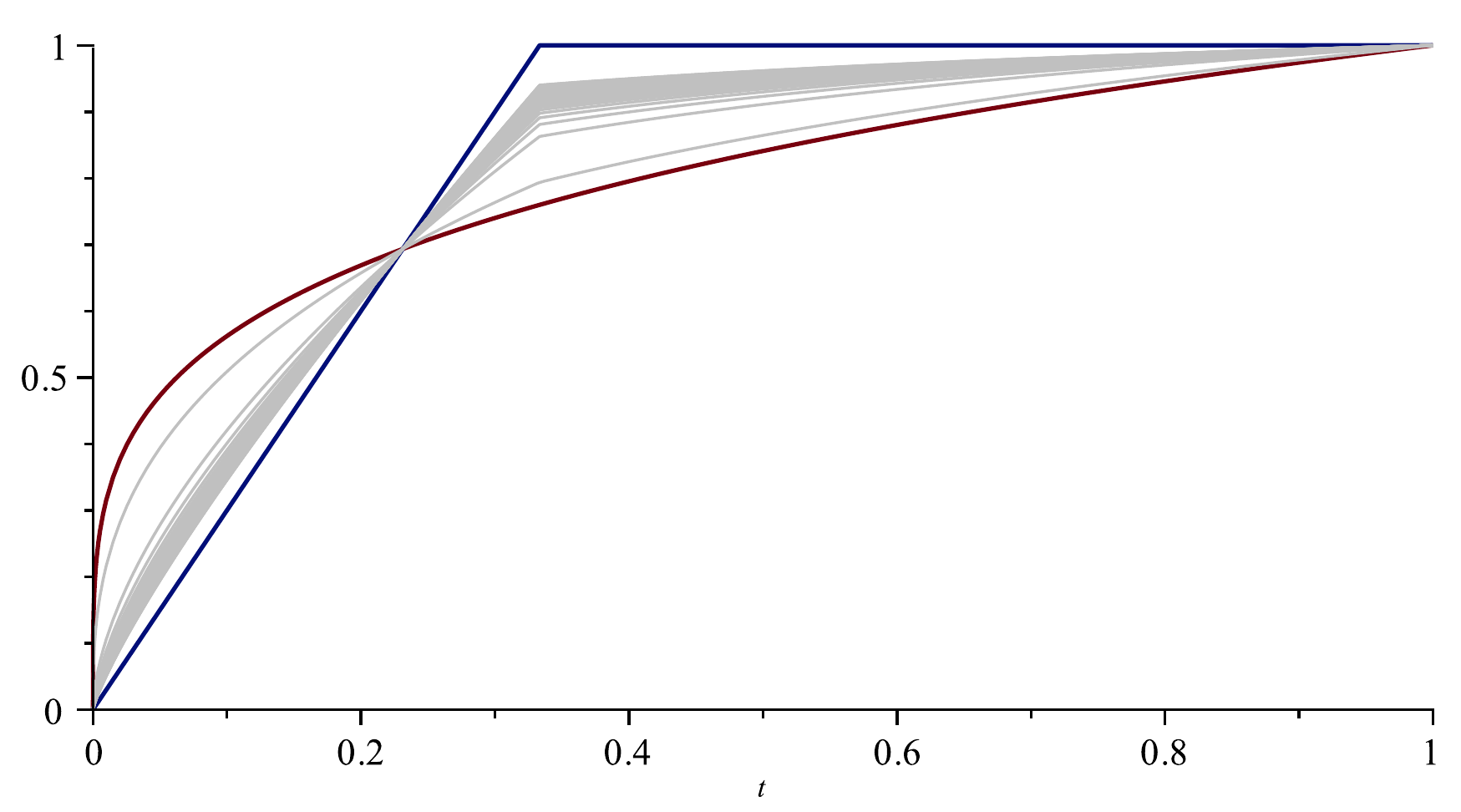}
\end{center}
\caption{
\label{fig:qc-interp}
Illustration of the interpolation between two 
    quasiconcave fundamental functions.
    The graph shows $\phi_{\mathrm{red}}(t)=t^{1/4}$ (in red) and 
    $\phi_{\mathrm{blue}}(t)= 3t\wedge 1$ (in blue).  
    The grey curves are obtained via 
    $\phi(t)=\breve{\phi}_a(\phi_{\mathrm{red}}(t),\phi_{\mathrm{blue}}(t))$ 
    where $\breve{\phi}_a$ is the perspective of 
    $\phi_a(t)=t^{1/a}$, with $a=\alpha^{1/4}$ and $\alpha$ ranges from 2 to 400 
    in steps of 10.  Small values of $a$ result in 
    $\phi$ being closer to $\phi_{\mathrm{red}}$ and larger 
    values result in  $\phi_a$ being closer to $\phi_{\mathrm{blue}}$.  
    Observe that  at the three points where $\phi_{\mathrm{red}}$ and 
    $\phi_{\mathrm{blue}}$ agree, so too does $\phi_a$.
}
\end{figure}

\end{example}

\section{Experiments}
\label{sec:experiments}
Coherent risk measures have already been successfully used in machine learning. For instance, \cite{williamson2019fairness} have employed them in a fairness context and demonstrated that using $\cvar$ on subgroup losses leads to them being more commensurate. In the context of machine learning, $\cvar$ has also been reinvented as ``average top-k loss'' \citep{fan2017learning}. \cite{curi2020adaptive} have proposed an adaptive sampling method for optimizing $\cvar$ in a batch setting. \cite{takeda2008nu} have established a close relation of $\cvar$ and the $\nu$-support vector machine. In reinforcement learning, coherent risk measures have been used \eg by \cite{singh2020improving,urpi2021risk,dabney2018implicit,tamar2015policy,vijayan2021}. Furthermore, distributionally robust optimization approaches based on $f$-divergence or Wasserstein ambiguity sets, which have been used extensively in machine learning, are subsumed in the framework of coherent risk measures \citep{rahimian2019distributionally}.

In our experiments, we aim to illustrate how spectral risk measures can lead to more robust solutions and attenuate inequality in the loss distribution. We focus on two kinds of problems: first, a coherent risk measure can act directly on the individual losses. We then have the risk minimization problem
\begin{equation}
    \argmin_f R(\ell(f(X),Y))
\end{equation}
for a risk measure $R$, a function $f$ from some hypothesis space, a loss function $\ell$, input $X$ and ground truth labels $Y$.
For its empirical counterpart, we use the empirical distribution of training losses. The risk measure $R$ then aggregates the observed losses, where each datum has a corresponding individual loss. Here, replacing the expectation $\mathbb{E}$ by a coherent risk measure $R$, in particular a spectral risk measure, has the effect of emphasizing large individual losses. As a consequence, a distribution of individual losses with less extreme losses (tail risk) will be preferred. There appears to be a fundamental trade-off between optimizing average loss versus reducing inequality. The precise nature of this trade-off is encoded in the choice of the fundamental function. In general, this setup is attractive in situations where relevant subgroups are not known or when a regulating agency disallows making decisions about people based on divisions into subgroups.

Second, we can apply coherent risk measures on subgroup losses. In a fairness context, we may wish to divide our data into ethically salient subgroups (\eg based on gender or race) and then ask for commensurate subgroup losses. In a technical context, for instance in multiclass classification, we may wish to achieve good performance not only on average, but also good performance for underrepresented classes in the training data. This is especially relevant if the distribution of the number of instances per class is heavy-tailed, as for example in natural species classification \citep{van2018inaturalist}.

Typically, the performance of a machine learning system is summarized by the average error on a test set. However, we think that this is a poor way of describing its performance, as it neglects the tail risk. In some settings, heavy-tailed risk must be avoided. This raises the question of a better performance representation which is sensitive to tail risks. We put forward two proposals.

\subsection{CVar Curves}
In light of the Kusuoka representation, the family $\cvar$ is the fundamental building block of all coherent risk measures. In a sense, $\cvar$ can be seen as measuring tail risks in purest form, as it merely integrates the $1-\alpha$ tail. We also have the property \citep[p. 61]{Bennett:1988aa} which is clear from the Kusuoka representation:
\begin{equation}
\label{eq:secondorderdominanceprop}
    \left(\forall \alpha \in [0,1): \cvar(X) \leq \cvar(Y)\right) \Rightarrow R(X) \leq R(Y)
\end{equation}
for \textit{any} ri function norm $R$ and $X,Y \in \mathcal{M}^+$.
Interestingly, the condition that $\forall \alpha \in [0,1): \cvar(X) \leq \cvar(Y) $ is equivalent to saying that $X$ is dominated by $Y$ in the \textit{second stochastic order} \citep{stochasticordercvar,bauerle2006stochastic}. Then \eqref{eq:secondorderdominanceprop} states that any ri function norm is consistent with the second stochastic order.\footnote{Note that the assumption that the measure space is  \textit{resonant} is crucial here \citep[p. 61]{Bennett:1988aa}; for a thorough discussion regarding atomic probability spaces see \citep{bauerle2006stochastic}.}

\begin{figure}
    \centering
    \begin{subfigure}[b]{0.37\textwidth}
    \includegraphics[width=\textwidth]{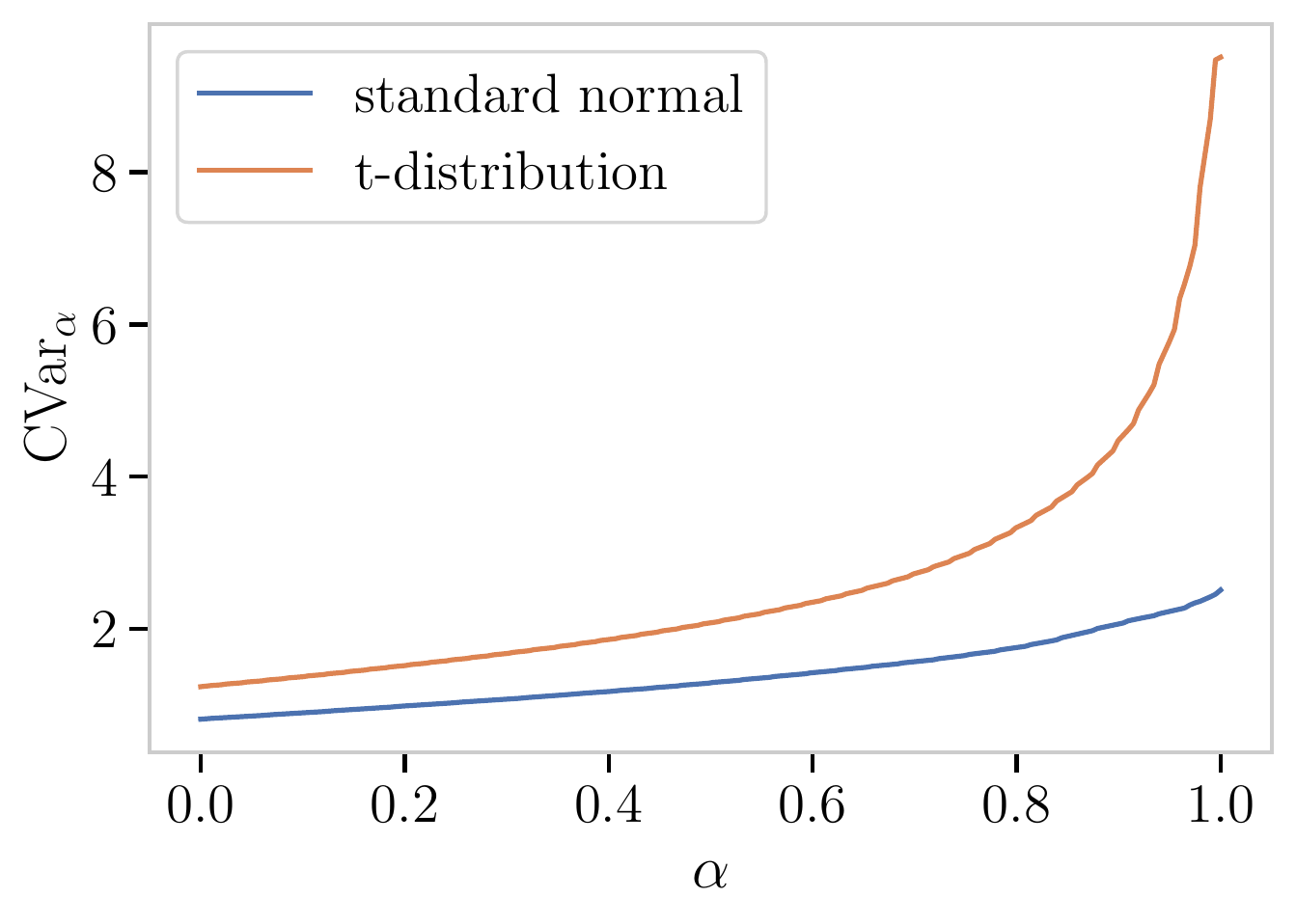}
    \end{subfigure}
    \begin{subfigure}[b]{0.37\textwidth}
    \includegraphics[width=\textwidth]{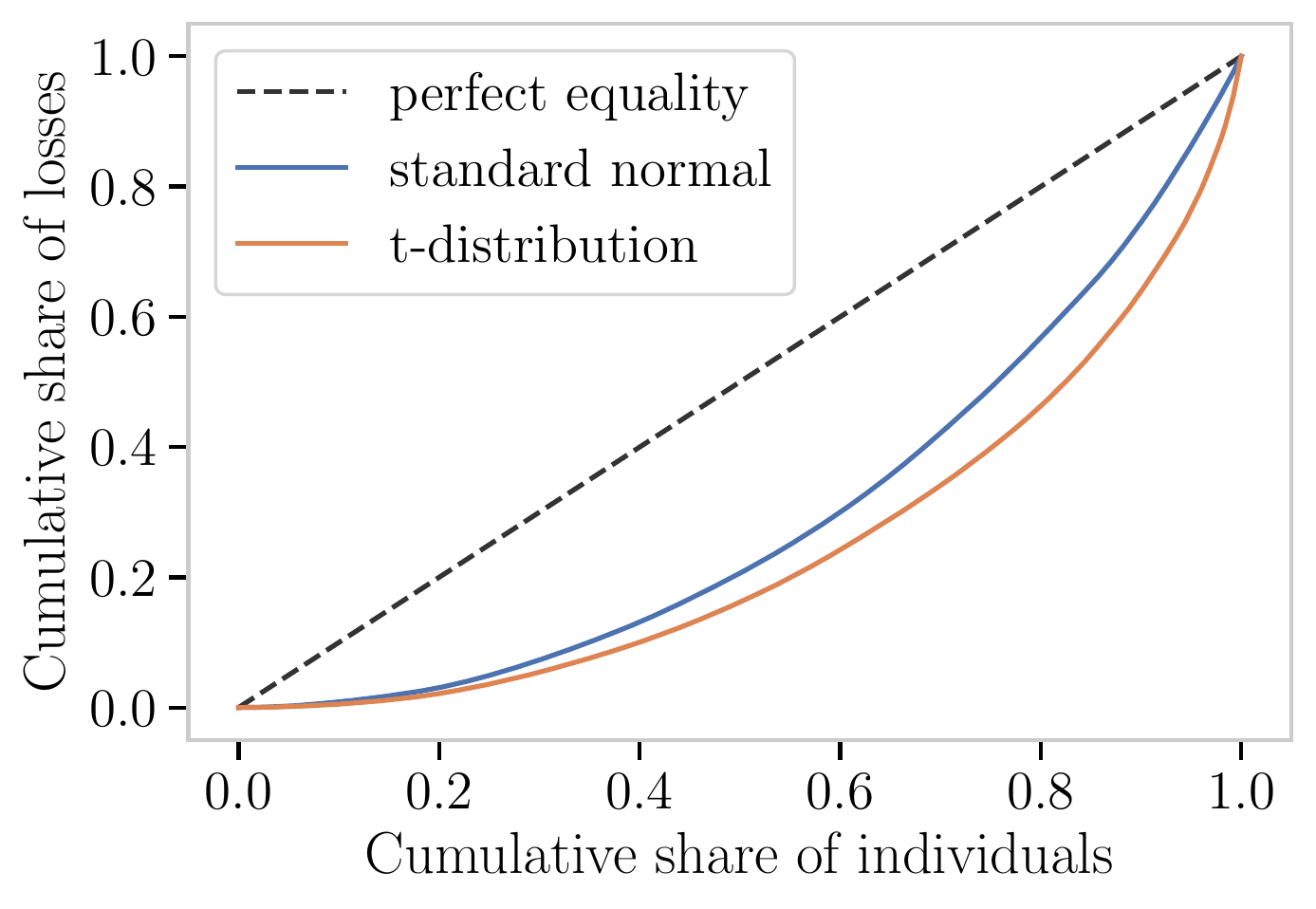}
    \end{subfigure}
    \caption{Left: $\cvar$ curves for $500$ randomly drawn samples from a standard normal and a $t$-distribution with $2$ degrees of freedom, respectively. Only nonnegative samples were kept. Both have approximately the same mean $CVar_{\alpha=0}$, but the $t$-distribution has substantially more weight in the tails. Right: empirical Lorenz curves for the same samples. Here, the curve of the standard normal is closer to the diagonal. The diagonal corresponds to perfect equality. The t-distribution exhibits higher inequality as compared to the standard normal. }
    \label{fig:norm_t_cvars}
\end{figure}

Due to this characterization, we propose to measure the performance by $\cvar$ loss curves.
The $x$-axis of this visualization corresponds to $\alpha \in [0,1)$ and the $y$-axis shows $\cvar(X)$ for some $X$. As an example, we draw $500$ samples from a standard normal distribution and a t-distribution with $2$ degrees of freedom. We keep only the nonnegative samples and interpret them as losses. The empirical $\cvar$ curves are shown in Figure~\ref{fig:norm_t_cvars}. Standard risk minimization takes only the value $CVar_{\alpha=0}$ into account, whereas we assert that the whole curve is relevant for measuring the performance.
In theory, the outcomes are unbounded for both distributions. In practice, however, we only ever observe finite values which enables plotting the $\cvar$ curves from samples. Another possibility is to introduce a cutoff value, so only values until for example $\alpha=0.99$ are plotted.

\subsection{Lorenz Curves}
A second possibility is to plot \textit{Lorenz curves}, which are widely used in economics to visualize inequality. The Lorenz curve of a random variable $X$ with quantiles $F_X^{-1}$ is defined as \citep{gastwirth1971general}:
\begin{equation}
    L_X(q) \coloneqq \frac{1}{\mathbb{E}[X]} \int_0^q F_X^{-1}(p) \d p, \quad 0 < q \leq 1; \quad L_X(0)=0.
\end{equation}
The empirical counterpart is defined in the obvious manner.
For perfect equality, \ie $X(\omega)=c$ $\forall \omega \in \Omega$, the Lorenz curve is the diagonal. Intuitively, $L(q)$ corresponds to the share of total loss which the individuals with the lowest $q$-percent of losses have. For an example, look at Figure~\ref{fig:norm_t_cvars}. At $q=0.9$, the empirical Lorenz curve of the $t$-distribution lies substantially below the curve of the standard normal. This means that the bottom $90\%$ of the $t$-distribution, \ie the individuals with the $90\%$ smallest losses, have a smaller share of the total loss than for the standard normal. This implies that for the $t$-distribution, a larger share of losses is in the $10\%$ tail. Hence the $t$-distribution has higher tail risk; given the same mean, a risk-averse decision maker would favor the standard normal.

One advantage of Lorenz curves is that even for unbounded random variables $X$, we can plot the full theoretical Lorenz curve, where $\cvar$ curves approach infinity. On the other hand, we find that in the context of losses, the interpretation is somewhat unintuitive. In economics, it is undesirable to belong to the bottom; in the context of losses, the bottom is constituted by the well-off. Another disadvantage is that absolute levels of loss are disregarded in this representation, whereas the $\cvar$ curves also give cardinal information. Therefore we find the $\cvar$ curves more useful to express the tradeoff between controlling average risk and tail risk.

Like the $\cvar$ curve, the Lorenz curve is also tightly linked to the second stochastic order \citep{muliere1989note}: given $X,Y\in \mathcal{M}$ with $\mathbb{E}[X]=\mathbb{E}[Y]$, it is easy to see that:
\[
\left(\forall \alpha \in [0,1): \cvar(X) \geq \cvar(Y) \right) \Leftrightarrow \left(\forall q \in (0,1]: L_X(q) \leq L_Y(q)\right).
\]
For a visualization, see Figure~\ref{fig:norm_t_cvars}. 
To concisely summarize performance across multiple runs of an experiment, we suggest using the \textit{Gini coefficient}, a single-number measure of inequality. The Gini coefficient of a Lorenz curve is defined as the ratio of the area between the diagonal (perfect equality) and the Lorenz curve over the total area under the diagonal. Therefore, if a distribution is perfectly equal, the Gini coefficient is $0$; for a perfectly unequal distribution, where a single $\omega$ receives the total loss, it is $1$.

\subsection{Spectral Risk Measures on Individual Losses}
Throughout, we focus on spectral risk measures in our experiments. First, we apply them to individual losses, \ie subgroups of size $1$. This setup is useful when we do not know relevant classes or care about individual loss in general. Consider, for instance, a self-driving car, which was mostly trained in snow-free environments. When the car is then deployed in a snowy environment, for which training data is scarce, its performance may be diminished. Since it can be a priori hard to partition data into fixed classes, we may employ a risk measure on the individual losses to account for risk aversion. Thus difficult training examples are emphasized. We illustrate this with a simple variant of principal component analysis (PCA), which could of course be replaced with a more sophisticated non-linear method. Since we focus on the risk measures, not the models themselves, we use simple methods for better interpretability of our results.

\subsubsection{Data}
\label{exp1:data}
We use the \texttt{MNIST}\footnote{http://yann.lecun.com/exdb/mnist/} and the \texttt{adult}\footnote{https://archive.ics.uci.edu/ml/datasets/Adult} data set. \texttt{MNIST} is a standard classification task. The problem is typically to classify grayscale images of handwritten digits, with $28 \times 28$ pixels, into the classes $0$-$9$. However, we view it as a dimensionality reduction task, where the goal is to compress the images. \texttt{MNIST} has $60{\small,}000$ training images and $10{\small,}000$ test images. 

The \texttt{adult} data set contains census data of $48{\small,}842$ persons with $14$ attributes and a binary target attribute, which specifies whether a person earns more or less than $50{\small,}000\$$ per year.
We disregard the binary target attribute and use the other $14$ attributes. 

To preprocess both \texttt{MNIST} and \texttt{adult}, we apply a \texttt{MinMaxScaler} with the feature range $[-1,1]$. We split the data into training and test sets. For \texttt{MNIST}, we use $6000$ training images and test on the remaining $54{\small,}000$ images. For \texttt{adult} we use $10{\small,}000$ data points as the training set and the remaining $38{\small,}842$ as the test set.

\subsubsection{Method}
Standard PCA can be solved using singular value decomposition. Recall that PCA minimizes the least squares reconstruction error. Assume our $n$ data points $x_i \in \mR^{m}$ with $m$ features are centered, \ie features have zero mean. Then the PCA objective is
\begin{equation}
    \min_{V_k} \sum_{i=1}^n \|x_i - V_k^\intercal V_k x_i\|_{\cL^2}^2 \quad \text{ so that } V_k V_k^\intercal = I_k
\end{equation}
where $V_k \in \mR^{k \times m}$ and $I_k$ is the identity matrix of size $k$. In contrast to the typical formulation in terms of an eigendecomposition or singular value decomposition, this formulation makes the individual losses explicit. Instead of only the expectation, we use different risk measures on the empirical distribution of reconstruction errors. Hence we obtain a variant of PCA (precisely, of a linear autoencoder) which is sensitive to large individual losses. We replace the objective by:
\begin{equation}
    \min_{V_k} R\left(\|X - V_k^\intercal V_k X\|_{\cL^2}^2\right)
\end{equation}
for a risk measure $R$ and where the distribution of the random loss variable follows the empirical distribution $\hat{P}_n$. 
We dropped the orthonormality constraint, which does not essentially alter the solution (\cf \cite{plaut2018principal}). We label this variant of PCA, where risk measures are employed, as PCA*.

We use the \texttt{pytorch} library to implement our experiments and the \texttt{Adam} optimizer with a learning rate of $0.001$. We initialize the matrix $V_k$ with the classical PCA solution, obtained from \texttt{sklearn.decomposition.PCA}. For \texttt{MNIST} we use $k=50$ components and for \texttt{adult} $k=5$. We train for $2000$ epochs with a learning rate of $0.001$. To avoid additional challenges from the stochasticity of mini-batches (see Section~\ref{sec:experiments_discussion}), we use the full data for each epoch. We repeat the experiments over $25$ independent runs with random training and test splits.

\subsubsection{Risk Measures}
\label{exp1:risks}
We compare the results when using $\cvar$, where $\alpha \in \{0.0,0.2,0.4,0.6,0.8\}$. Recall that $\cvarnoa_{0.0}=\mathbb{E}$. As $\alpha$ increases, sensitivity to tail risk increases. Moreover, we compare variations of the risk measure for integrated risk management ($\rim$), where $\alpha$ is fixed at $0.7$ and $\beta \in \{0.2,..,0.8,1.0\}$. Recall that
\begin{equation}
\rim(X) \coloneqq \beta \mathbb{E}[X] + (1-\beta)\cvar(X) 
\end{equation}
Hence $\beta=1$ yields the expectation, whereas $\beta=0$ would yield $\cvarnoa_{\alpha=0.7}$. As $\beta$ increases, sensitivity to tail risk decreases. In \texttt{pytorch}, this risk measure can be easily implemented by combining the \texttt{topk} and \texttt{mean} function, incurring virtually no computational overhead. 

\subsubsection{Results}
We here show $\cvar$ curves and Lorenz curves for the test losses on \texttt{MNIST} under the different risk measures. The curves are averaged over the independent runs. The results on \texttt{adult} are in Appendix~\ref{app:experiments}. In Appendix~\ref{app:experiments}, we also show boxplots of the Gini coefficients over the $25$ independent runs.
From the $\cvar$ curves, we observe that employing $\cvar$ or $\rim$ as a risk measure leads to lower tail risks as compared to the expectation. For moderate choices of $\alpha$ and $\beta$, we find that performance on average is hardly diminished, but there is substantial gain in tail performance. When choosing a high $\alpha$ such as $\alpha=0.8$, however, we clearly incur a cost in terms of average performance. From the Lorenz curves and Gini coefficient boxplots we observe that the expectation leads to the highest inequality of loss. As expected, increasing $\alpha$ for $\cvar$ and decreasing $\beta$ for $\rim$ (with fixed $\alpha$) gradually achieves a more equal distribution. This is most clearly visible in the Gini boxplots (Appendix~\ref{app:experiments}).

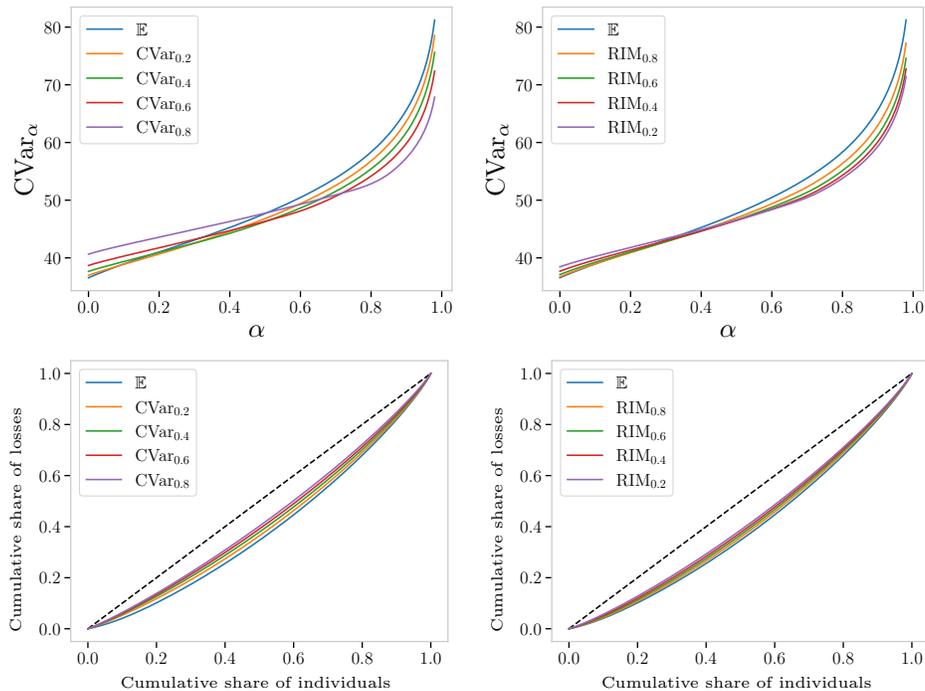
\begin{figure}
    \centering
    \begin{subfigure}[b]{0.37\textwidth}
        \begin{overpic}[width=\textwidth]{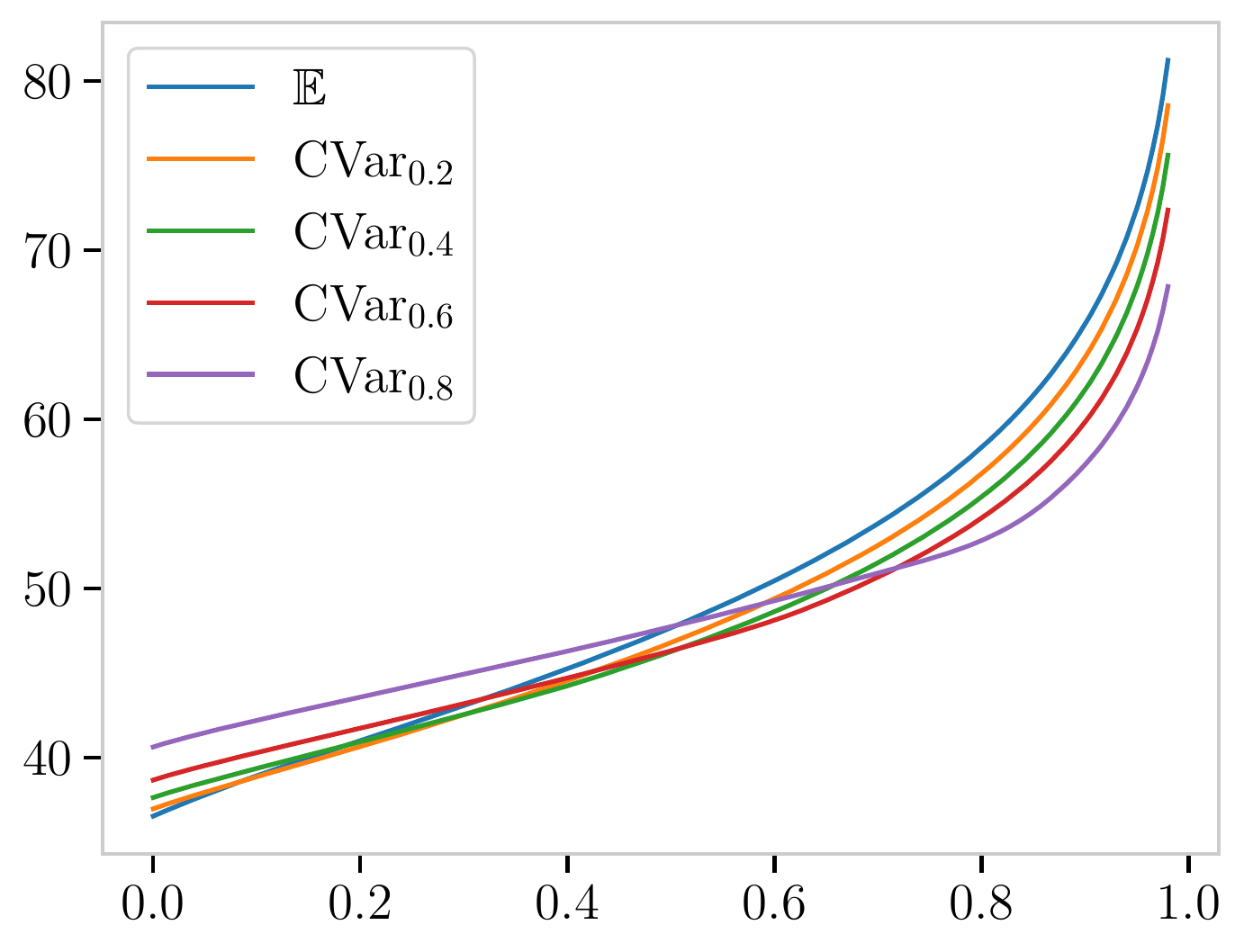}
            \put(50,-3){\small{$\alpha$}} 
            \put(-5,30){\rotatebox{90}{\small{$\cvar$}}} 
        \end{overpic}
    \end{subfigure}\hspace{0.5cm}
    \begin{subfigure}[b]{0.37\textwidth}
        \begin{overpic}[width=\textwidth]{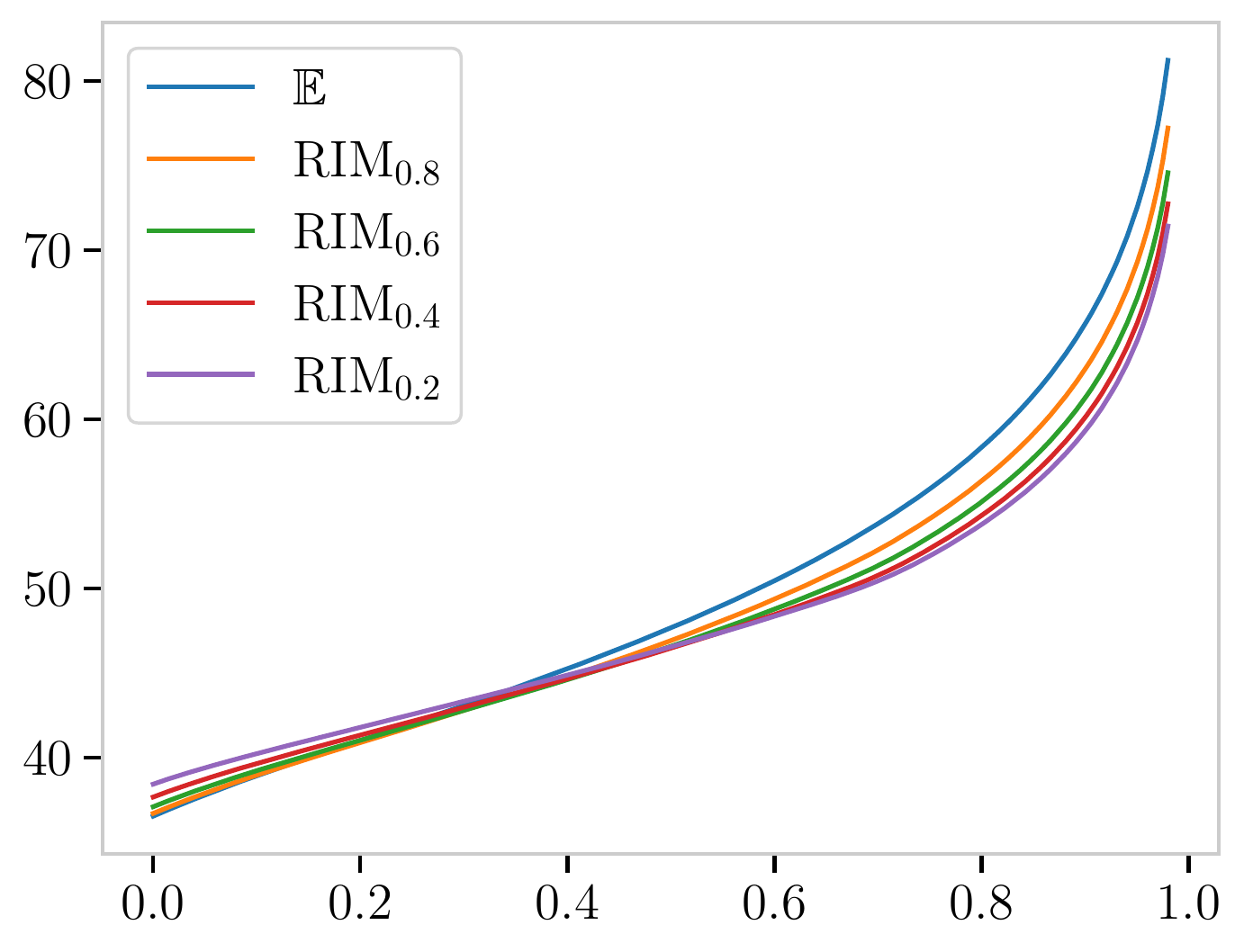}
            \put(50,-3){\small{$\alpha$}} 
            \put(-5,30){\rotatebox{90}{\small{$\cvar$}}} 
        \end{overpic}
    \end{subfigure}\\
    \vspace{0.4cm}
    \begin{subfigure}[b]{0.37\textwidth}
        \begin{overpic}[width=\textwidth]{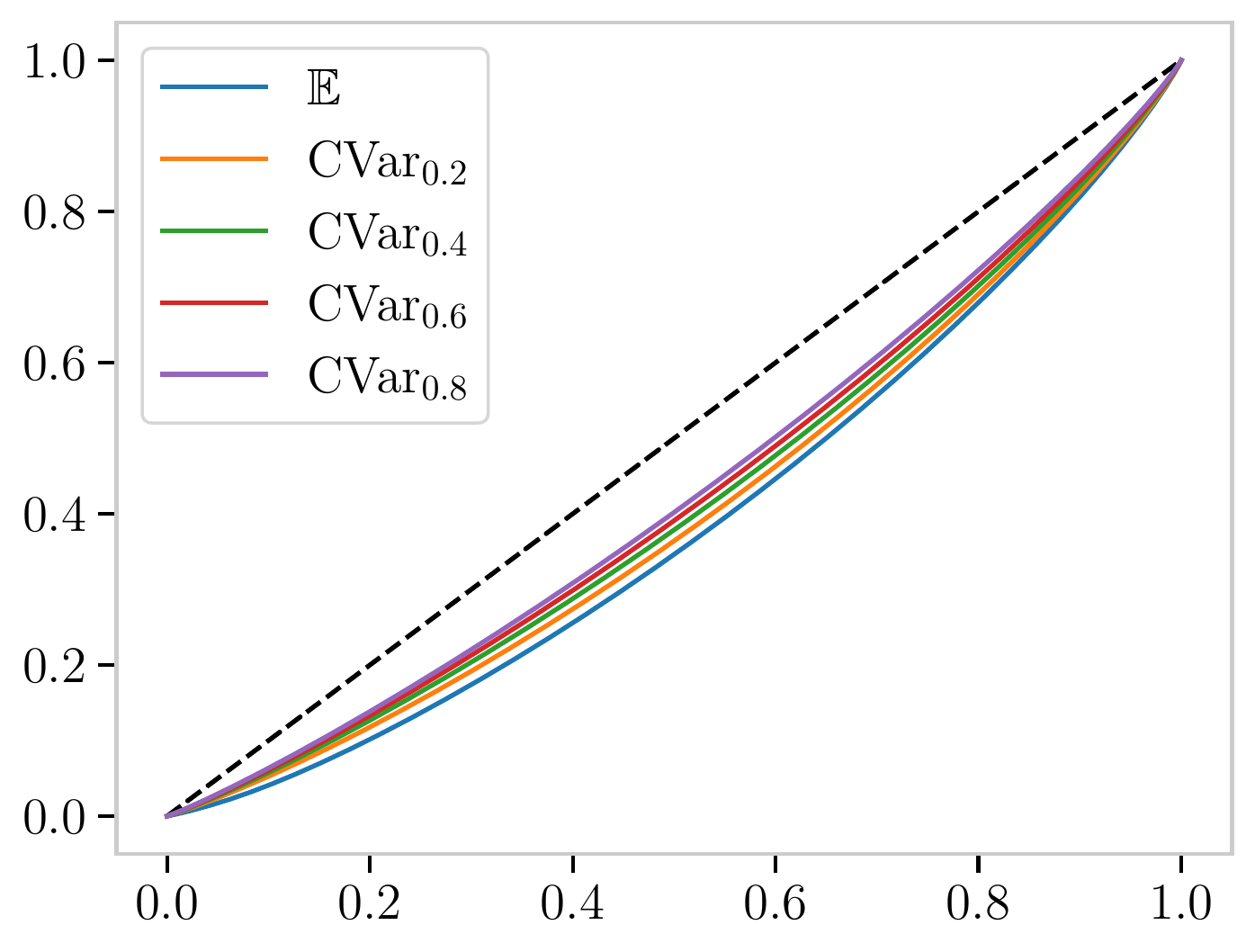}
            \put(21,-3){\tiny{Cumulative share of individuals}} 
            \put(-5,10){\rotatebox{90}{\tiny{Cumulative share of losses}}} 
        \end{overpic}
    \end{subfigure}
     \hspace{0.5cm}
     \begin{subfigure}[b]{0.37\textwidth}
        \begin{overpic}[width=\textwidth]{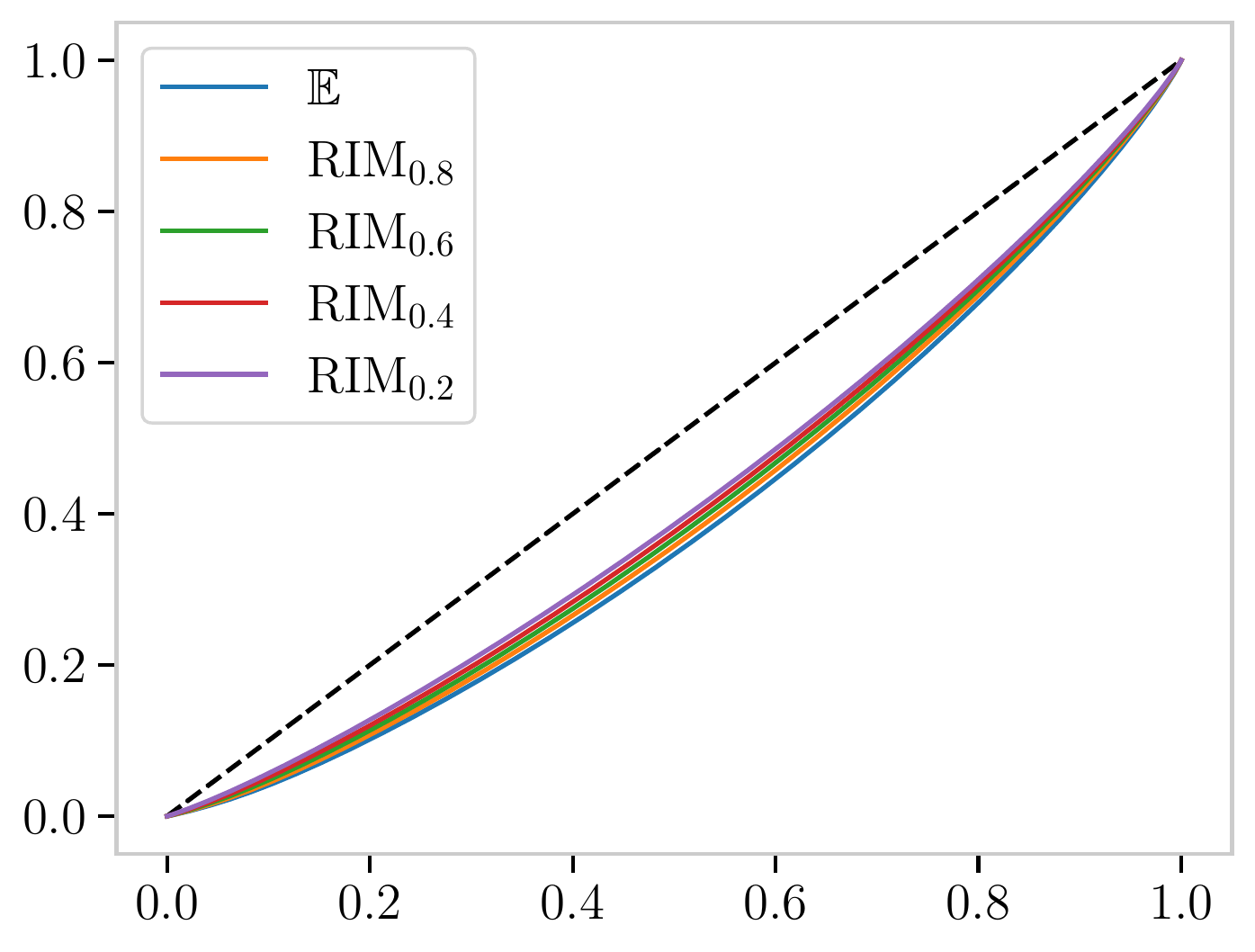}
            \put(21,-3){\tiny{Cumulative share of individuals}} 
            \put(-5,10){\rotatebox{90}{\tiny{Cumulative share of losses}}} 
        \end{overpic}
    \end{subfigure}
    \caption{PCA* results on \texttt{MNIST}. Top row: $\cvar$ curves of test losses for $\cvar$ risk measures (left) with different $\alpha$ and $\rimnoa$s risk measures (right) with $\alpha=0.7$ and different $\beta$, indicated by subscript. For better visibility of the differences, we cut off $\alpha$ at $0.98$. Bottom row: Lorenz curves of test losses for $\cvar$ (left) and $\rimnoa$s (right) with $\alpha=0.7$ and different $\beta$.}
    \label{fig:mnist_results}
\end{figure}

\subsection{Spectral Risk Measures on Subgroup Losses}
In this experiment, we use spectral risk measures on the aggregated losses of pre-specified subgroups. Denote the random variable which indicates the subgroup belonging as $S$. The objective in standard expected risk minimization can be written in a two-stage manner as
\begin{equation}
    \mathbb{E}[\ell(f(X),Y)] = \mathbb{E}_{S}\left[\mathbb{E}_{X,Y|S}[\ell(f(X),Y)]\right]
\end{equation}
using a conditional expectation. Our approach is to replace the outer expectation with a risk measure $R$, as in \cite{williamson2019fairness}. Within each subgroup, individuals are then treated as fungible and are identified with the subgroup aggregate, since the risk neutral expectation is used. Yet differences between subgroups are considered and punished in a risk-averse fashion, thereby favoring less spread in the distribution of subgroup losses. In our experiments, we employ the respective empirical versions of $R$ and $\mathbb{E}$, \ie with respect to the empirical distribution.

\subsubsection{Data}
In this experiment we perform multi-class classification on \texttt{MNIST} (see Section~\ref{exp1:data}) and linear regression with the loss $\ell_1(x,y) = |x-y|$ on \texttt{winequality}.
On \texttt{MNIST}, we create imbalance in the training data to simulate a setting of \textit{label shift}. We use the following number of random samples (without replacement) for a class with index $s \in [0,9]$:
\begin{equation}
    N(s) = 5000 \cdot \exp\left(\frac{-2 s}{10}\right)
\end{equation}
where we round to the nearest integer. The resulting distribution of class frequencies has a moderate imbalance, with frequencies $\{5000,4098,3351,2744,2246,1839,1505,1232,1009,826\}$, see Figure~\ref{fig:imbalance} in Appendix~\ref{app:experiments}. 
In the test data, we leave frequencies unchanged, so that the test images are approximately balanced with regard to class.

The task in \texttt{winequality} is to predict the perceived quality of a wine (on a numeric scale with integers from $3$ to $9$) by physiochemical properties (\eg fixed acidity and alcohol content). In total, there are $11$ input attributes. Here, we consider the two subgroups of red and white wine. In contrast to \texttt{MNIST}, we here purposefully rebalance the data set. The frequency of red and white examples is then the same. This experiment serves to illustrate that balancing data does not necessarily solve the problem of disparate subgroup losses. We use $80\%$ of the red wines ($1279$ examples) in the training set and correspondingly, $1279$ examples of white wine. The test set consists of $320$ red wines and $3619$ white wines. We preprocess both \texttt{MNIST} and \texttt{winequality} using a \texttt{MinMaxScaler} as in Section~\ref{exp1:data}.

\subsubsection{Method}
For \texttt{MNIST}, we use a simple multiclass logistic regression, \ie a cross entropy loss after a single linear layer. We pretrain for $2000$ epochs using the expectation as the risk measure and the \texttt{Adam} optimizer with a learning rate of $0.01$. Using this initialization, we then train for each risk measure for $5000$ epochs with a learning rate of $0.001$. 

For \texttt{winequality} we use a simple feedforward network (one hidden layer of $24$ units followed by a nonlinear $ReLu$ activation) with the $\ell_1$ loss. We train for $3000$ epochs using the \texttt{Adam} optimizer with a learning rate of $0.01$. Again, we use the full data in each epoch to avoid additional challenges due to stochastic mini-batches (see Section~\ref{sec:experiments_discussion}).
We compare the same risk measures as in the first experiment (Section~\ref{exp1:risks}). For both data sets we conduct $50$ independent runs. On \texttt{MNIST}, for each of these runs we randomly shuffle the assignment of the imbalanced frequencies to the classes.

\subsubsection{Results}
\begin{figure}
    \centering
    \begin{subfigure}[b]{0.45\textwidth}
    \includegraphics[width=\textwidth]{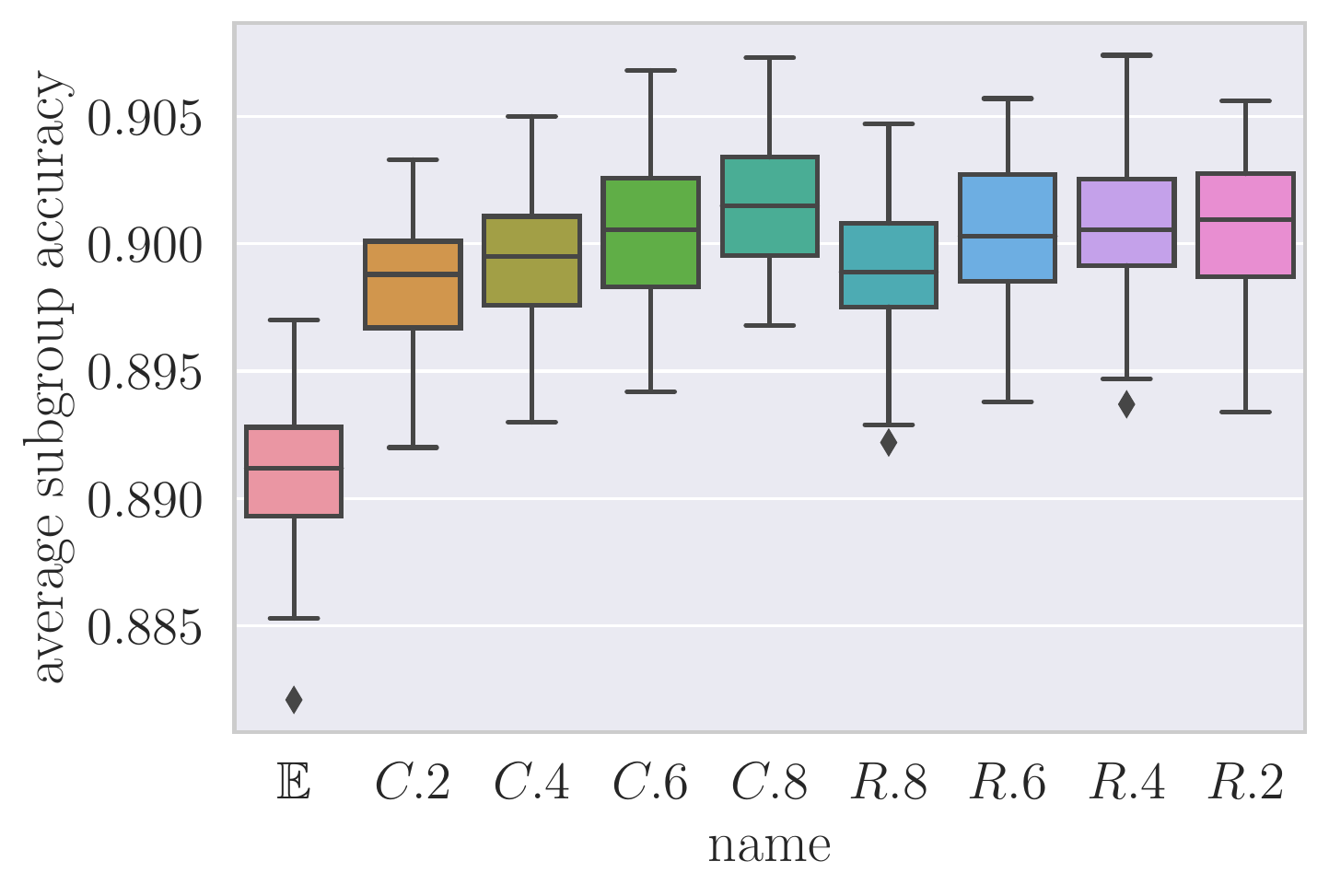}
    \end{subfigure}
    \begin{subfigure}[b]{0.45\textwidth}
    \includegraphics[width=\textwidth]{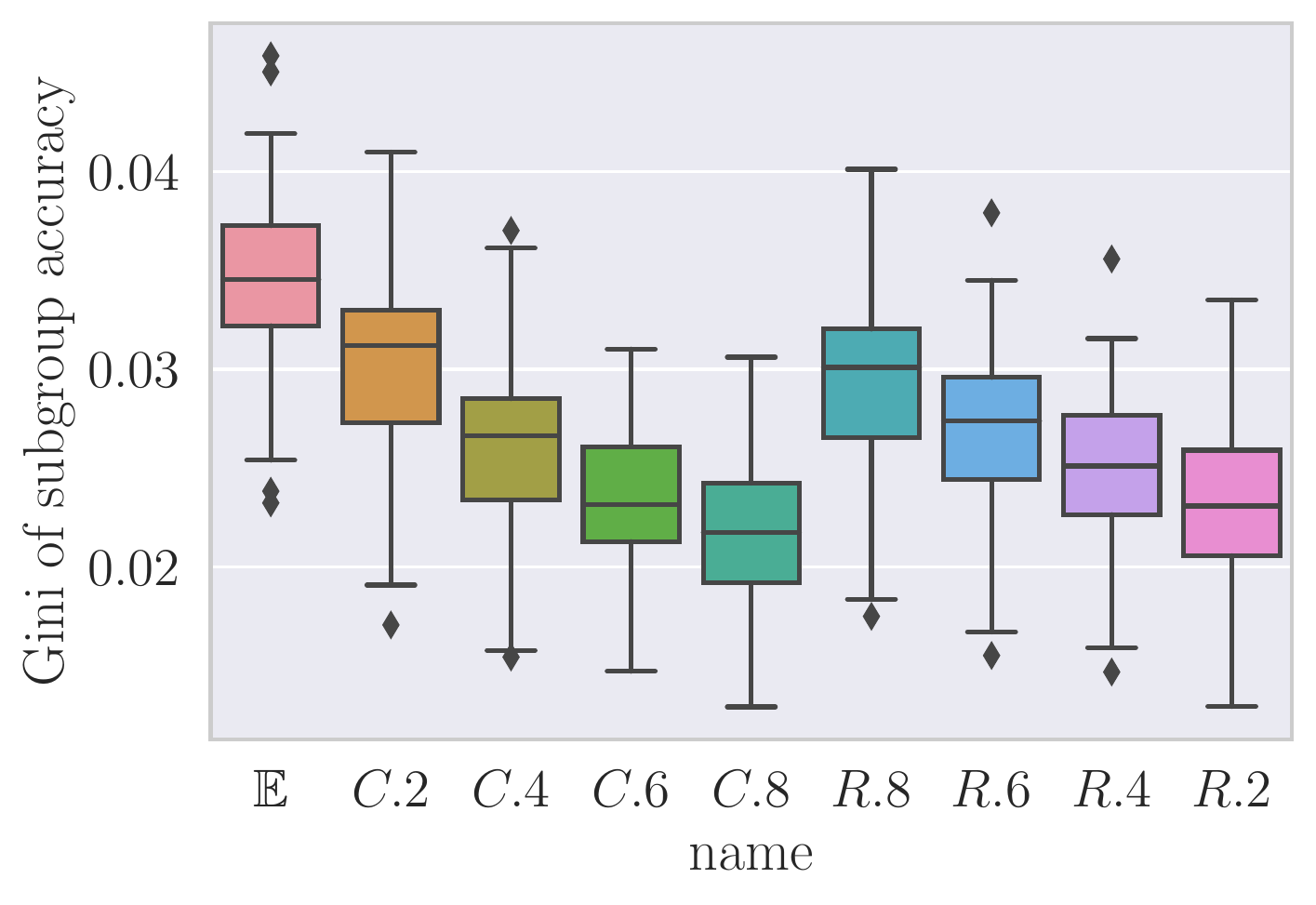}
    \end{subfigure}
    \caption{Results of logistic regression on \texttt{MNIST} across $50$ independent runs. Left: average subgroup accuracies. Right: Gini coefficients of subgroup accuracies. We abbreviate $\cvarnoa_{\alpha=0.f}$ as $C.f$ and $\rimnoa_{\alpha=0.7,\beta=0.f}$ as $R.f$.}
    \label{fig:logistic_mnist_res}
\end{figure}

For \texttt{MNIST}, we report the average subgroup test accuracies and Gini coefficients of subgroup accuracies in Figure~\ref{fig:logistic_mnist_res}.
Due to the data set shift setting, we find that the risk measures even lead to better \textit{average} subgroup performance on the test set. Furthermore, as is visible from the Gini coefficient boxplot, the inequality of subgroup accuracies is reduced.

\begin{figure}
    \centering
    \begin{subfigure}[b]{0.45\textwidth}
    \includegraphics[width=\textwidth]{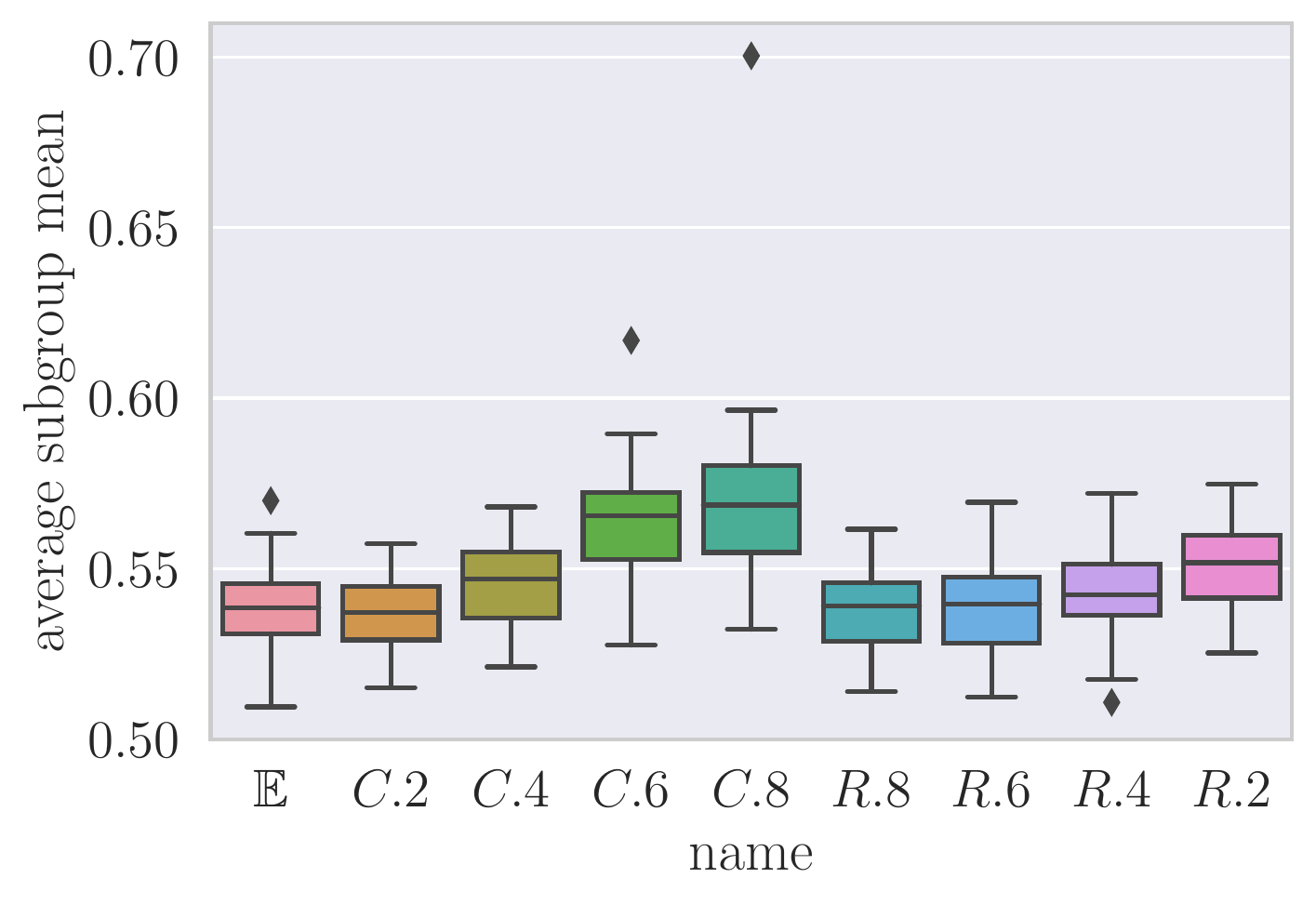}
    \end{subfigure}
    \begin{subfigure}[b]{0.45\textwidth}
    \includegraphics[width=\textwidth]{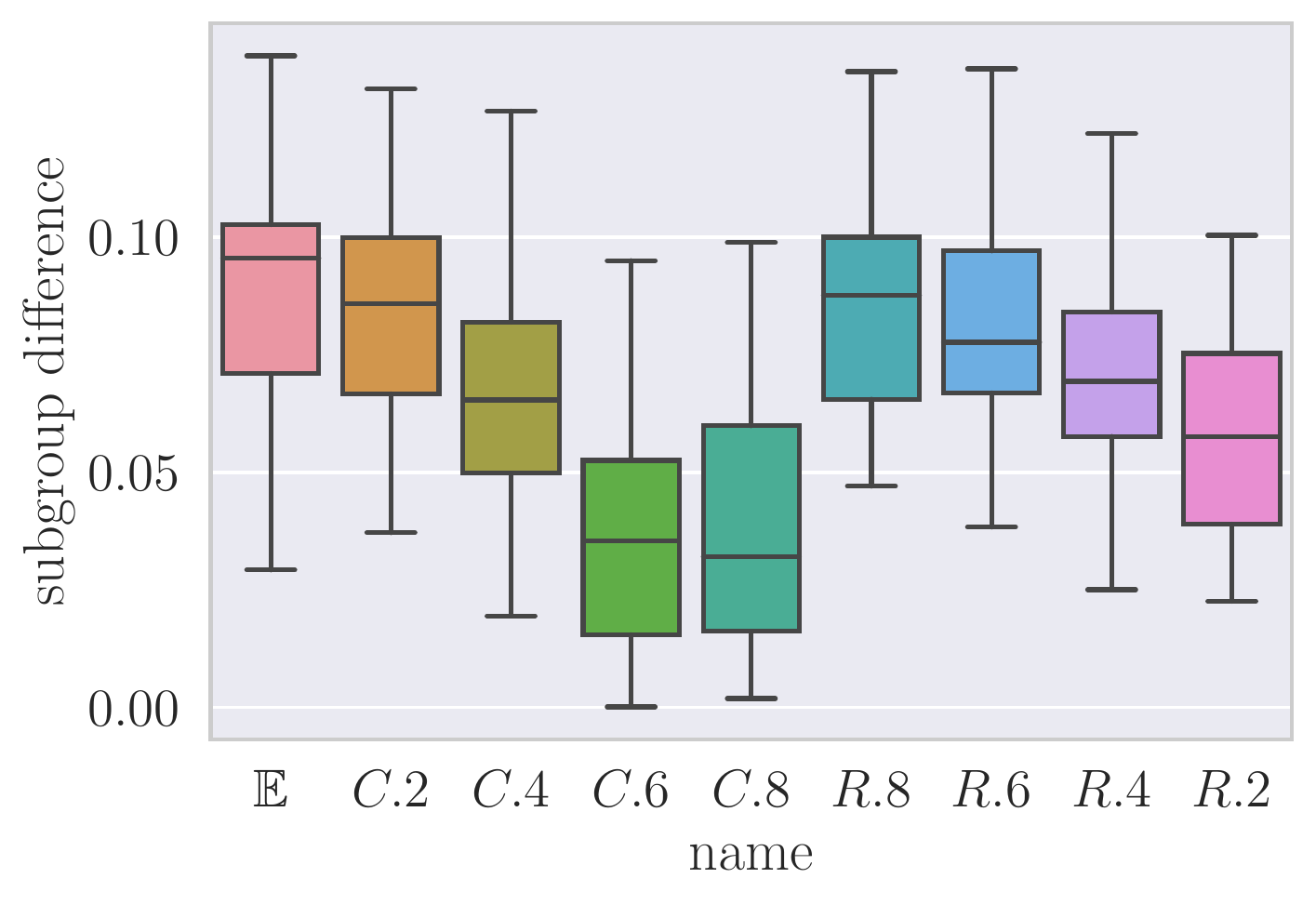}
    \end{subfigure}
    \caption{Results of linear regression on \texttt{winequality} across $50$ independent runs. Left: averages of the two subgroup error means on test data. Right: subgroup absolute error differences on test data across $50$ independent runs. We abbreviate $\cvarnoa_{\alpha=0.f}$ as $C.f$ and $\rimnoa_{\alpha=0.7,\beta=0.f}$ as $R.f$.}
    \label{fig:wine-diff}
\end{figure}

For \texttt{winequality} we show the average of the two subgroup means (red, white) and the absolute difference of the subgroup error means across the $50$ runs (Figure~\ref{fig:wine-diff}). In general, predictions for white wines incur a higher error than for red wines on average, even though the data is balanced. In this setting, the risk measures yield slightly higher errors on average. However, they reduce the difference between the two subgroup means.

\subsection{Discussion}
\label{sec:experiments_discussion}
We have seen that simple spectral risk measures can substantially improve tail performance and hence reduce inequality in the loss distribution. On the other hand, there is a natural trade-off between average and tail performance. In many settings, accounting for tail risk will imply some reduction in average performance. However, this is specific to the train-test data relationship. In the label shift setting on \texttt{MNIST}, we have seen that the increase in robustness by using a spectral risk measure can even lead to better average performance on the test set. Emphasizing ``difficult'' training examples (associated with high loss) in the optimization process guards against possible data set shift scenarios. For a spectral risk measure, the exact nature of this trade-off is directly encoded in the choice of the fundamental function.

However, there is yet another trade-off: that between \textit{robustness} and \textit{estimatibility}. We conjecture that it generally holds that higher tail sensitivity of a risk measure is accompanied with higher difficulty in estimating it from empirical samples. For $\cvar$ this is intuitive, since only a $1-\alpha$ fraction of the sample is actually used in the estimation. How this trade-off depends exactly on the fundamental function is, to our knowledge, as of now unclear. Since estimating tail-sensitive risk measures may lead to highly variable estimates, using risk measures in a mini-batch setting is problematic; but see \citet{curi2020adaptive} for an approach to optimize $\cvar$ in a batch setting. See also the recent review by \citet{laguel2021superquantiles} on the role of $\cvar$ in machine learning.
The use of other spectral risk measures in practice, which do not admit a simple representation as $\cvar$ or $\rim$, also raises challenges. To tackle this, \citet{mehta2023distributionally} have recently proposed a practical stochastic gradient-based method for optimizing spectral risk measures and $f$-divergence risk measures. \citet{leqi2022supervised} have obtained uniform convergence results that justify the optimization of a wide class of risk measures (including the spectrals) on the empirical distribution.

\section{Conclusion}
In this paper, we have questioned the assumption that the expectation is the only 
sensible functional to aggregate losses. Instead, we have considered a wide family of possible 
replacements, the coherent risk measures. These can be used to encode robustness, 
risk aversion and even fairness. The choice of risk measure is an additional choice to 
make for the ML engineer and it is orthogonal to the choice of loss function. 
Therefore we have aimed to stratify the space of possible risk measures. 
The fundamental function provides such a natural stratification. Depending on the 
application, it can be interpreted as an imprecise probability, a risk aversion profile 
or an inequality aversion profile. We have also seen that the fundamental function 
plays a major role in the combination of different risk measures which further 
justifies the appellation ``fundamental.''

Specifically, we have focused on the subclass of spectral risk measures which, 
as we have shown, are extremal risk measures with a given fundamental function, 
and are particularly convenient to work with. 
These occupy a prime position in the theory of coherent risk measures, 
can be motivated in different fashions and have been rediscovered multiple times. 
We assert that this convergence signals that they form a well-founded  and important class.

\subsection*{Acknowledgements}
This work was funded by the Deutsche Forschungsgemeinschaft (DFG, 
German Research Foundation) under Germany’s Excellence Strategy –- 
EXC number 2064/1 –- Project number 390727645.   
The authors thank the International Max Planck Research School for Intelligent Systems (IMPRS-IS) for supporting Christian Fröhlich.
Thanks to Rabanus Derr for many helpful discussions and comments.

\clearpage
\renewcommand{\theHsection}{A\arabic{section}}
\appendix
\section{Proofs}
\label{app:proofs}

\subsection{Proof of Theorem~\ref{theorem:spectraldistortion}}
\label{app:spectraldistortion}
\begin{proof}
\citep{gzyl2008relationship,ridaoui2016choquet}:
\begin{align}
\label{eq:spectral}
    R_\phi(X) &= \int_{-\infty}^{0} \left[\phi(S_X(x)) - 1\right] \d x + \int_0^\infty \phi\left(S_X(x)\right) \d x\\
    &= \int_{-\infty}^{0} \left[\phi(1-F_X(x)) - 1\right] \d x + \int_0^\infty \phi\left(1-F_X(x)\right) \d x\\
    &=^{t=F_X(x)} \int_{0}^{F_X(0)} \left[\phi(1-t)-1\right](F_X^{-1})'(t) \d t + \int_{F_X(0)}^1 \phi(1-t)(F_X^{-1})'(t) \d t\\
    &\label{eq:stareq}= \int_{0}^{F_X(0)} \phi'(1-t)F_X^{-1}(t) \d t + \int_{F_X(0)}^1 \phi'(1-t) F_X^{-1}(t) \d t\\
    &= \int_0^1 F_X^{-1}(1-t)\phi'(t) \d t = R_{(w)}(X),\\
\end{align}
where the step \eqref{eq:stareq} comes from partial integration and $F_X$ is assumed to be continuous. Hence we have the equivalence $\phi'(t)=w(1-t)$, where $w$ is the spectral weighting function. 
\end{proof}

\subsection{Proof of Theorem~\ref{theorem:minrep}}
\label{app:terep}
\begin{proof}
First observe that it is always true that $\mathbb{E}[Y] \leq R'(Y)$, due to the $\cL^1$ norm being the smallest of all ri norms (recall that we assume $R(\chi_\Omega)=1$, implying also $R'(\chi_\Omega)=1$ due to the associate relationship). Hence we never have any $\mathbb{E}[Y]>1$ in the unit ball of the associate norm. 
We label the logical proposition of allowing a representation in the form \eqref{eq:minrep} as $\TErep$, that is, the possibility of a representation of the form:
\begin{equation}
    R(X) = \sup\left\{\int_0^1 X^*(\omega) Y^*(\omega) \d \omega : \mathbb{E}[Y]=R'(Y)=1, Y \in \mathcal{M}^+\right\}.
\end{equation}
That $\TErep$ implies $\PTE$ is trivial:
\begin{align}
    R(X+c) &= \sup\left\{\int_0^1 (X+c)^*(\omega) Y^*(\omega) \d \omega : \mathbb{E}[Y]=R'(Y)=1, Y \in \mathcal{M}^+\right\}\\
    &= \sup\left\{\int_0^1 X^*(\omega) Y^*(\omega) \d \omega + c \int_0^1 Y^*(\omega) \d \omega : \mathbb{E}[Y]=R'(Y)=1, Y \in \mathcal{M}^+\right\}\\
    &= R(X) + c,
\end{align}
where we used the fact that $(X+c)^*=X^*+c$ holds even if $c<0$ as long as $X \geq 0$ and $X+c \geq 0$.
The other direction is more involved. We show $\PTE \Rightarrow \TErep$ by showing $\neg \TErep \Rightarrow \neg \PTE$. To show this, we need a technical lemma. Define the statement $A$ as: $\exists Z \in \mathcal{M}^+, Z > \epsilon$, for some $\epsilon > 0$, so that:
\begin{equation}
\label{eq:statementa}
    R(Z) = \sup\left\{\int_0^1 Z^*(\omega) Y^*(\omega) \d \omega : \mathbb{E}[Y] < 1, R'(Y)\leq1, Y \in \mathcal{M}^+\right\},
\end{equation}
that is, the supremum is not decreased when taking it only over the subset $\{\mathbb{E}[Y] < 1, R'(Y)\leq1\}$. However, the supremum need not be actually attained.

\begin{lemma}
\label{lemma:weakterep}
$\neg A \land \PTE \Rightarrow \TErep$, which is logically equivalent to $\neg \TErep \Rightarrow A \lor \neg \PTE$. 
\end{lemma}
\begin{proof}
So now assume $\neg A \land \PTE$. This means that $R$ is positive translation equivariant and that $\forall Z \in \mathcal{M}^+, Z > \epsilon$, for some $\epsilon > 0$, it holds
\begin{equation}
\label{eq:nota}
    R(Z) = \sup\left\{\int_0^1 Z^*(\omega) Y^*(\omega) \d \omega : \mathbb{E}[Y]=R'(Y)=1, Y \in \mathcal{M}^+\right\}.
\end{equation}
To see that this is indeed the negation of $A$, observe that $\neg A$ means $\forall Z \in \mathcal{M}^+, Z > \epsilon$, for some $\epsilon > 0$
\begin{equation}
\label{eq:eqnegation}
    R(Z) > \sup\left\{\int_0^1 Z^*(\omega) Y^*(\omega) \d \omega : \mathbb{E}[Y] < 1, R'(Y)\leq1, Y \in \mathcal{M}^+\right\}. 
\end{equation}
Negating the equality in \eqref{eq:statementa} here must yield ``strictly greater'', since the set on the righthand side in \eqref{eq:eqnegation} is a subset of the full envelope $\{Y : R'(Y) \leq 1\}$, which is implicit in the lefthand side of \eqref{eq:eqnegation}.
Now we continue to analyze the statement \eqref{eq:eqnegation}. Formally, we can write it as $\sup_C f(Y) > \sup_{C \setminus B} f(Y)$, where $B \subseteq C$. Here, $C \coloneqq \{Y : R'(Y) \leq 1\}$ and $B \coloneqq \{Y : \mathbb{E}[Y] = R'(Y)=1\}$. The function $f(Y) \coloneqq \int_0^1 Z^*(\omega) Y^*(\omega) \d \omega$ runs over the respective set in the subscript. But then $\sup_C f(Y) = \sup_B f(Y)$ holds, because $\sup_C f(Y) = \sup_{(C\setminus B) \cup B} f(Y) = \max(\sup_{C \setminus B} f(Y), \sup_B f(Y)) = \sup_B f(Y)$, where we used that by assumption $\sup_C f(Y) > \sup_{C \setminus B} f(Y)$. It is legitimate to ``decompose'' the supremum over the union into a maximum over the two suprema, \cf for instance \cite[p.\@\xspace 3]{hiriart2004fundamentals}. We conclude therefore that \eqref{eq:nota} is the negation of $A$. Intuitively, $\neg A$ is a slightly weakened form of $\TErep$. But we now show that when it is combined with $\PTE$, we can strengthen it and obtain $\TErep$.

Then for any $X \in  \mathcal{M}^+$ (for brevity, we drop explicitly writing $Y \in \mathcal{M}^+$):
\begin{align}
    R(X) &= \sup\left\{\int_0^1 X^*(\omega) Y^*(\omega) \d \omega : R'(Y)\leq1\right\} \quad \text{(general representation for any $R$)} \\
     \Rightarrow \forall \epsilon > 0 : R(X) &= \sup\left\{\int_0^1 X^*(\omega) Y^*(\omega) \d \omega + \epsilon \int_0^1 Y^*(\omega) \d \omega - \epsilon \int_0^1 Y^*(\omega) \d \omega: R'(Y)\leq1 \right\},
     \end{align}
     from which it follows due to PTE that:
     \[ \forall \epsilon > 0: R(X) = \sup\left\{\int_0^1 X^*(\omega) Y^*(\omega) \d \omega + \epsilon \int_0^1 Y^*(\omega) \d \omega :  \mathbb{E}[Y]=R'(Y)=1 \right\} - \epsilon.
     \]
     Since due to PTE: $R(X+\epsilon - \epsilon) = R(X+\epsilon) - \epsilon$ since $(X+\epsilon)-\epsilon\geq 0$. Taking the supremum over  $\{Y: \mathbb{E}[Y]=R'(Y)=1\}$  suffices due to $\neg A$. Then: 
     \begin{align}
     \forall \epsilon > 0 : R(X) &= \sup\left\{\int_0^1 X^*(\omega) Y^*(\omega) \d \omega + \epsilon \int_0^1 Y^*(\omega) \d \omega - \epsilon \int_0^1 Y^*(\omega) \d \omega:  \mathbb{E}[Y]=R'(Y)=1 \right\} \\
     \Rightarrow R(X) &= \sup\left\{\int_0^1 X^*(\omega) Y^*(\omega) \d \omega : \mathbb{E}[Y]=R'(Y)=1 \right\}.\\
\end{align}
Hence $R$ has a positive translation equivariant representation, \ie $\TErep$ holds. We have thus shown that $\neg A \land \PTE \Rightarrow \TErep$, which is equivalent to $\neg \TErep \Rightarrow A \lor \neg \PTE$.
\end{proof}

Recall that our goal is to show that $\neg \TErep \Rightarrow \neg \PTE$. Using the lemma \ref{lemma:weakterep}, that $\neg \TErep \Rightarrow A \lor \neg \PTE$, it only remains to show that $A \Rightarrow \neg \PTE$. So now assume $A$. This means that $\exists Z \in \mathcal{M}^+, Z > \epsilon$, for some $\epsilon > 0$, so that
\begin{equation}
    R(Z) = \sup\left\{\int_0^1 Z^*(\omega) Y^*(\omega) \d \omega : \mathbb{E}[Y] < 1, R'(Y)\leq1, Y \in \mathcal{M}^+\right\}.
\end{equation}
Write now $X + \epsilon = Z$, which is possible by assumption. Then $X \in \mathcal{M}^+$ and:
\begin{align}
    R(X+\epsilon) &= \sup\left\{\int_0^1 (X+\epsilon)^*(\omega) Y^*(\omega) \d \omega : \mathbb{E}[Y] < 1, R'(Y)\leq1, Y \in \mathcal{M}^+\right\}\\
    &\leq R(X) + \epsilon \sup\left\{\int_0^1 Y^*(\omega) \d \omega : \mathbb{E}[Y] < 1, R'(Y)\leq1, Y \in \mathcal{M}^+\right\} \quad \text{ (subadditivity)}\\
    &< R(X) + \epsilon.
\end{align}
Therefore $R$ is not $\PTE$ and the proof is complete.
\end{proof}

\subsection{Families of Fundamental Functions}
\label{app:families}
\subsubsection{Proof of Theorem~\ref{theorem_tm}.}
\label{app:tmnorm}
\begin{proof} Suppose $R$ is an ri norm which is PTE and has fundamental function $\phi$.
Write
\begin{equation}
     R(X) = \sup_{Z \in \mathcal{Z}}\left\{\int_0^1 X^*(\omega) Z'(\omega) \d \omega \right\}, \quad \phi(t) = \sup_{Z \in \mathcal{Z}}{Z(t)}.
\end{equation}
for some Kusuoka set $\mathcal{Z}$ of $R$. And define 
\begin{equation}
\forall t \in (0,1]: Z_t(x) \coloneqq 
\begin{cases}
       \phi(t) \frac{x}{t} & ,\text{ } x \leq t\\
    \frac{1-\phi(t)}{1-t} x + \frac{\phi(t)-t}{1-t} & ,\text{ } x > t
    \end{cases}
\end{equation}
Then the TM norm can be written as:
\begin{equation}
\label{tmnorm:kusuokarep}
    \|X\|_{TM_\phi} = \sup\left\{\int_0^1 X^*(\omega) Z_t'(\omega) \d \omega : t \in (0,1]\right\}.
\end{equation}
\begin{lemma}
\label{lemma:tmlemma1}
Using the above definitions, it holds that:
    \begin{equation}
        \forall t \in (0,1]: R(X) \geq \int_0^1 X^*(\omega) Z_t'(\omega) \d \omega.
    \end{equation}
\end{lemma}
\begin{proof}[of the Lemma]
    Consider some fixed $t \in (0,1]$. Since $\phi(t)=\sup_{Z \in \mathcal{Z}} Z(t)$, we can, to any $\varepsilon>0$, find some $Z_\varepsilon \in \mathcal{Z}$ so that $0 \leq \phi(t)-Z_\varepsilon(t)<\varepsilon$. Let $\varepsilon_n \downarrow 0$ be an arbitrary sequence converging to zero and denote by $Z_{\varepsilon_n}$ a corresponding sequence of selected concave functions from the Kusuoka set $\mathcal{Z}$. Next, define
    \begin{equation}
        h_{\varepsilon_n}(x) = \begin{cases}
       Z_{\varepsilon_n}(t) \frac{x}{t} & ,\text{ } x \leq t\\
    \frac{1-Z_{\varepsilon_n}(t)}{1-t} x + \frac{Z_{\varepsilon_n}(t)-t}{1-t} & ,\text{ } x > t
    \end{cases},  \quad 
        h_{\varepsilon_n}'(x) \coloneqq \begin{cases}
      \frac{Z_{\varepsilon_n}(t)}{t} & ,\text{ } x \leq t\\
    \frac{Z_{\varepsilon_n}(t)-1}{t-1} & ,\text{ } x > t
    \end{cases}
    \end{equation}
    where the derivative is defined first so that $h_{\varepsilon_n}(x) \coloneqq \int_0^x h_{\varepsilon_n}'(\omega) \d \omega$.
\end{proof}
We observe that by construction $Z_{\varepsilon_n} \geq h_{\varepsilon_n}$; the condition for Hardy's lemma is fulfilled, \ie $\int_0^x Z_{\varepsilon_n}'(\omega) \d \omega \geq \int_0^x h_{\varepsilon_n}'(\omega) \d \omega$. Therefore 
\[
\text{\RomanNumeralCaps{1}}(\varepsilon_n) \coloneqq \int_0^1 Z_{\varepsilon_n}'(\omega) X^*(\omega) \d \omega \geq \int_0^1 h_{\varepsilon_n}'(\omega) X^*(\omega) \d \omega  \eqqcolon \text{\RomanNumeralCaps{2}}(\varepsilon_n).
\]
By definition of $h_{\varepsilon_n}'$,
\[
\text{\RomanNumeralCaps{2}}(\varepsilon_n) = \frac{Z_{\varepsilon_n}(t)}{t} \int_0^t X^*(\omega) \d \omega + \frac{Z_{\varepsilon_n}(t)-1}{t-1} \int_t^1 X^*(\omega) \d \omega.
\]
Now consider $\text{\RomanNumeralCaps{1}}(\varepsilon_n)$. We know that $R(X) \geq \text{\RomanNumeralCaps{1}}(\varepsilon_n)$ by our choice of the $Z_{\varepsilon_n}$. Hence
\[
R(X) \geq \sup_{n \in \mathbb{N}}\left\{\int_0^1 Z_{\varepsilon_n}'(\omega) X^*(\omega) \d \omega \right\} = \limsup_{\varepsilon_n \downarrow 0} \text{\RomanNumeralCaps{1}}(\varepsilon_n) \geq \limsup_{\varepsilon_n \downarrow 0} \text{\RomanNumeralCaps{2}}(\varepsilon_n),
\]
since $0 \leq a_n \leq b_n$ implies $\limsup a_n \leq \limsup b_n$.

We find that
\[
\lim_{n \rightarrow \infty} \frac{Z_{\varepsilon_n}(t)}{t} \int_0^t X^*(\omega) \d \omega = \frac{\phi(t)}{t} \int_0^t X^*(\omega) \d \omega,
\]
since the integral term is constant, and as $n\rightarrow \infty$, $Z_{\varepsilon_n}(t) \rightarrow \phi(t)$ by construction (note that $t$ is fixed). Similarly
\[
\lim_{n \rightarrow \infty} \frac{Z_{\varepsilon_n}(t)-1}{t-1} \int_t^1 X^*(\omega) \d \omega = \frac{\phi(t)-1}{t-1} \int_t^1 X^*(\omega) \d \omega.
\]
Since both limits exist, the limit of their sum exists:
\[
\lim_{n \rightarrow \infty} \text{\RomanNumeralCaps{2}}(\varepsilon_n) = \frac{\phi(t)}{t} \int_0^t X^*(\omega) \d \omega +\frac{\phi(t)-1}{t-1} \int_t^1 X^*(\omega) \d \omega = \int_0^1 Z_t'(\omega) X^*(\omega) \d \omega. 
\] 
Therefore $R(X) \geq \int_0^1 Z_t'(\omega) X^*(\omega) \d \omega$ for fixed $t$.
Since this holds for all $t \in (0,1]$, we also get $R(X) \geq \sup_{t \in (0,1]}\left\{\int_0^1 X^*(\omega) Z_t'(\omega) \d \omega \right\} = \|X\|_{TM_\phi}$.

Thus, taking the embedding theorem into account, for any PTE ri norm (`coherent risk measure') $R$:
\begin{equation}
    \|X\|_{M_\phi} \leq \|X\|_{TM_\phi} \leq R(X) \leq \|X\|_{\Lambda_\phi} \quad \forall X \in \Lambda_\phi.
\end{equation}

We show explicitly that $\|\cdot\|_{TM_\phi}$ is positive translation equivariant.
Let $X \in \mathcal{M}^+$ and $c \in \mR$ so that $X+c \geq 0$:
\begin{align*}
    &\|X+c\|_{TM_\phi} \\
    &= \sup_{0<t<1} \left\{\frac{\phi(t)}{t}\int_0^t (X+c)^*(\omega) \d \omega
    + \frac{\phi(t)-1}{t-1} \int_t^1 (X+c)^*(\omega) \d \omega\right\}\\
    &= \sup_{0<t<1} \left\{\frac{\phi(t)}{t}\int_0^t X^*(\omega) \d \omega
    + \frac{\phi(t)-1}{t-1} \int_t^1 X^*(\omega) \d \omega
    + \frac{\phi(t)}{t}\int_0^t c \d \omega + \frac{\phi(t)-1}{t-1} \int_t^1 c \d \omega
    \right\}\\
    &= \sup_{0<t<1} \left\{\frac{\phi(t)}{t}\int_0^t X^*(\omega) \d \omega
    + \frac{\phi(t)-1}{t-1} \int_t^1 X^*(\omega) \d \omega
    + \frac{\phi(t)}{t} ct + \frac{\phi(t)-1}{t-1} (c-ct)
    \right\}\\
    &= \sup_{0<t<1} \left\{\frac{\phi(t)}{t}\int_0^t X^*(\omega) \d \omega
    + \frac{\phi(t)-1}{t-1} \int_t^1 X^*(\omega) \d \omega + c\right\}\\
    &= \|X\|_{TM_\phi} + c    
\end{align*}
and therefore the norm is PTE.

Recall that both the Dutch risk measure and the spectral MaxVar share the fundamental function $\phi(t)=2t-t^2$. We show that the Dutch risk measure is a special case of this $TM_\phi$ norm.
In this case, we have $\phi(t)/t=2-t$ and $(\phi(t)-1)/(t-1)=1-t$, $t \neq 1$. Therefore
\begin{align}
    \|X\|_{TM_\phi} &= \sup_{0 < t < 1} \left\{(2-t) \int_0^t X^*(\omega) \d \omega + (1-t) \int_t^1 X^*(\omega) \d \omega \right\}\\
    &= \sup_{0 < t < 1} \left\{2 \int_0^t X^*(\omega) \d \omega -t \int_0^1 X^*(\omega) \d \omega + \int_t^1 X^*(\omega) \d \omega - t\int_t^1 X^*(\omega) \d \omega \right\}\\
    &= \sup_{0 < t < 1} \left\{(1-t) \int_0^1 X^*(\omega) \d \omega + \int_0^t X^*(\omega) \d \omega \right\}\\
    &= \sup_{0 < t < 1} \left\{(1-t)\mathbb{E}[|X|] + t \cdot \cvarnoa_{1-t}(|X|) \right\} = \du(|X|).
 \end{align}
According to \citet[Corollary 5.1]{pichler2012uniqueness}, the last expression is equal to the Dutch risk measure.

We still need to show that $\|\cdot\|_{TM_\phi}$ is indeed a valid ri norm. First, consider it on the positive cone as an ri function norm. In \eqref{tmnorm:kusuokarep}, we have it expressed as a supremum over a set of Lorentz norms. Indeed a supremum over a non-empty but otherwise arbitrary set of ri function norms is a valid ri function norm (see Lemma~\ref{lemma:supoverrinorms}). Recall that we assume $R(\chi_\Omega)=1$ throughout the paper.
\begin{lemma}\label{lemma:supoverrinorms}
    Let $\{R_i : i \in \mathcal{I}\}$ be a non-empty family of ri function norms (Definitions~\ref{def:rifunctionnorm},\ref{def:rifunctionnormdef}). Then $R(X) \coloneqq \sup\{R_i(X) : i \in \mathcal{I}\}$ is an ri function norm.
\end{lemma}
\begin{proof}[of the Lemma]
Recall that we globally assume that any ri function norm $\tilde{R}$ satisfies $\tilde{R}(\chi_\Omega)=1$ (which clearly implies $R(\chi_\Omega)=1$ here).
    Properties~\ref{item:norm} and \ref{item:mon} are easy to check. For \ref{item:fatou} we want to show that \[
    (\forall i \in \mathcal{I} : 0 \leq X_n \uparrow X \thinspace\thinspace \mu\text{-a.e.} \Rightarrow R_i(X_n) \uparrow R_i(X)) \implies (0 \leq X_n \uparrow X \thinspace\thinspace \mu\text{-a.e.} \Rightarrow R(X_n) \uparrow R(X)).
    \]

    So let us assume $\forall i \in \mathcal{I} : 0 \leq X_n \uparrow X \thinspace\thinspace \mu\text{-a.e.} \Rightarrow R_i(X_n) \uparrow R_i(X)$. Then $R(X) = \sup_{i \in \mathcal{I}} R_i(X) = \sup_{i \in \mathcal{I}} \lim_{n \rightarrow \infty} R_i(X_n)$. First, assume $R(X) < \infty$. From the definition of $R(X)$ as the $\sup$, we know that $\forall \varepsilon>0 : \exists i(\varepsilon) \in \mathcal{I} : R_i(X) > R(X) - \varepsilon/2$.
    Second, we know that
    \[
    \forall \delta > 0: \forall i \in \mathcal{I} : \exists n_i \in \mathbb{N} : \forall n \geq n_i: R_i(X_n) > R(X) - \delta/2.
    \]
    Choosing $\varepsilon \coloneqq \delta$ and taking the corresponding $i(\delta)$ yields:
    \begin{align}
    &\forall \delta > 0 : \exists i(\delta) : \exists n_i \in \mathbb{N} : \forall n \geq n_i : R_i(X_n) > R_i(X) - \delta / 2 > R(X) - \delta/2\\
    &\Leftrightarrow \forall \delta > 0 : \exists n_i \in \mathbb{N}: \forall n \geq n_i : \sup_{i \in \mathcal{I}} R_i(X_n) > R(X) - \delta\\
    &\label{eq:onesideofineq}\Leftrightarrow \lim_{n \rightarrow \infty} \sup_{i \in \mathcal{I}} R_i(X_n) \geq R(X).
    \end{align}
The statement $\lim_{n \rightarrow \infty} R(X_n) = R(X)$ can be written as 
\[\lim_{n \rightarrow \infty} \sup_{i \in \mathcal{I}} R_i(X_n) = \sup_{i \in \mathcal{I}} \lim_{n \rightarrow \infty} R_i(X_n).
\] Since $\lim_{n \rightarrow \infty} \sup_{i \in \mathcal{I}} R_i(X_n) \leq  \sup_{i \in \mathcal{I}} \lim_{n \rightarrow \infty} R_i(X_n)$ 
is obvious, we have, taking this together with \eqref{eq:onesideofineq}, shown both inequalities and thus equality, \ie $\lim_{n \rightarrow \infty} R(X_n) = R(X)$. That the convergence is from below is clear; hence $R(X_n) \uparrow R(X)$. Finally, if $R(X)=\infty$ then it is obvious that $R(X_n) \uparrow \infty$.
    
As to \ref{item:technical}, note that $R_i(\chi_\Omega)=1 \thinspace \thinspace \forall i \in \mathcal{I}$ implies $R(\chi_\Omega)=1$, which by monotonicity implies $R(\chi_E)\leq 1$ for measurable $E$. Also, $\int_E X \d \mu \leq \int_\Omega X \d \mu = \mathbb{E}[X] \leq R_i(X) \forall i \in \mathcal{I}$ by assumption that $R_i(\chi_\Omega)=1$ and the embedding theorem. Hence choosing \eg $c=2$ gives $\int_E X \d \mu < c R(X)$ for any measurable $E$. Finally, the ri property is obvious.
\end{proof}
The theorem is thus proved.
\end{proof}

\subsubsection{Proof of Theorem~\ref{theorem_mlcvar}}
\label{app:theorem_mlcvar}
\begin{proof}
Let $\phi(t)=\min\left\{t/(1-\alpha),1\right\}$ for some $\alpha \in [0,1)$. Hence $\phi \in \Phi_{0+}$ We show that $\|X\|_{M_\phi} = \cvar(|X|) = \|X\|_{\Lambda_\phi}$. 
Clearly, the Lorentz norm for such $\phi$ is $\cvar$, as
\begin{equation}
    \|X\|_{\Lambda_\phi} = \frac{1}{1-\alpha} \int_0^{1-\alpha} X^*(\omega) \d \omega = \cvar(|X|).
\end{equation}

The Marcinkiewicz norm is
\begin{equation}
    \|X\|_{M_\phi} = \sup_{0 < t \leq 1}\left\{\frac{\phi(t)}{t} \int_0^t X^*(\omega) \d \omega \right\} \\
\end{equation}
We claim that the supremum is reached at $t=1-\alpha$. Then:
\begin{equation}
    \|X\|_{M_\phi} = \frac{1}{1-\alpha} \int_0^{1-\alpha} X^*(\omega) \d \omega = \|X\|_{\Lambda_\phi} = \cvar(|X|).
\end{equation}
Hence it remains to show that the supremum is indeed reached at $t=1-\alpha$. Let $t=1-\alpha+\epsilon$, $\epsilon > 0$. Then
\begin{equation}
    \frac{\phi(t)}{t} \int_0^t X^*(\omega) \d \omega = \phi(1-\alpha+\epsilon) X^{**}(1-\alpha+\epsilon) = 1 \cdot \cvarnoa_{\alpha-\epsilon}(|X|) < \cvar(|X|).
\end{equation}
Let $t<1-\alpha$. Then
\begin{equation}
    \frac{\phi(t)}{t} \int_0^t X^*(\omega) \d \omega = \frac{1}{1-\alpha} \int_0^t X^*(\omega) \d \omega < \frac{1}{1-\alpha} \int_0^{(1-\alpha)} X^*(\omega) \d \omega = \cvar(|X|),
\end{equation}
since $X^*$ is nonnegative. 
For $\alpha \rightarrow 1$, the Marcinkiewicz and the Lorentz norm both coincide with the $\cL^\infty$ norm. If $\alpha \rightarrow 1$, then $\phi(t)=\chi_{(0,1]} \notin \Phi_{0+}$.
\begin{align}
    \|X\|_{\Lambda_\phi} &= \int_0^1 X^*(\omega) \phi'(\omega) \d \omega  + X^*(0)\phi(0+) = X^*(0) = \|X\|_{\cL^\infty}\\
    \|X\|_{M_\phi} &= \sup_{0 < t \leq 1} \left\{1 \cdot \frac{1}{t} \int_0^t X^*(\omega) \d \omega \right\} = X^*(0) = \|X\|_{\cL^\infty}.
\end{align}

Therefore we have established that the coincidence of Marcinkiewicz and Lorentz norm holds, implying that there is then only a single coherent risk measure $\cvar$ with the given fundamental function. It remains to show the converse direction, $\|X\|_{M_\phi} = \|X\|_{\Lambda_\phi}$ only if $\phi$ is of $\cvar$-type, \ie $\phi(t)=\min\left\{t/(1-\alpha),1\right\}$ for some $\alpha \in [0,1)$ or $\phi(t) = \phi_\infty(t) = \lim_{\alpha \rightarrow 1} \min\left\{t/(1-\alpha),1\right\}$. We show that the Marcinkiewicz norm is only positive translation equivariant if $\phi$ is of that type. Since the Lorentz norm is always positive translation equivariant, the norms can only then coincide. Let $\{\phi_t\}$ be the Kusuoka set of the Marcinkiewicz norm constructed as before. 
\begin{equation}
    \forall t \in (0,1]: \phi_{t}(x) \coloneqq 
    \begin{cases}
    \phi(t) \frac{x}{t} & ,\text{ } x \leq t\\
    \phi(t) & ,\text{ } x > t.
    \end{cases}
\end{equation}

If the norm is PTE, we can reduce it (Theorem~\ref{theorem:minrep}) to a representation consisting only of those $\phi_{t'}$ with $\phi_{t'}(1)=1$. While the collection $\{\phi_t\}$ need not be the maximal Kusuoka set $\{t \mapsto \int_0^t Y^*(\omega) \d \omega : \|Y\|_{M_\phi' \leq 1}\}$, the proof of Theorem~\ref{theorem:minrep} is agnostic to the used Kusuoka representation; uniqueness is not assumed. But then, these $\phi_{t'}$ by their definition are 
\begin{equation}
    \phi_{t'}(x) \coloneqq 
    \begin{cases}
    \frac{x}{t'} & , \text{ }x \leq t'\\
    1 & , \text{ }x > t'
    \end{cases} \quad = \min\left\{x/(1-\alpha),1\right\}, \quad \alpha=1-t'.
\end{equation}
The fundamental function is then $\phi(x) = \sup_{t'}\phi_{t'}(x) = \sup_{t'}\min\left\{x/t',1\right\}$, and therefore $\phi$ is of $\cvar$ type. More specifically, either a single $t'=1-\alpha$ suffices in the representation or $\alpha \rightarrow 1$, for which an uncountable infinity of $t'$ is needed, \ie $\phi_{t'}(x) = \sup_{t' \rightarrow 0} \min\{\frac{x}{t'},1\}$, thus $\phi=\chi_{(0,1]}$.
\end{proof}

\subsubsection{Variational representation of $\rim$}
\label{app:rim}
Recall that (Theorem~\ref{theorem:rimwithdef}):
\begin{equation}
    \rim(X) \coloneqq \beta \mathbb{E}[X] + (1-\beta)\cvar(X).
\end{equation}
We show that $\rim$ admits the following variational representation:
\begin{equation}
    \rim(X) = \inf_{\mu \in \mR}{\mu + \mathbb{E} v(X-\mu)} \quad \forall X \in \mathcal{M},
\end{equation}
where the regret function $v$ is given by the piecewise linear 
\begin{equation}
    v(t) = \begin{cases}
    \beta t & t \leq 0\\
    \frac{\beta \alpha - 1}{\alpha-1} t & t > 0.
    \end{cases}
\end{equation}
\begin{proof}
We here translate a result from \citep{pflug2005measuring} to a loss-based formulation. Let $0 < \lambda_1 < 1 < \lambda_2$. Consider the function:
\begin{align}
    f_\mu(x) &= \mu + \lambda_2(x-\mu)^+ - \lambda_1(x-\mu)^-\\
    &= \mu + (\lambda_2 - \lambda_1)(x-\mu)^+ + \lambda_1(x-\mu)\\
    &= \mu + (\lambda_2 - \lambda_1)(x - \mu)^+ + \lambda_1 x - \lambda_1 \mu\\
    &= (\lambda_2 - \lambda_1)(x-\mu)^+ + \lambda_1 x + (1-\lambda_1)\mu \\
    &= \lambda_1 x + (1-\lambda_1) \left(\mu + \frac{\lambda_2 - \lambda_1}{1-\lambda_1} (x-\mu)^+ \right).
\end{align}
Now,
\begin{align}
    \rim(X) &= \inf_{\mu \in \mR} \mathbb{E}[f_\mu(X)]\\
    &= \lambda_1 \mathbb{E}[X] + (1-\lambda_1) \inf_{\mu \in \mR} \left(\mu + \frac{\lambda_2 - \lambda_1}{1-\lambda_1} \mathbb{E}(X-\mu)^+ \right)\\
    &= \beta \mathbb{E}[X] + (1-\beta)\cvar(X),
\end{align}
where $\lambda_1=\beta$ and $\frac{1}{1-\alpha} = \frac{\lambda_2 - \lambda_1}{1-\lambda_1}$, hence $\lambda_2 = \frac{\beta \alpha - 1}{\alpha-1}$.
\end{proof}

\subsection{Norm Equivalences and Tail Risk}
\label{app:normequivalences}

\subsubsection{Proof of Theorem~\ref{theorem:equivalence}}
\label{app:theorem_equivalence}
\begin{proof}
The proof in \citep{pichler2013natural} relies on translation equivariance, \ie that $\phi_{1\gamma}(1)=1 \thinspace \forall \gamma$ and $\phi_{2\zeta}=1 \thinspace \forall \zeta$ (\cf Section~\ref{sec:rikusuoka}). To enable comparison for not necessarily translation equivariant ri norms, as well, we give a new and shorter proof. First, assume that $\ronenorm,\rtwonorm$ have singleton Kusuoka sets $\mathcal{Z}_1=\{\phi_1\}$ and $\mathcal{Z}_2=\{\phi_2\}$, where $\phi_1,\phi_2$ are concave functions which are not required to satisfy $\phi_1(1)=1,\phi_2(1)=1$. Then let
\begin{equation}
    K \coloneqq \sup_{0 < \alpha \leq 1} \frac{\phi_2(\alpha)}{\phi_1(\alpha)}.
\end{equation}
Thus $\forall \alpha \in (0,1]$: $\phi_2(\alpha) \leq K \cdot \phi_1(\alpha)$. Then it directly follows from Hardy's lemma that 
\begin{equation}
    \rtwonormy = \int_0^1 X^*(\omega) \phi_2'(\omega) \d \omega \leq K \cdot \int_0^1 X^*(\omega) \phi_1'(\omega) \d \omega = K \cdot \ronenormy.
\end{equation}
Note that the use of the formal derivatives $\phi_1',\phi_2'$ is unproblematic even with kinks, since the Kusuoka sets are constructed as integrals of nonnegative decreasing functions. That is, we use the dash symbol here not as a differentiation operator, but as a mapping which assigns to a $\phi_1$ the function from which it was constructed as the integral of.
If $\mathcal{Z}_1$ and $\mathcal{Z}_2$ are not singletons, the constant $C$ according to \eqref{eq:K} is equal to \citep{pichler2013natural}
\[
C = \inf\{c>0: \forall \phi_{2\zeta} \in \mathcal{Z}_2: \exists \phi_{1\gamma} \in \mathcal{Z}_1: \forall \alpha \in (0,1]: \phi_{2\zeta}(\alpha) \leq C \cdot \phi_{1\gamma}(\alpha)\},
\]
which ensures that $\forall \epsilon>0 : \forall \phi_{2\zeta} \in \mathcal{Z}_2: \exists \phi_{1\gamma} \in \mathcal{Z}_1: \forall \alpha \in (0,1]$: $\phi_{2\zeta}(\alpha) \leq (C+\epsilon) \cdot \phi_{1\gamma}(\alpha)$.
But then,
\begin{align}
    \rtwonormy &= \sup\left\{\int_0^1 X^*(\omega) \phi_{2\zeta}'(\omega) \d \omega : \phi_{2\zeta} \in \mathcal{Z}_2 \right\}\\
    &\leq (C+\epsilon) \cdot \sup\left\{\int_0^1 X^*(\omega) \phi_{1\gamma}'(\omega) \d \omega : \phi_{1\gamma} \in \mathcal{Z}_1 \right\} = (C+\epsilon) \cdot \ronenormy,
\end{align}
where we applied Hardy's lemma to each of the replacements of $\phi_{2\zeta} \rightarrow \phi_{1\gamma}$ such that the above inequality holds. Let $\epsilon \downarrow 0$ to obtain $\rtwonormy \leq C \cdot \ronenormy$.
\end{proof}

\emph{The converse direction}: \cite{pichler2013natural} stated that $C=\infty$ \eqref{eq:K} implies non-equivalence of the norms. However, no proof was provided for the statement. In Theorem~\ref{theorem:mtmareequiv} we provide a counterexample. We raise the following point: our counterexample involves the Marcinkiewicz norm, which is not positive translation equivariant. In this specific case, the constant $C$ cannot be bounded because of the behaviour for large values of $\alpha$. This can only happen since the Kusuoka set does not satisfy $\forall \phi_1: \phi_1(\alpha) \geq \alpha$, which would be the case when $\ronenorm$ is PTE. In contrast, when both involved norms are PTE, $C=\infty$ can hold only because $\lim_{\alpha \rightarrow 0}$ grows unbounded. This then concerns the tail behaviour of the norms. We conjecture that when both norms are PTE, the original statement holds, \ie $C=\infty$ implies non-equivalence. However, we have been unable to prove this statement.

\subsubsection{Proof of Theorem~\ref{theorem:lmequiv}}
\label{app:normequivalences:lmequiv}
\begin{proof}
Denote by $\phi_t$ the family of functions 
\begin{equation}
    \forall t \in (0,1]: \phi_t(x) \coloneqq 
    \begin{cases}
    \phi(t) \frac{x}{t} & , \text{ }x \leq t\\
    \phi(t) & ,\text{ } x > t
    \end{cases}
\end{equation}
which generate a Kusuoka set of the Marcinkiewicz norm with fundamental function $\phi$. Then the problem of finding a $C$ as in Theorem~\ref{theorem:equivalence} reduces to
\begin{equation}
    C = \inf_{\phi_t} \sup_{0 < \alpha \leq 1} \frac{\phi(\alpha)}{\phi_t(\alpha)} = \inf_{\phi_t} \sup_{0 < \alpha \leq 1} \begin{cases}
    \frac{\phi(\alpha)}{\alpha} \frac{t}{\phi(t)} & ,\text{ } \alpha \leq t\\
    \frac{\phi(\alpha)}{\phi(t)} & ,\text{ } \alpha > t.
    \end{cases}
\end{equation}
Fix any $t$. Then certainly
\begin{equation}
    C \leq \sup_{0 < \alpha \leq 1} \frac{\phi(\alpha)}{\phi_t(\alpha)} = \sup_{0 < \alpha \leq 1} \begin{cases}
    \frac{\phi(\alpha)}{\alpha} \frac{t}{\phi(t)} & ,\text{ } \alpha \leq t\\
    \frac{\phi(\alpha)}{\phi(t)} & , \text{ }\alpha > t.
    \end{cases}
\end{equation}
The supremum over the second term (for $\alpha > t$) is bounded, since $t$ is fixed and $\phi$ is bounded by $[0,1]$. As to the first term ($\alpha \leq t$), since $\frac{\phi(\alpha)}{\alpha}$ is decreasing in $\alpha$ (due to the quasiconcavity of $\phi$), the supremum must occur\footnote{When writing \textit{occur}, we do not mean to imply that a supremum is actually attained. 
} for $\alpha \rightarrow 0$, or otherwise it occurs in the second term. Observe that $\phi'(0) = \lim_{\alpha \rightarrow 0} \frac{\phi(\alpha)}{\alpha}$ by definition of the difference quotient since $\phi(0)=0$. Therefore:
\begin{equation}
    \lim_{\alpha \rightarrow 0} \frac{\phi(\alpha)}{\alpha} \frac{t}{\phi(t)} = \frac{\phi'(0)}{1} \frac{t}{\phi(t)} < \infty.
\end{equation}
Therefore $C$ is finite. As an example of this statement, with the choice $\phi(t)=2t-t^2$, we obtain equivalence of the Dutch risk measure and MaxVar.

It remains to show that $K=1/(\phi(\frac{1}{\phi'(0)}))$ is a feasible constant. 
Obviously, this $K$ is finite under the assumption that $\phi'(0)<\infty$. Consider a linear function with slope $\phi'(0)$. It reaches $1$ at $t=1/\phi'(0)$. With this choice of $t$, we have
\begin{equation}
    \phi_t(\alpha) = \begin{cases}
    \frac{\phi(1/\phi'(0))}{1/\phi'(0)} \alpha & , \text{ }\alpha \leq t\\
    \phi(1/\phi'(0))  & ,\text{ } \alpha > t.
    \end{cases}
\end{equation}
Then
\begin{equation}
    \phi(\alpha) \leq K \cdot \phi_{1/\phi'(0)} = \begin{cases}
    \phi'(0) \cdot \alpha & ,\text{ } \alpha \leq 1/\phi'(0)\\
    1 & ,\text{ } \alpha > 1/\phi'(0).
    \end{cases} 
\end{equation}
To see that this holds, observe that $\forall \alpha \in (0,1]$ $\frac{\phi(\alpha)}{\alpha} \leq \phi'(0) = \lim_{t \rightarrow 0} \frac{\phi(t)}{t}$ due to quasiconcavity and also $\phi(\alpha) \leq 1$. Indeed $K$ is the smallest constant such that the statement $\exists t' \in (0,1]: \forall \alpha \in (0,1]$  $\phi(\alpha) \leq  K' \cdot \phi_{t'}(\alpha)$ holds. Assume $t' < t = 1/\phi'(0)$. 
For $\alpha > t'$ we require $ K' \cdot \phi(t') \geq 1$, otherwise the majorization does not hold as $\phi(\alpha)$ approaches $1$. But since $\phi(t') < \phi(t)$, this implies $K' > K$. Now assume $t' > t$. Clearly, we must require $(K' \cdot \phi_{t'})'(0) \geq \phi'(0)$ for the majorization to hold in the limit as $\alpha \rightarrow 0$ (this comes from letting $\alpha \rightarrow 0$ in the condition $K' \cdot \frac{\phi(t')}{t'}\alpha \geq \phi(\alpha)$). That is, $(K' \cdot \phi_{t'})'(0) = K' \cdot \frac{\phi(t')}{t'}$. By design, we have $K\cdot \phi_t'(0) = K \cdot \frac{\phi(t)}{t} =\phi'(0)$. Due to quasiconcavity and $t'>t$ we have $\frac{\phi(t')}{t'} > \frac{\phi(t)}{t}$. Hence $(K' \cdot \phi_{t'})'(0) \geq \phi'(0)$ implies $K' > K$.


However, we note that $K$ need not be the smallest constant such that $\|X\|_{\Lambda_\phi} \leq K \cdot \|X\|_{M_\phi}$ holds, but it is the smallest constant so that $\exists t' \in (0,1]: \forall \alpha \in (0,1]$ $\phi(\alpha) \leq  K' \cdot \phi_{t'}(\alpha)$, which guarantees the previous statement to hold.
\end{proof}

\subsubsection{Proof of Theorem~\ref{theorem:mlnotequiv}}
\label{app:normequivalences:mlnotequiv}
\begin{proof}
The result can be easily derived by combining two statements from \cite{rubshtein2016foundations}\footnote{Note that \citet{rubshtein2016foundations} denote the Marcinkiewicz space with fundamental function $V$ as $M_{V^*}$. With this notation, the associate space to the Lorentz space $\Lambda_V$ is $M_{V}$. In our notation, however, the associate relationship is $\Lambda_V' = M_{V^*}$, that is, the subscript indicates the fundamental function.}. According to \cite[p.\@\xspace 164]{rubshtein2016foundations}, a Lorentz space is separable if and only if $\phi(0+)=0$ ($\phi \in \Phi_{0+}$). On the other hand, if $\phi(0+)=0$ and $\phi'(0)=\infty$, the Marcinkiewicz space is not separable. Hence the two spaces do not coincide if $\phi'(0)=\infty$ and the norms are therefore not equivalent. 
This result implies that for a $\phi$ with $\phi'(0)=\infty$, not all ri norms are equivalent.

As a sanity check, we show that also $C=\infty$, which is a necessary condition for non-equivalence. Assume by contradiction that $C < \infty$, that is:
\begin{equation}
    C = \inf\left\{c > 0 :  \exists \phi_t: \forall \alpha \in (0,1]: \phi(\alpha) \leq c \cdot \phi_t(\alpha)\right\} < \infty.
\end{equation}
Then $\forall \epsilon > 0 : \exists t: \forall \alpha \in (0,1]: \phi(\alpha) \leq (C+\epsilon) \cdot \phi_t(\alpha)$. Fix some $\epsilon$. With this choice of $t$,
\begin{equation}
    1 \geq \sup_{0<\alpha\leq 1} {
    \begin{cases}
    \frac{\phi(\alpha)}{\alpha} \frac{t}{(C+\epsilon) \cdot \phi(t)} & ,\text{ } \alpha \leq t\\
    \frac{\phi(\alpha)}{(C+\epsilon) \cdot \phi(t)} & ,\text{ } \alpha > t .
    \end{cases}
    } 
\end{equation}
If the supremum occurs in the first term, it occurs as $\alpha \rightarrow 0$ due to the quasiconcavity of $\phi$. Recall that $\phi'(0) = \lim_{\alpha \rightarrow 0} \frac{\phi(\alpha)}{\alpha}$ by definition of the difference quotient. Therefore $1 \geq \sup_{0<\alpha\leq t} \frac{\phi(\alpha)}{\alpha} \frac{t}{(C+\epsilon) \cdot \phi(t)} = \lim_{\alpha \rightarrow 0} \frac{\phi(\alpha)}{\alpha} \frac{t}{(C+\epsilon) \cdot \phi(t)} = \phi'(0) \frac{t}{(C+\epsilon) \cdot \phi(t)}= \infty$, a contradiction. Thus no finite $C$ exists. 
Finally, the inequality $\|X\|_{M_\phi} \leq \|X\|_{\Lambda_\phi}$ stems from the embedding theorem.

\end{proof}

\subsubsection{Proof of Theorem~\ref{theorem:equivtol1}}
\label{app:normequivalences:equivtol1}
\begin{proof}
Assume that $\ronenorm=\cL^1$. Let $\mathcal{Z}_2=\{\phi_{2\zeta}\}$ be the Kusuoka set of $\rtwonorm$. Then, since $\{\alpha \mapsto \alpha\}$ is a Kusuoka set for the $\cL^1$ norm:
\begin{equation}
   C = \sup_{\phi_{2\zeta}} \sup_{0 < \alpha \leq 1} \frac{\phi_{2\zeta}(\alpha)}{\alpha}.
\end{equation}
Due to the concavity of $\phi_{2\zeta}$, the fraction is decreasing in $\alpha$ and hence
\begin{equation}
    C = \sup_{\phi_{2\zeta}} \lim_{\alpha \rightarrow 0} \frac{\phi_{2\zeta}(\alpha)}{\alpha} = \sup_{\phi_{2\zeta}} \phi_{2\zeta}'(0).
\end{equation}
It remains to show that $\phi_2'(0)<\infty \Rightarrow \sup_{\phi_{2\zeta}} \phi_{2\zeta}'(0)<\infty \quad \forall \phi_{2\zeta} \in \mathcal{Z}_2$. We have 
\begin{equation}
    \phi_2(\alpha) = \sup_{\mathcal{Z}_2}\{\phi_{2\zeta}(\alpha)\}.
\end{equation}
We know that
\begin{equation}
    \frac{\phi_2(t)}{t} \geq \frac{\phi_{2\zeta}(t)}{t} \quad \forall t \quad \forall \phi_{2\zeta},
\end{equation}
since we know the limit of the left hand side exists, we have
\begin{equation}
    \infty > \phi_2'(0) = \lim_{h \rightarrow 0} \frac{\phi_2(t)}{t} \geq \lim_{h \rightarrow 0} \frac{\phi_{2\zeta}(t)}{t} = \phi_{2\zeta}'(0) \quad \forall \phi_{2\zeta}.
\end{equation}

Therefore $C$ is finite and we obtain $\|X\|_{\cL^1} \leq \ronenormy \leq C \cdot \|X\|_{\cL^1} \quad \forall X \in \cL^1$. As a consequence, any two ri norms with finite $\phi_1'(0),\phi_2'(0)$ are equivalent.
\end{proof}

\subsubsection{Proof of Theorem~\ref{theorem:mtmareequiv}}
\label{app:normequivalences:mtmareequiv}
\begin{proof} First we show that $C=\infty$. Denote by $\phi_{TM,t},\phi_{M,t}$ the respective Kusuoka sets, constructed as:
\begin{equation}
    \forall t \in (0,1]: \phi_{TM,t}(x) \coloneqq 
    \begin{cases}
    \phi(t) \frac{x}{t} & ,\text{ } x \leq t\\
    \frac{1-\phi(t)}{1-t} x + \frac{\phi(t)-t}{1-t} & ,\text{ } x > t
    \end{cases},\quad
    \phi_{M,t}(x) \coloneqq 
    \begin{cases}
    \phi(t) \frac{x}{t} & ,\text{ } x \leq t\\
    \phi(t) & ,\text{ } x > t.
    \end{cases}
\end{equation}

The desired constant is
\begin{align}
    C &= \sup_{\phi_{TM,t}} \inf_{\phi_{M,t'}} \sup_{0 < \alpha \leq 1} \frac{\phi_{TM,t}(\alpha)}{\phi_{M,t'}(\alpha)}\\
    &= \inf\left\{c > 0 :  \forall \phi_{TM,t} \exists \phi_{M,t'}: \forall \alpha \in (0,1]: \phi_{TM,t}(\alpha) \leq c \cdot \phi_{M,t'}(\alpha)\right\}.
\end{align}
We show that no such finite constant can exist. Let some $t$ be given. We wish to find $t'$ such that $\forall \alpha \in (0,1]:$ $\phi_{TM,t}(\alpha) \leq K \cdot \phi_{M,t'}(\alpha)$. The argument in Theorem~\ref{theorem:lmequiv} shows that the smallest feasible $K$ with the Marcinkiewicz norm on the right hand side is $K=1/\phi_{TM,t}(1/\phi_{TM,t}'(0))$. Since $\phi_{TM,t}'(0)=\frac{\phi(t)}{t}$, $K=1/\phi_{TM,t}(\frac{t}{\phi(t)})$. Now $\frac{t}{\phi(t)} > t$ unless $t=1$ (or we have $\phi(t)=1$) so that by definition of $\phi_{TM,t}$
\begin{equation}
    \frac{1}{K} = \frac{1-\phi(t)}{1-t} \frac{t}{\phi(t)} + \frac{\phi(t)-t}{1-t}.
\end{equation}
For each fixed $t$, this is the best feasible constant (Theorem~\ref{theorem:lmequiv}) and it is finite. However, as $t\rightarrow 0$, we find that $\lim_{t \rightarrow 0} K = \frac{\phi(t)}{t} = \phi'(0) = \infty$. As $t \rightarrow 0$, we need not consider the case of $\phi(t)=1$, since in the limit this condition does not hold (noting that $\phi \in \Phi_{0+}$). We conclude that $C=\infty$.

We now show that, despite $C=\infty$, the Marcinkiewicz and PTE Marcinkiewicz norm are equivalent. Note that
\begin{align}
    \|X\|_{TM_\phi} &= \sup_{0 < t < 1}\left\{\frac{\phi(t)}{t}\int_0^t X^*(\omega) \d \omega
    + \frac{\phi(t)-1}{t-1}\int_t^1 X^*(\omega) \d \omega\right\}
    \\
    \label{eq:secondtermtm}
    &\leq \|X\|_{M_\phi} + \sup_{0 < t < 1}\left\{\frac{\phi(t)-1}{t-1}\int_t^1 X^*(\omega) \d \omega\right\}.
\end{align}
Knowing that $\|X\|_{M_\phi} \leq \|X\|_{TM_\phi}$, the norms could only \textit{not} be equivalent if $\|X\|_{M_\phi} < \infty$ for some $X$, while $\|X\|_{TM_\phi} = \infty$. Such an $X$ cannot exist, since the second term in \eqref{eq:secondtermtm} behaves nicely: $\int_t^1 X^*(\omega) \d \omega \leq \int_0^1 X^*(\omega) \d \omega < \infty$ due to $\cL^1$ being the largest ri space, in which any Marcinkiewicz space is embedded. Furthermore, the factor $\frac{\phi(t)-1}{t-1}$ does not exhibit pathological behaviour. Noting that $\phi$ is bounded from below and above, we only have to check the limits as $t \rightarrow 0$ and $t \rightarrow 1$:
\begin{equation}
    \lim_{t \rightarrow 0} \frac{\phi(t)-1}{t-1} = 1, \quad
    \lim_{t \rightarrow 1} \frac{\phi(t)-1}{t-1} = \phi'(1) < \infty.
\end{equation}
As $t \rightarrow 0$, we can apply the quotient rule, since both limits exist. As $t \rightarrow 1$, we use L'Hôpital's rule. Hence we conclude that the sets of functions, for which the Marcinkiewicz and PTE Marcinkiewicz norms are finite, coincide. Therefore the norms are equivalent \cite[p.\@\xspace 7]{Bennett:1988aa}.
\end{proof}

\subsection{Properties of Quasiconcave Functions}
\label{app:quasiconcave}

\subsubsection{Proof of Lemma~\ref{lemma:qcminmax}}
\label{app:quasiconcave:qcminmax}
\begin{proof}
	By quasiconcavity, and assumption, we have $\phi(0)=0$, $\phi(1)=1$
	and 
	\[
		0\le t_0\le t_1 \ \ \Rightarrow\ \ \phi(t_0)\le \phi(t_1)
		\mbox{\ and\ } \phi(t_0)/t_0 \ge \phi(t_1)/t_1.
	\]
	Taking $t_1=1$ we
	have $\phi(t_0)\le\phi(t_1)=1$. Taking  $1=t_0\le t_1$ implies
	$1=\phi(1)\le \phi(t_1)$. Furthermore, $t_0\le t_1=1$ implies
	$\phi(t_0)/t_0\ge 1$ which implies $\phi(t_0)\ge t_0$.
	Additionally, $1=t_0\le t_1$ implies $1\ge \phi(t_1)/t_1$ and thus
	$\phi(t_1)\le t_1$. Combining all these facts we have shown
	\begin{align*}
		t\le 1 \ \ \Rightarrow &\ \  t\le \phi(t)\le 1\\
		t\ge 1\ \ \Rightarrow &\ \ 1\le \phi(t)\le t.
	\end{align*}
	For $0\le t\le 1$, $t=1\wedge t$ and $1=1\vee t$.   For $t\ge
	1$, $1=1\wedge t$ and $t=1\vee t$.  Hence 
	Lemma~\ref{eq:bounds-on-quasiconcave-phi} holds.
\end{proof}

\subsubsection{Proof of Lemma~\ref{lemma:qc-min-max}}
\label{app:quasiconcave:qc-min-max}
\begin{proof}
	Since max and min are continuous, and the composition of continuous
	functions is continuous, we have that $t\mapsto \bigwedge_{i\in[n]}
	\phi_i(t)$ and
	$t\mapsto\bigvee_{i\in[n]}\phi_i(t)$ are continuous for all $t>0$. 

	Furthermore, min and max are increasing (i.e.~non-decreasing) in
	each argument, and the composition of
	increasing functions is increasing, and thus  $t\mapsto
	\bigwedge_{i\in[n]} \phi_i(t)$ and
	$t\mapsto\bigvee_{i\in[n]}\phi_i(t)$ are increasing.
	Suppose $t_0\le t_1$. Let $i^*=\argmax_i \phi_i(t_1)/t_1$.  Then
	\[
		\frac{\bigvee_{i\in[n]} \phi_i(t_0)}{t_0}\ge 
		\frac{\phi_{i^*}(t_0)}{t_0} \ge 
		\frac{\phi_{i^*}(t_1)}{t_1}=
		\frac{\bigvee_{i\in[n]} \phi_i(t_1)}{t_1},
	\]
	and thus $t\mapsto\left(\bigvee_{i\in[n]}\phi_i(t)\right)/t$ 
	is decreasing.  A similar argument holds for $\bigwedge_i \phi_i$.  
\end{proof}

\subsubsection{Proof of Lemma~\ref{lemma:perspective}}
\label{app:quasiconcave:perspective}
\begin{proof}
	(If): For $\alpha>0$, $\breve{\psi}(\alpha x,\alpha y)=\alpha y\psi(\alpha
	x/(\alpha y))=\alpha\breve{\psi}(x,y)$, and thus $\breve{\psi}$ is
	positively homogeneous. Furthermore,
	$\breve{\psi}(x,y)$ is nondecreasing in each argument as we now show.
	Let $y\in\reals_{\ge 0}$ be arbitrary but fixed and consider
	$x\mapsto\breve{\psi}(x,y)=y\psi(x/y)$.  This is nondecreasing since
	$\psi$ is nondecreasing. Now let $x\in\reals_{\ge 0}$ be arbitrary
	but fixed and let $g(y)\coloneqq y\psi(x/y)$. Observe that
	$z\mapsto g(1/z)=\psi(xz)/z$ which is nonincreasing (since $\psi$
	is quasiconcave). Hence $y\mapsto g(y)$ is nondecreasing. Thus
	$\breve{\psi}$ is nondecreasing in both of its arguments concluding
	the demonstration that $\breve{\psi}\in\Pscr$.

	(Only if): If $x\le y$ then $\psi(x)=\breve{\psi}(x,1)\le
	\breve{\psi}(y,1)=\psi(y)$ and thus $\psi$ is nondecreasing.  Since
	$\breve{\psi}$ is positively homogeneous and nonzero, $\psi(s)\ne
	0$ unless $s=0$.  Finally, for $x\le y$, 
	\[
		\frac{\psi(x)}{x}=\frac{\breve{\psi}(x,1)}{x}=
		\breve{\psi}\left(1,\frac{1}{x}\right)
		\ge
		\breve{\psi}\left(1,\frac{1}{y}\right)=
		\frac{\breve{\psi}(y,1)}{y}=\frac{\psi(y)}{y},
	\]
	which shows that $x\mapsto\psi(x)/x$ is nonincreasing,
	demonstrating that $\psi$ is quasiconcave.
\end{proof}

\subsubsection{Proof of Lemma~\ref{lemma:composition-of-quasiconcave}}\label{app:quasiconcave:composition-of-quasiconcave}
\begin{proof}
	(If):
	Since $\psi$ is quasiconcave, it is continuous everywhere except at
	the origin and thus so is $t\mapsto
	\phi_1(t)\psi(\phi_0(t)/\phi_1(t))$.
	Furthermore, since $\phi_0$ and
	$\phi_1$ are nondecreasing, and $\breve{\psi}$ is nondecreasing in each
	argument, then $f_{\phi_0,\phi_1}(t)$ is nondecreasing in $t$.
	Finally, we need to show that 
	\begin{equation}\label{eq:psi-on-t-decreasing}
		t_0\le t_1 \ \Rightarrow\
		\frac{\breve{\psi}(\phi_0(t_0),\phi_1(t_0))}{t_0} \ge
		\frac{\breve{\psi}(\phi_0(t_1),\phi_1(t_1))}{t_1}.
	\end{equation}
	Since $\breve{\psi}$ is positively homogeneous,
	\eqref{eq:psi-on-t-decreasing} is equivalent to 
	\[
		t_0\le t_1 \ \Rightarrow\ 
                \breve{\psi}\left(\frac{\phi_0(t_0)}{t_0},
		\frac{\phi_1(t_0)}{t_0}\right) \ \ge\ 
		\breve{\psi}\left(\frac{\phi_0(t_1)}{t_1}, 
		\frac{\phi_1(t_1)}{t_1}\right).
	\]
	But by assumption on $\phi_0$ and $\phi_1$ we have
	\[
		t_0\le t_1 \ \Rightarrow\ 
		\frac{\phi_0(t_0)}{t_0} \ge \frac{\phi_0(t_1)}{t_1} \ \
		\mbox{\ and\ }\ \ 
		\frac{\phi_1(t_0)}{t_0} \ge \frac{\phi_1(t_1)}{t_1}.
	\]
	Since $\breve{\psi}$  is nondecreasing in each argument, we can thus
	conclude that  \eqref{eq:psi-on-t-decreasing} holds, thus demonstrating
	the final property needed to show quasiconcavity of
	$f_{\phi_0,\phi_1}$.

	(Only if):  Using the definition of $\breve{\psi}$ we have 
	 $f_{\phi_0,\phi_1}(t)=
	\phi_1(t)\psi\left(\frac{\phi_0(t)}{\phi_1(t)}\right)$.  We need to
	show that
	$[ \forall\phi_0,\phi_1\in\Qscr,\ 
	f_{\phi_0,\phi_1}\in\Qscr] \Rightarrow\psi\in\Qscr$.
	Now $f_{\phi_0,\phi_1}\in\Qscr$ means that
	\begin{enumerate}
		\item $t_0\le t_1\Rightarrow
			\phi_1(t_0)\psi\left(\frac{\phi_0(t_0)}{\phi_1(t_0)}\right)
			\le
			\phi_1(t_1)\psi\left(\frac{\phi_0(t_1)}{\phi_1(t_1)}\right)$.
		\item $t_0\le t_1\Rightarrow
			\frac{\phi_1(t_0)}{t_0} 
			      \psi\left(\frac{\phi_0(t)0)}{\phi_1(t_0)}\right)
			\ge
			\frac{\phi_1(t_1)}{t_1} 
			     \psi\left(\frac{\phi_0(t_1)}{\phi_1(t_1)}\right)$.
		     \item  $f_{\phi_0,\phi_1}(t)=0 \Leftrightarrow t=0$.
	\end{enumerate}
	Choose $\phi_1(t)=t$ and $\phi_0\in\Qscr$. Then condition 2 above requires
	that 
	\[
		t_0\le t_1 \Rightarrow
		\psi\left(\frac{\phi_0(t_0)}{t_0}\right) \ge
		\psi\left(\frac{\phi_0(t_1)}{t_1}\right).
	\]
	But since $\phi_0\in\Qscr$, $t_0\le t_1\Rightarrow
	\frac{\phi_0(t_0)}{t_0} \ge \frac{\phi_0(t_1)}{t_1}$ and thus $\psi$
	must be nondecreasing.

	Now choose $\phi_1(t)=1$ and $\phi_0(t)=t$.  Then the second
	condition implies
	\[
		t_0\le t_1 \Rightarrow \frac{\psi(t_0)}{t_0}\ge
		\frac{\psi(t_1)}{t_1}.
	\]
	Furthermore, with the same choice for $\phi_0$ and $\phi_1$, we
	have $f_{\phi_0,\phi_1}(t)=0\Leftrightarrow t=0$ which implies that
	$\psi(t)=0\Leftrightarrow t=0$. We have thus shown all the required
	properties of quasiconcavity for $\psi$.
\end{proof}

\subsection{Interpolation Functors and their Fundamental Functions}
\label{app:interpolation}
\subsubsection{Proof of Lemma~\ref{lemma:ptefunctor}}
\label{app:interpolation:lemma:ptefunctor}
\begin{proof}
Recalling the definition of PTE (\ref{eq:PTE-def}),
we show that for any $X\in\mathcal{M}^+$ and $c\in\reals$ such that $X+c\ge 0$ that 
$\|X+c\|_{\Lambda_{\phi(\bar{\Xcal})}}=\|X\|_{\Lambda_{\phi}(\bar{\Xcal})}+c$, where $\bar{\Xcal}=(\Xcal_0,\Xcal_1)$.  
First consider the case that $c\ge 0$. 
Writing $\|\cdot\|_\Lambda=\|\cdot\|_{\Lambda_{\phi}(\bar{\Xcal})}$, 
$\|\cdot\|_0=\|\cdot\|_{\Xcal_0}$ and $\|\cdot\|_1=\|\cdot\|_{\Xcal_1}$  for brevity we have
\begin{align*}
\|X\|_\Lambda+c &= \inf\left\{\sum_{k}^K \phi(\|X_k\|_0,\|X_k\|_1) +c 
        \colon {X_k\in\Xcal_0+\Xcal_1,\  K \in \mathbb{N},\ X=\sum_{k}X_k}\right\}\\
    &=\inf \left\{\sum_{k}^K \phi(\|X_k\|_0,\|X_k\|_1) +\phi(c,c) 
         \colon {X_k\in\Xcal_0+\Xcal_1,\  K \in \mathbb{N},\ X=\sum_{k}X_k}\right\}\\
    \intertext{since $\phi(1,1)=1$ and $\phi$ is positively homogeneous,}
    &=\inf \left\{\sum_{k}^K \phi(\|X_k\|_0,\|X_k\|_1) +\sum_{k}\phi\left(\textstyle\frac{c}{K},\textstyle\frac{c}{K}\right) 
        \colon {X_k\in\Xcal_0+\Xcal_1,\  K \in \mathbb{N},\ X=\sum_{k}X_k}\right\}\\
    &=\inf \left\{\sum_{k}^K \left[\phi(\|X_k\|_0,\|X_k\|_1) +\phi\left(\textstyle\frac{c}{K},\textstyle\frac{c}{K}\right) \right] 
        \colon {X_k\in\Xcal_0+\Xcal_1,\  K \in \mathbb{N},\ X=\sum_{k}X_k}\right\}\\
    &\le\inf \left\{\sum_{k}^K \phi(\|X_k\|_0+\textstyle\frac{c}{K},\|X_k\|_1+\textstyle\frac{c}{K}) 
        \colon {X_k\in\Xcal_0+\Xcal_1,\  K \in \mathbb{N},\ X=\displaystyle\sum_{k}X_k}\right\} \\
    \intertext{since $\phi(z_1)+\phi(z_2)\le\phi(z_1+z_2) $ for arbitrary $z_1,z_2\in\reals_+^2$
        because $\phi$ is a concave gauge function \cite[Proposition 2.1]{barbara1994concave}}
    &= \inf \left\{\sum_{k}^K \phi(\|X_k+\textstyle\frac{c}{K}\|_0,\|X_k+\textstyle\frac{c}{K}\|_1) 
        \colon X_k\in\Xcal_0+\Xcal_1,\  K \in \mathbb{N},\ X=\displaystyle\sum_{k}X_k\right\}\\
    \intertext{since $\|\cdot\|_0$ and $\|\cdot\|_1$ are positive translation equivariant. 
        Now let $X_k'=X_k+\textstyle\frac{c}{K}$, for $k$ and thus $X_k=X_k'-\textstyle\frac{c}{K}$, $k$.  Thus}
    &= \inf \left\{\sum_{k}^K \phi(\|X_k\|_0,\|X_k\|_1) 
        \colon X_k'\in\Xcal_0+\Xcal_1,\ K \in \mathbb{N},\ X=\sum_{k}(X_k'-\textstyle\frac{c}{K})\right\}\\
    &= \inf \left\{\sum_{k}^K \phi(\|X_k\|_0,\|X_k\|_1)  
        \colon X_k'\in\Xcal_0+\Xcal_1,\  K \in \mathbb{N},\ X=\left(\sum_{k}X_k'\right)-c\right\}\\
    &= \inf \left\{\sum_{k}^K \phi(\|X_k\|_0,\|X_k\|_1) 
        \colon X_k'\in\Xcal_0+\Xcal_1,\  K \in \mathbb{N},\ X+c=\left(\sum_{k}X_k'\right)\right\}\\
    &=\|X+c\|_\Lambda.
\end{align*}
Since $\|\cdot\|_\Lambda$ is a norm it satisfies the triangle inequality 
$\|X+c\|_\Lambda\le\|X\|_\Lambda+\|c\|_\Lambda=\|X\|_\Lambda+c$. Thus combining with the above  we have 
\[
    \|X\|_\Lambda +c \le \|X+c\|_\Lambda \le \|X\|_\Lambda+c
\]
and thus $\|X+c\|_\Lambda=\|X\|_\Lambda +c$ as required.

If instead  we have $c<0$ but $X+c\ge 0$, then there exists some $c_0\ge 0$ such that $c_0\ge -c$
and $X=X_0+c_0$, with $X_0\in\mathcal{M}_+$. Consequently,
\[
\|X+c\|_\Lambda = \|X_0+c_0+c\|_\Lambda=\|X_0+(c_0+c)\|_\Lambda\stackrel{(*)}{=} \|X_0\|_\Lambda+(c_0+c) \stackrel{(**)}{=} \|X_0+c_0\|_\Lambda +c =\|X\|_\Lambda+c,
\]
where  (*)  holds from the case already shown, since $X_0\ge 0$ and $(c_0+c)\ge 0$, and (**) holds similarly (since $c_0\ge 0$).
\end{proof}

\subsubsection{Proof of Lemma~\ref{lemma:ff-abstract-marcinkiewicz}}
\label{app:interpolation:lemma:ff-abstract-marcinkiewicz}
\begin{proof}
	For $t>0$, we need to compute 
	\[
		\phi_{M_{\bar{\phi}}(\bar{\Xcal})}(t)=\|\chi_{[0,t]}\|_{M_{\bar{\phi}}(\bar{\Xcal})} = \sup_{s_0,s_1 \ge 0}
		\frac{K(s_0,s_1,\chi_{[0,t]},\bar{\Xcal})}{\bar{\phi}^*(s_0,s_1)}.
	\]
	Thus for arbitrary $s_0,s_1$ we need to determine
	\begin{align*}
		K(s_0,s_1,\chi_{E_t},\bar{\Xcal}) & = 
		\inf (s_0\|X_0\|_{\Xcal_0} + s_1 \|X_1\|_{\Xcal_1}),\\
	\intertext{with the infimum taken over all $X_0,X_1$ such
		$X_0+X_1=\chi_{E_t}$, where $E_t$ is chosen such that 
		$\mu(E_t)=t$.    Since by Theorem \ref{theorem:sandwich}, the Marcinkiewicz norm minorises any 
		ri norm with fundamental function $\phi_\Xcal$: 
		$\|X_i\|_{M_{\bar{\phi}}} \le \|X\|_\Xcal$, we have} 
	 K(s_0,s_1,\chi_{E_t},\bar{\Xcal}) & \ge \inf_{X_0+X_1=\chi_{E_t}}
		s_0\|X_0\|_{M_{\phi_0}}+s_1\|X_1\|_{M_{\phi_1}},\\
	\intertext{and since $\|X\|_{M_{\bar{\phi}}}=\displaystyle\sup_{0<r<\infty}
		X^{**}(r)\bar{\phi}(r)$, we have}
	 K(s_0,s_1,\chi_{E_t},\bar{\Xcal}) &\ge 
	        \inf_{X_0+X_1=\chi_{E_t}} s_0\left(\sup_{0<r<\infty}
		X_0^{**}(r)\phi_0(r)\right) +s_1\left(\sup_{0<r<\infty}
		X_1^{**}(r)\phi_1(r)\right)\\
	&\ge \inf_{X_0+X_1=\chi_{E_t}} \sup_{0<r<\infty} 
		\left(s_0\phi_0(r)X_0^{**}(r)+s_1\phi_1(r) X_1^{**}(r)\right)\\
		\intertext{and by choosing $r=t$ we obtain}
	K(s_0,s_1,\chi_{E_t},\bar{\Xcal})&\ge \inf_{X_0+X_1=\chi_{E_t}} (c_0 X_0^{**}(t) +c_1
		X_1^{**}(t)),\\
	\intertext{where $c_0=s_0\phi_0(t)$ and $c_1=s_1\phi_1(t)$
		are constants (since $t$ is fixed),}
	& = \inf_{X_0+X_1=\chi_{E_t}} \left( (c_0 X_0)^{**}(t) +
		(c_1 X_1)^{**}(t)\right).\\
	\intertext{But  $(f+g)^{**}(t) \le f^{**}(t) + g^{**}(t)$
		for all $t>0$ and any $f,g$, and so}
	 K(s_0,s_1,\chi_{E_t},\bar{X}) &\ge  
	      \inf_{X_0+X_1=\chi_{E_t}} (c_0 X_0 +c_1 X_1)^{**}(t)).\\
	\intertext{Since for any $f$ we have $f^*(t)\le f^{**}(t)$, for all
		$t$, we have}
	K(s_0,s_1,\chi_{E_t},\bar{\Xcal}) & \ge
	       \inf_{X_0+X_1=\chi_{E_t}} (c_0 X_1 +c_1 X_1)^*(t)\\
	&= \inf_{X_0+X_1=\chi_{E_t}} \inf\{\lambda\colon \mu_{c_0 X_0+c_1
		X_1}(\lambda)\le t\}\\
	&=\inf_{X_0+X_1=\chi_{E_t}} \inf\{\lambda\colon
		\mu\{s\in\reals\colon(c_0 X_0+c_1 X_1)(s)>\lambda\}\le t\}.\\
	\intertext{Now let $A_\lambda\coloneqq \mu\{s\in\reals\colon(c_0 X_0+c_1
		X_1)(s)>\lambda\}$ and 
		$B_\lambda\coloneqq \mu\{s\in\reals\colon(c_0\wedge c_1)(X_0
		+X_1)(s)>\lambda\}$.  Since $(c_0\wedge c_1)(X_0+X_1)=(c_0\wedge c_1)X_0
		+(c_0\wedge c_1)X_1\le c_0 X_0 +c_1 X_1$ we have that
		$B_\lambda\le A_\lambda$ for all $\lambda$. Furthermore,
		$\lambda\mapsto A_\lambda$ and $\lambda\mapsto B_\lambda$
		are nonincreasing and thus
		$\inf\{\lambda\colon A_\lambda \le t\} \ge
		\inf\{\lambda\colon B_\lambda \le t\}$, and hence}
	K(s_0,s_1,\chi_{E_t},\bar{\Xcal}) &\ge \inf_{X_0+X_1=\chi_{E_t}} \inf\{\lambda\colon
		\mu\{s\in\reals\colon(c_0\wedge
		c_1)(X_0+X_1)(s)>\lambda\}\le t\}\\
	&= \inf_{X_0+X_1=\chi_{E_t}} (c_0\wedge c_1)(X_0+X_1)^*(t)\\
	&= (c_0\wedge c_1) \chi_{[0,t]}(t)\\
	&=c_0\wedge c_1\\
	&=s_0\phi_0(t) \wedge s_1\phi_1(t).
	\end{align*}

	The infimum in the definition of $K$ is in fact attained by
	choosing 
	$X_0=\alpha\chi_{[0,t]}$ and
	$X_1=(1-\alpha)\chi_{[0,t]}$ for some $\alpha\in[0,1]$.  

	In this case we have
	\begin{align*}
		& \inf_{\alpha\in[0,1]} s_0\|\alpha\chi_{E_t}\|_{\Xcal_0} +
		s_1\|(1-\alpha)\chi_{E_t}\|_{\Xcal_1}\\
		=& \inf_{\alpha\in[0,1]} s_0\alpha\phi_0(t)+s_1(1-\alpha)\phi_1(t)\\
		=& s_0\phi_0(t)\wedge s_1\phi_1(t).
	\end{align*}
	Thus $K(s_0,s_1,\chi_{[0,t]},\bar{\Xcal}) = s_0\phi_0(t)\wedge s_1\phi_1(t).$

	Consequently
	\[
	\|\chi_{[0,t]}\|_{M_{\bar{\phi}}(\bar{\Xcal})} = \sup_{s_0,s_1\ge 0}
		\frac{s_0\phi_0(t) \wedge s_1\phi_1(t)}{\bar{\phi}^*(s_0,s_1)}.
	\]
	Noting that both numerator and denominator are positively
	homogeneous in $(s_0,s_1)$ it suffices to enforce $s_0+s_1=1$ and
	thus by setting $s_0=s$ and $s_1=(1-s)$ for $s\in[0,1]$,
	\[
		\|\chi_{[0,t]}\|_{M_{\bar{\phi}}(\bar{\Xcal})} = \sup_{s\in[0,1]} 
		\frac{s\phi_0(t) \wedge (1-s)\phi_1(t)}{\bar{\phi}^*(s,1-s)}.
	\]
	Now $\bar{\phi}^*(\alpha,\beta)=\beta\bar{\psi}(\alpha/\beta)$ for some
	$\bar{\psi}\in\Qscr$ and so $\bar{\phi}^*(s,1-s)=(1-s)\bar{\psi}(s/(1-s))$.
	Furthermore $s\phi_0(t)\wedge (1-s)\phi_1(t)$ can be written as 
	\begin{align*}
		s\phi_0(t) \ \ \ \ &\mbox{if\ } s\phi_0(t)\le(1-s)\phi_1(t)\\
		(1-s)\phi_1(t) \ \ \ \ &\mbox{if\ }
		s\phi_0(t)\ge(1-s)\phi_1(t).
	\end{align*}
	Setting $\gamma\coloneqq \phi_1(t)/\phi_0(t)$, we have
	$s\phi_0(t)\le (1-s)\phi_1(t) \Leftrightarrow s/(1-s) \le \gamma
	\Leftrightarrow  s\le \gamma/(1+\gamma)$.  Hence
	\[
	s\phi_0(t)\wedge (1-s)\phi_1(t) = \left\{\begin{array}{ll}
		s\phi_0(t),\ \ \  & s\le \frac{\gamma}{1+\gamma}\\
		(1-s)\phi_1(t),\ \ \  & s\ge \frac{\gamma}{1+\gamma}
	\end{array}\right. .
	\]
	Hence 
	\[
	\phi_{M_{\bar{\phi}}(\bar{\Xcal})}(t) = \|\chi_{[0,t]}\|_{M_{\bar{\phi}}(\bar{\Xcal})} 
	      = \min\left( \sup_{s\le
			\frac{\gamma}{1+\gamma}}
			\frac{s\phi_0(t)}{(1-s)\bar{\psi}\left(\frac{s}{1-s}\right)} , 
			\sup_{s\ge \frac{\gamma}{1+\gamma}} \frac{(1-s)
			\phi_1(t)}{(1-s)\bar{\psi}\left(\frac{s}{1-s}\right)}\right).
	\]
	Since $\phi_0(t),\phi_1(t)\ge 0$, we only need to determine
	\[
		a\coloneqq\sup_{0\le s\le \frac{\gamma}{1+\gamma}} f(s) 
		\ \ \ \mbox{and}\ \ \
		b\coloneqq\sup_{\frac{\gamma}{1+\gamma} \le s <\infty} g(s),
	\]
	where $f(s)=\frac{s}{(1-s)\bar{\psi}(s/(1-s))}$ and
	$g(s)=\frac{1}{\bar{\psi}(s/(1-s))}$, and $\phi_{M_{\bar{\phi}}(\bar{\Xcal})}(t)=a\phi_0(t)\wedge
	b\phi_1(t)$. Considering $f$ first, and setting
	$t\coloneqq s/(1-s)$ and so $s=t/(1+t)$ we have
	\[
		a=\sup_{t\in[0,\gamma]} \frac{t}{\bar{\psi}(t)}.
	\]
	But $\bar{\psi}$ is quasiconcave and thus $t\mapsto\bar{\psi}(t)/t$ is positive and  
	nonincreasing
	and so $t\mapsto t/\bar{\psi}(t)$ is nondecreasing and the supremum is attained
	at $t=\gamma$ and $a=\gamma/\psi(\gamma)$. Similarly for $g$, we have
	\[
		b=\sup_{t\ge \gamma} \frac{1}{\bar{\psi}(t)}.
	\]
	Since $\bar{\psi}$ is quasiconcave,  it is positive and nondecreasing and so $t\mapsto
	1/\bar{\psi}(t)$ is nonincreasing and the supremum is attained at $t=\gamma$ and
	$b=1/\bar{\psi}(\gamma)$. Recalling $\gamma=\phi_1(t)/\phi_0(t)$, we have
	\begin{align}
		\|\chi_{[0,t]}\|_{M_{\bar{\phi}}(\bar{\Xcal})} &=
		\min\left(\frac{\gamma\phi_0(t)}{\bar{\psi}(\gamma)},
		\frac{\phi_1(t)}{\bar{\psi}(\gamma)}\right) =
		\min\left(\frac{\phi_1(t)}{\bar{\psi}(\gamma)},
		    \frac{\phi_1(t)}{\bar{\psi}(\gamma)}\right)\\
		&=
		\frac{\phi_1(t)}{\bar{\psi}(\phi_1(t)/\phi_0(t))}=
		\frac{\phi_1(t)\phi_0(t)}{\bar{\phi}^*(\phi_1(t),\phi_0(t))}.
	\end{align}
	But $\bar{\phi}^*$ is positively homogeneous, and thus 
	\begin{align}
		\|\chi_{[0,t]}\|_{M_{\bar{\phi}}(\bar{\Xcal})} &= \frac{\phi_1(t)\phi_0(t)}{\bar{\phi}^*(\phi_1(t),\phi_0(t))} = 
		\phi_1(t)\phi_0(t)
		\bar{\phi}\left(\frac{1}{\phi_1(t)},\frac{1}{\phi_0(t)}\right) \\ &=
		\bar{\phi}\left(\frac{\phi_1(t)\phi_0(t)}{\phi_1(t)},
		\frac{\phi_1(t)\phi_0(t)}{\phi_0(t)}\right) =
		\bar{\phi}(\phi_0(t),\phi_1(t)).
	\end{align}
\end{proof}

\subsubsection{Proof of Lemma~\ref{lemma:ff-abstract-lorentz}}
\label{app:interpolation:lemma:ff-abstract-lorentz}
\begin{proof}
	We have
	\begin{align*}
		\phi_{\Lambda_{\bar{\phi}}(\bar{\Xcal})}(t) &=\|\chi_{E_t}\|_{\Lambda_{\bar{\phi}}(\bar{\Xcal})},\\
		\intertext{where $E_t$ is such that $\mu(E_t)=t$,} 
		&= \inf_{(X_k)_k\colon \chi_{E_t}=\sum_k X_k} 
			\sum_k \bar{\phi}(\|X_k\|_{\Xcal_0},\|X_k\|_{\Xcal_1}).\\
		\intertext{Taking the particular choice $X_1=\chi_{[0,t]}$ and $X_k=0$
		for $k>1$, gives an upper bound on the infimum:}
		\phi_{\Lambda_{\bar{\phi}}(\bar{\Xcal})}(t) & \le 
			\bar{\phi}(\|\chi_{E_t}\|_{\Xcal_0},\|\chi_{E_t}\|_{\Xcal_1})\\
		& = \bar{\phi}(\phi_0(t),\phi_1(t)).
	\end{align*}
\end{proof}

\section{(Non)-Expected Utility Theories}
\label{app:utility}
Coherent risk measures in finance are intimately connected with generalized utility theories in rational choice theory, situated in the context of economics. These theories offer formal axiomatic bases for rational decision making under uncertainty. Our motivation is that we take the following two ideas seriously: empirical risk minimization (ERM) in machine learning is a decision problem not only under \textit{risk}, but also under \textit{ambiguity}, and a loss function is an outcome-contingent \textit{disutility} \citep{berger1985statistical}.

In the ERM problem, the decision maker, \ie the machine learning engineer, faces the problem of summarizing the loss distribution in a single number, which is then employed in a minimization routine. This summary is typically the expectation, reducing to summation under the empirical distribution. Loss is a disutility in the sense that the decision maker wants to have as little as possible of it. Therefore, modulo a sign flip, loss minimization can be described in the framework of expected utility theory, where the aim is to maximize utility in an uncertain setting.

First, if we knew the `true' probability distribution, risk minimization is indeed a decision problem under risk. We can model a risky situation with a probability distribution. In an economics context, the analogy is a choice for the decision maker between different \textit{lotteries} with known probabilities. Think for instance of a coin flip. Probability theory was historically developed to handle such decisions under risk, where probabilities are ``well-behaved'': relative frequencies are stable and can be known, \eg by combinatorial arguments \citep{hacking1990taming}.

In contrast to risk is the challenge of ambiguity or, in the extreme, Knightian uncertainty. These
are ``non-probabilized'' forms of uncertainty \citep{etner2009decision} and cannot be captured by a single probability distribution. In \textit{empirical} risk minimization, we have no good reason to believe that the observed data perfectly represents the `true' distribution. Moreover, in a dynamically changing environment, such a stable distribution may not exist \citep{gorban2017statistical}. Hence it may be better to assume a whole set of probability distributions to represent the belief of the decision maker, from a subjectivist view, or to represent the behavior of the loss sequence, from a frequentist view\footnote{For a frequentist interpretation of coherent upper probabilities, see \citep{walley1982towards,frohlich2023towards}. Whereas precise probabilities model converging sequences of relative frequencies, coherent upper probabilities can be linked to sequences whose relative frequencies diverge within an interval, whose boundaries are given by the lower and upper probability.}.

In the presence of risk and ambiguity, different attitudes are conceivable: a decision maker might be risk loving, risk-neutral, risk-averse and ambiguity-loving, ambiguity-neutral, ambiguity-averse. We shall focus on a risk-averse and ambiguity-averse attitude. First, we frame the standard ERM problem in the framework of expected utility theory. Throughout, we make the translation to a loss-based formulation. Our aim is to demonstrate the limitations of the classical approach and illuminate attractive alternatives. In particular, we will find yet more ways to arrive at the classes of coherent and spectral risk measures. These new perspectives offer additional motivation for why employing a coherent or spectral risk measure in place of the expectation is normatively permissible and motivated.

\subsection{Expected Utility}
Classical expected utility theory comes in two flavors: objective and subjective. In the objective setting developed by \cite{von1947theory}, the decision maker chooses between lotteries, which yield specified losses/rewards with known probabilities. In contrast, in the subjective setting of \cite{savage1954foundations}, the decision maker does not know the probability measure a priori. The two formulations differ mainly in interpretation. Mathematically they are closely related. For ease of exposition, we focus on von Neumann's framework. We refer to \cite{follmer2016stochastic} for a detailed account. Denote by $C$ a set of possible consequences. Typically, $C=\mR$ and we interpret the elements as monetary outcomes. Assume some $\sigma$-algebra $\mathcal{F}$ is given on $C$. Let $\mathcal{P}$ denote the set of probability distributions over $C$ with finite support.
\begin{equation}
    \mathcal{P} = \{P : \mathcal{F} \rightarrow [0,1] : P(\{c\}) \neq 0 \text{ only for finitely many } c \in C, P(C)=1\}.
\end{equation}
An element $P \in \mathcal{P}$ is called a \textit{lottery}. We characterize a decision maker by her preference relation $\succcurlyeq$ on $\mathcal{P}$. The meaning of $X \succcurlyeq Y$ is that the lottery $P$ is preferred over the lottery $Q$. Similarly, $P \sim Q$ denotes indifference and $P \succ Q$ strict preference. We say that a preference relation $\succcurlyeq$ is represented by a functional $R: \mathcal{P} \rightarrow \mR$ if
\begin{equation}
    P \succcurlyeq Q \Longleftrightarrow R(P) \leq R(Q),
\end{equation}
Assume that $\succcurlyeq$ satisfies the following axioms:
\begin{enumerate}[label=\textbf{N\arabic*.}, ref=N\arabic*]
  \item \label{item:N1} $\forall P,Q \in \mathcal{P}: P \succcurlyeq Q $ or $Q \succcurlyeq P$ or both. \quad (completeness)
  \item \label{item:N2} $\forall P,Q,S \in \mathcal{P}:  P\succcurlyeq Q$, $Q \succcurlyeq S \Rightarrow P \succcurlyeq S$ \quad (transitivity)
  \item \label{item:N3} $\forall P,Q,S \in \mathcal{P} : \exists \alpha, \beta \in (0,1): \alpha P + (1-\alpha) S \succ Q \succ \beta P + (1-\beta) S$ \quad (Archimedean)
  \item  \label{item:N4} $\forall P,Q,S \in \mathcal{P}: \forall \alpha \in (0,1]: P \succ Q \Rightarrow \alpha P + (1-\alpha) S \succ \alpha Q + (1-\alpha)S$ \quad (independence)
\end{enumerate}
While the first three axioms are relatively uncontroversial and common to different theories of rational choice, it is the independence axiom which characterizes the theory. The independence axiom, essentially equivalent to the \textit{sure thing principle} in \cite{savage1954foundations}, is an additivity principle. Intuitively, it means that the common component $(1-\alpha)S$ does not matter for the ranking \citep{al2010subjective}. Another way to express it is that preferences must be separable across mutually exclusive events \citep{denuit2006risk}.
\begin{theorem}
\citep{von1947theory}. If and only if \ref{item:N1}-\ref{item:N4} are satisfied, the preference relation $\succcurlyeq$ allows an affine representation with a loss function $\ell: C \rightarrow \mR$, unique up to an affine transformation:
\begin{equation}
    P \succcurlyeq Q \Longleftrightarrow \int_C \ell(\omega) \d P(\omega) \leq \int_C \ell(\omega) \d Q(\omega) \Longleftrightarrow \mathbb{E}_P[\ell] \leq \mathbb{E}_Q[\ell].
\end{equation}
\end{theorem}
Instead of a utility function $u$, which is typically used in the literature, we have expressed the theorem using its mirror image, the loss function $\ell(\omega) = -u(-\omega)$.\footnote{This would correspond to also flipping the gain vs.\@\xspace loss orientation of the $\omega$ (see \eg \citet{rockafellar2013fundamental}). This presupposes that $C$ supports a ``$-$'' operation. An alternative would be to set $\ell(\omega)=-u(\omega)$, which would keep the orientation.} Instead of expected utility maximization, our decision maker aims for expected loss minimization. 

The celebrated theories of \cite{von1947theory} and \cite{savage1954foundations} have become deeply entrenched in economics and spread into other disciplines in the course of the 20th century. The crucial ingredient is the independence axiom, which corresponds to additivity of the representation. The structure of such a representation implies a strict separation of \textit{belief} and \textit{action} (or \textit{taste}, in the language of \cite{al2010subjective}). Belief is embodied by the probability distribution; action relates to the choice of the loss function, which specifies the attitude of a decision maker towards outcomes. These two separate domains are then conjoined using the expectation operator. We remark that this separability is at the basis of challenges which have been raised against classical expected utility, such as the Ellsberg's urns \citep{ellsberg1961risk} and Allais'  paradox \citep{allais1953comportement}. The non-expected utility theories which we will consider refrain from making the separation to this extent.

Related is the issue of risk aversion: in expected utility theory, attitudes toward wealth (outcomes $c \in C$) and probabilities are forever bound together. The standard definition of (weak) risk aversion is that 
\begin{align}
    \forall P \in \mathcal{P}: \mathbb{E}_P[id] &\succcurlyeq P,\\
    \Leftrightarrow \ell\left(\int_C \omega \d P(\omega)\right) &\leq \int_C \ell(\omega) \d P(\omega),
\end{align}
where $id$ is the identity function and $\mathbb{E}_P[id]$ is a constant lottery which yields the loss $\mathbb{E}_P[id]$ with probability $1$. It is a classical result that risk aversity holds if and only if $\ell$ is a convex function, that is, when $u$ is concave. For an expected utility decision maker, risk aversity is synonymous with diminishing marginal utility of wealth \citep{denuit2006risk}, expressed via the utility function. Diminishing marginal utility is the phenomenon that an increase at a higher wealth level is valued less than the same increase at a lower wealth level. This is modelled by a concave utility function, which has a convex loss function as its mirror image. In machine learning, it is customary to employ convex loss functions (with respect to the predictions, not necessarily the parameters) such as the squared loss. This captures the wish to increase punishment the farther the prediction is from the ground truth. Thus, one has in a sense automatically implemented this form of risk aversion. Since there is no sensible unit of `wealth' in machine learning to establish the analogy to economics, we instead fix a loss function a priori. Then we consider the loss values as making up the space of consequences $C$ and apply expected utility theory with the identity $id$ as a ``loss function''. In this way, classical expected utility yields the familiar problem of expected risk minimization.

We find this kind of risk aversion too weak; it does not actually capture a risk-averse attitude \citep{buchak2013risk}. Aversion to risk in the sense of unpredictability seems prima facie different from diminishing marginal utility towards wealth, yet in classical expected utility theory they are conflated. A decision maker who has a diminishing attitude towards the amount of some commodity even under certainty seems prima facie rational \citep[see][for this line of argument]{buchak2013risk}. When then uncertainty enters the picture, the decision maker might display additional risk aversion, a disinclination to take a risky bet on the commodity of interest, which is not exhausted by the concave utility function. To us, risk aversion amounts to encoding an attitude towards the \textit{probability} itself: a decision maker might prefer a distribution which is less spread-out over a distribution with higher spread, even given that they have the same mean. Risk aversion is asymmetric, however: unexpected high gain is not as problematic as unexpected high loss. It is not clear why the only reason for this preference should arise from diminishing marginal utility instead of from an aversion to the inherent risk.

Another criticism of classical expected utility is \textit{overprecision}, the ``excessive faith that you know the truth'' \citep{moore}. 
We alluded to this problem in Section~\ref{sec:riskuncertainty} and Section~\ref{sec:walley}. Using a single probability measure expresses precise belief, when instead sometimes a degree of ignorance is warranted by the available evidence. \cite{gilboa2009always} write: ``The Bayesian approach is lacking because it is not rich enough to describe one’s degree of confidence in one’s assessments''. Here, the Bayesian approach refers to Savage's axiomatization. Along similar lines, Keynes observes that ``new evidence will sometimes decrease the probability of an argument, but it will always increase its `weight''' \citep[p.\@\xspace 78]{keynes1921treatise}. A decision maker following Savage's axioms has a precise belief concerning the probability that  right now \textit{24 men in Bulgaria are standing on their heads} \citep{schoenfield2012chilling}, down to arbitrary precision. Furthermore, she would be willing to take bets both on and against this event, where the betting rate is the specified precise probability. In contrast to this behaviour, a lack of knowledge rather warrants \textit{ambiguity aversion}, a certain pessimism in the face of non-probabilized uncertainty. Hence we now turn to \textit{maxmin expected utility}, closely related to imprecise probability.

\subsection{Maxmin Expected Utility}
An influential generalization of expected utility, \textit{maxmin expected utility}, has been put forward by \cite{gilboa1989maxmin}. Following \cite{anscombe1963definition}, they work with a two-stage model, comprising both objective and subjective probabilities. Let $\Omega$ denote a set of outcomes. Let $S$ denote a set comprising the \textit{states of nature} and let $\mathcal{F}$ an algebra of subsets of $S$ closed under finite intersections. By $\mathcal{P}$, we denote the set of probability distributions on $\Omega$ with finite support, \ie \textit{lotteries} with objective probabilities:
\begin{equation}
    \mathcal{P} = \{P : \Omega \rightarrow [0,1] : P(\omega) \neq 0 \text{ only for finitely many } \omega, \sum_{\omega \in \Omega} P(\omega)=1 \}.
\end{equation}
An \textit{act} is a function $X: S \rightarrow \mathcal{P}$ belonging to some specified convex set of acts $\cL$, which includes constant functions. We denote the set of constant acts as $\cL_c$. Convex combinations of acts are performed pointwise: let $X,Y \in \cL$. Then $\alpha X + (1-\alpha)Y = \omega \mapsto \alpha X(\omega) + (1-\alpha)Y(\omega)$. The goal is to obtain a subjective probability about acts, sometimes called \textit{horse lotteries} by leveraging the objective probabilities through the preference relation. 
\cite{gilboa1989maxmin} impose the following axioms:
\begin{enumerate}[label=\textbf{M\arabic*.}, ref=M\arabic*]
  \item \label{item:M1} $\forall X,Y \in \cL: X \succcurlyeq Y $ or $X \succcurlyeq Y$ or both. \quad (completeness)
  \item \label{item:M2} $\forall X,Y,Z \in \cL:  X\succcurlyeq Y$, $Y \succcurlyeq Z \Rightarrow X \succcurlyeq Z$ \quad (transitivity)
  \item \label{item:M3} $\forall X,Y,Z \in \cL$: if $X \succ Y$ and $Y \succ Z$ then $\exists \alpha,\beta \in (0,1):$\\$\alpha X + (1-\alpha)Z \succ Y$ and $Y \succ \beta X + (1-\beta)Z$ \quad (continuity)
  \item \label{item:M4} If $\forall \omega \in \Omega: \omega' \mapsto X(\omega) \succcurlyeq \omega' \mapsto Y(\omega)$ then $X \succcurlyeq Y$ \quad (monotonicity) [sic]
  \item  \label{item:M5} $\forall X,Y \in \cL, c \in \cL_c, \alpha \in (0,1): X \succ Y \Rightarrow \alpha X + (1-\alpha) c \succ \alpha Y + (1-\alpha)c$ \quad (c-independence)
  \item  \label{item:M6} $\forall X,Y \in \cL, \alpha \in (0,1): X \sim Y \Rightarrow \alpha X + (1-\alpha)Y \succcurlyeq X$\quad (ambiguity aversion)
  \item  \label{item:M7} not for all $X,Y \in \cL: X \succcurlyeq Y$ (non-degeneracy)
\end{enumerate}
where $X \sim Y$ denotes the indifference relation, \ie $X \succcurlyeq Y$ and $Y \succcurlyeq X$, and $\succ$ is the strict part of the relation. Certainty independence (c-independence) is strictly weaker than independence; it only requires the separability with respect to constants. As a consequence of this axiom, \cite{gilboa1989maxmin} obtained the following representation.
\begin{theorem} \citep{gilboa1989maxmin}.
If and only if the preference relation $\succcurlyeq$ satisfies M1-M7 then it allows a representation of the form 
\begin{equation}
    X \succcurlyeq Y \Longleftrightarrow \max_{Q \in \mathcal{Q}} \int \ell \circ X \d Q \leq \max_{Q \in \mathcal{Q}} \int \ell \circ Y \d Q
\end{equation}
for a loss function $\ell: \mathcal{P} \rightarrow \mR$ defined at the level of lotteries and a non-empty closed convex set $\mathcal{Q}$ of finitely additivity probability measures on $\mathcal{F}$.
\end{theorem}
Because of the translation to losses, it would be more appropriate to call it \textit{minmax} expected loss in our context. Observe that this is essentially nothing but a two-stage formulation of Walley's upper previsions (coherent risk measures) and with a loss function entering the picture. When the loss function is the identity and the functional $R(X) = \max_{Q \in \mathcal{Q}} \int \ell \circ X \d Q$ is applied only to acts which yield degenerate constant lotteries, \ie $X(s) = c_s \in \mathcal{P}$, we recover an upper prevision (coherent risk measure). Maxmin expected utility has also recently been formalized in a single-stage subjective setting \citep{al2010subjective}. We have chosen to present the two-stage formulation since the axiomatization is simpler there.

A decision maker who adopts the maxmin axioms takes a worst-case stance towards a set of probability measures considered as relevant candidates. For this, the crucial axiom is \textit{ambiguity aversion}. The intuition behind it is that, in financial terms, hedging against ambiguity is preferred. Consider some $X \sim Y$, which are both ambiguous, \ie objective probabilities are not known. Ambiguity aversion states that then a convex combination $\alpha X + (1-\alpha)Y$ is weakly preferred over $X$ or $Y$. Possibly, $X$ acts as a hedge against $Y$ or vice versa, that is, $X$ tends to yield losses for those states where $Y$ tends to yields gains. In extreme cases, a convex combination of such acts can even reduce the ambiguous situation to a risky one with known probabilities (see \eg \citep{follmer2015axiomatic} or \citep{etner2009decision} for examples). On the other hand, no hedging is possible when $X$ and $Y$ are comonotone, since they then share the same rank ordering of outcomes\footnote{We here rely on the intuitive understanding of perfect rank correlation. For the definition of comonotonicity when outcomes are lotteries see \citep{schmeidler1989subjective}.}.
Ambiguity aversion states that, irrespective of the concrete $X$ and $Y$, hedging can at least never be strictly worse for the decision maker. The next theory, a close cousin of maxmin expected utility, takes the idea that comonotonicity prevents hedging seriously.

\subsection{Choquet Expected Utility}
\label{sec:choquet}
While maxmin expected utility is closely related to coherent risk measures and upper previsions, \textit{Choquet expected utility} contains as important special cases the class of spectral risk measures. The theory was originally developed by \cite{schmeidler1989subjective} and like maxmin expected utility was set in the two stage model of \cite{anscombe1963definition}. 
However, for easier exposition we present the single-stage version of \cite{chateauneuf1994modeling}, translated to losses. This is in contrast to maxmin expected utility, where the single-stage version is significantly more complicated than the two-stage version. Consider a space $\Omega$ and a $\sigma$-algebra $\mathcal{F}$. The preference relation $\succcurlyeq$ is defined on the set $\cL$ of bounded, real-valued measurable functions on $\Omega$. \citet{chateauneuf1994modeling} proposes the following axioms (translated to losses):
\begin{enumerate}[label=\textbf{CH\arabic*.}, ref=CH\arabic*]
  \item \label{item:CH1} Completeness, transitivity and non-degeneracy
  \item \label{item:CH2} If $\forall \omega \in \Omega: Y(\omega) \geq X(\omega) \Rightarrow X \succcurlyeq Y$ \quad (monotonicity)
  \item \label{item:CH3} Continuity with respect to monotone uniform convergence, see \citep{chateauneuf1994modeling}.
  \item  \label{item:CH4} $\forall X,Y,Z \in \cL: $ If $X$ and $Z$ are comonotone, $Y$ and $Z$ are comonotone and  $X \sim Y$,\\ then $X + Z \sim Y + Z$ \quad (comonotonic independence),
\end{enumerate}
Compared to maxmin expected utility, certainty independence has here been strengthened to comonotonic independence and uncertainty aversion has been dropped. Recall again that two functions $X$ and $Y$ are comonotone if 
\begin{equation}
    (X(\omega)-X(\omega'))(Y(\omega)-Y(\omega')) \geq 0 \quad \forall \omega,\omega' \in \Omega
\end{equation}
A constant function is comonotone with any other function and therefore comonotonic independence implies certainty independence. In the words of \cite{chateauneuf1994modeling}, ``comonotonic independence requires the direction of preference to be retained under adding payments, provided hedging is not involved''. When $X$ and $Y$ are comonotone, neither can work as a hedge against the other due to perfect rank correlation. As a consequence, the ambiguity cannot be reduced in favor of risk. \cite{chateauneuf1994modeling} obtains the following representation result. Recall that a capacity is a set function with $\overline{\mu}(\emptyset)=0$ and $\overline{\mu}(\Omega)=1$,\footnote{The normalization $\overline{\mu}(\emptyset)=0$ is required for any capacity. Capacities with $\overline{\mu}(\Omega)=1$ are also called \textit{normalized capacities}. We impose $\overline{\mu}(\Omega)=1$ throughout, however, and simply call it a capacity.} which is monotone.
\begin{theorem} \label{theorem:choquet} \citep{chateauneuf1994modeling}.
If and only if $\succcurlyeq$ satisfies the above axioms, there exists a capacity $\overline{\mu}$ on $\mathcal{F}$ such that
\begin{equation}
    X \succcurlyeq Y \Longleftrightarrow \int X \d \overline{\mu} \leq \int Y \d \overline{\mu},
\end{equation}
where the \textit{Choquet integral} with respect to the capacity $\overline{\mu}$ is defined as
\begin{equation}
    \int X \d \overline{\mu} \coloneqq \int_{-\infty}^{0} \left[\overline{\mu}(\{X \geq x\}) - 1\right]\d x + \int_0^\infty \overline{\mu}(\{X \geq x\}) \d x.
\end{equation}
\end{theorem}
\begin{remark}\normalfont
In this single-stage formulation, no loss/utility function has entered the picture. Typically, Choquet expected utility refers to representations of the form $\int u \circ X \d \overline{\mu}$ with a utility function $u$. For instance, \cf the axiomatization of \cite{schmeidler1989subjective}. To us, this is not a relevant difference since in our machine learning setup the random variable $X$ directly represents a loss.
\end{remark}
 Compare this to \eqref{eq:distortion}, where the capacity is given as the composition of a concave function and a probability measure.
The capacity then determines whether the decision maker is ambiguity-averse, -neutral or -loving. Consider the uncertainy aversion axiom
\begin{enumerate}[resume,label=\textbf{CH\arabic*.}, ref=CH\arabic*]
  \item \label{item:CH5} $\forall X,Y,Z \in \cL:$ If $X \sim Y$ and $Y$ and $Z$ are comonotone, then $X + Z \succcurlyeq Y + Z$
\end{enumerate}
The intuition is that $Z$ cannot act as a hedge against $Y$, but it could possibly hedge against $X$, so the direction of preference turns at least weakly in favor of $X+Z$. This axiom is in some sense a combination of comonotonic independence and uncertainty aversion.
\begin{theorem} \citep{chateauneuf1994modeling}. 
If and only if $\succcurlyeq$ satisfies CH1-CH3 and CH5, then the representation of Theorem~\ref{theorem:choquet} holds and the capacity $\overline{\mu}$ is furthermore submodular, that is:
\begin{equation}
\overline{\mu}(A\cup B) +\overline{\mu}(A\cap B) \leq \overline{\mu}(A)+\overline{\mu}(B).
\end{equation}
\end{theorem}
To get an intuition for the Choquet integral, let us consider a finite space $\Omega=\{\omega_1,..,\omega_n\}$. Assume $X$ is a step function which takes on the values $x_1 \leq x_2 \leq .. \leq x_n$. Let $x_0=0$. Then the Choquet integral can be written as
\begin{equation}
    \int X \d \overline{\mu} = \sum_{i}^n (x_i - x_{i-1}) \overline{\mu}(\{X \geq x_i\}).
\end{equation}
If the capacity is a probability measure, this reduces to the usual expectation. The decision maker starts with the lowest loss value $x_1$ and then successively adds up the increments $x_i - x_{i-1}$, but weighted with the capacity. In particular a capacity need not be additive for disjoint events, which allows to model interaction effects such as hedging against ambiguity.

Consider the special case of a submodular capacity, which represents uncertainty aversion. For finite $\Omega$, submodularity is equivalent to this property of \textit{diminishing marginal returns}:
\begin{equation}
    \forall A \subseteq B \subset \Omega, c \notin B: \overline{\mu}(A \cup \{c\}) - \overline{\mu}(A) \geq  \overline{\mu}(B \cup \{c\}) - \overline{\mu}(B) .
\end{equation}
This expresses that adding an element to a smaller set results in a greater increase in decision weight. Consequently, large losses (where $\overline{\mu}(\{X \geq x\})$ is small) are emphasized.
Whereas risk aversion is expressed by a concave utility function in \citep{von1947theory}, ambiguity aversion is a submodular attitude towards \textit{probability} itself. Submodular capacities are also called concave, since they exhibit a similar diminishing marginal returns property as concave functions. Furthermore, recall that if the capacity is given as the composition of an increasing function and a probability measure, the capacity is submodular if and only if the function is concave (Section~\ref{sec:distortion}).

Choquet expected utility is closely related to maxmin expected utility. If and only if the capacity is submodular, then the Choquet integral is convex \citep{alfonsi2015simple} and the representation takes a maxmin form (minmax, in loss-based formulation), where the envelope is given by the \textit{core} of the capacity
\begin{equation}
    \operatorname{core}(\overline{\mu}) = \left\{P : P(A) \leq \overline{\mu}(A) \text{ } \forall A \in \mathcal{F} \text{, } P \text{ finitely additive probability measure}\right\}
\end{equation}
\begin{equation}
    \int X \d \overline{\mu} = \sup_{P \in \operatorname{core}(\overline{\mu})}\left\{\int_{-\infty}^{\infty} X \d P\right\}.
\end{equation}
The ambiguity aversion is directly related to the convexity of the functional.
The close relationship between maxmin expected utility (MMEU) and Choquet expected utility (CEU) has been concisely summarized by \cite{klibanoff2001characterizing}:
\begin{quote}
    Fundamentally, CEU decision makers view uncertainty in terms of (roughly) how states are ordered by an act's utility payoff. Given a set of acts which all induce the same ordering, a CEU decision maker acts exactly like an expected utility (and thus uncertainty neutral) decision maker. MMEU decision makers, in contrast, may view uncertainty not only in terms of ordering of states, but also in terms of how much better the payoff is in one state as opposed to another.
\end{quote}

Both MMEU and CEU are theories about uncertainty in the sense of ambiguity. A capacity in CEU contains both a component of \textit{belief} and \textit{action} \citep{diecidue2001intuition}, where \textit{action} refers to a decision attitude. However, for a general capacity, these components cannot be separated. This intertwining empowers Choquet expected utility to tackle problems of ambiguity in a broad sense; yet it also renders it somewhat impractical. Revisiting Ellsberg's urns (Section~\ref{sec:riskuncertainty}), CEU can indeed describe the ambiguity-averse preferences which most decision makers exhibit in this scenario \citep{schmeidler1989subjective}. Ellsberg's urns are challenging because they not only violate expected utility, but also \textit{probabilistic sophistication} \citep{etner2009decision,machina1992more}. A probabilistically sophisticated decision maker acts in accordance with a belief which can be captured by a probability measure, but uses it in a manner that can extend beyond classsical expected utility. Hence belief and action are still intertwined to some degree. For instance, a risk-averse decision maker might express beliefs with an underlying probability measure but decides in a way so as to put more weight on worse outcomes. In the setting of CEU, probabilistic sophistication implies that the capacity is given by a composition $\overline{\mu}=\phi \circ P$ of an increasing function and a probability measure. Indeed, we may equate probabilistic sophistication with law invariance (rearrangement invariance). If a probabilistically sophisticated CEU decision maker satisfies CH5, then $\phi$ is concave and the Choquet integral is therefore a spectral risk measure. However, the typical preference behaviour in Ellsberg's urns cannot be modelled by such a functional \citep{schmeidler1989subjective}. Some authors therefore identify probabilistic sophistication with ambiguity neutrality \citep{epstein1999definition}.
We think that this goes too far: for instance, a law-invariant spectral risk measure expresses risk aversion by aversion to hallucinated ambiguity. There is still the assumption of a base measure, on which belief rests, but the action component constructs an ambiguity set around this base measure. This amounts to blurring the line between risk aversion and ambiguity aversion. Under law invariance, their mathematical form is equivalent and we may interpret risk aversion as a form of ambiguity aversion with respect to an artificially constructed (`hallucinated') ambiguity set. We emphasize that we do not claim that risk and ambiguity are equivalent, but rather that risk aversion can be modelled via aversion to hallucinated ambiguity. Furthermore, in light of the Kusuoka representation, any coherent risk measure is a combination of the two, since it can be described as an ambiguity set over a risk spectrum. 

In summary, we advocate thinking of a direct relation between risk aversion and ambiguity: at one extreme of the spectrum, where the supremum risk measure embodies maximal risk aversion, it has the corresponding interpretation of the maximal ambiguity set, consisting of all\footnote{Probability measures which are absolutely continuous with respect to a base measure.} probability measures. At the other extreme, the expectation is risk neutral and is represented by the singleton ambiguity set $\{1\}$. Hence, whether a law invariant coherent risk measure should be seen as modelling risk or ambiguity depends on the context and the modelling intentions of the decision maker.
Therefore we will now examine Choquet expected utility under probabilistic sophistication (law invariance), where the capacity can be decomposed into a belief and action attitude. In the concave case, this yields the class of spectral risk measures.


\subsection{Rank Dependent Expected Utility}
The crucial difference between CEU and \textit{rank dependent expected utility} (RDEU) is the additional requirement of law invariance (probabilistic sophistication, rearrangement invariance). Therefore, RDEU is typically viewed as CEU under risk. We pointed out, however, that this can also be viewed as theory of hallucinated ambiguity. Different authors have arrived at variants of RDEU  \citep{yaari1987dual,wang1995insurance,quiggin2012generalized,buchak2013risk}, which turned out to approximately coincide. RDEU represents preferences by law invariant Choquet integrals:
\begin{equation}
    X \succcurlyeq Y \Longleftrightarrow \int \ell \circ X \d \thinspace (\phi \circ P) \leq \int \ell \circ Y \d \thinspace (\phi \circ P),
\end{equation}
with a loss function $\ell$ and where the capacity is specialized as the composition of an increasing function $\phi$ and a probability measure $P$. Hence, risk aversion (submodularity) of the capacity is equivalent to the concavity of $\phi$ (\cf Section~\ref{sec:distortion}). If the loss function is the identity, as we take it in machine learning\footnote{Note again that in our machine learning setup $X(\omega)$ already corresponds to a loss value.}, then we recover exactly the class of spectral risk measures.

RDEU is \textit{rank dependent}, since the weight of a certain outcome in the decision not only depends on its probability via $P$, but also on how it is ranked with respect to other outcomes. This enables the decision maker to express a desire for \textit{distributional objectives} \citep{lopes1984risk}. Given a fixed mean, decision makers may prefer a less spread-out distribution as compared to a more spread-out one. The exact nature of this tradeoff is encoded in the function $\phi$, which can be considered a risk aversion profile. In the context of machine learning, it allows us to emphasize the largest losses to increase robustness.




Of particular interest to us is the rank dependent account of \cite{buchak2013risk}, which is called \textit{risk weighted expected utility}. \cite{buchak2013risk} aims to provide argumentative ground for why risk attitudes via rank dependence are normatively permissible, instead of only empirically adequate. Other authors are less clear on this issue or take a different stance. For instance, a slight variant of RDEU, \textit{prospect theory}, is only defended as a descriptive theory \citep{tversky1992advances}. Furthermore, \cite{buchak2017taking} has also considered the theory in the setting of social choice, which is relevant to fair machine learning. The possible application of rank dependence in this context has been hinted at by other authors \citep{schmeidler1989subjective,quiggin2012generalized}, but not elaborated.

\subsection{Rational and Social Choice with Spectral Risk Measures}
Rational choice is about an individual decision maker in a context where the decision affects only that individual. This can be modelled with a state space $\Omega$, where each $\omega \in \Omega$ represents a possible state of the world. A gamble $X: \Omega \rightarrow \mathbb{R}$ assigns to each state a resulting loss to the decision maker, given that this state is realized. In classical probability, such a gamble is evaluated via the expectation. This is the standard ML problem, where the engineer aims to minimize loss. By contrast, social choice concerns collective decision by a combination of individual preferences. This is closer to the model for a fair ML problem, where the individuals are salient subgroups. However, in ML the engineer chooses the loss function for everyone, whereas in the ``real world'' setting, individuals might have different loss functions.

The structural analogy is that a state $\omega$ in a rational choice problem corresponds to an individual (subgroup) in social choice \citep{buchak2017taking} and a gamble then describes a social arrangement (``who gets what''). Expected utility theories ask the question \textit{how should an individual value a gamble?} and the classical theory gives the expectation as the unique answer, whereas we have demonstrated that there exists a variety of interesting alternatives. Social choice theory asks: which social arrangements are to be preferred (or \textit{fair})? Due to the structural analogy, it is not surprising that similar answers have been given. Most prominently, expected utility theory in rational choice has average utilitarianism as its social counterpart \citep{buchak2017taking}. The analogy also yields an interesting interpretation for probability: an individual considers its possible ``future selves'', which would result from each outcome, which makes the question of how to value the gamble equivalent to the problem of finding a fair distribution among those future selves.

Relevant to this discussion is the distinction between \textit{aggregate} (or groupist) and \textit{individual} risk \citep{dawid2017individual}. The former is what we know well from probability theory: statements about relative frequencies are aggregate statements. When tossing a fair coin, on what basis do we assign the probability $p=0.5$ that it will land heads? Typically, the reasoning proceeds from the aggregate to the individual here. 
A frequentist explanation is that we have observed many coin tosses and the relative frequency of heads stabilized around $0.5$ (although such a statement can only be made in the limit of infinitely many tosses, which itself is problematic). A Bayesian may appeal to a symmetry principle, because there is no reason to favour either heads or tails for a fair coin, one should assign the degree of belief $0.5$ that it will land heads. Such a notion might strike one as individualistic. In practice, however, Bayesian inference is typically with respect to an exchangeable information base – an aggregate. The Bayesian might have flipped the coin many times, considered the sequence exchangeable \citep{de2017theory}, updated their beliefs accordingly and hence arrived at a probability of $0.5$. Similarly, if the Bayesian used prior knowledge from other fair coins, which they had experience with, this has an ‘aggregate flavour’. \cite{dawid2017individual} concludes by stating that the group to individual inference direction remains problematic and elusive.

While the above example of a coin seems innocent, it is problematized in ethical contexts, where the individual coin toss is replaced by an event that concerns a human. A fair ML problem can be phrased as distributing loss (in the ML sense) over individuals or subgroups of ethically fungible individuals. Such a subgroup (\eg men, women), according to the designer, is then viewed as an individual in the given context. 
One possibility of expressing imperfect fairness in this context is that we demand that subgroup losses are commensurate, \ie they should not differ much.
Under the assumption of mutual disinterest, an individual (subgroup) is concerned only with its own risk. When the aggregate risk is low, \ie the average individual risk is low, this is no consolation for any individual, who does not care for the average. When an inference is based on an aggregate, how can we control individual risk? Classical probability, firmly based on an aggregate conception due to its ‘casino origin’, is of no help.


As a corollary, we find that the concept of individual risk has mirror images in rational and social choice. On the one hand, inequality aversion can be understood as risk aversion, as it is a focus on the worse outcomes. On the other hand, a risk-averse individual is one that is inequality-averse with respect to future selves. In essence, this means that the individual is concerned with its own individual risk, instead of merely its aggregate risk of future selves. A decision maker who uses the expectation is risk-neutral and cares only about aggregate risk; here, the aggregate is to be understood as formed from the possible outcomes for that single individual. This is reasonable, when an experiment is repeated under stable conditions indefinitely and the possibility of a catastrophic event (\eg going bankrupt) is excluded. However, in real ML problems, this is not the case and often individuals only have a single shot, for instance at getting a loan. When choosing between a sure gain of $c$ or a lottery which yields $2c$ with probability $0.5$ and $0$ otherwise, almost all individuals choose the sure gain \citep{cappelen2013just}. Our interpretation of this robust pattern is that they care about their individual risk and adopt a pessimistic attitude. Standard expected utility theory would model this via a concave attitude towards wealth. But even if $c$ would already be in units of loss, as is the case in ML, we are inclined to think that it is still preferable to have less spread: because we care about individual risk.

Spectral risk measures, provide a partial resolution to this conflict between aggregate and individual, between average utilitarianism and subgroup fairness. The extreme points of the family of spectral risk measures are $\cvar$. Here, $\alpha=0$ recovers the risk (inequality) neutral expectation. On the other hand, $\alpha=1$ embodies the maximally risk (inequality)-averse attitude. In distributive justice, the corresponding theory is the Rawlsian maximin principle \citep{rawls}, where only the position of the worst-off counts. The parameter $\alpha$ offers a smooth interpolation between these two ends of the spectrum. The behaviour $\cvar$ is extreme in the sense that it neglects all outcomes below the $1-\alpha$ tail of losses. Finer control is possible by employing any spectral risk measure, where the tradeoff aggregate vs. individual is encoded in the shape of the fundamental function $\phi$.

\section{Experiments}
\label{app:experiments}
\begin{figure}
    \centering
     \begin{subfigure}[b]{0.37\textwidth}
        \begin{overpic}[width=\textwidth]{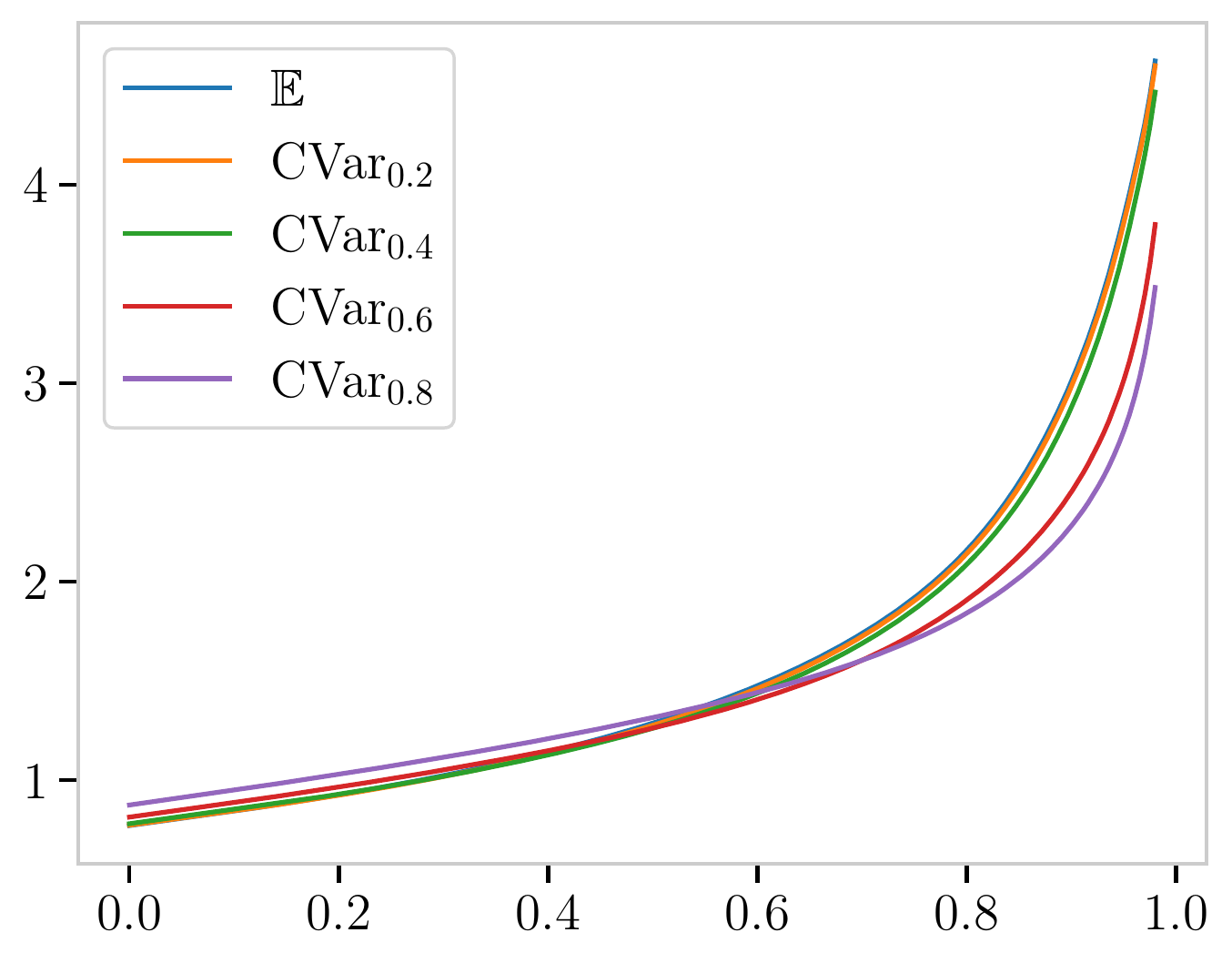}
            \put(50,-3){\small{$\alpha$}} 
            \put(-5,30){\rotatebox{90}{\small{$\cvar$}}} 
        \end{overpic}
    \end{subfigure}\hspace{0.5cm}
    \begin{subfigure}[b]{0.37\textwidth}
        \begin{overpic}[width=\textwidth]{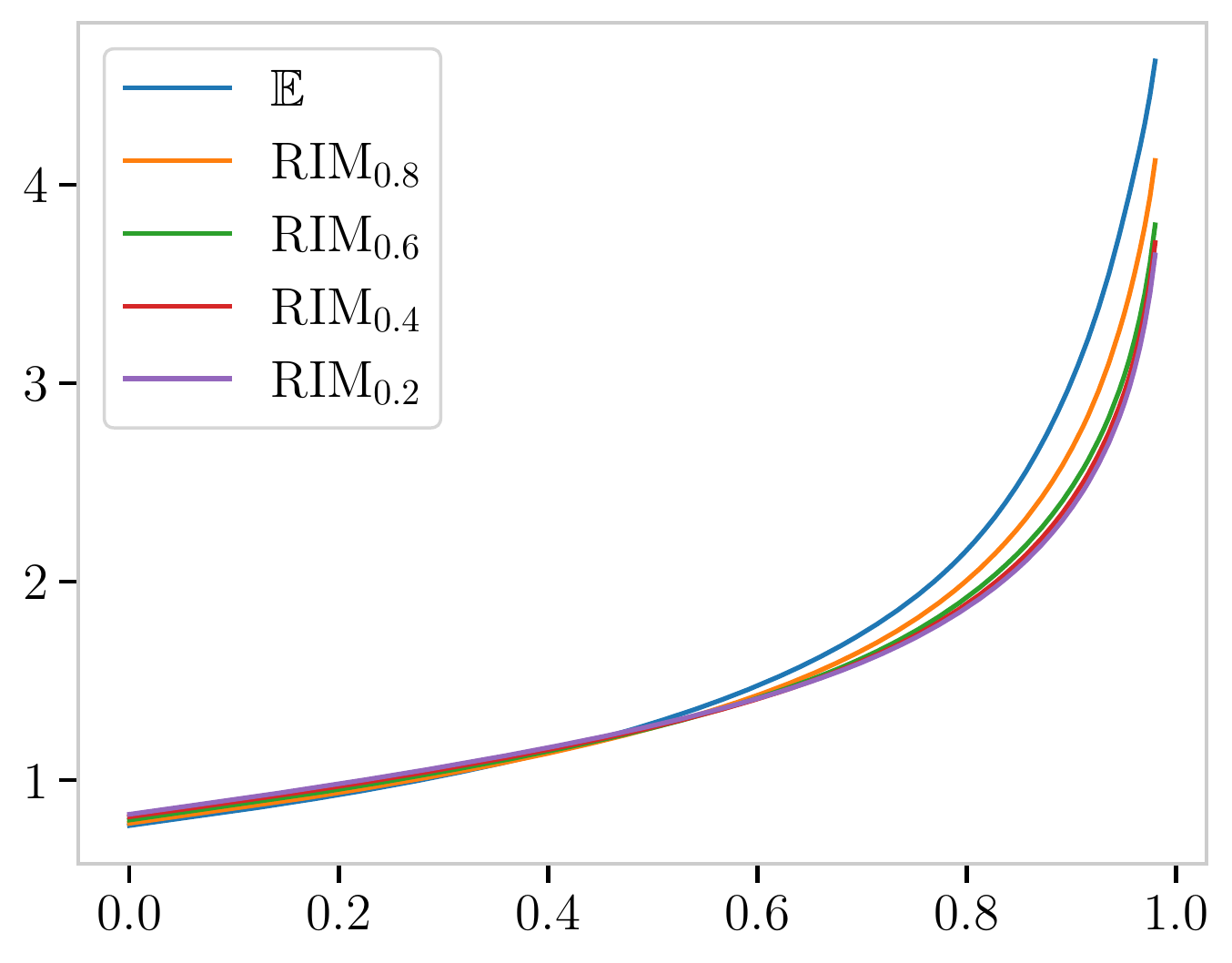}
            \put(50,-3){\small{$\alpha$}} 
            \put(-5,30){\rotatebox{90}{\small{$\cvar$}}} 
        \end{overpic}
    \end{subfigure}\\
    \vspace{0.4cm}
    \begin{subfigure}[b]{0.37\textwidth}
        \begin{overpic}[width=\textwidth]{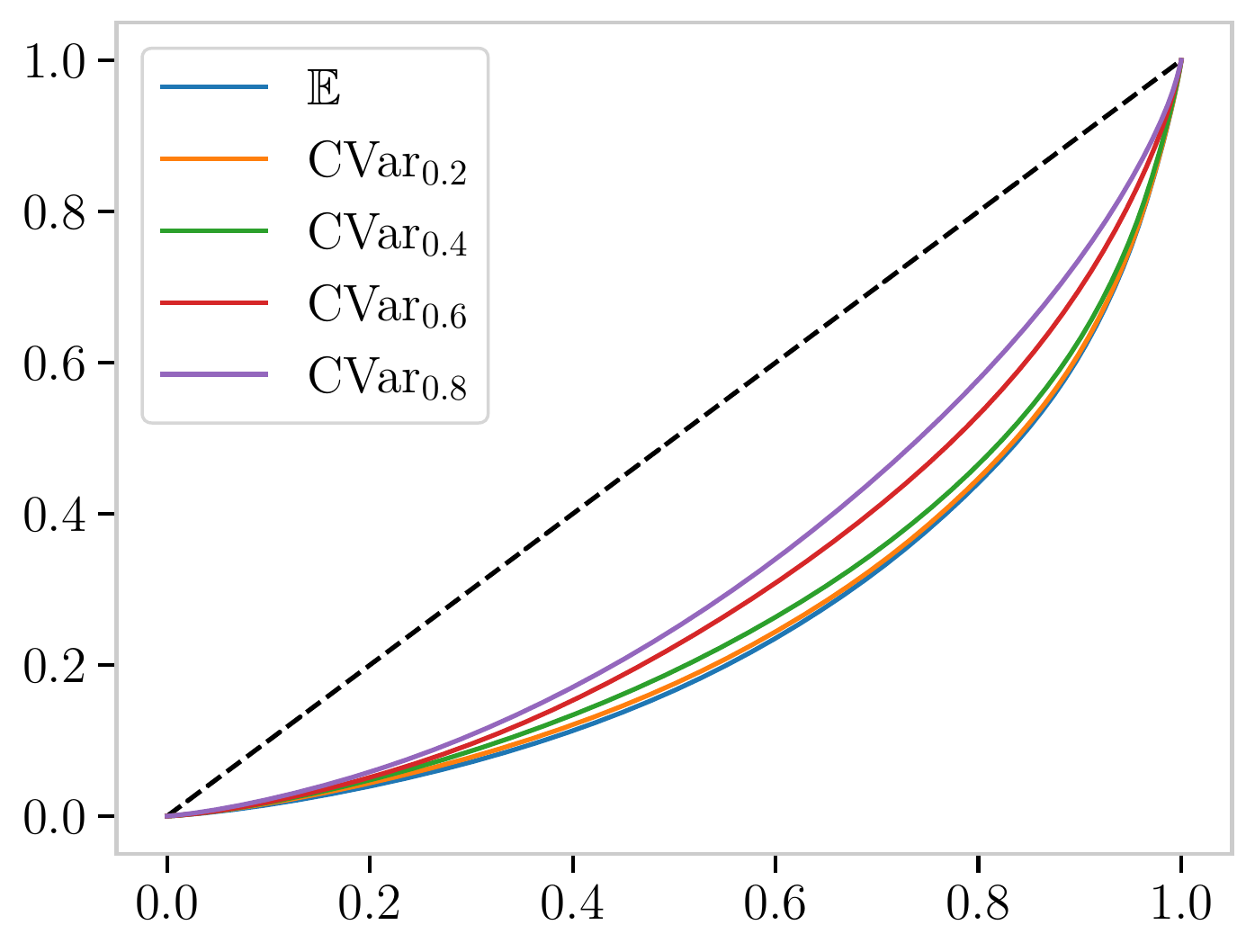}
            \put(21,-3){\tiny{Cumulative share of individuals}} 
            \put(-5,10){\rotatebox{90}{\tiny{Cumulative share of losses}}} 
        \end{overpic}
    \end{subfigure}
     \hspace{0.5cm}
     \begin{subfigure}[b]{0.37\textwidth}
        \begin{overpic}[width=\textwidth]{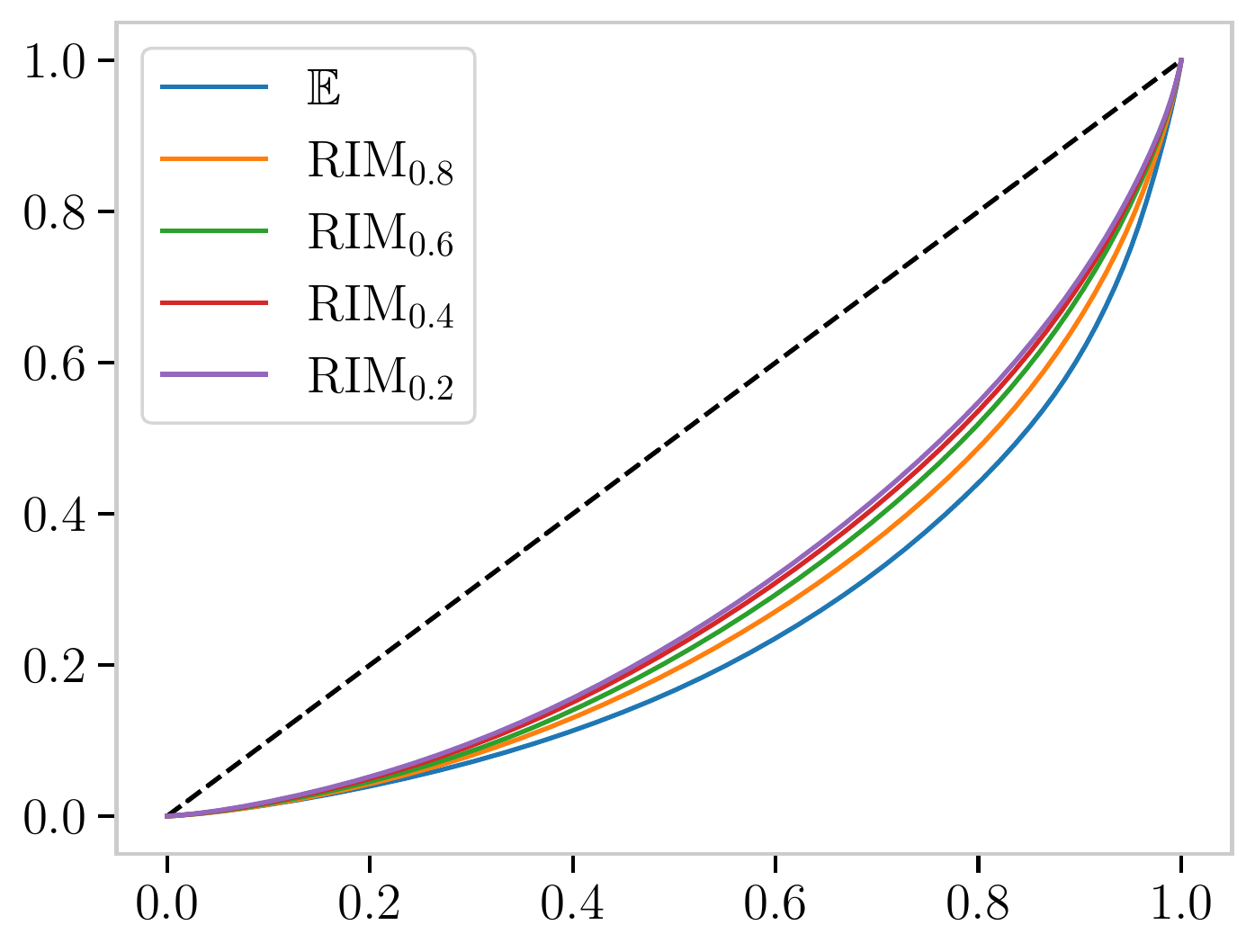}
            \put(21,-3){\tiny{Cumulative share of individuals}} 
            \put(-5,10){\rotatebox{90}{\tiny{Cumulative share of losses}}} 
        \end{overpic}
    \end{subfigure}
    \caption{PCA* results on \texttt{adult}. Top row: $\cvar$ curves of test losses for $\cvar$ risk measures (left) with different $\alpha$ and $\rimnoa$s risk measures (right) with different $\beta$, indicated by subscript, where $\alpha=0.7$. For better visibility of the differences, we cut off $\alpha$ at $0.98$. Bottom row: Lorenz curves of test losses for $\cvar$ (left) and $\rimnoa$s (right) with $\alpha=0.7$ and different $\beta$.}
    \label{fig:adult_results}
\end{figure}

\begin{figure}
    \centering
    \begin{subfigure}[b]{0.45\textwidth}
    \includegraphics[width=\textwidth]{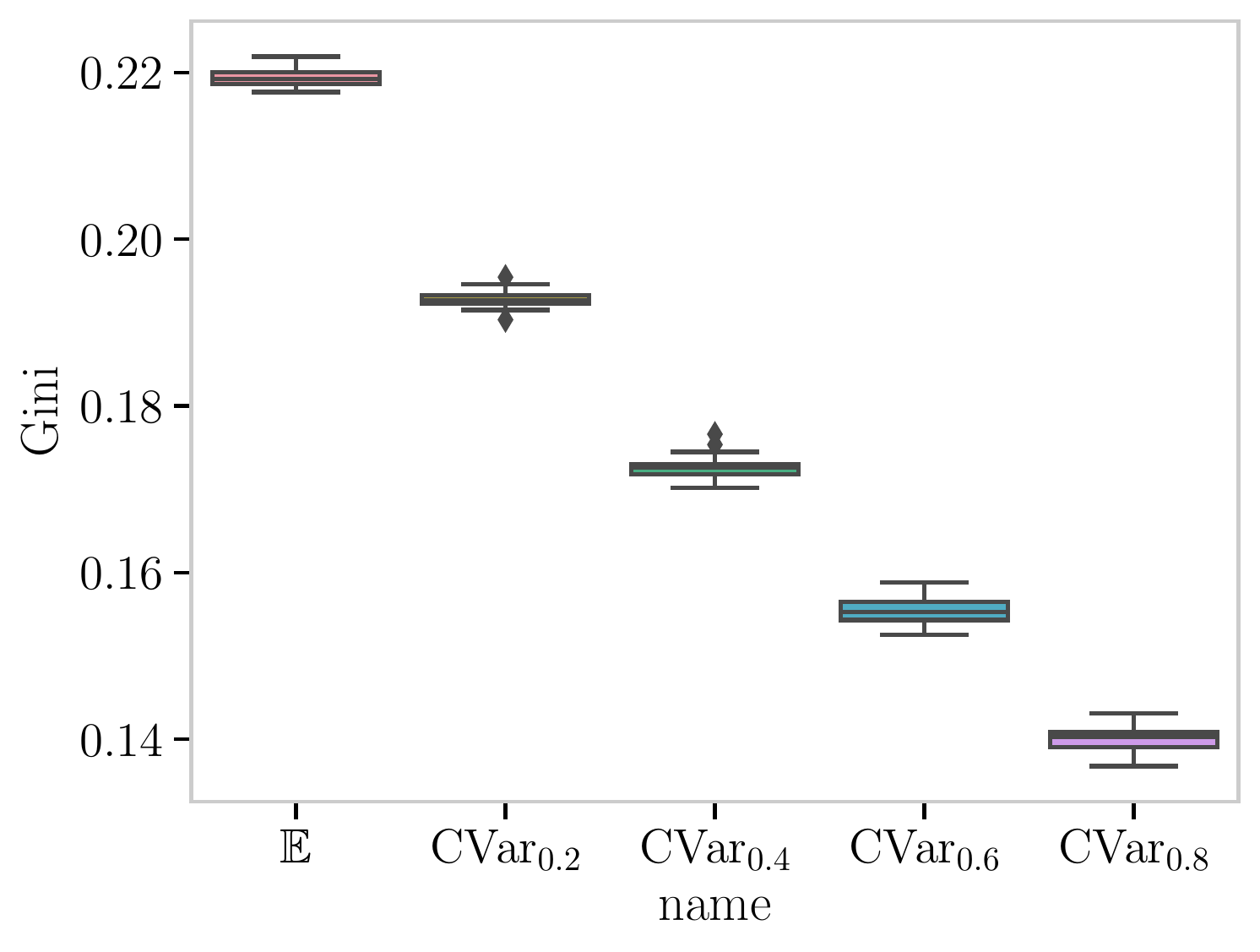}
    \end{subfigure}
    \begin{subfigure}[b]{0.45\textwidth}
    \includegraphics[width=\textwidth]{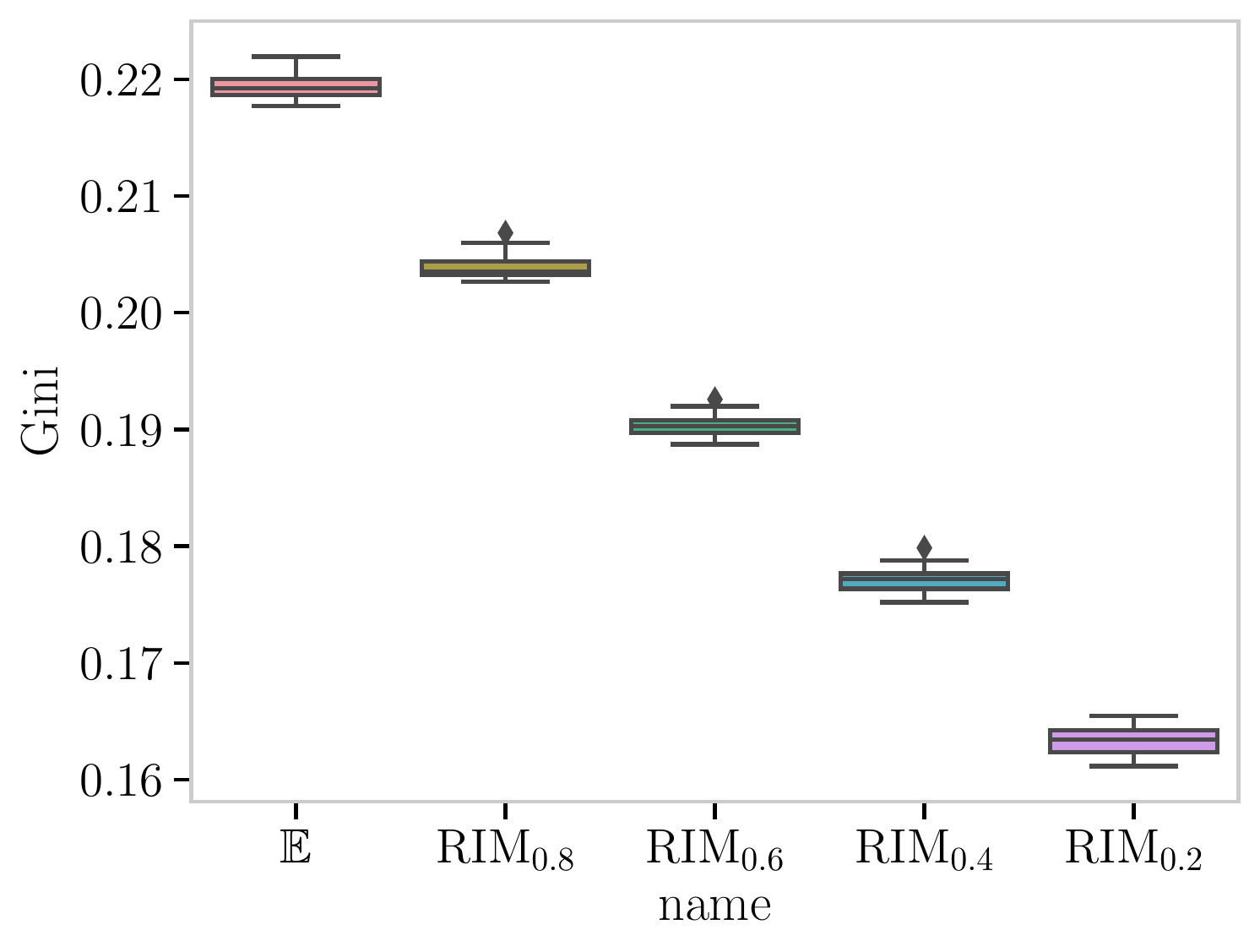}
    \end{subfigure}
    \begin{subfigure}[b]{0.45\textwidth}
    \includegraphics[width=\textwidth]{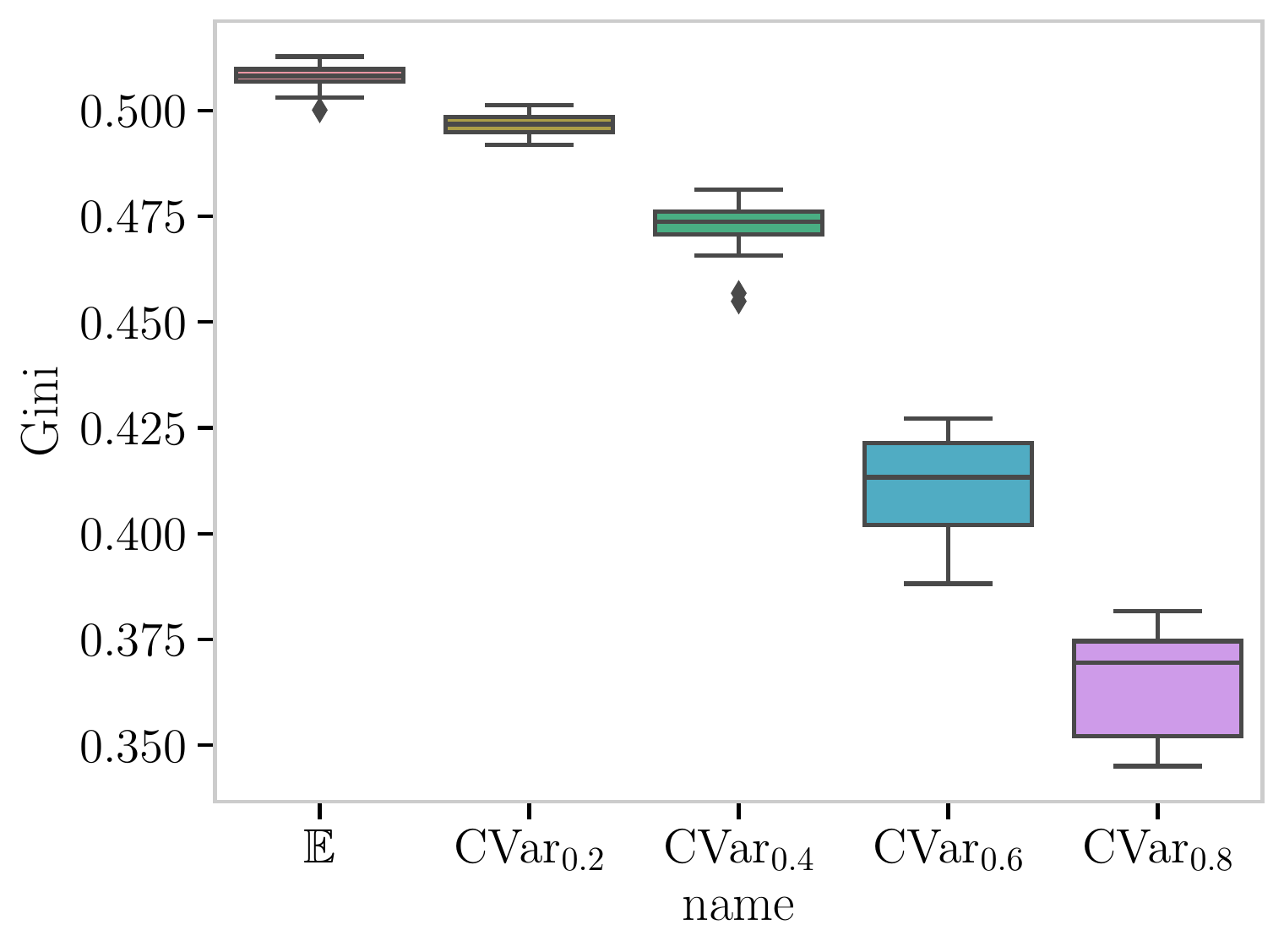}
    \end{subfigure}
    \begin{subfigure}[b]{0.45\textwidth}
    \includegraphics[width=\textwidth]{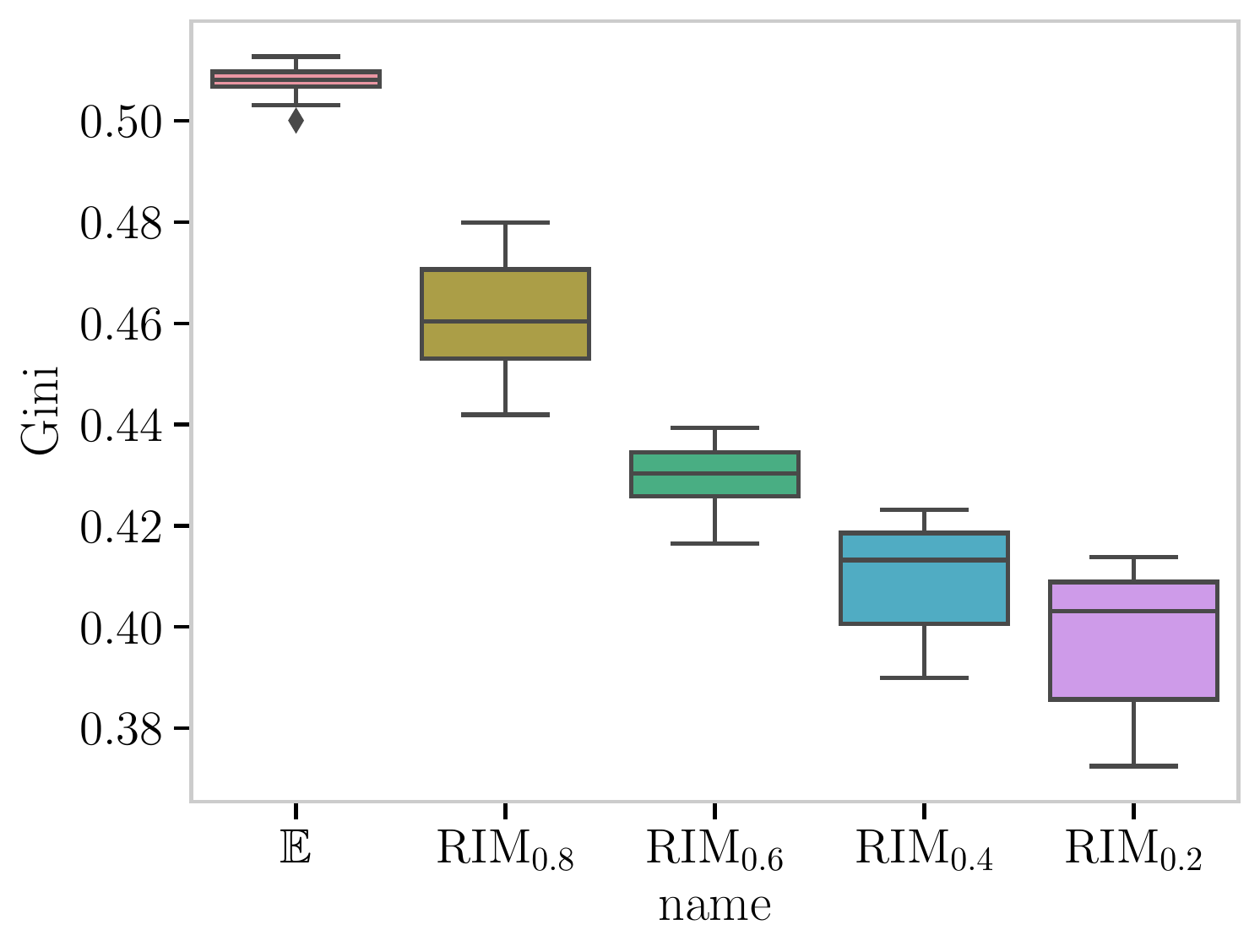}
    \end{subfigure}
    \caption{PCA* Gini coefficients for \texttt{MNIST} (top) and \texttt{adult} (bottom) over $25$ runs.}
    \label{fig:pca_ginis}
\end{figure}

\begin{figure}
    \centering
    \includegraphics[width=0.55\textwidth]{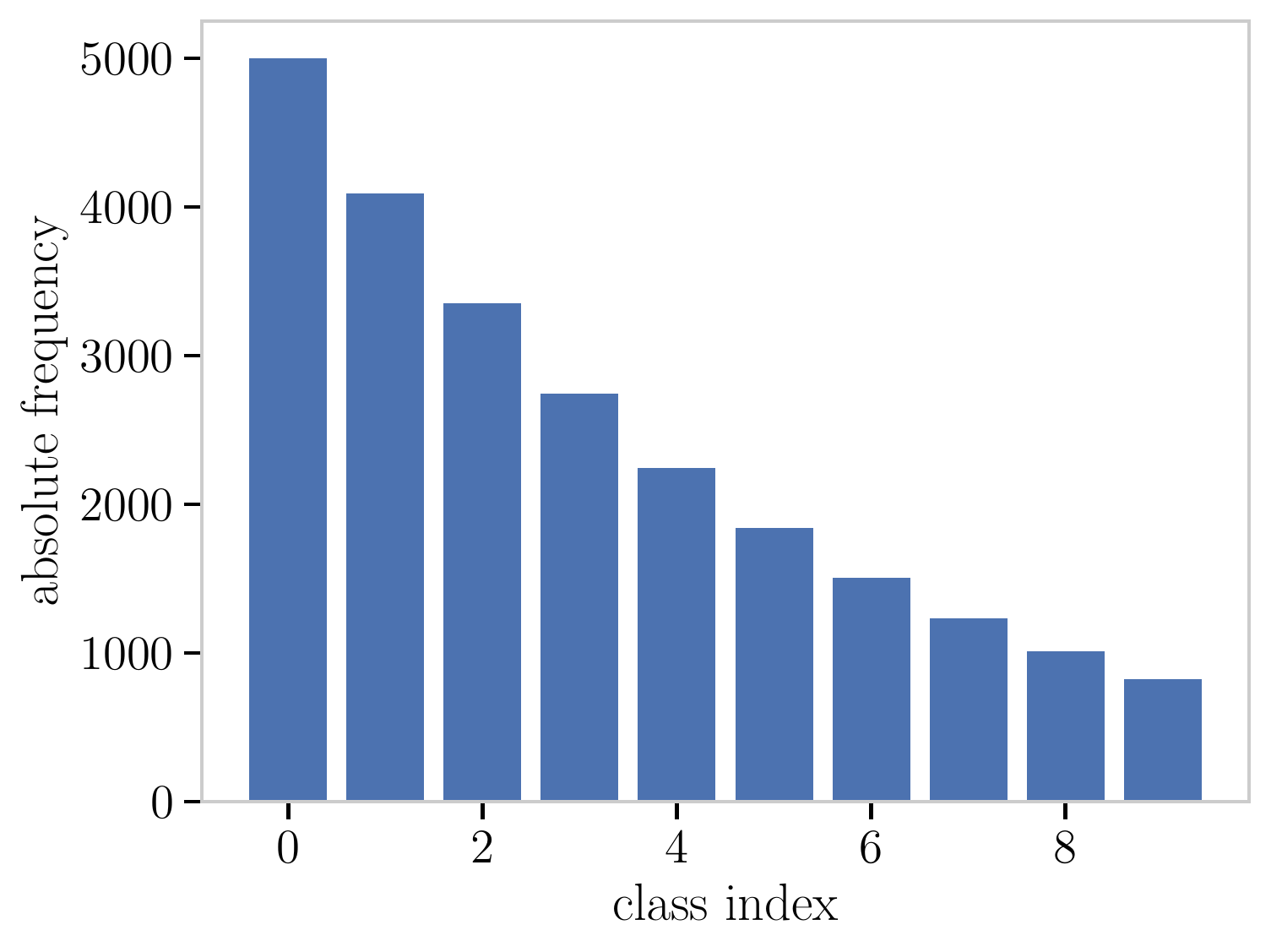}
    \caption{Class frequencies of imbalanced \texttt{MNIST}. Before each iteration, the assignments of digit class to frequency are randomly shuffled.}
    \label{fig:imbalance}
\end{figure}

Here we report additional results, not shown in the main paper. See Figures~\ref{fig:adult_results}, \ref{fig:pca_ginis}, \ref{fig:imbalance}.

\clearpage

\bibliography{bib}
\end{document}